\PassOptionsToPackage{notheorems}{jmlrutils}
\PassOptionsToClass{12pt}{jmlr}
\documentclass{colt2026}

\usepackage{times}
\usepackage{amsthm,mathtools}
\usepackage{adjustbox}
\usepackage{booktabs}
\usepackage{enumitem}

\makeatletter
\patchcmd{\@jmlr@check@packages}
  {\@ifpackageloaded{amsthm}{%
     \ClassError{jmlr}{Package `amsthm' detected.\MessageBreak
      This package conflicts with the jmlr class}{}}{}}
  {}{}{\PackageError{colt2027}{Could not adapt the JMLR amsthm check}{}}
\makeatother

\newtheorem{theorem}{Theorem}
\newtheorem{proposition}[theorem]{Proposition}
\newtheorem{corollary}[theorem]{Corollary}
\newtheorem{lemma}[theorem]{Lemma}
\newtheorem{conjecture}[theorem]{Conjecture}
\theoremstyle{definition}
\newtheorem{definition}[theorem]{Definition}
\newtheorem{example}[theorem]{Example}
\newtheorem{remark}[theorem]{Remark}


\newcommand{\cP}{\mathcal{P}}
\newcommand{\cC}{\mathcal{C}}
\newcommand{\cT}{\mathcal{T}}
\newcommand{\E}{\mathbb{E}}
\newcommand{\Prob}{\mathbb{P}}
\newcommand{\R}{\mathbb{R}}
\newcommand{\Sym}{\mathfrak{S}}
\newcommand{\ov}{\mathrm{ov}}
\DeclareMathOperator{\tr}{tr}
\DeclareMathOperator{\Var}{Var}

\title{Shared information governs multi-view planted matching}
\author[Wang Hu]{%
 \Name{Ruiqi Wang} \Email{riw010@ucsd.edu}\\
 \addr Department of Physics, University of California San Diego, La Jolla, CA 92093, USA
 \AND
 \Name{Jingtong Hu} \Email{jhu130@ucr.edu}\\
 \addr Department of Mathematics, University of California, Riverside, Riverside, California 92521, USA}
\hypersetup{
  pdftitle={Shared information governs multi-view planted matching},
  pdfauthor={Ruiqi Wang, Jingtong Hu},pdfsubject={},pdfkeywords={}
}

\begin{document}
\maketitle
\thispagestyle{plain} %

\begin{abstract}
We study when the rows of $K$ shuffled views of the same $n$ objects can be matched, where each object is observed once in every view through a Gaussian feature vector with known covariance. For every fixed $K$, all views can be matched up to $o(n)$ errors if the shared information $J$ of one object's $K$ observations exceeds $(1+\epsilon)\log n$, and not if it is below $(1-\epsilon)\log n$; this can hold although no two views can be matched on their own. Below the threshold, two views can still be matched if some set of views containing both has shared information above $\log n$; otherwise every estimator, even one that reads all views, has vanishing overlap between them. Matching every row is governed by the weakest bipartition of the views: with weak correlations the threshold is $2\log n$ for the mutual information across it. For a weak common signal observed through independent noise, the two thresholds are $4/K$ and $4/(K-1)$ in units of $\sum_\ell\rho_\ell^2/\log n$, which answers a question of Wee, Yang, Fan and Mao. Beyond weak correlations the bipartition condition is not sufficient: rare subpopulations impose a R\'enyi condition for every partition into at least three blocks. We conjecture that the two conditions together characterize exact recovery for Gaussian laws, and prove this for three views with equal correlations and in several further cases, in part with computer assistance. With equal correlations, layered screening attains $4/K$ for $K=3,4$ in $O(Kn^2d)$ operations, and for general laws a recursive decoder attains $J$. The almost-exact thresholds extend to known product laws and to objects observed in random subsets of the views.

\end{abstract}
\section{Introduction}\label{sec:intro}

The same $n$ objects are often listed several times, in different and unknown orders: keypoints seen by several cameras, records of the same individuals in several databases, cells measured in several modalities. Each list, or \emph{view}, holds one feature vector per object, and the features of an object are correlated across views. The task is to find out which rows belong together. We ask when almost all rows, or all rows, of $K$ views can be matched, in a model where the $K$ feature vectors of an object are jointly Gaussian with a known covariance and the objects are independent.

For two views the answer is governed by the mutual information $I$ between the two observations of an object~\citep{cullina2018fundamental,dai2019database,dai2020achievability,shirani2019concentration,kunisky2022strong,wang2022random,dai2023gaussian,wee2026geometric}. For weakly correlated Gaussian databases, \citet[Theorem~1]{dai2020achievability} show that all but $n^\lambda$ rows, for some fixed $\lambda<1$, can be matched when $I\ge(1+\epsilon)\log n$. In the high-dimensional isotropic model, all but $o(n)$ rows can be matched above this level, and below $(1-\epsilon)\log n$ no estimator matches a positive fraction of rows~\citep[Theorem~2.1 and Corollary~2.2]{wee2026geometric}; the results of~\citet[Theorems~2 and~3]{yarandi2026sharp} on attributed graphs give the same transition when the graphs are uncorrelated. Matching every row needs $I$ above $2\log n$~\citep[Theorems~1 and~2]{dai2019database}.

\begin{figure}[t]
\centering
\includegraphics[width=\linewidth]{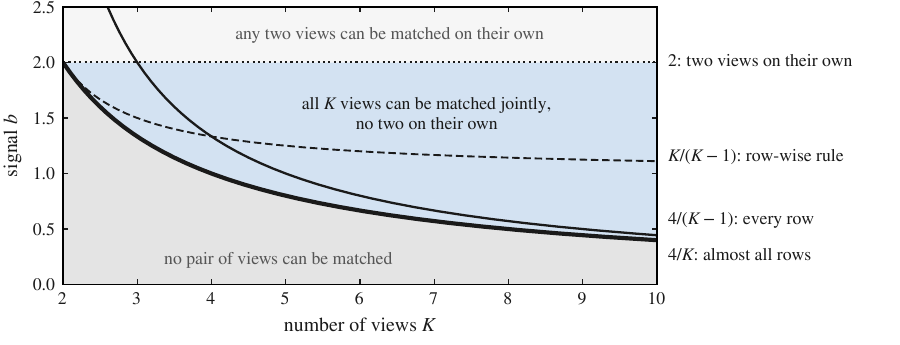}
\caption{A common signal with weak correlations, $b=\sum_\ell\rho_\ell^2/\log n$. Above $4/K$ almost all rows of all views can be matched, and below it no positive fraction of any pair of views; every row can be matched above $4/(K-1)$ and not below (Corollary~\ref{cor:equi}). The row-wise rule of~\citet{wee2026geometric} needs $b>K/(K-1)$, and decoders that match pairs of views need $b>2$ (Theorem~\ref{thm:ladder}).}\label{fig:concept}
\end{figure}

With more views, matching them jointly can succeed where every pair fails. Suppose that each view observes a common signal through independent noise, and measure the signal by $b=\sum_\ell\rho_\ell^2/\log n$, where $\rho_\ell$ is the correlation between two views in coordinate $\ell$. With weak correlations two views have $I\approx(b/2)\log n$, so they can be matched from their own rows only if $b>2$. We show that all $K$ views can be matched as soon as $b>4/K$ (Figure~\ref{fig:concept}). For $4/K<b<2$ every pair of views is then matched on almost all rows, although no pair can be matched alone and every decoder that matches pairs and composes them fails. \citet{wee2026geometric}, who introduced this model with isotropic noise, showed that maximizing the sum of pairwise inner products one anchor row at a time succeeds for $b>K/(K-1)$ and asked whether this level is optimal~\citep[Remark~2.6]{wee2026geometric}. The same model had been studied as multiple feature matching by~\citet{even2025computational}. In our high-dimensional regime their bounds, stated up to multiplicative constants, place both recovery thresholds at $b$ of order $1/K$; they also give a polynomial-time algorithm and low-degree lower bounds that indicate a statistical-computational gap when the number of views is large. For $K$ correlated Gaussian graphs the threshold was found by~\citet{vassaux2025feasibility}: it is $2/K$ times the two-graph threshold.

For a general Gaussian law, the quantity that replaces $I$ is the \emph{shared information}. For a partition $\cP$ of the views, let $D_\cP$ be the divergence between the law of one object's tuple and the product of its block marginals, and let $J=\min_\cP D_\cP/(|\cP|-1)$; for two views $J=I$.

\paragraph{Results.} Throughout, $K$ is fixed and thresholds are stated with fixed margins. The results fall into three groups.

\smallskip\noindent\emph{Thresholds for Gaussian features} (Sections~\ref{sec:almost}--\ref{sec:beyond}).
\begin{enumerate}[leftmargin=*,itemsep=1pt,topsep=2pt]
\item \textbf{Almost exact recovery is governed by the shared information.} All relative permutations can be recovered up to $o(n)$ errors if $J\ge(1+\epsilon)\log n$, and by no estimator if $J\le(1-\epsilon)\log n$ (Theorem~\ref{thm:main}).
\item \textbf{Which pairs survive.} Below the threshold, two views $a,b$ can still be matched almost exactly if $J_B>\log n$ for some set of views $B\ni a,b$. Otherwise every estimator, even one that reads all views, has vanishing overlap between $a$ and $b$ (Theorem~\ref{thm:pairwise}).
\item \textbf{Exact recovery with weak correlations is governed by the weakest bipartition.} Let $I_*$ be the least mutual information between the two sides of a bipartition of the views. With weak correlations, maximum likelihood recovers every row if $I_*\ge(2+\epsilon)\log n$, and for every Gaussian law no estimator does if $I_*\le(2-\epsilon)\log n$ (Theorem~\ref{thm:exact}). In the common-signal model the two thresholds are $b=4/K$ and $b=4/(K-1)$ (Corollary~\ref{cor:equi}). This answers the question of Wee et al.; their row-wise rule is sharp at $K/(K-1)$, because a wrong anchor row can collect the inner products among the other observations of a single object (Theorem~\ref{thm:ladder}).
\item \textbf{Beyond weak correlations, rare subpopulations matter.} For a fixed coordinate law, $I_*>2\log n$ and $J>\log n$ together do not suffice. Subpopulations of $n^\theta$ atypical objects impose a R\'enyi condition of order $(q-1)/q$ for every partition into $q\ge3$ blocks (Theorem~\ref{thm:rare}). For three views with equal negative correlations this condition is the exact threshold; at correlation $-1/3$ it is $d/\log n\approx10.17$, where the other two conditions hold from $9.87$ on (Theorem~\ref{thm:three}). We conjecture that the swap and R\'enyi conditions characterize exact recovery for Gaussian coordinate laws (Conjecture~\ref{conj:c1prime}) and prove this in several further cases (Theorem~\ref{thm:cases}).
\end{enumerate}

\smallskip\noindent\emph{Decoders and their limits} (Section~\ref{sec:algorithms}). Partition tests evaluate $O_K(n^K)$ candidate tuples, so for fixed $K$ the almost-exact threshold is reached in polynomial time; the question is how far the exponent can be lowered.
\begin{enumerate}[resume,leftmargin=*,itemsep=1pt,topsep=2pt]
\item \textbf{Faster decoders.} With equal correlations, layered screening attains $4/K$ for $K=3,4$, and $b>1$ for $K\ge5$, with $O(Kn^2d)$ operations on whitened coordinates (Theorem~\ref{thm:screen}). For general Gaussian laws, decoding along a tree of weakest partitions attains $J$, with $K-1$ two-view matchings when the partitions are binary (Theorem~\ref{thm:recursive}).
\item \textbf{Limits of specific fast decoders.} We prove no general computational lower bound. For $b<1$, a decoder that tests the triples in the product of two pairwise candidate lists of each anchor row needs $n^{3-b-o(1)}$ tests to reach a positive fraction of the true triples, which is more than quadratic, and $n^{3-b+o(1)}$ tests suffice; for $K=5$, rounding after either of the first two passes of Sinkhorn soft averaging has vanishing overlap on the first view it updates. For exact recovery, the half-edge dual prices that certify the linear-programming relaxation of the multi-index assignment when $I_*>3\log n$ are infeasible when $2\log n<I_*<3\log n$; whether the relaxation itself is exact there is open (Proposition~\ref{prop:barriers}).
\end{enumerate}

\smallskip\noindent\emph{Other observation models} (Section~\ref{sec:other}). The almost-exact thresholds survive several changes of the observation model, and the two obstructions to exact recovery do not depend on Gaussianity.
\begin{enumerate}[resume,leftmargin=*,itemsep=1pt,topsep=2pt]
\item \textbf{Known laws, missing views, unknown covariance.} The almost-exact and pairwise thresholds hold for every known product law with a fixed coordinate law (Theorem~\ref{thm:known}, Corollary~\ref{cor:product}) and for objects observed in random subsets of the views (Proposition~\ref{prop:overlap}). Unknown marginal covariances do not change them, but an unknown orthogonal cross-covariance does (Proposition~\ref{prop:unknown}).
\item \textbf{Exact recovery for laws that are not Gaussian.} The swap and rare-subpopulation conditions remain necessary (Theorem~\ref{thm:f2-converse}). Under weak dependence the exact threshold is the least partition divergence divided by the number of blocks, which a three-view parity law shows can be set by a partition into three blocks (Theorem~\ref{thm:f2-weak}).
\end{enumerate}

\paragraph{Techniques.} Theorem~\ref{thm:main} follows from a first-moment bound over the ways a candidate tuple can mix objects, and a counting converse. Three other arguments need more. The converse of Theorem~\ref{thm:pairwise} must rule out estimators that use views which can themselves be matched to $a$ or to $b$; we add shared discrete labels that make every node of a tree of weakest partitions decodable, and compare the result with a product law. For Theorem~\ref{thm:exact}, a union bound over wrong alignments reaches only $J\ge2\log n$. Alignments that break the same true pairs share the score of those pairs, so we charge its lower tail once per pattern of broken pairs; \citet[Appendix~C.3]{dai2023gaussian} use a related separation for two databases. Both directions of Theorem~\ref{thm:three} work with the tilted law that attains the R\'enyi rate. The converse (Theorem~\ref{thm:rare}) marks a random subpopulation whose rows follow this law and reveals everything else; converses for community recovery in correlated block models reveal all but a data-dependent set in a similar way~\citep{gaudio2022exact,racz2024harnessing}. The upper bound restricts maximum likelihood to alignments in which no set of tuples is too tilted, and controls the remaining Gaussian integrals by expanding around a fixed point of belief propagation and charging short cycles to the count of alignments.

\paragraph{Proofs and computer assistance.} Each section states its results and sketches the main arguments, and the proofs are in the appendices. Some of the results on exact recovery beyond weak correlations (Theorems~\ref{thm:three} and~\ref{thm:cases}) rest on finite-dimensional inequalities that are verified in exact rational arithmetic or by outward-rounded interval arithmetic. The statements say where this is the case, and the scripts and their recorded outputs are distributed with the paper as ancillary files. Three results come with proof outlines only: Proposition~\ref{prop:barriers}(c) and the two results of Appendices~\ref{app:beyond:axial} and~\ref{app:beyond:analytic}.

\paragraph{Related work.} \emph{Several databases.} For three databases, the dissertation of Zeynep~K~\citep[Chapter~7]{k2024database} analyzes exact recovery with a joint typicality decoder, which rules out wrong triples according to which databases they split, and a Fano bound; Theorem~\ref{thm:rare} adds a necessary condition coming from rare subpopulations. In the model of~\citet{even2025computational}, constrained $K$-means maximizes the sum of pairwise inner products within tuples, which is the maximum-likelihood estimate (Remark~\ref{rem:posterior}), so Theorem~\ref{thm:exact} gives the sharp constant for their exact-recovery problem in high dimension. Detection of correlated Gaussian databases is studied by~\citet{k2022detecting}, and multifile record linkage by~\citet{aleshin2023multifile}.

\emph{Several graphs.} \citet{vassaux2025feasibility} show that $K$ correlated Gaussian graphs can be aligned exactly when $n\rho^2/\log n>8/K$, and that below this level the fraction of vertices on which all $K$ permutations are simultaneously correct vanishes, so exact and almost exact recovery share one threshold. \citet{ameen2026converse} prove converses by revealing all but one correspondence; for graphs, their last-matching bound $\relpenalty=10000 n\rho^2/\log n>4/(K-1)$ is not tight when $K>2$. Theorem~\ref{thm:exact}(b) is a bound of the same kind, and for point features with weak correlations it is tight: the global threshold $4/K$ lies below the local one $4/(K-1)$, so the weakest bipartition decides exact recovery. Growing numbers of graphs, detection and attributed graphs are studied by~\citet{even2026statistical,ameen2025detecting,yarandi2026sharp}. In point-feature models every coordinate of a row moves with the row, whereas in graph models a permutation acts on both endpoints of an edge, so the graph proofs do not transfer directly.

\emph{Shared information.} $J$ is the multivariate mutual information of~\citet[Eq.~(4.1)]{chan2015multivariate}, also called shared information~\citep{narayan2016multiterminal}. It is the secret-key capacity of $K$ terminals with public discussion~\citep{csiszar2004secrecy,chan2008tightness,nitinawarat2012gaussian}, and for pairwise independent networks it equals the strength of the graph of edge informations~\citep{nitinawarat2010pin}. Info-clustering~\citep{chan2016info} thresholds the shared information of subsets, and \citet{raman2017universal} use it to cluster and register images. All-or-nothing transitions are also known in sparse regression, tensor PCA and graph alignment~\citep{reeves2019all,nilesweed2020all,wu2022settling,vassaux2025feasibility}, and planted matching with general weights is studied in~\citet{moharrami2021planted,ding2023planted}.

\section{Model and notation}\label{sec:model}

Fix $K\ge2$ and let $n\to\infty$; all other quantities may depend on $n$, and $\log$ is the natural logarithm. Object $i\in[n]$ carries a tuple $X_i=(X_i^1,\ldots,X_i^K)$ with $X_i^a\in\R^{d_a}$. The tuples are i.i.d.\ with a Gaussian law $P$ whose mean and covariance $\Sigma$ are known to the statistician; we take the mean to be zero. View $a$ is the matrix $Y^a\in\R^{n\times d_a}$ whose row $\pi_a(i)$ is $X_i^a$. The permutations $\pi_1,\ldots,\pi_K$ are i.i.d.\ uniform on $\Sym_n$ and independent of the tuples. Only relative positions matter, so we relabel objects by their row in the anchor view $1$; after this relabeling $\pi_1=\mathrm{id}$, and $\Pi=(\pi_2,\ldots,\pi_K)$ is uniform on $\Sym_n^{K-1}$.

For $B\subseteq[K]$, $P_B$ denotes the law of $X^B=(X^a)_{a\in B}$. For a partition $\cP$ of $[K]$ into $|\cP|\ge2$ blocks, let
\begin{equation}\label{eq:J}
D_\cP=D\Big(P\,\Big\|\,\bigotimes_{B\in\cP}P_B\Big),\qquad J=\min_{|\cP|\ge2}\frac{D_\cP}{|\cP|-1}.
\end{equation}
If $\Sigma\succ0$, then $D_\cP=\frac12\log\big(\prod_{B\in\cP}\det\Sigma_{BB}/\det\Sigma\big)$; the divergence may be infinite for degenerate $\Sigma$. For $K=2$, $J$ is the mutual information $I(X^1;X^2)$, and in general it is the shared information of $X^1,\ldots,X^K$~\citep{chan2015multivariate,narayan2016multiterminal}. For $B\subseteq[K]$ with $|B|\ge2$, $J_B$ is defined by~\eqref{eq:J} with $P_B$ in place of $P$, and $J_B=\infty$ if $|B|=1$. Whitening each view gives the joint covariance $R$ with blocks $R_{ab}=\Sigma_{aa}^{-1/2}\Sigma_{ab}\Sigma_{bb}^{-1/2}$; the \emph{weak-correlation regime} is $\|R-I\|_{\mathrm{op}}\to0$.

An estimator outputs maps $\widehat\pi_a:[n]\to[n]$, possibly randomized and not necessarily bijective. Its errors are $M_a=|\{i:\widehat\pi_a(i)\ne\pi_a(i)\}|$. Recovery is \emph{simultaneously almost exact} if $\max_{2\le a\le K}M_a/n\to0$ in probability, and \emph{exact} if $\max_{2\le a\le K}M_a=0$ with probability tending to one. For views $a\ne b$, the true correspondence from rows of $Y^a$ to rows of $Y^b$ is $\sigma_{ab}=\pi_b\circ\pi_a^{-1}$, and the overlap of a map $f:[n]\to[n]$ is $\ov(f)=n^{-1}|\{j:f(j)=\sigma_{ab}(j)\}|$. All thresholds are stated with fixed margins: a condition such as $J>\log n$ means $\liminf_nJ/\log n>1$.

\section{Almost exact recovery}\label{sec:almost}

\begin{theorem}[Sharp threshold]\label{thm:main}
\begin{enumerate}[label=(\alph*),leftmargin=*]
\item If $\liminf_{n}J/\log n>1$, the partition-test decoder defined below recovers all relative permutations simultaneously almost exactly.
\item Let $\cP$ be any partition with $m\ge2$ blocks, let $B_1$ be the block containing view $1$, and choose a representative $r_B\in B$ for each block, with $r_{B_1}=1$. For every $n\ge2$ and every estimator,
\begin{equation}\label{eq:converse}
\frac{1}{(m-1)n}\sum_{B\ne B_1}\E M_{r_B}\ \ge\ 1-\frac{D_\cP}{(m-1)\log n}-\frac{2}{\log n}.
\end{equation}
If $D_\cP=\infty$ the bound is vacuous. In particular, if $\limsup_{n}J/\log n<1$, then simultaneously almost exact recovery is impossible.
\end{enumerate}
\end{theorem}

\paragraph{Partition tests.} Fix an anchor row $i$ and a candidate tuple $t=(i,j_2,\ldots,j_K)$. Given $\Pi$, the views of $t$ split into blocks according to which object they come from. If this \emph{identity partition} is $\cP$, the features of $t$ have law $Q_\cP=\bigotimes_{B\in\cP}P_B$, because distinct objects are independent. The \emph{partition-test decoder} accepts $t$ if
\[
L_\cP(t)=\log\frac{dP}{dQ_\cP}(t)>(1+\gamma)(|\cP|-1)\log n\qquad\text{for every partition $\cP$,}
\]
with a small constant $\gamma>0$. Each anchor row proposes an accepted tuple if there is one, conflicts are resolved in each view separately, and the rows left over are matched arbitrarily. A false tuple with identity partition $\cP$ passes the $\cP$-test with probability at most $n^{-(1+\gamma)(|\cP|-1)}$, by Markov's inequality applied to $e^{L_\cP}$. At most $n^{|\cP|-1}$ tuples have this identity partition for a given anchor row, since the anchor's block is fixed and each other block picks one object. A union bound, which needs no independence, leaves $o(1)$ false acceptances per anchor row. The true tuple passes because $L_\cP$ has mean $D_\cP\ge(|\cP|-1)J$ and variance at most $2K^2D_\cP$ (Lemma~\ref{lem:llr}), so Chebyshev's inequality suffices. Resolving conflicts and completing costs at most twice the number of bad rows in each view (Appendix~\ref{app:main}).

\paragraph{Converse.} Reveal the correspondences inside each block of $\cP$. What remains are $m-1$ independent uniform permutations, whose mutual information with the data is at most $nD_\cP$. A permutation that differs from an estimate in $t$ positions can be described with about $t\log n$ nats, which gives~\eqref{eq:converse}.

\paragraph{Below the threshold.} Some correspondences may still be recoverable. Which ones is decided by the shared information of the subsets of views.

\begin{theorem}[Pairwise matchability]\label{thm:pairwise}
For views $a\ne b$ let $J^*_{ab}=\max_{B\supseteq\{a,b\}}J_B$.
\begin{enumerate}[label=(\alph*),leftmargin=*]
\item If $\liminf_n J^*_{ab}/\log n>1$, the partition-test decoder applied to the views of a maximizing $B$ gives a permutation $\widehat\sigma_{ab}$ with $\ov(\widehat\sigma_{ab})\to1$ in probability.
\item If $\limsup_n J^*_{ab}/\log n<1$, then $\sup_f\E\,\ov(f)\to0$, where the supremum is over all estimators $f:[n]\to[n]$ of $\sigma_{ab}$ that may use all $K$ views and their own randomness.
\end{enumerate}
In particular, if $\limsup_nJ/\log n<1$ and, for all large $n$, some partition attaining the minimum in~\eqref{eq:J} puts $a$ and $b$ in different blocks, then every estimator has vanishing overlap between $a$ and $b$.
\end{theorem}

For a threshold $c$, the inclusion-maximal subsets with $J_B>c$ are the clusters of the info-clustering of~\citet{chan2016info}; Theorem~\ref{thm:pairwise} says that, with fixed margins, the clusters at level $c=\log n$ are exactly the groups of views whose pairs are matchable, even when the estimator reads all views. The last sentence of the theorem is an all-or-nothing statement. A pair separated by a weakest partition lies in no subset whose shared information exceeds $J$~\citep[Theorems~5.2--5.3 and Proposition~5.3]{chan2015multivariate}, so $J^*_{ab}=J$ (Appendix~\ref{app:f1-pairwise} gives a direct proof for general laws) and the last sentence follows from~(b).

\paragraph{Proof idea for the converse.} Part~(b) does not follow from Theorem~\ref{thm:main}(b), because the estimator may use views that can themselves be matched to $a$ or to $b$. Suppose that $J^*_{ab}\le(1-2\delta)\log n$. Starting from $[K]$, split every set of views $B$ with $J_B\le(1-\delta)\log n$ along a partition attaining $J_B$. Every set that contains both $a$ and $b$ is split, so the two views end in different leaves of the resulting tree. Reveal the correspondences inside the $r$ leaves. At every internal node $B$, give each object, for every pair of children of $B$, one uniform label in $[M_B]$ seen by both children, with $M_B$ chosen so that the shared information of the node rises to $(1+\eta)\log n$. Every node can then be decoded by partition tests once its children are aligned, so the permutations between the leaves have conditional entropy $o(n\log n)$ given the augmented data. Adding divergences along the tree puts the augmented data within divergence $((r-1)\eta+o(1))\,n\log n$ of the product $\mathsf Q$ of the leaf marginals. On the other hand, an estimate with expected overlap $u>0$ between $a$ and $b$, which may be taken to be a permutation, pairs rows so that, at their lowest common node $B_0$, at least $un$ labels of $a$ and $b$ agree in expectation, whereas under $\mathsf Q$ the number of agreements is $\mathrm{Bin}(n,1/M_{B_0})$. Tilting $\mathsf Q$ by this number shows that the two laws are at divergence at least $(4u\delta/K-o(1))\,n\log n$, which does not depend on $\eta$; a small $\eta$ gives a contradiction (Appendix~\ref{app:f1-pairwise}; Appendix~\ref{app:pairwise} runs the argument with Gaussian coordinates in place of labels).

\paragraph{Other laws.} The Gaussian law enters Theorems~\ref{thm:main} and~\ref{thm:pairwise} only through the concentration of $L_\cP$ on true tuples. Section~\ref{sec:other} shows that both thresholds hold for known laws under a lower-tail condition, in particular for every fixed coordinate law with no moment assumption, and for objects that are observed in random subsets of the views.

\section{Exact recovery with weak correlations}\label{sec:exact}

Recovering every row requires more than $J>\log n$. A single wrong pair of rows suffices for an error, and the cheapest such error swaps two objects on one side of a bipartition of the views. Let
\begin{equation}\label{eq:Istar}
I_*=\min_{\varnothing\ne S\subsetneq[K]}I\big(X^S;X^{S^c}\big)
\end{equation}
be the minimum mutual information across a bipartition. In the weak-correlation regime this, and not $J$, is the threshold.

\begin{theorem}[Exact recovery]\label{thm:exact}
Fix $\epsilon\in(0,2)$.
\begin{enumerate}[label=(\alph*),leftmargin=*]
\item Suppose that $h=\|R-I\|_{\mathrm{op}}\to0$ for the whitened joint covariance $R$ of Section~\ref{sec:model}. If $I_*\ge(2+\epsilon)\log n$ for all large $n$, the maximum-likelihood estimate recovers all relative permutations exactly with probability tending to one.
\item For every Gaussian law, if $I_*\le(2-\epsilon)\log n$ for all large $n$, then for every estimator the probability of recovering all relative permutations exactly tends to zero. More precisely, for every pair of views $a,b$ separated by a minimizing bipartition, the probability that an estimate of $\sigma_{ab}$ is correct in every row tends to zero, even if all correspondences inside the two sides are revealed.
\end{enumerate}
\end{theorem}

Part~(b) does not need weak correlations. In the weak regime the two thresholds are ordered: splitting off each block of a $p$-block partition in turn cuts every separated pair exactly twice, and divergences add over cut pairs to first order (Proposition~\ref{prop:strength} below), so $D_\cP\ge(1+o(1))\,pI_*/2$ and $J\ge(1+o(1))I_*/2$.

\begin{remark}[Growing number of views]\label{rem:growK}
The converses of Theorems~\ref{thm:main} and~\ref{thm:exact} hold for every $K=K_n$. For general laws, Theorem~\ref{thm:main}(a) holds when $K\log K=o(\log n)$. In the weak regime $h\to0$, which in the common-signal model below means $(K-1)\rho_{\max}\to0$, both Theorem~\ref{thm:main}(a) and Theorem~\ref{thm:exact}(a) hold when $\log K=o(\log n)$, because, for a positive minimum cut, the number of partitions whose cut weight is at most $u$ times the minimum cut is at most $K^{9u}$. The condition on $K$ cannot be dropped. For $K=\lceil n^\beta\rceil$ with fixed $\beta>0$, take a star of $K-1$ independent two-view channels, each with mutual information $(2+\eta)\log n$ for a fixed $\eta<\beta$. Then $h\to0$ and $I_*=(2+\eta)\log n$, but each leaf is matched exactly with probability at most $1-n^{-\eta-o(1)}$, so exact recovery of all $n^\beta$ leaves fails (Appendix~\ref{app:growingviews}).
\end{remark}

\paragraph{A common signal.} Suppose that, after an invertible linear map applied identically to every view, the coordinates $\ell=1,\ldots,d$ are independent and the $K$ observations of an object in coordinate $\ell$ have law $N(0,R_K(\rho_\ell))$, where $R_K(s)=(1-s)I_K+s\mathbf 1\mathbf 1^\top$ and $\rho_\ell\in[0,1)$. This holds whenever $X_i^a=Z_i+E_i^a$ with a common center $Z_i\sim N(0,S)$ and i.i.d.\ view noise $E_i^a\sim N(0,N)$, $N\succ0$; then $\rho_\ell=\lambda_\ell/(1+\lambda_\ell)$ for the eigenvalues $\lambda_\ell$ of $N^{-1/2}SN^{-1/2}$. Let $A=\sum_\ell\rho_\ell^2$ and $\rho_{\max}=\max_\ell\rho_\ell$, so that $h=(K-1)\rho_{\max}$.

\begin{corollary}[The constants $4/K$ and $4/(K-1)$]\label{cor:equi}
In this model the singleton partition attains the minimum in~\eqref{eq:J}, and if $\rho_{\max}\to0$ then $J=(K/4+O_K(\rho_{\max}))A$ and $I_*=((K-1)/2+O_K(\rho_{\max}))A$. Hence, with fixed margins in $A/\log n$, simultaneously almost exact recovery is possible above $4/K$ and impossible below, in which case every pair of views has vanishing overlap; exact recovery is possible above $4/(K-1)$ and impossible below. In the model of~\citet{wee2026geometric}, where $X\sim N(0,I_d)$, the views are $X+\tau Z^a$, $\tau^4=d/(b\log n)$ and $d/\log n\to\infty$, one has $A/\log n\to b$. The almost-exact threshold is therefore $b=4/K$, strictly below their guarantee $K/(K-1)$ for every $K\ge3$, and the exact threshold is $b=4/(K-1)$.
\end{corollary}

\paragraph{Weak signal and spanning trees.} For nonnegative weights $w=(w_{ab})_{a<b}$ on the complete graph on $[K]$, the \emph{strength} is
\[
s(w)=\min_{|\cP|\ge2}\frac{\sum_{a<b \text{ in different blocks of }\cP}w_{ab}}{|\cP|-1}.
\]
By the Tutte--Nash-Williams theorem~\citep{tutte1961,nashwilliams1961}, applied to rational weights scaled to integers and extended by continuity, $s(w)$ equals the largest total weight of a fractional packing of spanning trees with capacities $w$~\citep{cunningham1985,nitinawarat2010pin}.

\begin{proposition}[$J$ is a graph strength to first order]\label{prop:strength}
Let $h=\|R-I\|_{\mathrm{op}}$ and $I_{ab}=I(X^a;X^b)=-\frac12\log\det(I_{d_a}-R_{ab}R_{ab}^\top)$. There is a constant $C_K$, independent of the dimensions $d_a$, such that whenever $h\le1/C_K$, every partition satisfies
\[
D_\cP=(1+\tau_\cP)\sum_{a<b\text{ in different blocks}}I_{ab},\qquad |\tau_\cP|\le C_Kh,
\]
and hence $J=(1+\tau)\,s\big((I_{ab})_{a<b}\big)$ and $I_*=(1+\tau')\min_S\sum_{a\in S,\,b\notin S}I_{ab}$ with $|\tau|,|\tau'|\le C_Kh$.
\end{proposition}

The error is relative to the cut itself, so the approximation stays accurate when the weakest cut is much weaker than the rest. In the weak regime the almost-exact threshold asks the graph of pairwise informations to pack $\log n$ worth of spanning trees, and the exact threshold asks its minimum cut to carry $2\log n$. Matching along a single spanning tree needs every edge of the tree to carry $\log n$ (Appendix~\ref{app:weak}).

\paragraph{Proof idea: converse.} Let $\{S,S^c\}$ attain $I_*$ and reveal all correspondences inside $S$ and inside $S^c$. What remains is a two-database problem between $X^S$ and $X^{S^c}$. In canonical coordinates with correlations $r_j$, exchanging two rows $i,k$ of one side lowers the log-likelihood by $\Delta_{ik}=\sum_jw_j(U_{ij}-U_{kj})(V_{ij}-V_{kj})$, with $w_j=r_j/(1-r_j^2)$. Tilting by $e^{-\Delta_{ik}/2}$ makes every canonical pair independent and has normalizing constant $e^{-I_*}$, which gives $\Prob(\Delta_{ik}<0)\ge\frac38e^{-I_*-\sqrt{8I_*}}$, while two exchanges that share a row both win with probability at most $e^{-3I_*/2}$. If $I_*\le(2-\epsilon)\log n$, the second moment method yields an exchange that beats the truth, so maximum likelihood, which maximizes the probability of exact recovery, fails.

\paragraph{Proof idea: achievability.} The log-likelihood is a sum of pairwise scores (Remark~\ref{rem:posterior}). An alternative to the truth wins when the scores $F$ of the false pairs of rows it creates are at least the scores $T$ of the true pairs it breaks. Group the alternatives by their \emph{pattern}, which records for each object which of its views stay together. A pattern determines $T$. If it splits $m=n^\alpha$ objects into $m+g$ groups in total, there are at most $e^{L}$ such patterns and at most $e^{M}$ alternatives with a given pattern, where
\[
L=(1-\alpha)\,m\log n+O_K(m),\qquad M=\alpha\,g\log n ,
\]
because an object is one of $n$ but each group that leaves its object has only $m$ possible destinations. In the weak regime, with $\mu=\E T$, the pair $(T-\mu,F)$ is jointly sub-Gaussian with a variance proxy $V$ such that $\mu^2/V\ge D:=2(1+\epsilon_0)(m+g)\log n$ for a fixed $\epsilon_0>0$, uniformly over patterns, and $\E F=0$; this uses $I_*\ge(2+\epsilon)\log n$ and the fact that a partition into $p$ blocks separates at least $p/2$ times the weight of the weakest bipartition. The bound $\Prob(F\ge T)\le e^{-D/4}$ for a single alternative beats their number $e^{L+M}$ when $M\le L$, but not in general: used for all patterns it would require $J\ge2\log n$. When the alternatives outnumber the patterns, we use that they share $T$: its lower tail is charged once per pattern, and the union over the $e^M$ alternatives is taken only on the event that $T$ is typical. For $a+s=1$ with $a\le s$,
\[
\Prob(T<s\mu)\le e^{-a^2D/2},\qquad \Prob(F\ge T\ge s\mu)\le e^{-(s^2+a^2)D/2},
\]
and both $a^2D/2\ge L$ and $s^2D/2\ge M$ can be met, with room to spare, because by the Cauchy--Schwarz inequality $(\sqrt L+\sqrt M)^2\le(1+o(1))(m+g)\log n\le D/2$ (Appendix~\ref{app:exact}).

\section{Exact recovery beyond weak correlations}\label{sec:beyond}

Outside the weak regime, $I_*>2\log n$ is not sufficient for exact recovery. One reason is that $J>\log n$ is also necessary and can be the stronger condition.

\begin{remark}[A near-linear constraint]\label{rem:general}
Take scalar views with
$R=\frac{K}{K-1+\delta}\big(I_K-(1-\delta)uu^\top\big)$, $u=K^{-1/2}\mathbf 1$ and $\delta=n^{-2c}$ with $c>0$ fixed, a near-linear constraint among the views. Every nontrivial partition then has $D_\cP=c\log n+O_K(1)$, so $I_*=c\log n+O_K(1)$ while $J=c\log n/(K-1)+O_K(1)$, and exact recovery is possible for $c>\max\{2,K-1\}$ and impossible for $c<\max\{2,K-1\}$ (Appendix~\ref{app:nearconstraint}). For $K\ge4$ and $2<c<K-1$ the minimum cut exceeds $2\log n$, but $J<\log n$ and even almost exact recovery fails.
\end{remark}

A second obstruction is not captured by $I_*$ and $J$ together. It appears at an intermediate scale: a set of $n^\theta$ objects, $0<\theta<1$, whose features happen to be atypical, among which the views share too little information to be matched. No error pattern on a bounded number of objects detects it.

In the rest of this section each tuple consists of $d$ i.i.d.\ coordinates with a fixed density $p$ on $\R^K$, one scalar per view, and $I_*$, $J$ are per coordinate, so the conditions above read $dI_*>2\log n$ and $dJ>\log n$. For $0<\alpha<1$ let $D_\alpha(p\|v)=(\alpha-1)^{-1}\log\int p^\alpha v^{1-\alpha}$ be the R\'enyi divergence of order $\alpha$.

\begin{definition}[Rare-subpopulation rate]\label{def:rare}
For a partition $\cP$ of $[K]$ with $q\ge2$ blocks let
\[
R_\cP=\frac{1}{q-1}\min_{v=\bigotimes_{B\in\cP}v_B}D_{(q-1)/q}\big(p\,\big\|\,v\big),
\]
the minimum over densities that factor over the blocks of $\cP$. Equivalently,
\[
R_\cP=\min_{Q}\Big\{D(Q\|p)+\frac{1}{q-1}\,D\Big(Q\,\Big\|\bigotimes_{B\in\cP}Q_B\Big)\Big\},
\]
where $Q_B$ are the block marginals of $Q$.
\end{definition}

The second form explains the first. About $n^\theta=n\,e^{-dD(Q\|p)}$ objects have empirical coordinate law close to $Q$, and by analogy with Theorem~\ref{thm:main} at scale $n^\theta$, matching them across the blocks of $\cP$ needs $dD(Q\|\bigotimes_BQ_B)/(q-1)>\theta\log n$. For the minimizing $Q$ this is $dR_\cP>\log n$. The two forms agree by the identity
\[
D(Q\|p)+\frac{D(Q\|v)}{q-1}=\frac{D(Q\|Q_v)}{\alpha}+\frac{D_\alpha(p\|v)}{q-1},\qquad \alpha=\frac{q-1}{q},\quad Q_v\propto p^{\alpha}v^{1-\alpha},
\]
minimized first over $Q$ and then over $v$. At a fixed reference $v$, the minimization over $Q$ is the variational formula for R\'enyi divergence of~\citet[Theorem~30]{vanerven2014renyi}. For finite alphabets, \citet[Theorems~9--10 and Proposition~15]{tomamichel2018operational} express the optimal error exponents for testing against products of $q$ factors through the same minimized R\'enyi divergence at orders above $(q-1)/q$, so the order used here is the lower endpoint of their range. Taking $Q=p$ gives $R_\cP\le D_\cP/(q-1)$, so the condition below is at least as strong as the one in Theorem~\ref{thm:main}.

\begin{theorem}[Rare subpopulations]\label{thm:rare}
Fix $K$, a density $p$ and a partition $\cP$ with $q\ge2$ blocks. If $dR_\cP\le(1-\epsilon)\log n$ for all large $n$, every estimator recovers all relative permutations with probability tending to zero. For Gaussian $p$ and two blocks $S,S^c$, $R_\cP\ge I(X^S;X^{S^c})/2$, so only partitions into at least three blocks add to Theorem~\ref{thm:exact}(b).
\end{theorem}

For two blocks the new condition is thus implied by the swap condition. For two views this holds for every law, Gaussian or not. If $a$ is the vector of squared singular values of the kernel $\sqrt{p(x,y)}$, an $m$-cycle has per-object Chernoff exponent $-\log\|a\|_m$~\citep{cullina2018fundamental}. This exponent is nondecreasing in $m$. For a swap ($m=2$) it is half the order-two cycle mutual information of~\citet{cullina2018fundamental}, which is half the mutual information for Gaussian laws. As $m\to\infty$ it tends to $-\log\max_ia_i$, which equals $R_{1|2}$, the order-$1/2$ dependence measure of~\citet{lapidoth2019two}. With three or more blocks the failing configurations are general permutations among $n^\theta$ objects. Their entropy, not the probability of any fixed pattern, causes the failure. In sparse graph alignment, exact recovery is similarly blocked by vertices that are isolated in an intersection graph~\citep{cullina2017exact,ameen2025aligning}; here the small group consists of objects with atypical coordinates.

The new condition binds already for three views, and there it is the threshold.

\begin{theorem}[Three views with equal negative correlations]\label{thm:three}
Let $K=3$ and $p=N(0,R)$ with unit variances and all correlations equal to $r\in(-1/2,0)$, and let $x\in(0,1/2)$ be the solution of $r=-3x/(2+x+2x^2)$. Then
\[
R_{1|2|3}=-\tfrac14\log(1-2x)-\log\big[(1+x)(1-\tfrac x2)\big]<\frac{I_*}{2}.
\]
For every fixed $\epsilon>0$, maximum likelihood recovers all relative permutations exactly with probability tending to one if $d\ge(1/R_{1|2|3}+\epsilon)\log n$, and every estimator fails with probability tending to one if $d\le(1/R_{1|2|3}-\epsilon)\log n$.

At $r=-1/3$ one has $x=2-\sqrt3$, $I_*=\frac12\log\frac32$, $J=\frac14\log\frac{27}{16}$ and
\[
R_{1|2|3}=\tfrac12\log\tfrac{16}{27}+\tfrac34\log\tfrac{3+2\sqrt3}{4}\approx0.09835 .
\]
The threshold $1/R_{1|2|3}\approx10.17$ exceeds $2/I_*\approx9.87$ and $1/J\approx7.64$: for $9.9\le d/\log n\le10.1$ both $dI_*>(2+\epsilon)\log n$ and $dJ>(1+\epsilon)\log n$ hold for a fixed $\epsilon>0$, yet exact recovery fails.
\end{theorem}

The lower bound is Theorem~\ref{thm:rare}. For the upper bound at $r=-1/3$ the proof uses only explicit decimal inequalities, the tightest with a margin below one percent. For general $r$ the proof is computer-assisted: the parameter range is covered by four pieces, and on each piece finitely many inequalities are certified in exact rational or outward-rounded interval arithmetic, together with machine enumerations of short walks (Appendix~\ref{app:beyond:negative}). We conjecture that for Gaussian laws there is no other obstruction.

\begin{conjecture}[Exact recovery for Gaussian coordinate laws]\label{conj:c1prime}
Let $K$ be fixed and $p=N(0,R)$ with $R\succ0$. Put
\[
L(R)=\min\Big\{\frac{I_*}{2},\ \min_{|\cP|\ge3}R_\cP\Big\}.
\]
With fixed margins, maximum likelihood recovers all relative permutations exactly if $dL>\log n$, and no estimator does if $dL<\log n$.
\end{conjecture}

Impossibility is Theorems~\ref{thm:exact}(b) and~\ref{thm:rare}; what is open is sufficiency. The conjecture agrees with every case settled above. In the weak regime $L=(1+o(1))I_*/2$, in agreement with Theorem~\ref{thm:exact}. For the near-linear constraint of Remark~\ref{rem:general}, the singleton partition gives $R_\cP=c\log n/(K-1)+O_K(1)$, which reproduces the threshold $\max\{2,K-1\}$. For $K=2$ only the swap condition remains. Theorem~\ref{thm:three} is the conjecture for three views with equal negative correlations. The next two statements describe when the new condition can bind and list further cases in which the conjecture is proved.

\begin{proposition}[Structure of the new condition]\label{prop:structure}
Fix $K$.
\begin{enumerate}[label=(\alph*),leftmargin=*]
\item \emph{Bounded patterns.} For every density $p$, every error pattern on a bounded number of objects has Chernoff exponent at least $\tfrac12\min_SI^\circ_2(S)$ per moved object, with equality for a swap across a minimizing cut. Here $I^\circ_2$ is the cycle mutual information of order two of~\citet{cullina2018fundamental}, which equals the mutual information for Gaussian laws.
\item \emph{Gaussian references.} For Gaussian $p$ the minimizer in the first form of Definition~\ref{def:rare} is unique and is a centered Gaussian density.
\item \emph{When the new condition binds.} For $K=3$ with nonnegative correlations, and for every $K$ when the views are a common Gaussian factor with positive loadings plus independent noise, $L=I_*/2$. For $K=4$ there are laws with all pairwise correlations positive for which $L$ is smaller than both $I_*/2$ and $J$. For every Gaussian law with $0<J\le I_*/2$, some partition has $R_\cP<J$.
\end{enumerate}
\end{proposition}

\begin{theorem}[Further cases of the conjecture]\label{thm:cases}
Conjecture~\ref{conj:c1prime} holds in the following cases.
\begin{enumerate}[label=(\roman*),leftmargin=*]
\item $K=3$ with equal correlations $r\in(0,1)$; there $L=I_*/2$.
\item $K=4$ with equal correlations $-1/4$; there $L=R_{1|2|3|4}$ and $9.8511971<1/L<9.8512024$, while $dI_*>2\log n$ and $dJ>\log n$ require only $d/\log n>8.511$.
\item $K=3$ Gaussian chains, $r_{13}=r_{12}r_{23}$ up to relabeling the views; there $L=I_*/2=\min\{I_{12},I_{23}\}/2$.
\item Every $R$ with unit variances, $I_*>0$ and $\|R-I\|_{\mathrm{op}}\le\gamma_K$, for a constant $\gamma_K>0$ that depends only on $K$.
\end{enumerate}
More generally, Appendix~\ref{app:beyond:criterion} gives a condition on $R$, verifiable by finitely many interval-arithmetic checks on matrices of size at most $K$, under which the conjecture holds for $R$. It holds in case~(iv), in case~(ii) and at $r=-1/3$ in Theorem~\ref{thm:three}; it fails for negatively equicorrelated laws close to singularity, where Theorem~\ref{thm:three} is proved by other means. The proofs of~(i), for $9/20\le r\le7/8$, and of~(ii) are computer-assisted.
\end{theorem}

\begin{proposition}[Critical window]\label{prop:window}
For $K=3$ and equal correlations $-1/3$ put $s_n=dR_{1|2|3}-\log n$. Exact recovery succeeds with probability tending to one if $s_n\to+\infty$ and fails with probability tending to one if $s_n+\tfrac34\log\log n\to-\infty$.
\end{proposition}

\paragraph{Two further families, with proofs in outline.} Appendix~\ref{app:beyond:axial} states the conjecture for a band of axially symmetric three-view covariances, $r_{12}=r_{13}=r\in[-11/30,-3/10]$ and $r_{23}=r+\delta$ with every $\delta$ for which $R\succ0$; at $r=-1/3$ each of the three terms of $L=\min\{R_{1|2|3},I_1/2,I_2/2\}$, with $I_i=I(X_i;X_{[3]\setminus\{i\}})$, is the minimum on some interval of $\delta$. Appendix~\ref{app:beyond:analytic} states that, for three views with equal correlations $-9/23\le r<0$ or $-1/2<r\le-1/2+2^{-12}$, Theorem~\ref{thm:three} has a proof without interval arithmetic, and that the conjecture then also holds at every correlation matrix in some neighborhood of $R$. For these two results this version gives the structure of the proofs and the new ingredients, not all details, and we do not count them among the proved cases above.

\paragraph{Proof idea: lower bound.} Let $v$ attain the first form of $R_\cP$ and let $Q$, with density proportional to $p^{(q-1)/q}v^{1/q}$, be the tilted law, which attains the second; optimizers of the two forms chosen separately need not be compatible. Then $D(Q\|p)\le R_\cP$, so $\theta\ge\epsilon$ under the hypothesis of Theorem~\ref{thm:rare}. On an event $A_d$ of $Q^{\otimes d}$-probability $1-o(1)$, mark each object independently with probability $e^{-d(D(Q\|p)+\delta)}\prod_{j\le d}(Q/p)(x_j)\le1$. The $m\ge n^{\theta-O(\delta)}$ marked rows are then i.i.d.\ from $Q^{\otimes d}(\cdot\mid A_d)$. Reveal the marked positions, all unmarked rows and their correspondences, and the correspondences inside each block. What remains are $q-1$ uniform bijections among $m$ objects. If the marked rows were drawn from $v^{\otimes d}$, which factors over the blocks, they would carry no information about these bijections, so every estimator would succeed with probability at most $(m!)^{-(q-1)}$. Changing measure costs at most $e^{d(D(Q\|v)+\delta)+o(1)}$ per marked row, too little to compensate when $dR_\cP\le(1-\epsilon)\log n$ (Appendix~\ref{app:beyond:rare}). Revealing everything outside a data-dependent set also drives the exact community-recovery converses for correlated block models of~\citet{gaudio2022exact} and~\citet{racz2024harnessing}, and \citet[Appendix~E.3]{ding2023planted} restrict a planted-matching converse to the vertices whose planted weights lie in the common support of the two weight laws; here the set comes from a likelihood-ratio marking, so the retained rows are i.i.d.\ from the tilted law. Block-product reference laws, with the same normalization by $|\cP|-1$, also give the multiterminal secret-key converses of~\citet{tyagi2015converses}. For Proposition~\ref{prop:window} the marking window has width $O(1)$ in the log-likelihood ratio, which costs a factor $d^{-1/2}$ in the number of marked objects and produces the term $\tfrac34\log\log n$.

\paragraph{Proof idea: upper bound at $r=-1/3$.} A union bound over wrong alignments does not reach the threshold: its sum is dominated by rare events on which many tuples are tilted and very many alignments beat the truth at once. Let $q_3\propto p^{2/3}(\phi^{\otimes3})^{1/3}$ be the tilted law for the partition into singletons, with $\phi$ its optimal Gaussian reference. The decoder maximizes the likelihood over the alignments that pass a test: no set $S$ of their tuples may have $\sum_{i\in S}\log(q_3^{\otimes d}/p^{\otimes d})(x_i)$ above $\log\binom n{|S|}+\eta|S|\log n$. The true alignment passes with probability tending to one, and the test caps the number of tuples that look like draws from $q_3$ at the size $n^\theta$ of the heuristic above. A wrong alignment differs from the truth on the components of a bipartite graph between objects and tuples. In a component with $m$ objects, $k$ of which are split into three, the probability that it beats the truth while both alignments pass the test is at most
\[
\exp\bigl\{\log\tbinom nk+\eta k\log n-d\,[(m-k)R_2+2kR_3]\bigr\}\;\mathcal I_G^{\,d},
\]
where $R_3=R_{1|2|3}$, $R_2>R_3$ is the rate of a partition into two blocks, and $\mathcal I_G$ is a Gaussian integral over the edges of the component. Combined with the geometric mean of the true and false likelihoods, the test replaces $\sqrt p$ by the tilted law at every object that is split into three, which is what produces the R\'enyi order $2/3$. If $\mathcal I_G\le1$, the bound beats the number of components, about $n^{m}m^{k}$, as soon as $dR_3>\log n$. Cycles raise $\mathcal I_G$. Writing it at a fixed point of Gaussian belief propagation gives $\mathcal I_G\le\det(I-B)^{-1/2}$ for a matrix $B$ of non-backtracking steps with absolute row sums below $0.556$, and we charge its closed walks to the edges that close short cycles. An edge that closes a short cycle when the component is built edge by edge has only $O(1)$ possible endpoints, which saves $\log m$ in the count. We may assume $d<10.2\log n$. Each such edge then costs at most $0.0488\,d<\frac12\log n$ in the integral and saves $\log m\ge\frac12\log n$ when $m\ge\sqrt n$. Smaller components are handled directly by the Bhattacharyya exponent, written in terms of the principal angles between the true and the candidate row spaces.

\paragraph{Proof idea: the other cases.} For other negative correlations the same scheme is used, but the charge of a short cycle and the budget of an edge both depend on $r$, and a single estimate does not cover the family. For moderate $|r|$ the two-step norm bound above is refined by enumerating the short non-backtracking returns in the possible neighborhoods of a marked edge. Near the singular end $r\to-1/2$ these norm bounds fail, and they are replaced by comparisons of effective resistances between the graph and its covering tree, through the weighted Ihara--Bass formula. For positive equal correlations the binding condition is the swap condition; the range $9/20\le r\le7/8$ uses an exponential tilt of two sums of squares of the true rows, after which only the first two moments of the squared sines of the principal angles enter, and for $r\ge7/8$ each view is first aligned to the anchor by two-view maximum likelihood and the joint likelihood is then maximized near that alignment. The criterion of Theorem~\ref{thm:cases} packages the scheme for a general covariance: local tilted factors for every partition, an invariant box for the belief-propagation messages, and a budget inequality between the cost of a marked edge and the counting gain.

\section{Decoders and computation}\label{sec:algorithms}

We first compare three natural ways of combining the views, which settles the question of~\citet{wee2026geometric}, and then give two decoders that evaluate far fewer than the $n^K$ candidate tuples of Theorem~\ref{thm:main}. The last part of the section shows where some natural fast decoders stop. Table~\ref{tab:intro} collects the levels for equal correlations.

\begin{theorem}[Hierarchy]\label{thm:ladder}
Consider the model of~\citet{wee2026geometric}, with $d/\log n\to\infty$ and $A/\log n\to b$.
\begin{enumerate}[label=(\roman*),leftmargin=*]
\item \emph{Tree decoders.} Fix a spanning tree $\cT$ on $[K]$, estimate each $\sigma_{ac}$, $\{a,c\}\in\cT$, from $Y^a$ and $Y^c$ only, and compose along $\cT$. For $b>2$, a suitable two-view estimator makes this simultaneously almost exact. For $b<2$ every such decoder has vanishing overlap on every edge of $\cT$.
\item \emph{Row-wise pairwise sum}~\citep[Algorithm~1]{wee2026geometric}. For each anchor row $i$, the decoder proposes a maximizer of $\sum_{1\le a<c\le K}\langle Y^a_{j_a},Y^c_{j_c}\rangle$ over $(j_2,\ldots,j_K)\in[n]^{K-1}$, with $j_1=i$; conflicts are then resolved and the result is completed to permutations. If $b>K/(K-1)$ it is simultaneously almost exact~\citep[Theorem~2.5]{wee2026geometric}. If $b<K/(K-1)$, then for each fixed anchor row $i$ the maximizer satisfies $j_a\ne\pi_a(i)$ for all $a\ge2$ with probability tending to one. Consider any completion that, in each view $a\ge2$, (C1)~keeps exactly one of the anchor rows proposing each proposed row and (C2)~matches the remaining anchor rows bijectively to the unproposed rows, where both choices are functions of the table of raw proposals and of independent randomness. Then every anchor permutation $\widehat\pi_a$ has vanishing expected overlap, and so does every composed estimate $\widehat\pi_c\circ\widehat\pi_a^{-1}$ between two non-anchor views (Appendix~\ref{app:ladder}).
\item \emph{Partition tests.} For $b>4/K$ the decoder of Theorem~\ref{thm:main} is simultaneously almost exact; for $b<4/K$ no decoder is (Corollary~\ref{cor:equi}).
\end{enumerate}
\end{theorem}

\begin{remark}[The posterior depends only on pairwise scores]\label{rem:posterior}
If $\Omega=\Sigma^{-1}$, then under the uniform prior $\log p(\Pi\mid Y)=C(Y)-\sum_{a<c}\sum_i(Y^a_{\pi_a(i)})^\top\Omega_{ac}Y^c_{\pi_c(i)}$, because the quadratic terms of each view are invariant under its row permutation. In the model of~\citet{wee2026geometric}, $\Omega_{ac}$ is a negative multiple of the identity for $a\ne c$, so the joint maximum a posteriori estimate maximizes the same sum as Algorithm~1, but over all permutations jointly. The gap between $K/(K-1)$ and $4/K$ comes from maximizing one anchor row at a time, which lets several anchor rows claim the same internally consistent object. The conditions on completion in~(ii) cannot be dropped in general: a completion that rereads the data can recover more (Remark~\ref{rem:completion}).
\end{remark}

\begin{table}[htb]
\centering
\caption{Levels in units of $b$ for a common signal with equal correlations in the weak limit. Each decoder succeeds above its level. Below $4/(K-1)$ no estimator recovers every row, and below $4/K$ none is almost exact; the last two decoders fail below their levels in the sense of Theorem~\ref{thm:ladder}. Layered screening needs $K\ge3$; operation counts are on whitened coordinates.}\label{tab:intro}
\vspace{2pt}
\small
\begin{tabular}{lllcc}
\toprule
decoder & recovery & operations & level & $K=5$\\
\midrule
maximum likelihood (Thm.~\ref{thm:exact}) & exact & --- & $4/(K-1)$ & $1$\\
partition tests (Thm.~\ref{thm:main}) & almost exact & $O_K(n^K+n^2d)$ & $4/K$ & $0.8$\\
layered screening (Thm.~\ref{thm:screen}) & almost exact & $O(Kn^2d)$ & $4/\min\{K,4\}$ & $1$\\
row-wise pairwise sum (Thm.~\ref{thm:ladder}) & almost exact & $O_K(n^K+n^2d)$ & $K/(K-1)$ & $1.25$\\
tree decoder (Thm.~\ref{thm:ladder}) & almost exact & $O(Kn^2d)$ & $2$ & $2$\\
\bottomrule
\end{tabular}
\end{table}

At $K=5$ and $b=1.1$, for example, maximum likelihood recovers every row and layered screening recovers all but $o(n)$ rows in quadratic time, while the row-wise sum and tree decoders fail (Table~\ref{tab:intro}).

\paragraph{Layered screening.} Partition tests need not look at all $n^K$ tuples. Let $c(\cP)$ be the number of pairs of views separated by $\cP$. In the model of Corollary~\ref{cor:equi}, the log-likelihood ratio of a true tuple for a partition $\cP$ of its first $s$ views has mean $(b/2+o(1))\,c(\cP)\log n$, while a false tuple whose identity partition is $\cP$ exceeds $\alpha\,c(\cP)\log n$ with probability at most $n^{-\alpha c(\cP)}$. Tuples can therefore be grown one view at a time from short lists, and the false prefixes that survive stay few.

\begin{theorem}[Layered screening]\label{thm:screen}
In the model of Corollary~\ref{cor:equi}, suppose that $\rho_{\max}\to0$ and that $A/\log n\to b<\infty$, and fix $3\le m\le K$. If $b>4/m$, the decoder below recovers all relative permutations simultaneously almost exactly with $O_m(Kn^2d+n^{\gamma_m})$ real operations on the whitened coordinates, where
\[
\gamma_m=\max\Big\{2,\ 1-\frac2m+\frac{\lfloor(m+1)^2/4\rfloor}{m}\Big\}.
\]
With $m=\min\{K,4\}$ the cost is $O(Kn^2d)$ and the guarantee is $b>4/3$ for $K=3$ and $b>1$ for $K\ge4$; for $K=3,4$ this is the threshold $4/K$.
\end{theorem}

When the coordinates are not already independent, whitening is a separate precomputation of $O(d^3+Knd^2)$ operations; in the model of~\citet{wee2026geometric} it is not needed. Fix $\alpha\in(2/m,b/2)$ and $\beta\in(1,mb/2)$. The decoder (i) lists, for every anchor row and every view $a\in\{2,\ldots,m\}$, the rows of view $a$ whose two-view log-likelihood ratio with the anchor row exceeds $\alpha\log n$, and discards anchor rows with a list longer than $\lceil n^{1-2/m}\rceil$; (ii) for $s=3,\ldots,m$, extends each surviving prefix by the anchor's list for view $s$ and keeps the extension if it passes every partition test on its $s$ views at level $\alpha c(\cP)\log n$, aborting if more than $2n^{u_s}$ prefixes survive an intermediate level, where $u_s=\max\{1,s-s(s-1)/m\}$; (iii) keeps a maximal set of disjoint $m$-tuples and completes; and (iv) matches each remaining view to the normalized sum of the $m$ aligned rows, whose mutual information with that view is $(m/2+o(1))A$, by a two-view test at level $\beta\log n$ and a maximal matching (Appendix~\ref{app:screen}).

\paragraph{Proof idea for Theorem~\ref{thm:screen}.} In the model of Corollary~\ref{cor:equi}, a false prefix of $s$ views whose identity partition $\cP$ has $p$ blocks passes the test for its own partition with probability at most $n^{-\alpha c(\cP)}$, and there are at most $n^p$ such prefixes. Since $c(\cP)\ge(p-1)(s-p/2)$ and ${p-\frac2m(p-1)(s-p/2)}$ is convex in $p$, it is at most the larger of its values at $p=1$ and $p=s$, which is $u_s$; as $\alpha>2/m$, each partition loses a further factor $n^{-(\alpha-2/m)c(\cP)}$, so the expected number of false prefixes that survive level $s$ is $o(n^{u_s})$. True prefixes pass, because each test has mean $(b/2+o(1))c(\cP)\log n$ and variance $O(\log n)$. Since $u_m=1$, the last level leaves $o(n)$ false $m$-tuples, and each of them blocks at most $m$ true tuples in a maximal disjoint selection. The lists have expected length at most $1+n^{1-\alpha}$, so discarding the anchor rows whose lists are longer than $n^{1-2/m}$ loses $o(n)$ of them, and extending at most $2n^{u_s}$ prefixes by such lists gives the exponent $\gamma_m$. Each remaining view is attached by a two-view test against the normalized sum of the $m$ aligned rows, which carries all their information about it.

The guarantee is asymptotic. For $n\le8192$ the gap $(b/2-\alpha)\log n$ is comparable to the $O(\sqrt{\log n})$ fluctuations of the scores, and in our simulations layered screening recovers fewer whole tuples than message passing or Sinkhorn scaling, which have no guarantees (Appendix~\ref{app:numerics}).

\paragraph{Decoding along the weakest partition.} For $U\subseteq[K]$, a weakest partition of $U$ is a partition attaining $J_U$. Choosing a weakest partition of $[K]$, then of each of its blocks, and so on down to single views gives a tree; let $p^*$ be the largest number of blocks at a node.

\begin{theorem}[Recursive weakest-partition decoding]\label{thm:recursive}
Let $\Sigma\succ0$ and $J\ge(1+\epsilon)\log n$ for a fixed $\epsilon>0$. Decode the tree from the leaves up. At each node, concatenate the aligned rows of each child into a super-view, accept the tuples of super-view rows that pass, for every partition $\mathcal R$ of the children, the test at level $(1+\gamma)(|\mathcal R|-1)\log n$, for a fixed $\gamma\in(0,\epsilon)$, keep a maximal set of disjoint accepted tuples and complete. This recovers all relative permutations simultaneously almost exactly, with $O_K(n^{p^*})$ operations after $O_K(d^3+nd^2+n^2d)$ precomputation, $d=\sum_ad_a$. If every weakest partition in the tree has two blocks, $K-1$ two-view matchings of super-views suffice, and if moreover each of these bipartitions is the unique weakest partition of its node, the tree is found by greedily merging the two current clusters whose concatenations have the largest mutual information.
\end{theorem}

Every node inherits the margin, because $J_B\ge J_U$ for each block $B$ of a weakest partition of $U$ and the super-views of these blocks have shared information exactly $J_U$. Errors propagate additively: a node with $p$ children has at most $F+pT+(p+1)\sum_jE_{B_j}$ wrong rows, where $F$ counts true tuples that fail a test, $T$ false tuples that pass, and $E_{B_j}$ the wrong rows of the children (Appendix~\ref{app:recursive}).

\begin{figure}[t]
\centering
\includegraphics[width=0.6\linewidth]{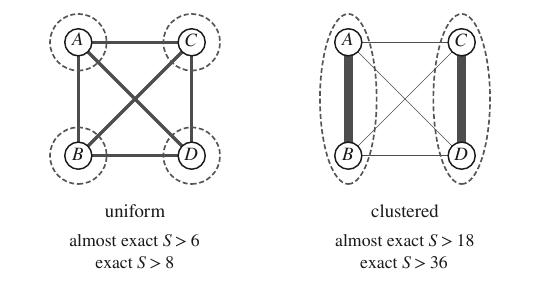}
\caption{The two four-view designs of Example~\ref{ex:cluster}, with the same total signal $S$. Edge widths are proportional to the pairwise mutual information, and dashed loops mark the partition that attains $J$.}\label{fig:designs}
\end{figure}

\begin{example}[Clustered views]\label{ex:cluster}
Take $K=4$ views $A,B,C,D$ (Figure~\ref{fig:designs}) in which every coordinate of an object's four observations is an independent $N(0,R)$ vector with unit variances, in the weak regime $d/\log n\to\infty$, with total signal $S=(d/\log n)\sum_{a<b}\rho_{ab}^2$. In the \emph{uniform} design all six correlations are equal and positive. In the \emph{clustered} design $\rho_{AB}=\rho_{CD}=s>0$ and the four cross correlations are $s/4$. The thresholds in $S$ are:
\begin{center}
\begin{tabular}{lcccc}
\toprule
design & tree decoder & row-wise sum & almost exact ($J$) & exact ($I_*$)\\
\midrule
uniform & $12$ & $8$ & $6$ & $8$\\
clustered & $72$ & $22.5$ & $18$ & $36$\\
\bottomrule
\end{tabular}
\end{center}
In the clustered design $\{AB,CD\}$ attains both $J$ and $I_*$. The pairs $AB$ and $CD$ are matchable once $S>4.5$, while for $S<18$ every cross pair has vanishing overlap (Theorem~\ref{thm:pairwise}). Since the cross term of the block log-likelihood ratio is a multiple of $\langle A_i+B_{\alpha(i)},C_j+D_{\delta(j)}\rangle$, the threshold $18$ is also attained by three two-view assignments, $A$ to $B$, $C$ to $D$, and the summed rows $A+B$ to $C+D$, in time $O(n^3+n^2d)$ (Appendix~\ref{app:recursive}).
\end{example}

\begin{remark}[Summing rows can lose the threshold]\label{rem:rowsum}
In the coordinate model of Example~\ref{ex:cluster}, let $\rho_{AB}=\rho_{CD}=2r$ and $\rho_{AC}=r$ with $r\to0$, let the other three correlations vanish, and let $dr^2/\log n\to3$. The weakest partition is $\{AB,CD\}$ with $J/\log n\to3/2$, but the summed rows $A+B$ and $C+D$ have mutual information $(3/8+o(1))\log n$. Every estimator that sees only the summed rows of the two correctly aligned clusters then has vanishing overlap between them, whereas Theorem~\ref{thm:recursive} with the block likelihood ratio succeeds.
\end{remark}

\begin{remark}[Further computational results]\label{rem:compute}
(a) \emph{Heterogeneous laws.} Layered screening extends to weak laws with unequal pairwise informations $w_{ab}\log n$. For a three-view seed with $(w_{12},w_{13},w_{23})=(0.78,0.75,0.72)$ and a fourth view with weights $(0.58,0.48,0.02)$ to the seed, every pair is below the two-view threshold, so no tree decoder succeeds, yet the seed has shared information $1.125\log n$, the fourth view carries $1.08\log n$ about it, and screening makes $O(n^2)$ candidate checks after an $O(n^2d)$ score precomputation.
(b) \emph{Exact recovery in polynomial time.} When $I_*\ge(2+\epsilon)\log n$ in the weak regime, partition tests at level $(|\cP|+\eta)\log n$ accept only correct tuples and leave $O(n^{1-c})$ objects unmatched. The linear-programming relaxation of the multi-index assignment, with $n^K$ variables, is exact when $I_*\ge(3+\epsilon)\log n$.
(c) \emph{Low-degree polynomials.} With equal correlations, $(\log n)^{41+40\eta}\ll K=n^{o(1)}$ and $d\gg K\log n$, polynomials of degree $(\log n)^{1+\eta}$ estimate whether two rows hold the same object only at chance level, in normalized mean squared error, when $b=(4+\epsilon)/K$, although recovery is information-theoretically possible; this is Theorem~2.2 of~\citet{even2025computational} in our parametrization. For fixed $b>0$ and large $K$, degree $O_b(\log n)$ makes this error $o(1/n)$, so a barrier of this kind requires $b_n\to0$.
Proofs are in Appendix~\ref{app:computationalremarks}.
\end{remark}

\paragraph{Where natural fast decoders stop.} For fixed $K\ge5$ and $4/K<b<1$, partition tests achieve almost exact recovery in polynomial time, but whether $O(n^2\,\mathrm{poly}(d,K))$ operations suffice is not known. We prove no general lower bound for this question. Parts~(a) and~(c) below show that two natural fast approaches stop short of the window, and part~(b) shows that the dual certificate behind Remark~\ref{rem:compute}(b) cannot reach the exact threshold.

\begin{proposition}[Barriers for natural decoders]\label{prop:barriers}
\begin{enumerate}[label=(\alph*),leftmargin=*]
\item \emph{Explicit enumeration of pairwise candidates.} In the model of Corollary~\ref{cor:equi} with all $\rho_\ell$ equal, $d/\log n\to\infty$ and $A/\log n\to b\in[0,1)$, let every anchor row $i$ have candidate lists $L_{i,2}$ and $L_{i,3}$ in two other views, where each list is a function of the anchor, its own view and independent randomness and may have any data-dependent length. Suppose a decoder visits every element of $L_{i,2}\times L_{i,3}$ for each anchor row $i$ that it scans. If it makes at most $n^{3-b-\delta}$ visits with probability tending to one, for a fixed $\delta>0$, then it scans only $o_\Prob(n)$ true triples. Lists of length $\lceil n^{1-b/2+\kappa}\rceil$ by raw inner product cover $n-o_\Prob(n)$ true triples with $O(n^{3-b+2\kappa})$ visits, for every fixed $\kappa>0$.
\item \emph{Half-edge dual prices.} For a Gaussian law with $\|R-I\|_{\mathrm{op}}\to0$ and $I_*/\log n\to c\in(2,3)$, the dual prices that certify the linear-programming relaxation of Remark~\ref{rem:compute}(b) when $I_*\ge(3+\epsilon)\log n$, half of the true score of each pair of views at each row, violate some dual constraint with probability tending to one.
\item \emph{Sinkhorn soft averaging} (proof in outline). In the model of~(a) with $K=5$, consider in real arithmetic the soft-averaging decoder of Appendix~\ref{app:numerics}: pairwise tables scaled by a fixed number of Sinkhorn steps~\citep{sinkhorn1964relationship,sinkhorn1967concerning}, Gauss--Seidel updates of each view's table from the field $S_{a1}+\sum_{c\ne a}S_{ac}Q_c$, where $S_{ac}$ is the matrix of normalized inner products between views $a$ and $c$, and rounding by the top $W$ candidates of each anchor row with diagonal completion. In the first and the second pass, the table of the first updated view gives the true row a score $(\sqrt b+o_\Prob(1))\sqrt{\log n}$ and a weight tending to zero, and in the second pass every anchor row has a wrong score above $t\sqrt{\log n}$ for each fixed $t<\sqrt2$. After any positive diagonal scaling, the top $W\le Cn^\omega$ candidates, $\omega<1-b/2$, contain the true row for only $o_\Prob(n)$ anchor rows, so rounding this view after either pass has vanishing overlap. If the first update instead uses held-out coordinates, no estimator that reads the whole held-out field, together with the training data of the other views, has nonvanishing expected overlap.
\end{enumerate}
\end{proposition}

Part~(a) is a change of measure on short lists: replacing the true pairs of a set of $\eta n$ anchor rows by independent rows makes the remaining correspondence uniform given the lists, so short lists rarely contain the truth, and long lists cost $w^2$ visits per anchor row. Part~(b) is a second-moment count of two-object false tuples across a minimizing cut, whose violation has probability $n^{-(2/3+o(1))c}$; it concerns this certificate only, and whether the relaxation itself is exact for $2\log n<I_*\le3\log n$ is open. Part~(c) follows from a comparison of the first field with a two-view experiment whose correlation is too weak, and from rectangle bounds on the wrong scores that hold for every nonnegative column-stochastic table (Appendix~\ref{app:barriers}). Sinkhorn soft averaging is among the decoders that do well at finite size in our simulations (Appendix~\ref{app:numerics}); its default third pass is not covered, and recovery in quadratic time for $b<1$ remains open. Parts~(a) and~(b) are proved in full in Appendix~\ref{app:barriers}; for part~(c) the appendix states the three lemmas on which the proof rests and gives their proofs in outline.

\section{Other observation models}\label{sec:other}

The results so far assume Gaussian features with a known covariance, observed in every view. This section relaxes these assumptions one at a time. The almost-exact and pairwise thresholds keep their form for known laws that are not Gaussian and for objects that are missing from some views; unknown marginal covariances are harmless, but an unknown cross-covariance is not; and for exact recovery the two necessary conditions of Sections~\ref{sec:exact} and~\ref{sec:beyond} hold for every law with a density.

\subsection{Known observation laws}\label{sec:other-known}

Gaussianity enters Theorems~\ref{thm:main} and~\ref{thm:pairwise} only through the concentration of the information density of true tuples. Let the feature of each view take values in an arbitrary measurable space and let the tuple law $P_n$ be known. For a set $B$ of views and a partition $\cC$ of $B$, let $D_{B,\cC}$ be the divergence between the marginal $P_{n,B}$ and the product of its block marginals, define $J_B$ and $J^*_{ab}$ from these divergences as before, and let $L_{B,\cC}$ be the logarithm of the ratio of the densities of the two laws with respect to their sum. Infinite divergences and infinite values of $L_{B,\cC}$ are allowed.

\begin{theorem}[Known observation laws]\label{thm:known}
The bound~\eqref{eq:converse} of Theorem~\ref{thm:main}(b) holds for every known law $P_n$. Suppose moreover that, for every set $B$ of at least two views, every partition $\cC$ of $B$ into at least two blocks and every fixed $\eta>0$,
\begin{equation}\label{eq:lowertail}
P_{n,B}\big(L_{B,\cC}<\min\{D_{B,\cC},K\log n\}-\eta\log n\big)=o(1).
\end{equation}
Then Theorem~\ref{thm:main}(a) and Theorem~\ref{thm:pairwise} hold for $P_n$, with the same decoders.
\end{theorem}

\begin{corollary}[Fixed coordinate law]\label{cor:product}
Let each tuple have the known law $p^{\otimes d}$, where $p$ is a fixed law on a product of $K$ measurable spaces and $d/\log n\to c\in(0,\infty)$, and define $J(p)$ and $J^*_{ab}(p)$ with $p$ in place of $P_n$. Simultaneously almost exact recovery is possible if $cJ(p)>1$ and impossible if $cJ(p)<1$. For views $a\ne b$, if $cJ^*_{ab}(p)>1$ there is a permutation $\widehat\sigma_{ab}$ with $\ov(\widehat\sigma_{ab})\to1$ in probability, and if $cJ^*_{ab}(p)<1$ then $\sup_f\E\,\ov(f)\to0$ as in Theorem~\ref{thm:pairwise}(b).
\end{corollary}

The cap $K\log n$ in~\eqref{eq:lowertail} lies above every level $(1+\gamma)(|\cC|-1)\log n$ at which the decoder tests, so the condition only asks that true scores rarely fall far below their mean up to that level; values above the cap, including infinite ones, do not matter. At a finite divergence it follows from $\Var L_{B,\cC}=o(D_{B,\cC}^2)$ by Chebyshev's inequality, and for Gaussian laws from the variance bound of Lemma~\ref{lem:llr}. For a fixed $p$ the information density is a sum of $d$ independent terms, so~\eqref{eq:lowertail} follows from the weak law of large numbers, applied to truncated terms when a divergence is infinite; no moment of the information density is needed. The same argument uses only $d\to\infty$, so if $d/\log n\to\infty$, simultaneously almost exact recovery is possible whenever $J(p)>0$, and views $a,b$ are matchable whenever $J^*_{ab}(p)>0$; if $d/\log n\to0$ and $J(p)<\infty$, almost exact recovery is impossible by the first part of Theorem~\ref{thm:known}. Exact recovery behaves differently: for laws that are not Gaussian a partition into three blocks can bind even under weak dependence (Example~\ref{ex:f2-parity}), so there is no analogue of Corollary~\ref{cor:product} in terms of the weakest bipartition; Section~\ref{sec:other-nongauss} gives the converse for general laws and the threshold under weak dependence. When the law changes with $n$, some condition of the kind~\eqref{eq:lowertail} is necessary for almost exact recovery. If every view copies the same $d$ independent Bernoulli$(1/n)$ bits, then $J$ exceeds $\log n$ as soon as $d\ge\kappa n$ for a fixed $\kappa>1$, but the all-zero rows, a fraction close to $e^{-d/n}$ of each view, cannot be matched, and in this family almost exact recovery holds exactly when $d/n\to\infty$. The proofs and the analysis of this family are in Appendix~\ref{app:f1-known}.

\subsection{Partially observed objects}\label{sec:other-partial}

Suppose that object $i$ is observed only in a random set $M_i$ of views, drawn from a fixed law $\mu$ independently of the features, that each view shuffles its visible rows uniformly, and that the sets $M_i$ are hidden; rows with $|M_i|=1$ are outliers that belong to no tuple. An estimator outputs groups of rows with at most one row per view. The loss is the number of pairs of rows that are placed in a common group by exactly one of the estimate and the truth, so an outlier costs one for every row it is grouped with. When every object is seen in all views, the loss of a group estimate bounds the number of misaligned rows in each view of the permutations it induces, and permutations with $M$ misaligned rows in all induce groups with loss at most $2(K-1)M$, so the two criteria agree. For $M\subseteq[K]$ let $J_M$ be the shared information of the marginal law on the views in $M$.

\begin{proposition}[Partial overlap]\label{prop:overlap}
In the Gaussian model with a known law, with fixed margins, the loss is $o_{\Prob}(n)$ if $\min\{J_M:\mu(M)>0,\ |M|\ge2\}>\log n$ and is not $o_{\Prob}(n)$ if this minimum is below $\log n$. For a type $M$ with $\mu(M)>0$ and views $a,b\in M$, let $J^*_{ab}(M)=\max\{J_B:\{a,b\}\subseteq B\subseteq M\}$. If $J^*_{ab}(M)>\log n$, the same decoder recovers the pair $a,b$ for all but $o_{\Prob}(n)$ objects of type $M$; if $J^*_{ab}(M)<\log n$, every estimator recovers it for a vanishing expected fraction of these objects. The threshold is not a $\mu$-average of partition divergences: for three views, each present independently with probability $1/2$, and equal correlations $1/2$, there are dimensions at which the natural average exceeds $\log n$, objects seen in all three views are recovered almost exactly, and every pair of views has vanishing overlap on the objects seen only in that pair.
\end{proposition}

The decoder runs the partition tests on every set of at least two views, which for fixed $K$ means $O_K(n^K)$ candidate tuples, joins any two rows that lie in a common accepted tuple, outputs each connected group that contains at most one row per view, and outputs the rows of every other group as single rows. Each falsely accepted tuple affects at most $K$ objects (Appendix~\ref{app:overlap}).

\subsection{Unknown covariance}\label{sec:other-unknown}

All decoders so far use the known covariance. Marginal covariances can be estimated from a few rows, but an unknown cross-covariance cannot in general be learned from shuffled data at the information threshold.

\begin{proposition}[Unknown covariance]\label{prop:unknown}
\begin{enumerate}[label=(\alph*),leftmargin=*]
\item \emph{Unknown marginals.} Let the marginal covariances $S_a$ of the views be unknown, and let the joint correlation matrix $R\succ0$ obtained by whitening each view with $S_a^{-1/2}$ be known. Let $d=\sum_ad_a$, $c_n=\lambda_{\min}(R)$ and $u_n=(d+\log n)\log^2(2d)$. If $u_n=o(nc_n)$ and $\liminf_nJ/\log n>1$, simultaneously almost exact recovery is possible: whiten each view with the sample covariance of $o(n)$ reserved rows and run the partition tests. If the marginal condition numbers are bounded, $u_n=d+\log n$ suffices.
\item \emph{Unknown noise level.} In the model of~\citet{wee2026geometric} the thresholds $4/K$ and $4/(K-1)$ of Corollary~\ref{cor:equi} hold when $\tau$ is unknown.
\item \emph{Unknown orthogonal cross-covariance.} Let $K=2$, let both views have identity covariance and cross-covariance $\rho O$ with $O\in O(d_0)$ unknown, and fix $\epsilon\in(0,2)$. If $n\rho^4\le(2-\epsilon)\log n$, then for every estimator, averaged over Haar-distributed $O$, the probability of exact recovery is at most $\exp[-(\epsilon/4+o(1))n\log n]$. Under the same average, if $n\rho^4=o(\log n)$, the probability that the overlap exceeds any fixed $\alpha>0$ tends to zero. In particular, if $d_0\gg\sqrt{n\log n}$ and $\rho^2=2b\log n/d_0$ with $b>1$ fixed, then $J\sim b\log n$, yet every estimator has vanishing expected overlap along some sequence of matrices $O$.
\end{enumerate}
\end{proposition}

Part~(c) gives the necessary order of the second term in the sufficient condition
\[
\rho^2\ge C\max\Big\{\frac{\log n}{d_0},\sqrt{\frac{\log n}{n}}\Big\}
\]
of~\citet[Theorem~2.3]{niu2026procrustes}, in the fully observed Gaussian model. \citet[Section~1.1]{huang2026rgg} report a threshold of the same order for a thresholded random geometric graph model, which observes less. The proof of~(c) bounds the second moment of the likelihood ratio by a Haar integral and controls it with the first fundamental theorem of invariant theory for the orthogonal group (Appendix~\ref{app:unknown}).

\subsection{Exact recovery for laws that are not Gaussian}\label{sec:other-nongauss}

The two obstructions to exact recovery, swaps across a cut and rare subpopulations, do not use Gaussianity. Let each object's tuple consist of $d=d_n$ i.i.d.\ coordinates with a known law $P=P_n$ that has a density $p$ with respect to a product of $\sigma$-finite measures. For a cut $S|S^c$ let $Z_S$ be the operator with kernel $\sqrt{p(x_S,x_{S^c})}$, let $I^\circ_2(S)=-\log\tr[(Z_S^*Z_S)^2]$, and put $\chi_*=\min_SI^\circ_2(S)$, which equals $I_*$ for Gaussian laws. With $R_\cP$ as in Definition~\ref{def:rare}, let
\[
L(P)=\min\Big\{\frac{\chi_*}{2},\ \min_{|\cP|\ge3}R_\cP(P)\Big\}.
\]

\begin{theorem}[General converse]\label{thm:f2-converse}
Let $P_n=P$ be fixed, or let $d_n\to\infty$. If $d_nL(P_n)\le(1-\epsilon)\log n$ for a fixed $\epsilon\in(0,1)$ and all large $n$, every estimator recovers all relative permutations with probability tending to zero. The swap part holds for a cut with cycle information $\chi$, possibly varying with $n$, whenever $d_n\chi\le2(1-\epsilon)\log n$ and $d_n(\chi+\chi^2)=o((\log n)^2)$, even if all correspondences inside the two sides of the cut are revealed.
\end{theorem}

\begin{theorem}[Weak dependence]\label{thm:f2-weak}
Let $U=\bigotimes_aP_a$ be the product of the view marginals, suppose that $h_n=\|dP_n/dU-1\|_\infty\to0$, and put $B(P_n)=\min_{|\cP|\ge2}D_\cP/|\cP|$. With fixed margins, the maximum-likelihood estimate recovers all relative permutations with probability tending to one if $d_nB(P_n)>\log n$, and every estimator fails if $d_nB(P_n)<\log n$. Moreover $L(P_n)=(1+O_K(h_n))B(P_n)$.
\end{theorem}

The uniform bound on the density ratio excludes Gaussian laws, so Theorem~\ref{thm:f2-weak} complements Theorem~\ref{thm:exact} rather than extending it. In the weak Gaussian regime the minimum of $D_\cP/|\cP|$ is attained at a bipartition to first order, since $D_\cP\ge(1+o(1))|\cP|I_*/2$; for other laws it need not be.

\begin{example}[Three-view parity]\label{ex:f2-parity}
Let $X_a\in\{-1,+1\}$ and $p_h(x_1,x_2,x_3)=2^{-3}(1+hx_1x_2x_3)$ with $h=h_n\to0$. Every proper marginal is uniform, so every partition into at least two blocks has the same divergence $D(h)\sim h^2/2$. Hence $B=D/3$, while every bipartition carries $D$ and $\chi_*\sim D$. If $d_nD(h_n)/\log n\to c\in(2,3)$, both $d_nI_*$ and $d_n\chi_*$ exceed $2\log n$ with a fixed margin, yet exact recovery fails; for $c>3$ maximum likelihood succeeds.
\end{example}

The converse combines the swap argument of Theorem~\ref{thm:exact}(b), which here counts likelihood ties as well as strict improvements, with the rare-subpopulation argument of Theorem~\ref{thm:rare}. The weak-dependence threshold follows from a quadratic approximation of all partition divergences, joint concentration of the likelihood scores and the two-level count of Theorem~\ref{thm:exact}(a) (Appendices~\ref{app:f2-converse} and~\ref{app:f2-weak}). Whether $d_nL(P_n)>(1+\epsilon)\log n$ suffices for general laws is open.

\section{Discussion}\label{sec:discussion}

\paragraph{Thresholds.} All thresholds are stated with fixed margins. The critical windows around $J=\log n$ and $I_*=2\log n$ are open, and Proposition~\ref{prop:window} leaves a gap of order $\log\log n$ in the one window we study. Beyond weak correlations, exact recovery for Gaussian laws is characterized only in the cases of Theorems~\ref{thm:three} and~\ref{thm:cases}; whether the swap and rare-subpopulation conditions together suffice for every Gaussian law (Conjecture~\ref{conj:c1prime}) is open. What our proofs lack for a general covariance is a uniform bound on the Gaussian integral of components with many short cycles. The criterion of Theorem~\ref{thm:cases} reduces it to finitely many inequalities for each covariance, but these fail near singular covariances, where in the equicorrelated case the conjecture is proved by a different comparison. The margins in these proofs are small, below one percent at correlation $-1/3$, and several of the proofs are computer-assisted.

\paragraph{Computation.} For fixed $K$, almost exact recovery has no statistical-computational gap in the usual sense, since partition tests run in polynomial time; the question is the exponent. With equal correlations, layered screening lowers it to $2$ for $K=3,4$, but for larger $K$ it attains $4/K$ only with an exponent that grows linearly in $K$. Can $J$ be attained with $O(n^2\,\mathrm{poly}(d,K))$ operations for every covariance? Below $b=1$, Proposition~\ref{prop:barriers} shows that scanning products of pairwise candidate lists needs $n^{3-b-o(1)}$ tests and, for $K=5$, that the first two passes of Sinkhorn soft averaging fail; these are statements about specific decoders, not lower bounds for all quadratic-time algorithms. When $K$ grows, low-degree polynomials fail at $b=(4+\epsilon)/K$ but succeed at every fixed $b>0$ (Remark~\ref{rem:compute}), so the computational threshold for $b_n\to0$ is open, as is exact recovery in polynomial time for $K\ge3$ when $2\log n<I_*\le3\log n$ in the weak regime, where the half-edge dual certificate of the linear program fails (Proposition~\ref{prop:barriers}(b)).

\paragraph{Other models.} Unknown marginal covariances can be estimated from a few rows, but an unknown cross-covariance cannot in general (Proposition~\ref{prop:unknown}); the critical constant for an unknown orthogonal cross-covariance is open. For known laws the almost-exact and pairwise thresholds hold under the lower-tail condition~\eqref{eq:lowertail}, which a fixed coordinate law satisfies (Theorem~\ref{thm:known}). Its necessity for changing laws, and achievability of exact recovery at the rate $L$ of Theorem~\ref{thm:f2-converse} for laws that are neither Gaussian nor weakly dependent, are open.

\paragraph{Simulations.} Appendix~\ref{app:numerics} reports simulations with $n\le8192$. Of the $33$ predictions that were recorded before the runs, $32$ held; the exception, a finite-size prediction for the row-wise sum, is explained there. The guarantees of this paper are asymptotic: at these sizes layered screening recovers fewer whole tuples than message passing or Sinkhorn scaling, for which we have no guarantees.

\section*{Use of AI assistants}
This work used AI assistants (Claude, Anthropic; Codex, OpenAI) for literature search, drafting and independently checking proofs, numerical checks and writing. The authors directed the work, checked every statement and proof, and take full responsibility for the content.

\bibliography{refs}

\appendix
\setlength{\emergencystretch}{3em}
\section{Proof of Theorem~\ref{thm:main}}\label{app:main}

\subsection{The log-likelihood ratio of a partition}

Fix a partition $\cP$ with $m$ blocks and write $Q_\cP=\bigotimes_{B\in\cP}P_B$. Restrict each block to the linear support of $P_B$ and whiten it. Then $Q_\cP$ has identity covariance and $P$ has a covariance $C$ whose diagonal blocks are identities. If the support of $P$ is a proper subspace of the support of $Q_\cP$, then $D_\cP=\infty$ and $Q_\cP$ gives the support of $P$ probability zero; the decoder then uses the test ``the tuple lies in the support of $P$'', which true tuples pass almost surely and tuples with identity partition $\cP$ fail almost surely. Otherwise $C\succ0$; let $\xi_1,\ldots,\xi_N>0$ be its eigenvalues. Since $\tr C$ equals the dimension, $\sum_j(\xi_j-1)=0$.

\begin{lemma}\label{lem:llr}
Let $L_\cP=\log(dP/dQ_\cP)$ and $D_\cP<\infty$. Then $\E_{Q_\cP}e^{L_\cP}=1$, $\E_PL_\cP=D_\cP$ and $\Var_PL_\cP\le2m^2D_\cP\le 2K^2D_\cP$.
\end{lemma}

\begin{proof}
The first identity holds because $P\ll Q_\cP$. In the whitened coordinates
\[L_\cP(x)=-\tfrac12\log\det C-\tfrac12x^\top(C^{-1}-I)x.\]
Under $P$ write $x=C^{1/2}g$ with $g$ standard and diagonalize $C$, so that
\[L_\cP=-\tfrac12\textstyle\sum_j\log\xi_j+\tfrac12\sum_j(\xi_j-1)Z_j^2,\]
with $Z_j$ i.i.d.\ standard normal. Because $\sum_j(\xi_j-1)=0$,
\[
\E_PL_\cP=-\tfrac12\sum_j\log\xi_j=\tfrac12\sum_j(\xi_j-1-\log\xi_j)=D_\cP,\qquad \Var_PL_\cP=\tfrac12\sum_j(\xi_j-1)^2.
\]
For any block vector $z$, with $W=C^{1/2}g$, the Cauchy--Schwarz inequality gives
\[z^\top Cz=\E\Big(\sum_Bz_B^\top W_B\Big)^2\le m\sum_B\E(z_B^\top W_B)^2=m\|z\|^2,\]
so $\xi_j\le m$. The function $u-1-\log u$ vanishes with its derivative at $u=1$ and has second derivative $u^{-2}\ge m^{-2}$ on $(0,m]$. Hence $\xi-1-\log\xi\ge(\xi-1)^2/(2m^2)$, so $\sum_j(\xi_j-1)^2\le4m^2D_\cP$, and the bound on the variance follows.
\end{proof}

\subsection{Achievability}

Choose $\gamma>0$ such that $D_\cP\ge(1+2\gamma)(|\cP|-1)\log n$ for every partition and all large $n$. For each anchor row $i$ and each $t=(i,j_2,\ldots,j_K)\in\{i\}\times[n]^{K-1}$, let $x(t)=(Y^1_i,Y^2_{j_2},\ldots,Y^K_{j_K})$. Accept $t$ if $x(t)$ lies in the support of $P$ and $L_\cP(x(t))>(1+\gamma)(|\cP|-1)\log n$ for every $\cP$ with $D_\cP<\infty$. Each row proposes one accepted tuple, if any, and abstains otherwise. In each view $a\ge2$ separately, among rows proposing the same column one is kept, and the rows left without a column are matched to the unused columns by an arbitrary bijection.

\emph{True tuples.} By Lemma~\ref{lem:llr}, the mean of $L_\cP$ exceeds the threshold by at least $\gamma D_\cP/(1+2\gamma)$. Chebyshev's inequality bounds the probability that the true tuple fails the $\cP$-test by $2K^2(1+2\gamma)^2/(\gamma^2D_\cP)=O(1/\log n)$. There are at most $B_K$ partitions, where $B_K$ is the Bell number.

\emph{False tuples.} Fix $\Pi$ and an anchor row. A tuple whose identity partition is $\cP$ has law $Q_\cP$. If $D_\cP=\infty$ it is rejected almost surely; otherwise Lemma~\ref{lem:llr} and Markov's inequality give $\Prob(L_\cP>(1+\gamma)(|\cP|-1)\log n)\le n^{-(1+\gamma)(|\cP|-1)}$. Given $\Pi$, such a tuple is determined by the objects of the $|\cP|-1$ blocks not containing the anchor, so there are at most $n^{|\cP|-1}$ of them. Since $\mathbf 1_{\cup E_t}\le\sum\mathbf 1_{E_t}$ pointwise, the probability that some false tuple of this row is accepted is at most $\sum_\cP n^{-\gamma(|\cP|-1)}\le B_Kn^{-\gamma}$.

\emph{Completion.} Call a row good if its true tuple is accepted and no false tuple is, and let $\mathcal B_n$ be the number of bad rows; then $\E\mathcal B_n=o(n)$. Good rows propose their true columns, which are distinct. In a fixed view, a good row loses its column only to a bad row, and each bad row takes at most one column. Hence $M_a\le2\mathcal B_n$ for every $a\ge2$ simultaneously, and Markov's inequality completes the proof.

\emph{Margins.} The proof uses $\gamma$ only through $D_\cP-(1+\gamma)(|\cP|-1)\log n\ge c\,D_\cP$ for a constant $c>0$ and through $B_Kn^{-\gamma}\to0$. Both hold for any $\gamma_n\to0$ with $\gamma_n\log n\to\infty$, for instance $\gamma_n=(\log n)^{-1/2}$, whenever $\liminf J/\log n>1$, so the decoder needs no knowledge of how far $J$ exceeds $\log n$. A fixed additive margin $\eta>0$, with thresholds $(|\cP|-1+\eta)\log n$ as in Appendix~\ref{app:numerics}, moves the threshold. The proof above goes through when $D_\cP\ge(1+\epsilon)(|\cP|-1+\eta)\log n$ for every $\cP$. If instead $D_\cP\le(1-\epsilon)(|\cP|-1+\eta)\log n$ for some $\cP$, Lemma~\ref{lem:llr} and Chebyshev's inequality show that the true tuple fails the $\cP$-test with probability tending to one, while each row accepts a false tuple with probability at most $B_Kn^{-\eta}$, so the expected number of anchor rows making a proposal is $o(n)$.

Suppose the selection among accepted tuples is equivariant under row relabeling. For example, the row can maximize the singleton-partition log-likelihood ratio as in G6, with uniform tie-breaking if needed. If conflict resolution and completion use only the raw proposals and independent randomness, as in G4's index-order completion in Appendix~\ref{app:numerics}, the completion argument in the proof of Theorem~\ref{thm:ladder}(ii) gives vanishing expected overlap for the final permutations. Completion that rereads the data can do better (Remark~\ref{rem:completion}). For three views with equal correlations, in the weak limit, the two-block tests need $b>1+\eta$ and the singleton test needs $b>2(2+\eta)/3$, so the level is $\max\{1+\eta,\,2(2+\eta)/3\}$, that is, $41/30$ for $\eta=0.05$.

\subsection{Converse}

Let $\cP=\{B_1,\ldots,B_m\}$, $1\in B_1$, with representatives $r_j=r_{B_j}$ and $r_1=1$. For $a\in B_j$ write $G_a=\pi_a\circ\pi_{r_j}^{-1}$. The map from $(\pi_a)_{a\ge2}$ to $\big((G_a)_{a\notin\{r_1,\ldots,r_m\}},(\pi_{r_j})_{j\ge2}\big)$ is a bijection of finite groups. Under the uniform prior, $U=(\pi_{r_2},\ldots,\pi_{r_m})$ is therefore uniform on $\Sym_n^{m-1}$ and independent of $G=(G_a)$. Give $G$ to the statistician. Knowing $G$, reorder each view $a\in B_j$ by $G_a^{-1}$, so that row $k$ of the block database $Y'_j$ is $(Y^a_{G_a(k)})_{a\in B_j}$. If $k=\pi_{r_j}(i)$, this is $X_i^{B_j}$. Since $(Y',G)$ determines $Y$, every estimator can be run in this enlarged experiment.

Let $Q=\bigotimes_j P_{B_j}^{\otimes n}$, under which the block databases are independent with i.i.d.\ rows. For each $u$, after reordering rows by $u$ the law of $Y'$ given $U=u$ and $G$ consists of $n$ i.i.d.\ tuples from $P$, and $Q$ is invariant under such reordering. Hence $D(P_{Y'\mid u,G}\|Q)=nD_\cP$. Averaging over $u$,
\[
I(U;Y'\mid G)+D(P_{Y'\mid G}\|Q)=nD_\cP,\qquad\text{so}\qquad I(U;Y'\mid G)\le nD_\cP .
\]
For a fixed map $\hat u:[n]\to[n]$, at most $\binom nt t!\le n^t$ permutations differ from $\hat u$ in exactly $t$ positions: choose the positions, after which the images of the other positions are fixed and the remaining $t$ rows can use only the remaining $t$ columns. Encoding first each error count and then the permutation,
\[
H(U\mid\widehat U,G)\le(m-1)\log(n+1)+\log n\sum_{j\ge2}\E M_{r_j}.
\]
The estimator's own randomness is independent of everything else, so data processing gives $I(U;\widehat U\mid G)\le I(U;Y'\mid G)\le nD_\cP$. With $H(U\mid G)=(m-1)\log n!$, $\log n!\ge n\log n-n$ and $\log(n+1)\le n$, we obtain~\eqref{eq:converse}. If $\limsup J/\log n<1$, apply~\eqref{eq:converse} to a minimizing partition for each $n$. The right side is then bounded below by a positive constant, while simultaneously almost exact recovery would force the bounded ratios $M_a/n$, and hence their expectations, to vanish. \qed

\section{Proof of Theorem~\ref{thm:pairwise}}\label{app:pairwise}

\begin{lemma}[From maps to permutations]\label{lem:prune}
If an estimator $f:[n]\to[n]$ has $\E\,\ov(f)=q$, then there is a randomized permutation-valued estimator $\tilde f$ with $\E\,\ov(\tilde f)\ge q^2$.
\end{lemma}

\begin{proof}
For each column $j$ with $k_j\ge1$ preimages, keep one preimage chosen uniformly at random. Then match the remaining rows and columns by any bijection. Let $c$ be the number of rows on which $f$ is correct, and $H$ the set of their columns. Each $j\in H$ has exactly one correct preimage, and $|H|=c$. The expected number of correct rows kept is $\sum_{j\in H}k_j^{-1}\ge c^2/\sum_{j\in H}k_j\ge c^2/n$. Taking expectations and using $\E c^2\ge(\E c)^2$ proves the claim.
\end{proof}

\begin{proof}[Proof of the all-or-nothing statement in Theorem~\ref{thm:pairwise}]
Suppose the claim fails. By Lemma~\ref{lem:prune}, there are $u>0$, a subsequence and permutation estimators $f$ of $\sigma_{ab}$ with $\E\,\ov(f)\ge u$. There are finitely many partitions, so we may pass to a further subsequence on which a fixed minimizing partition $\cP^*=\{B_1,\ldots,B_m\}$ separates $a$ and $b$ and $J/\log n\to j_0<1$. Put $g=1-j_0$. Use view $a$ as the anchor, with $a\in B_1$ and $b\in B_2$ as the representatives of their blocks.

\emph{Collapsing.} Reveal the within-block correspondences $G$, as in the converse. This turns the $K$ views into $m$ super-views, with super-tuple law $P$ grouped by $\cP^*$, and unknown permutations $\Pi=(\pi_{r_2},\ldots,\pi_{r_m})$ that are i.i.d.\ uniform. As in Appendix~\ref{app:main}, $G$ is independent of $\Pi$ and of the collapsed data, whose rows are shuffled using only $\Pi$. The collapsed data and $G$ determine the original observations. A partition $\cC$ of the super-views into $h$ blocks induces a coarsening $\bar\cC$ of $\cP^*$, so
\begin{equation}\label{eq:collapse}
D^{\circ}_{\cC}=D_{\bar\cC}\ge(h-1)J,\qquad D^{\circ}_{\mathrm{sing}}=D_{\cP^*}=(m-1)J.
\end{equation}
Here $D^\circ$ refers to the collapsed old data and $\mathrm{sing}$ to the partition of the super-views into singletons. Since $D^\circ_{\mathrm{sing}}<\infty$, the super-tuple has a positive definite covariance in suitable coordinates on each super-view.

\emph{Augmenting.} Fix $\varepsilon>0$ and put $H_n=(1+\varepsilon)\log n-J>0$ and $L_n=4H_n/m$. Add fresh coordinates $\ell=1,2,\ldots$ to every super-view, carried by the same rows. In coordinate $\ell$ the $m$ super-views of an object have law $N(0,R_m(s_\ell))$, independently across objects, coordinates and the old data. Take $s_\ell=r_n=(\log n)^{-1/2}$ for $\lfloor L_n/r_n^2\rfloor$ coordinates and one more coordinate absorbing the remainder, so that $\sum_\ell s_\ell^2=L_n$ and $\max_\ell s_\ell\to0$. Dropping the fresh coordinates recovers the old experiment, so $f$ still has $\E\,\ov(f)\ge u$. For a partition $\cC$ with $h$ blocks and $e(\cC)$ separated pairs of super-views, expanding $\log\det R_v(s)=(v-1)\log(1-s)+\log(1+(v-1)s)$ gives $D^{+}_{\cC}=\frac12e(\cC)L_n+O_m(\max_\ell s_\ell\,L_n)$. Moving elements from smaller to larger blocks shows $e(\cC)\ge(h-1)(m-h/2)\ge(h-1)m/2$, with equality for singletons. Divergences add over independent coordinates, so by~\eqref{eq:collapse}
\[
\frac{D^\circ_\cC+D^+_\cC}{h-1}\ge(1+\varepsilon)\log n-o(\log n),\qquad D^\circ_{\mathrm{sing}}+D^+_{\mathrm{sing}}=(m-1)(1+\varepsilon)\log n+o(\log n).
\]
By Theorem~\ref{thm:main}(a) applied to the $m$ super-views, some estimator recovers $\Pi$ from the augmented data $Y$ with $\E M_j=o(n)$ for every $j$.

\emph{Upper bound.} Let $P_{\mathrm{mix}}$ be the law of $Y$ and $Q$ the law under which the super-view databases are independent with i.i.d.\ rows from their marginals. As in the converse, $I(\Pi;Y)+D(P_{\mathrm{mix}}\|Q)=n(D^\circ_{\mathrm{sing}}+D^+_{\mathrm{sing}})$. Moreover $H(\Pi\mid Y)\le(m-1)\log(n+1)+\log n\sum_j\E M_j=o(n\log n)$, and $H(\Pi)=(m-1)\log n!$. Therefore
\begin{equation}\label{eq:upper}
\limsup D(P_{\mathrm{mix}}\|Q)/(n\log n)\le(m-1)\varepsilon .
\end{equation}
This step uses the joint entropy of all $m-1$ permutations. Recovering the designated pair alone would not cancel the leading term.

\emph{Lower bound.} Extend both $P_{\mathrm{mix}}$ and $Q$ by an independent copy of $G$ and the estimator's independent seed, with the same laws on both sides. This leaves their divergence unchanged. Let $W^1,W^2$ be the fresh coordinates of super-views $1$ and $2$, and
\[T=\sum_\ell s_\ell\sum_k W^1_{k\ell}W^2_{f(k)\ell},\]
where $f$ uses the old data reconstructed with $G$ and its seed. Given $\Pi$, the old data, $G$ and the seed, each product has mean $s_\ell$ if $f(k)=\sigma_{ab}(k)$ and $0$ otherwise, so $\E_{P_{\mathrm{mix}}}T=n\,\E\ov(f)\,L_n\ge nuL_n$. Under $Q$, given the old data, $G$ and the seed, the fresh coordinates are independent standard normals. Since $f$ is a bijection, the $n$ products in each coordinate are independent. With $\E e^{vZ_1Z_2}=(1-v^2)^{-1/2}$ for independent standard normals $Z_1,Z_2$ and $|v|<1$,
\[
\log\E_Qe^{tT}=-\frac n2\sum_\ell\log(1-t^2s_\ell^2)=\frac n2t^2L_n+O\Big(nt^4\sum_\ell s_\ell^4\Big)=\frac n2t^2L_n+o(n\log n).
\]
The Donsker--Varadhan inequality $D(P_{\mathrm{mix}}\|Q)\ge t\,\E_{P_{\mathrm{mix}}}T-\log\E_Qe^{tT}$ with $t=u$ gives
\begin{equation}\label{eq:lower}
\liminf D(P_{\mathrm{mix}}\|Q)/(n\log n)\ge\tfrac{2u^2}{m}(1+\varepsilon-j_0)\ge\tfrac{2u^2}{m}g .
\end{equation}

\emph{Conclusion.} The constants $u$, $g$ and $m$ are fixed before $\varepsilon$ is chosen. Taking $\varepsilon<2u^2g/(m(m-1))$ makes~\eqref{eq:upper} and~\eqref{eq:lower} incompatible.

If the singleton partition of $[K]$ attains the minimum, it separates every pair.
\end{proof}

\begin{proof}[Proof of Theorem~\ref{thm:pairwise}]
\emph{Part (a).} Apply Theorem~\ref{thm:main}(a) inside a set $B$ attaining
$J^*_{ab}$ and compose the estimates from its anchor to $a$ and $b$.
The composed permutation has at most the sum of their errors. If $B$
varies with $n$, apply the argument on the finitely many subsequences
with a fixed maximizing set.

\emph{Weakest partitions.}
Suppose $S\subseteq B$ meets $v\ge2$ blocks of a weakest partition
$\cC_B$ of $B$ with $m_B$ blocks and finite $J_B$. Merge the $v$
intersected blocks. By the chain rule the divergence decreases by the
total correlation among those full blocks. The coarsened divergence
is at least $(m_B-v)J_B$, with value zero when $v=m_B$, so the decrease
is at most $(v-1)J_B$. Projecting the $v$ blocks to their intersections
with $S$ can only decrease their total correlation. The induced
partition of $S$ therefore gives $J_S\le J_B$. In particular, when a
weakest partition of $[K]$ separates $a,b$ and $J$ is subcritical, every
set containing the pair has $J_S\le J$; part (b) then gives the final
claim. At a fixed level $c$, the same property shows that recursive
splitting of nodes with $J_B\le c$ produces the inclusion-maximal sets
of shared information above $c$ and any remaining singletons.

\emph{Part (b): the tree and augmentation.}
Appendix~\ref{app:f1-pairwise} runs the same argument for general known laws, with shared uniform labels in place of the Gaussian layers used here.
Choose $\delta>0$ so that eventually
$J^*_{ab}\le(1-2\delta)\log n$. Starting at $[K]$, split each
nonsingleton $B$ with $J_B\le(1-\delta)\log n$ according to a weakest
partition $\cC_B$. For fixed $K$ there are only finitely many such
trees, so on any subsequence of positive overlap we may pass to a
further subsequence with a fixed tree. Every node containing both
$a,b$ is split, so the leaf partition $\mathcal L$ separates them.
Choose the first view of
each node under the priority order $a,b,\text{other views}$ as its
representative. Reveal the within-leaf alignments $G$, and align each
leaf to its representative. With the leaf of $a$ as anchor, the $r-1$
permutations $\Pi_{\mathcal L}$ of the other leaf representatives are
independent uniform permutations, independent of $G$. This follows by
rewriting the original permutations as the within-leaf alignments and
the representative permutations, a bijection. The aligned leaves and
$G$ determine the original observations.

At each internal node $B$, add one independent layer of Gaussian
coordinates to the $m_B=|\cC_B|$ child representatives, with the same
row permutations as their old features. In coordinate $\ell$, the
tuple for one object has covariance $R_{m_B}(s_\ell)$, and we choose
\[
L_B=\sum_\ell s_\ell^2
   =\frac{4}{m_B}\big((1+\varepsilon)\log n-J_B\big),
\qquad
0\le s_\ell\le(\log n)^{-1/2},
\]
for a fixed $\varepsilon>0$ to be chosen. Copies of the maximal
correlation and a last smaller coordinate realize this sum. The new
layers are mutually independent and independent of the old data
before shuffling. If the children are internally aligned, a partition
of their super-views into $v$ groups has old divergence at least
$(v-1)J_B$ by coarsening. Suppose that it cuts $e$ child pairs. The
equicorrelation determinant expansion used in
Appendix~\ref{app:weak} gives additional divergence
$eL_B/2+O_K(\max_\ell s_\ell\,L_B)$. The smallest cut for $v$ groups
puts $m_B-v+1$ children in one group and one in each remaining group,
so $e\ge(v-1)m_B/2$. Each normalized divergence is therefore at least
$(1+\varepsilon)\log n-o(\log n)$. For the partition into all $m_B$
singletons the combined divergence is
$(m_B-1)(1+\varepsilon)\log n+o(\log n)$. All remainders are uniform
because $K$ is fixed.

\emph{Entropy of the augmented experiment.}
Write $Y_{\mathrm{aug}}$ for the shuffled augmented leaf databases and
$G$. Let $P_{\mathrm{mix}}$ be their law, and let $Q$ be the product of
the leaf database marginals, with an independent copy of $G$. Each
augmented layer has its child representatives in distinct leaves. The
old leaf divergence telescopes down the tree by the KL chain rule,
with contribution $(m_B-1)J_B$ at each internal node. In particular it
is finite even for degenerate old laws. One can project each leaf onto
its Gaussian support, since finiteness implies that the joint support
equals the product of the leaf supports. Adding the new layers and
using $\sum_B(m_B-1)=r-1$ gives the one-object divergence
\[
D_{\mathcal L}^{\mathrm{aug}}
   =(r-1)(1+\varepsilon)\log n+o(\log n).
\]
Conditional on the leaf permutations, undoing them gives $n$
independent true tuples, and leaves $Q$ invariant. Thus
\begin{equation}\label{eq:tree-aug-kl}
D(P_{\mathrm{mix}}\|Q)+I(\Pi_{\mathcal L};Y_{\mathrm{aug}})
    =nD_{\mathcal L}^{\mathrm{aug}} .
\end{equation}
Rewrite $\Pi_{\mathcal L}$ bijectively as the relative permutations
between child representatives at each internal node. Expand its
conditional entropy in an order that lists descendants first. For
the term at $B$, conditioning on the earlier permutations gives the
true alignments within its children. Theorem~\ref{thm:main}(a) applies
to their old coordinates and the new layer at $B$. The decoder ignores
all other new coordinates, including those added by ancestors to
representatives in $B$. The resulting estimators have expected error
$o(n)$ per child. Given an estimate and its number $M$ of errors,
there are at most $n^M$ possible true permutations. Consequently each
node contributes at most $(m_B-1)\log(n+1)+o(n\log n)$ to the
conditional entropy. Summing and using
$H(\Pi_{\mathcal L})=(r-1)\log(n!)$ in~\eqref{eq:tree-aug-kl} gives
\begin{equation}\label{eq:tree-aug-upper}
\frac{D(P_{\mathrm{mix}}\|Q)}{n\log n}
    \le(r-1)\varepsilon+o(1).
\end{equation}

\emph{Positive overlap would force a larger divergence.}
If estimators using the old observations have expected overlap at
least $\eta>0$ along a subsequence, Lemma~\ref{lem:prune} turns them
into permutation estimates $\widehat f$ with expected overlap at
least $u=\eta^2$. The Cauchy--Schwarz step in that lemma is
$C^2\le(\sum_{j\in H}k_j)(\sum_{j\in H}1/k_j)\le
n\sum_{j\in H}1/k_j$ for $C$ correct rows; Jensen gives the claimed
expected overlap after dividing the retained count by $n$.
At the lowest node $B_0$ containing both $a,b$, these views represent
different children. With $W^a,W^b$ their fresh coordinates at $B_0$,
put
\[
T=\sum_\ell s_\ell\sum_i W^a_{i\ell}W^b_{\widehat f(i),\ell}.
\]
Under $P_{\mathrm{mix}}$, conditioning on the old observations, the
true permutations and the estimator's seed gives a mean of $L_{B_0}$
per correct pair. Hence $\E_{P_{\mathrm{mix}}}T\ge nuL_{B_0}$. Under
$Q$, conditional on the old observations and the independent seed,
the products are independent because $\widehat f$ is bijective, so
for fixed $t$,
\[
\log\E_Q e^{tT}
   =-\frac n2\sum_\ell\log(1-t^2s_\ell^2)
   =\tfrac n2t^2 L_{B_0}+o(n\log n).
\]
Attaching the same independent seed to both experiments leaves their
divergence unchanged. The entropy inequality with $t=u$ now gives
\[
\frac{D(P_{\mathrm{mix}}\|Q)}{n\log n}
 \ge \frac{u^2L_{B_0}}{2\log n}-o(1)
 \ge \frac{2u^2\delta}{K}-o(1).
\]
Choose $0<\varepsilon<2u^2\delta/[K(K-1)]$ to contradict
\eqref{eq:tree-aug-upper}. Taking near-optimal estimators on a
subsequence establishes the supremum claim.
\end{proof}

\section{Proofs of Corollary~\ref{cor:equi} and Proposition~\ref{prop:strength}}\label{app:weak}

\begin{proof}[Proof of Corollary~\ref{cor:equi}]
The linear map is invertible and commutes with row permutations, so it does not change the problem. Let $F_v=-\frac12\sum_\ell\log\det R_v(\rho_\ell)$ for $v\le K$, so that $F_1=0$ and $D_\cP=F_K-\sum_{B\in\cP}F_{|B|}$. The increment $F_v-F_{v-1}=-\frac12\sum_\ell\big[\log(1-\rho_\ell)+\log\frac{1+(v-1)\rho_\ell}{1+(v-2)\rho_\ell}\big]$ is nondecreasing in $v$. For $p$ blocks, moving an element from a smaller block to a larger one does not decrease $\sum_BF_{|B|}$, so $\sum_BF_{|B|}\le F_{K-p+1}$. The average of the last $p-1$ increments is at least the average of all $K-1$ of them, so $F_K-F_{K-p+1}\ge(p-1)F_K/(K-1)$. Hence $D_\cP/(p-1)\ge F_K/(K-1)$, with equality for singletons. When $(K-1)\rho_{\max}\le1/2$, Taylor expansion gives $F_v=\frac12\binom v2A+O_K(\rho_{\max}A)$, so $J=F_K/(K-1)=(K/4+O_K(\rho_{\max}))A$. The thresholds and the all-or-nothing statement follow from Theorems~\ref{thm:main} and~\ref{thm:pairwise}.

For a cut with sides of sizes $s$ and $K-s$,
\[
F_K-F_s-F_{K-s}=\tfrac12s(K-s)A+O_K(\rho_{\max}A).
\]
Minimizing over $1\le s<K$ gives
$I_*=((K-1)/2+O_K(\rho_{\max}))A$. In these coordinates the operator norm of
the off-diagonal joint correlation matrix is $(K-1)\rho_{\max}$, which tends to
zero. Theorem~\ref{thm:exact} thus gives the exact threshold
$A/\log n=4/(K-1)$ with fixed margins.

In the model of~\citet{wee2026geometric}, after dividing each view by $(1+\tau^2)^{1/2}$ every coordinate has $\rho=(1+\tau^2)^{-1}$. If $\tau^4=d/(b\log n)$ and $d/\log n\to\infty$, then $\tau\to\infty$ and $A/\log n=d\rho^2/\log n=b\,(\tau^2/(1+\tau^2))^2\to b$.
\end{proof}

\begin{proof}[Proof of Proposition~\ref{prop:strength}]
Write $R=I+E$, where $E$ has zero diagonal blocks and $\|E\|_{\mathrm{op}}=h$. For a partition $\cP$, let $B_\cP$ be the block-diagonal part of $R$ along $\cP$, $F=R-B_\cP$ and $W=B_\cP^{-1/2}FB_\cP^{-1/2}$. Then $D_\cP=-\frac12\log\det(I+W)$, $\tr W=0$, and $\|W\|_{\mathrm{op}}=O(h)$, since the spectrum of $B_\cP$ lies in $[1-h,1+h]$. Expanding $-\log(1+x)+x=x^2/2+O(x^3)$ over the eigenvalues of $W$ and using $|\tr W^j|\le\|W\|^{j-2}_{\mathrm{op}}\tr W^2$ gives $D_\cP=\frac14(1+O(h))\|W\|_F^2$. The bounds on the singular values of $B_\cP^{-1/2}$ give $\|W\|_F^2=(1+O(h))\|F\|_F^2=(2+O(h))\sum_{a<b\text{ separated}}\|R_{ab}\|_F^2$. Finally $I_{ab}=-\frac12\log\det(I-R_{ab}R_{ab}^\top)=\frac12(1+O(h^2))\|R_{ab}\|_F^2$. All constants depend only on $K$, and the errors are relative to the cut in question. Dividing by $|\cP|-1$ and minimizing gives $J=(1+O(h))s(I)$. A cut with zero weight has $D_\cP=0$ on both sides.

For a bipartition $\{S,S^c\}$, $D_{\{S,S^c\}}=I(X^S;X^{S^c})$.
Applying the same relative approximation to each bipartition gives
\[
(1-C_Kh)\min_S w(S,S^c)
 \ \le\ I_*\ \le\
(1+C_Kh)\min_S w(S,S^c),
\qquad w(S,S^c)=\sum_{a\in S,\,b\notin S}I_{ab}.
\]
This also covers a zero minimum cut. For the lower bound minimize the
lower estimate over all cuts. For the upper bound evaluate the upper
estimate at a minimum-weight cut.

The Tutte--Nash-Williams theorem equates strength and fractional tree packing. For integer weights, $k$ edge-disjoint spanning trees fit if and only if every partition into $p$ blocks is crossed by at least $k(p-1)$ edges. Apply this to rational weights scaled by a common denominator, let the scale grow, and extend to real weights by monotone approximation.
\end{proof}

\section{Proof of Theorem~\ref{thm:exact}}\label{app:exact}

\subsection{Achievability in the weak-correlation regime}

\paragraph{Grouping candidates by lost true edges.}
Whiten each view and condition on the true permutations. Relabel the rows so that the truth is the identity in each view. Write a candidate as $\tau=(\mathrm{id},\tau_2,\ldots,\tau_K)$, where $\tau_a(i)$ is the source row used by view $a$ in candidate tuple $i$. By Remark~\ref{rem:posterior}, the part of the Gaussian log-likelihood that depends on $\tau$ is
\[
  \sum_{i=1}^n\sum_{a<b}
  (X^a_{\tau_a(i)})^\top W_{ab}X^b_{\tau_b(i)},
  \qquad W_{ab}=-(R^{-1})_{ab}.
\]
Here and below $X$ denotes the whitened features. The covariance $R$ is positive definite for all sufficiently large $n$ since $h=\|R-I\|_{\mathrm{op}}\to0$.

For each source row $i$, let $\cP_i$ group two views together if their copies of that row go to the same candidate tuple, equivalently if their values of $\tau_a^{-1}(i)$ agree. Call $H=(\cP_i)_{i=1}^n$ the pattern of $\tau$. Set
\[
  U=\{i:|\cP_i|\ge2\},\qquad m=|U|,
  \qquad Q=\sum_{i\in U}(|\cP_i|-1).
\]
A nonidentity permutation moves at least two source rows, which must be split from the anchor view, so a wrong candidate has $m\ge2$. Also $m\le Q\le(K-1)m$. The number of patterns with $m$ split rows is at most $\binom nm B_K^m$, where $B_K$ is the Bell number. For a fixed pattern, each inverse permutation fixes every row outside $U$ and therefore maps $U$ into itself. In a split row $i$, the block containing view~$1$ goes to tuple $i$, and each other block can go to at most $m$ tuples. These choices determine the inverse permutations, giving at most $m^Q$ candidates with pattern $H$.

Let $T_H$ be the sum of the true edge scores discarded by the candidate, that is, the scores between views in different blocks of $\cP_i$ for all $i\in U$. Let $F_\tau$ be the sum of the false edge scores introduced by $\tau$. The score difference from the truth is $F_\tau-T_H$. For each view pair, the number of lost true edges equals the number of added false edges, since both assignments use exactly $n$ edges. Each false edge connects independent source rows, so $\E F_\tau=0$.

\paragraph{Joint moment bound.}
For a partition $\cP$ put $c_R(\cP)=\sum_{a<b\text{ separated by }\cP}\|R_{ab}\|_F^2$. For a pattern $H$ define
\[
\begin{aligned}
  A_H &= \sum_{i\in U}c_R(\cP_i),\\
  V_H^0 &= \sum_{i\in U}\sum_{a<b\text{ separated by }\cP_i}
                 \|W_{ab}\|_F^2,\\
  \mu_H &= \E T_H
         = \sum_{i\in U}\sum_{a<b\text{ separated by }\cP_i}
                 \langle W_{ab},R_{ab}\rangle_F .
\end{aligned}
\]

\begin{lemma}[Joint moment bound]\label{lem:exact-mgf}
There exist $h_K>0$ and $C_K>0$ depending only on $K$ such that, if $h\le h_K$, then
\[
  |\mu_H-A_H|+|V_H^0-A_H|\le C_Kh A_H .
\]
With $V_H=(1+C_Kh)V_H^0$, for every pattern $H$, every candidate $\tau$ with that pattern and all $|x|,|y|\le2$,
\begin{equation}\label{eq:exact-mgf}
  \log\E e^{xF_\tau+y(T_H-\mu_H)}
  \le \tfrac12 V_H (x^2+y^2).
\end{equation}
\end{lemma}

\begin{proof}
For a partition $\cP$, let $D$ be the block-diagonal part of $R$ in that partition, $E=R-D$, and $\mathcal O$ the Frobenius-orthogonal projection onto entries between different blocks. Set $\Omega=R^{-1}$. The inverse identity gives
\[
\begin{aligned}
  \mathcal O(-\Omega)-E
    &=\mathcal O(D^{-1}E\Omega-E)\\
    &=\mathcal O\big((D^{-1}-I)E\Omega+E(\Omega-I)\big).
\end{aligned}
\]
Since $\|D-I\|_{\mathrm{op}}\le h$, the inverse deviations of $D$ and $R$ from the identity have operator norm $O(h)$ when $h$ is small. Thus the displayed difference has Frobenius norm $O(h)\|E\|_F$. Applying Cauchy--Schwarz to the scalar products and the triangle inequality to the norms, and summing over the split rows, yields the claimed relative errors for $\mu_H$ and $V_H^0$.

Write $T_H$ and $F_\tau$ as quadratic forms in the stacked Gaussian features, with symmetric matrices $M_T$ and $M_F$. Each edge between views $a$ and $b$ contributes $W_{ab}/2$ at one block and its transpose at the opposite block. Both squared Frobenius norms equal $V_H^0/2$, by the equality of added and lost edge counts for each view pair. The two matrices are Frobenius-orthogonal because the edges of $M_T$ connect the same source row and those of $M_F$ connect different source rows. Each matrix has at most $K-1$ nonzero blocks in every block row, with block operator norms $O(h)$. The scalar symmetric matrix of block norms bounds the operator norm of the original matrix and has maximum row sum $O_K(h)$. Thus $\|M_T\|_{\mathrm{op}}+\|M_F\|_{\mathrm{op}}=O_K(h)$.

Let $\mathcal R=I_n\otimes R$ and $C=\mathcal R^{1/2}(xM_F+yM_T)\mathcal R^{1/2}$. For the stated range of $x,y$, choose $h_K$ so that $\|C\|_{\mathrm{op}}\le1/4$. The centered Gaussian quadratic-form formula gives
\[
\begin{aligned}
  \log\E e^{xF_\tau+y(T_H-\mu_H)}
    &=-\tfrac12\log\det(I-2C)-\tr C\\
    &\le(1+O_K(h))\|C\|_F^2\\
    &\le\tfrac12(1+C_Kh)V_H^0(x^2+y^2).
\end{aligned}
\]
The first inequality expands $-\tfrac12\log(1-2z)-z=z^2+O(|z|^3)$ for each eigenvalue $z$ of $C$. The second uses $\|\mathcal R\|_{\mathrm{op}}\le1+h$ and the orthogonality of $M_T$ and $M_F$. A single $C_K$ works in both claims independently of $n$, the feature dimensions, and the pattern.
\end{proof}

\paragraph{The minimum cut controls every pattern.}
Let $c_*=\min_{\varnothing\ne S\subsetneq[K]}c_R(\{S,S^c\})$. Separating each block of a $p$-block partition $\cP$ from the rest counts each cut edge twice. Hence
\begin{equation}\label{eq:exact-pattern-weight}
  c_R(\cP)\ge\tfrac p2 c_*,
  \qquad A_H\ge\tfrac12(m+Q)c_* .
\end{equation}
Proposition~\ref{prop:strength} and $I_{ab}=(1+O(h^2))\|R_{ab}\|_F^2/2$ give
$D_\cP=(1+O_K(h))c_R(\cP)/2$ uniformly in $\cP$ and $I_*=(1+O_K(h))c_*/2$.
It follows that $A_H\ge(1-O_K(h))(m+Q)I_*$, and for $p=|\cP|\ge2$,
\[
  D_\cP\ge(1-O_K(h))\tfrac p2 I_*,
  \qquad J\ge(1-O_K(h))\tfrac12 I_* .
\]
Under the upper-side hypothesis, $c_*>0$. Lemma~\ref{lem:exact-mgf} gives $D_H:=\mu_H^2/V_H=(1+O_K(h))A_H$. Using~\eqref{eq:exact-pattern-weight} and $I_*\ge(2+\epsilon)\log n$, there is a fixed $\epsilon_0>0$ such that for all large $n$ and every wrong pattern,
\begin{equation}\label{eq:exact-effective}
  D_H\ge2(1+\epsilon_0)(m+Q)\log n .
\end{equation}
For large $n$, $\mu_H>0$ and $\mu_H/V_H=1+O_K(h)\le2$ uniformly. All moment parameters below have absolute value at most $\mu_H/V_H$.

\paragraph{A two-level union bound.}
Set $w=m\log n$, $\alpha=\log m/\log n$, and $q=Q/m$. Write $L=(1-\alpha)w$ for the leading term in the logarithm of the pattern count and $M=\alpha q w$ for the logarithm of the bound on the number of candidates per pattern. Indeed, $\binom nm B_K^m\le\exp(L+O_K(m))$ and $m^Q=e^M$. Fix $0<\delta\le\epsilon_0/2$. We bound the probability that any candidate of a given pattern ties or beats the truth by $2e^{-L-\delta w}$.

If $M\le L$, apply~\eqref{eq:exact-mgf} with $x=-y=\mu_H/(2V_H)$ to obtain $\Prob(F_\tau\ge T_H)\le e^{-D_H/4}$. Since $\alpha\le1/(1+q)$ and $q\ge1$,
\[
  w\le L+M=[1+\alpha(q-1)]w
  \le\frac{2q}{1+q}w\le\frac{1+q}{2}w.
\]
By~\eqref{eq:exact-effective}, $D_H/4\ge(1+\epsilon_0)(L+M)$. The union bound over the $e^M$ candidates gives at most $e^{-L-\delta w}$.

If $M>L$, Cauchy--Schwarz and $L+M/q=w$ give
\[
\begin{aligned}
  (\sqrt{L+\delta w}+\sqrt{M+\delta w})^2
  &\le(1+q)(1+\delta+\delta/q)w\\
  &\le D_H/2 .
\end{aligned}
\]
Define $a=\sqrt{2(L+\delta w)/D_H}$ and $s=1-a$. Here $M>L$ implies $L\le qw/(1+q)$, so
\[
  a^2\le
  \frac{q/(1+q)+\delta}{(1+\epsilon_0)(1+q)}
  \le\frac{1/4+\delta/2}{1+\epsilon_0}\le\tfrac14 .
\]
Thus $s\ge a$ and $s\sqrt{D_H/2}\ge\sqrt{M+\delta w}$. Apply~\eqref{eq:exact-mgf} first with $x=0$ and $y=-a\mu_H/V_H$. The common event $T_H<s\mu_H$ has probability at most $e^{-D_Ha^2/2}=e^{-L-\delta w}$. On the event $F_\tau\ge T_H\ge s\mu_H$, we have
\[
  sF_\tau-a(T_H-\mu_H)\ge(s^2+a^2)\mu_H .
\]
Indeed, the left side is at least $(s-a)T_H+a\mu_H$. Using $x=s\mu_H/V_H$ and $y=-a\mu_H/V_H$ in~\eqref{eq:exact-mgf} bounds the probability of this event for one candidate by $e^{-D_H(s^2+a^2)/2}\le e^{-L-M-2\delta w}$. Summing this bound over at most $e^M$ candidates and counting the common lower tail once gives the claimed $2e^{-L-\delta w}$.

Summing over patterns and $m$ gives
\[
  \Prob(\text{some wrong candidate ties or beats the truth})
    \le 2\sum_{m=2}^n\exp\{O_K(m)-\delta m\log n\}
    =o(1).
\]
The bound is uniform in the true permutations, proving part~(a). A related separation of the true-score lower tail from the probability that a wrong score wins is used by \citet[Appendix~C.3]{dai2023gaussian} in their analysis of partial recovery for two databases.

\subsection{Converse for a pair across a bipartition}

\paragraph{Revealing the within-side correspondences.}
Take any nontrivial bipartition $\{S,S^c\}$, with $1\in S$ by relabeling its sides, and set $I=I(X^S;X^{S^c})$. Let $b_0\in S^c$ and $U=\pi_{b_0}$. Reveal $G=(G_a)_{a\in[K]}$, where $G_a=\pi_a$ for $a\in S$ and $G_a=\pi_a\circ U^{-1}$ for $a\in S^c$. The original permutation vector is in bijection with $(G,U)$, so $U$ is uniform conditional on $G$ and independent of the feature tuples. Using $G$ to concatenate the features on each side gives two databases with true correspondence $U$. The original observations can be reconstructed from the two databases and $G$.

For $a\in S$ and $b\in S^c$, the true correspondence is $G_b\circ U\circ G_a^{-1}$. Thus an estimator returning the correct correspondence gives the correct $U$ by composition with $G_b^{-1}$ and $G_a$. For the reverse direction, first invert any permutation-valued output and replace nonbijective outputs arbitrarily, which preserves every exact success. It suffices to bound the probability of recovering $U$ in this more informative problem.

\paragraph{Canonical correlations and a winning exchange.}
Center and whiten the two sides on their marginal supports. Separate orthogonal changes of coordinates from a singular value decomposition of the cross-covariance then give independent paired standard Gaussian coordinates with correlations $r_j\in[0,1]$, together with independent residual coordinates. These invertible transformations on the supports commute with row permutations. Finiteness of $I$ excludes $r_j=1$, since such a pair is supported on a line of measure zero under its product marginals. Thus
\[
  I=-\tfrac12\sum_j\log(1-r_j^2),
  \qquad w_j=\frac{r_j}{1-r_j^2}.
\]
If $I=0$, $U$ is independent of the databases even given $G$, and its probability of exact recovery is at most $1/n!$. Now suppose $I>0$. As the prior is uniform conditional on $G$, maximum a posteriori estimation is maximum likelihood and is optimal for exact recovery. Condition on any true $U$ and relabel the second database to make it the identity. Denote the canonical coordinates of object $i$ by $(\xi_{ij},\eta_{ij})$. The log-likelihood of the truth minus that of the exchange of rows $i,k$ in the second database is
\[
  \Delta_{ik}=\sum_j w_j(\xi_{ij}-\xi_{kj})(\eta_{ij}-\eta_{kj}) .
\]
Any exchange with $\Delta_{ik}<0$ makes the truth strictly suboptimal.

Let $A_j=(\xi_{1j}-\xi_{2j})/\sqrt2$ and $B_j=(\eta_{1j}-\eta_{2j})/\sqrt2$. The pairs are independent across $j$, each with standard Gaussian marginals and correlation $r_j$. Put $S_{12}=\Delta_{12}/2=\sum_j w_j A_jB_j$. Multiplying their joint density by $e^{-S_{12}}$ cancels the cross terms. The total mass is $\prod_j\sqrt{1-r_j^2}=e^{-I}$, and in the normalized tilted law the two entries of each pair are independent $N(0,1-r_j^2)$. Under the tilt, $S_{12}$ has a continuous symmetric law and variance $A_S:=\sum_j r_j^2\le2I$. Chebyshev's inequality and symmetry give tilted probability at least $3/8$ for $-2\sqrt{A_S}\le S_{12}<0$. Undoing the tilt yields
\begin{equation}\label{eq:exact-swap}
  p:=\Prob(\Delta_{12}<0)
  \ge\tfrac38 e^{-I-2\sqrt{A_S}}
  \ge\tfrac38 e^{-I-\sqrt{8I}} .
\end{equation}

\paragraph{Overlapping exchanges and the second moment.}
For exchanges $(1,2)$ and $(1,3)$, the contribution of coordinate $j$ to $(\Delta_{12}+\Delta_{13})/2$ is $w_j/2$ times the bilinear form of the three rows with the path Laplacian. This Laplacian has eigenvalues $0,1,3$. Applying the same diagonalizing orthogonal transform to both databases' row vectors gives independent Gaussian pairs $(A,B)$, each with correlation $r_j$. For $t\in\{1/2,3/2\}$ and $\lambda_j=r_j^2/(1-r_j^2)$, direct integration gives
\[
  \E e^{-t w_j AB}=(1+\lambda_j(2t-t^2))^{-1/2}.
\]
In fact the off-diagonal precision entry changes from $-w_j$ to $(t-1)w_j$, and the displayed factor is the square root of the ratio of the two precision determinants. Both parameters give $2t-t^2=3/4$. Markov's inequality and concavity of $\log$ yield
\begin{equation}\label{eq:exact-overlap-swap}
\begin{aligned}
  p_{\mathrm{both}}&:=\Prob(\Delta_{12}<0,\Delta_{13}<0)\\
    &\le\E e^{-(\Delta_{12}+\Delta_{13})/2}\\
    &=\prod_j(1+3\lambda_j/4)^{-1}
      \le e^{-3I/2},
\end{aligned}
\end{equation}
where $\log(1+3\lambda/4)\ge(3/4)\log(1+\lambda)$ and $\sum_j\log(1+\lambda_j)=2I$. These bounds hold for all finite Gaussian cut informations.

Let $N=\sum_{i<k}\mathbf{1}\{\Delta_{ik}<0\}$. Exchanges on disjoint rows are independent, and the number of ordered pairs of distinct exchanges sharing a row is $6\binom n3$. Hence $\E N=\binom n2p$ and $\Var N\le\binom n2p+6\binom n3p_{\mathrm{both}}$. Equations~\eqref{eq:exact-swap} and~\eqref{eq:exact-overlap-swap} give
\[
\begin{aligned}
  \Prob(N=0)
  &\le \frac{\Var N}{(\E N)^2}
   \le O\left(\frac{1}{n^2p}+\frac{p_{\mathrm{both}}}{np^2}\right)\\
  &\le O\left(n^{-2}e^{I+\sqrt{8I}}
       +n^{-1}e^{I/2+2\sqrt{8I}}\right).
\end{aligned}
\]
If $I\le(2-\epsilon)\log n$, the two terms are $n^{-\epsilon+o(1)}$ and $n^{-\epsilon/2+o(1)}$, respectively, as upper bounds. All bounds are uniform in $G$, the true $U$, the canonical spectrum, and the pair across the cut. Maximum likelihood therefore fails with probability tending to one, even given the oracle information. Choosing a minimizing cut gives part~(b), including its pairwise conclusion.

\section{Proofs for exact recovery beyond weak correlations}\label{app:beyond}

Appendices~\ref{app:beyond:rare}--\ref{app:beyond:window} prove the results of Section~\ref{sec:beyond}. Several steps are finite computer-assisted checks, carried out in exact rational arithmetic or in outward-rounded interval arithmetic. The scripts and the tables that they output are distributed with this paper as ancillary files, and all file paths in this appendix are relative to that directory. The scripts use only the Python standard library, and the \path{README} file there gives the commands that regenerate every table. Appendices~\ref{app:beyond:axial} and~\ref{app:beyond:analytic} state two further results with proofs in outline.

\subsection{The rare-subpopulation converse and the constants at correlation \texorpdfstring{$-1/3$}{-1/3}}
\label{app:beyond:rare}

\noindent\textbf{Definition~\ref{def:rare} and its variational form.}
Let $p$ be a density and let $\cP$ have $q\ge2$ blocks. For
$\alpha=(q-1)/q$, the rare-subpopulation rate has the equivalent forms
\begin{equation}\label{eq:rare:two-forms}
 \begin{split}
 R_\cP
 &=\frac{1}{q-1}\min_{v=\bigotimes_B v_B}D_\alpha(p\|v)\\
 &=\min_Q\left\{D(Q\|p)+\frac{1}{q-1}
                 D\Big(Q\,\Big\|\bigotimes_B Q_B\Big)\right\}.
 \end{split}
\end{equation}
Here $Q_B$ denotes the block marginal of $Q$.

\begin{proof}
Write $\beta=1-\alpha=1/q$, $A(v)=\int p^\alpha v^\beta$ and
$q_v=p^\alpha v^\beta/A(v)$ for $A(v)>0$. For any $Q$ with finite
objective against $p,v$, substitution of the tilted density gives
\begin{equation}\label{eq:rare:kl-identity}
 D(Q\|p)+\frac{D(Q\|v)}{q-1}
   =\frac{1}{\alpha}D(Q\|Q_v)
      +\frac{D_\alpha(p\|v)}{q-1}.
\end{equation}
This follows from
$\log q_v=\alpha\log p+\beta\log v-\log A(v)$ on the common support and
$D_\alpha(p\|v)=-q\log A(v)$. A reference with $A(v)=0$ gives infinite
objective for every $Q$. When $A(v)>0$, both divergences on the left are
finite at $Q_v$. For instance, for $0<\eta<\alpha$,
\[
 \int q_v\,(q_v/p)^{\eta/\beta}
   = A(v)^{-1-\eta/\beta}\int p^{\alpha-\eta}v^{\beta+\eta}
   \le A(v)^{-1-\eta/\beta}.
\]
This controls the positive part of $\log(q_v/p)$ under $Q_v$. Interchanging
$p,v$ and $\alpha,\beta$ gives the corresponding bound for $v$. The
negative parts are integrable by $u\log(1/u)\le 1/e$ on $0<u\le1$,
integrated against the reference density. Thus the minimum over $Q$
for fixed $v$ is attained by $Q_v$. This is the variational formula for
R\'enyi divergence of~\citet[Theorem~30]{vanerven2014renyi}. The KL chain rule gives
\[
 D(Q\|v)=D\Big(Q\,\Big\|\bigotimes_B Q_B\Big)+\sum_B D(Q_B\|v_B).
\]
In particular the minimizing $v$ for fixed $Q$ is the product of its
block marginals. These two minimizations give~\eqref{eq:rare:two-forms}
with infima.

To see attainment, maximize $A(v)$ in the variables $f_B=v_B^{1/q}$.
Each $f_B$ belongs to the nonnegative unit ball of $L^q$. Its weak
compactness follows from reflexivity and weak closedness of the positive
cone. The functional
$\int p^\alpha\prod_B f_B$ is jointly weakly continuous on the product
of these balls. In fact, $p^\alpha\in L^{1/\alpha}$ can be approximated
there by finite sums of indicators of rectangles of finite measure.
For each such sum the integral depends continuously on finitely many
weak coordinates of the $f_B$. H\"older bounds the approximation error
uniformly by the $L^{1/\alpha}$ error, since
$\|\prod_B f_B\|_{L^q}=\prod_B\|f_B\|_{L^q}\le1$ on the product space.
The maximum is positive, as is seen by taking positive block densities.
Positive homogeneity then forces $\|f_B\|_{L^q}=1$ for every $B$ at the
maximum. The associated $v$ and $Q_v$ attain the joint minimum, so
$v=\bigotimes_B(Q_v)_B$ by the chain rule. Taking $Q=p$ in the second
form gives $R_\cP\le D_\cP/(q-1)$.
\end{proof}

\noindent\textbf{Theorem~\ref{thm:rare}.}
Fix $K$, a density $p$ and a partition $\cP$ with $q\ge2$ blocks. If
$dR_\cP\le(1-\epsilon)\log n$ for all large $n$, the probability that
any estimator recovers all relative permutations tends to zero. For
Gaussian $p$ and $q=2$, the necessary swap bound already implies the
necessary rare-subpopulation bound for exact recovery.

\begin{proof}
Let $v$ and $Q=Q_v$ attain~\eqref{eq:rare:two-forms} and put
$D=D(Q\|p)$ and $T=D(Q\|v)$. These are finite and
$R_\cP=D+T/(q-1)$. If $R_\cP=0$ then both vanish, whence $p=Q=v$.
Revealing the within-block alignments leaves $q-1$ independent
uniform permutations with no information across the blocks, and the
probability of guessing them all is $(n!)^{-(q-1)}$.

Suppose $R_\cP>0$. By reducing the fixed margin if needed, we may take
$0<\epsilon<1$. Increasing $d$ to
$\max\{d,\lfloor\sqrt{\log n}\rfloor\}$ preserves the hypothesis for
large $n$ and can only increase the optimal success probability.
Thus we may assume $d\to\infty$, and we have $d=O(\log n)$.
Fix $\delta>0$ small enough that
$qd\delta\le(q-1)\epsilon\log n/2$ for all large $n$. For one row of
$d$ coordinates let $A_d$ be the intersection of the events
\[
 \log\frac{dQ^{\otimes d}}{dP^{\otimes d}}\le d(D+\delta),
 \qquad
 \log\frac{dQ^{\otimes d}}{dV^{\otimes d}}\le d(T+\delta).
\]
Here $P,V$ have densities $p,v$, and the density ratios can be
restricted to the support of $Q^{\otimes d}$. The law of large numbers gives
$Q^{\otimes d}(A_d)\to1$. Independently for each object, mark it
with probability
\[
 e^{-d(D+\delta)}
 \frac{dQ^{\otimes d}}{dP^{\otimes d}}\,{\bf1}_{A_d}.
\]
This lies in $[0,1]$ by the first defining inequality. A row
conditional on being marked has law
$Q_A=Q^{\otimes d}(\,\cdot\,\mid A_d)$.
The number $m$ of marked rows is binomial with mean
$ne^{-d(D+\delta)}Q^{\otimes d}(A_d)$, which grows at least as a
positive power of $n$ because $D\le R_\cP$.
Binomial concentration therefore gives, with probability tending to one,
\begin{equation}\label{eq:rare:marked-size}
 \log m=\log n-d(D+\delta)+O(1).
\end{equation}

For every block choose one representative view, choosing view $1$
for its own block. Reveal all within-block correspondences, the
sets of marked positions in the representative views, and the
correspondences of all unmarked positions between the representatives.
To see that the within-block maps and representative permutations are
independent before this revelation, compose each nonrepresentative
permutation with the inverse representative permutation. This is a
bijection on the finite set of permutation assignments. Conditional
on the anchor's marked set, its unmarked images, and the sets of
marked images, each other representative has a uniform bijection
between the anchor's marked set and its own marked set, independently
of the others. The marks are generated from the unpermuted rows
with independent coins, so these bijections are independent of
the marked rows before shuffling. Conditional on the marked anchor
positions and on the unmarked data, the marked rows are still
independent with law $Q_A$. Thus the remaining problem consists of
$m$ such rows and $q-1$ independent uniform bijections. A similar
revelation outside a data-dependent set of vertices underlies the
exact community-recovery converses of~\citet{gaudio2022exact}
and~\citet{racz2024harnessing} for correlated block models; here the
set comes from a likelihood-ratio marking, which makes the retained
rows i.i.d.\ from the tilted law.

With the revealed information fixed, compare its conditional
distribution to the one with the same prior on these bijections and
independent rows from $V^{\otimes d}$ before shuffling. As $v$ factors
over blocks, the shuffled data in this reference law are independent
of the bijections. Any estimator has reference success probability
at most $(m!)^{-(q-1)}$, including after averaging its randomization.
The second defining inequality of $A_d$ gives
\[
 \frac{dQ_A}{dV^{\otimes d}}
 \le \frac{e^{d(T+\delta)}}{Q^{\otimes d}(A_d)}.
\]
Multiplying across rows bounds the conditional success probability
under the planted law by the exponential of
\[
 m\{d(T+\delta)-\log Q^{\otimes d}(A_d)\}-(q-1)\log(m!).
\]
On~\eqref{eq:rare:marked-size}, use $\log(m!)\ge m\log m-m$ to
bound this exponent by
\[
 m\{(q-1)(dR_\cP-\log n)+qd\delta+O(1)\}
 \ \le\ -\tfrac14(q-1)\epsilon m\log n
\]
for large $n$. Thus the success probability with the revealed
information tends to zero, and the same holds in the original problem.

For the final assertion, consider any two-block cut $S|S^c$. The
nonnegative Hilbert--Schmidt kernel $z=\sqrt p$, viewed as an operator
from one block's $L^2$ to the other's, has singular values
$\sigma_i$ with $a_i=\sigma_i^2$ and $\sum_i a_i=1$. Maximizing
$\int z(x,y)\sqrt{v_S(x)}\sqrt{v_{S^c}(y)}\,dx\,dy$ over block-product
references gives $\|z\|_{\rm op}=\sqrt{\max_i a_i}$. Indeed, the
operator norm is attained by a pair of unit singular vectors, and taking
absolute values cannot decrease their pairing with the nonnegative
kernel. The resulting nonnegative unit vectors are square roots of
probability densities. Hence
\[
 R_{S|S^c}=-\log\max_i a_i=H_\infty(a)
 \ \ge\ \tfrac12 H_2(a),\qquad H_2(a)=-\log\sum_i a_i^2 .
\]
The inequality uses $\sum_i a_i^2\ge(\max_i a_i)^2$. The swap
exponent across the cut is $H_2(a)=I^\circ_2(S)$, which is
$I(X_S;X_{S^c})$ for Gaussian $p$ by
Appendix~\ref{app:beyond:patterns}. Consequently
$R_{S|S^c}\ge I_*/2$ in the Gaussian case. Integrating around an
$m$-cycle gives the midpoint coefficient
$\tr[(z^*z)^m]=\sum_i a_i^m$. The cycle is conjugate to its inverse
under a relabeling of the $m$ objects, so the Chernoff coefficient
is symmetric in the parameter and is minimized at the midpoint by
log-convexity. Its exponent per moved object is $-\log\|a\|_m$,
which increases to $H_\infty(a)$.
\end{proof}

\noindent\textbf{The constants at $r=-1/3$ and the failure window in Theorem~\ref{thm:three}.}
Let $K=3$ and $p=N(0,R)$ with unit variances and all three correlations
equal to $-1/3$. The per-coordinate rates are
$I_*=\tfrac12\log(3/2)$, $J=\tfrac14\log(27/16)$, and
$R_{1|2|3}\approx0.09835$. If $d=\lceil\kappa\log n\rceil$ with
$9.866<\kappa<10.167$, the total information quantities
$dI_*$ and $dJ$ exceed $(2+\epsilon)\log n$ and $(1+\epsilon)\log n$,
respectively, for some fixed $\epsilon>0$. Exact recovery still
fails with probability tending to one.

\begin{proof}
The eigenvalues of $R$ are $4/3,4/3,1/3$, so $\det R=16/27$.
Every two-view principal minor is $8/9$. The Gaussian determinant
formula gives $I_*=\tfrac12\log[(8/9)/(16/27)]$ and the singleton
partition has normalized information $-\tfrac14\log\det R$.
The latter is smaller than $I_*$, so it equals $J$.

For the singleton reference take $V=vI$ with
$v=(1+\sqrt3)/3$. The tilted covariance is
$S=[2R^{-1}/3+I/(3v)]^{-1}$. It has equal diagonal entries, and
$\tr S=3v$ is equivalent to
\[
 1=\frac{1}{6v+1}+\frac{4}{3v+2},
\]
which the stated $v$ satisfies. The sufficiency of this marginal
equation for optimality is proved in Appendix~\ref{app:beyond:gaussian}.
The eigenvalues of $S^{-1}$ are $\sqrt3/2$ with multiplicity
two and $(3+\sqrt3)/2$ with multiplicity one. Hence
$v\det(S^{-1})=(3+2\sqrt3)/4$. The affinity formula
\eqref{eq:rare:gauss-affinity} now gives
\begin{equation}\label{eq:rare:example-rate}
 R_{1|2|3}=\frac12\log\frac{16}{27}
             +\frac34\log\frac{3+2\sqrt3}{4}.
\end{equation}

\paragraph{Numerical certificate.}
The rational interval check in
\nolinkurl{scripts/check_d1_k3_sharp.py}
encloses \eqref{eq:rare:example-rate} and its reciprocal.
The \texttt{R3} and \texttt{inverse\_R3} rows of
\nolinkurl{d1_k3_sharp/rows.csv} give
$0.098353188221<R_{1|2|3}<0.098353188222$ and
$10.167438575994<R_{1|2|3}^{-1}<10.167438575995$.
The lower inequality $\kappa I_*>2$ follows from
$2/I_*<9.866$, which follows by keeping the first three positive
terms in the series
$\log(3/2)=2\sum_{j\ge0}(1/5)^{2j+1}/(2j+1)$ and comparing
rational numbers. Also $J>I_*/2$, since $27/16>3/2$. The
upper bound $\kappa<10.167<R_{1|2|3}^{-1}$ gives a fixed margin
in Theorem~\ref{thm:rare} for all large $n$, proving the failure claim.
\end{proof}

\subsection{Bounded patterns}
\label{app:beyond:patterns}

\noindent\textbf{Proposition~\ref{prop:structure}(a).}
For every density $p$, every error pattern on a bounded number of
objects has Chernoff exponent at least $\tfrac12\min_S I^\circ_2(S)$
per moved object, with equality for a swap across a minimizing cut.
Here $I^\circ_2$ is the order-two cycle mutual information of
\citet{cullina2018fundamental}. For Gaussian laws it equals mutual
information across the cut.

\begin{proof}
Use the product base measure over views and rows. Anchor the first
view and let $\pi=(\pi_1,\ldots,\pi_K)$ with $\pi_1=\mathrm{id}$.
Retain the $k$ rows moved in at least one view, a set invariant under
each $\pi_a$. For one coordinate write
$x_i=(x_{ia})_{a=1}^K$, $y_i=(x_{\pi_a(i),a})_{a=1}^K$,
$p_0=\prod_i p(x_i)$, and $p_\pi=\prod_i p(y_i)$. With
$z=\sqrt p$, let
\[
 H_\pi(t)=\int p_0^{1-t}p_\pi^t,\qquad
 C_\pi=-\log\inf_{0<t<1}H_\pi(t).
\]
The midpoint integral has integrand $\prod_i z(x_i)z(y_i)$.
For each row $i$ put $S_i=\{a:\pi_a(i)=i\}$, a nonempty proper
subset of $[K]$. Variables in $S_i$ occur in the same pair
$z(x_i)z(y_i)$ and can be integrated out as one group. The
resulting factor is
\[
 G_i(u,v)=\int z(s,u)z(s,v)\,ds,\qquad
 (u,v)=(x_{i,S_i^c},y_{i,S_i^c}).
\]
This is the kernel of $Z_{S_i}^*Z_{S_i}$, where $Z_{S_i}$ is
the Hilbert--Schmidt operator with kernel $z$ across cut $S_i|S_i^c$.
Thus $\|G_i\|_2^2=\tr[(Z_{S_i}^*Z_{S_i})^2]=e^{-I^\circ_2(S_i)}$.

Every remaining variable occurs in two distinct factors, once in
the first argument of $G_j$ and once in the second argument of
$G_{\pi_a^{-1}(j)}$ for variable $x_{ja}$. Integrate one variable
at a time. If a variable $x$ occurs in factors $F_i,F_j$, then
Cauchy--Schwarz replaces $\int F_iF_j\,dx$ by the upper bound
$(\int F_i^2\,dx)^{1/2}(\int F_j^2\,dx)^{1/2}$, with all
other variables fixed. Each factor retains its full $L^2$ norm
under this replacement. Tonelli applies to the nonnegative
integrands. After integrating all variables, this gives
\[
 C_\pi \ \ge\ -\log H_\pi(1/2)
 \ \ge\ \tfrac12\sum_{i=1}^k I^\circ_2(S_i)
 \ \ge\ \tfrac k2 \min_S I^\circ_2(S).
\]
For a swap of two objects across any cut, exchange the views on
the side not containing the anchor. The permutation is an
involution, so a change of variables gives
$H_\pi(t)=H_\pi(1-t)$. H\"older gives log-convexity, and the
minimum is at $1/2$. Here the midpoint integral is exactly
$\int G(u,v)G(v,u)=\|G\|_2^2$, giving the exponent
$I^\circ_2(S)$ for two moved objects. Minimizing over $S$
achieves the lower bound.

For $N(0,R)$, apply the orthogonal sum-and-difference transform
to the two rows in a swap. If $D_S$ is diagonal with $-1$ on
the exchanged side and $+1$ on the other, then the true and
swapped covariance matrices in these coordinates are
$\Sigma_0=\operatorname{diag}(R,R)$ and
$\Sigma_1=\operatorname{diag}(R,D_SRD_S)$. They have equal
determinant. Their Gaussian midpoint coefficient is
\[
 (\det\Sigma_0\det\Sigma_1)^{1/4}
       \det[(\Sigma_0+\Sigma_1)/2]^{-1/2}.
\]
The average on the difference coordinate is
$(R+D_SRD_S)/2=R_{SS}\oplus R_{S^cS^c}$, so the negative
logarithm of the midpoint coefficient is
\[
 \tfrac12\log\frac{\det R_{SS}\det R_{S^cS^c}}{\det R}
   =I(X_S;X_{S^c}).
\]

For $d$ independent coordinates the midpoint error bound for a
particular candidate is $H_\pi(1/2)^d$. On the event that the
candidate likelihood is at least the true likelihood, the true
density is bounded by their geometric mean. For each fixed $k$ there
are at most $\binom nk(k!)^{K-1}$ candidates. Thus if
$d\min_S I^\circ_2(S)\ge(2+\epsilon)\log n$, a union bound over
$2\le k\le k_0$, for any fixed $k_0$, rules out all such competitors
with probability tending to one.
\end{proof}

\subsection{The Gaussian product reference}
\label{app:beyond:gaussian}

\noindent\textbf{Proposition~\ref{prop:structure}(b).}
For a centered Gaussian $p$ with positive-definite covariance, the
minimizer in Definition~\ref{def:rare} is unique and is a centered
Gaussian block-product density.

\begin{proof}
Take $\beta=1/q$, $\alpha=1-\beta$ and $p=N(0,R)$. For a centered
Gaussian reference with block-diagonal precision
$W=\bigoplus_B W_B\succ0$, let $A(W)=\int p^\alpha v^\beta$.
Writing $C=\alpha R^{-1}$, Gaussian integration gives
\begin{equation}\label{eq:rare:gauss-affinity}
 A(W)^2=(\det R)^{-\alpha}
    \frac{(\det W)^\beta}{\det(C+\beta W)}.
\end{equation}
If $c=\lambda_{\min}(C)>0$, the determinant ratio is bounded
by $\prod_i \lambda_i(W)^\beta/(c+\beta\lambda_i(W))$ because
$C+\beta W\succeq cI+\beta W$. Each factor is bounded and tends
to zero at both ends of $(0,\infty)$. Therefore the nonempty
positive superlevel sets of $A(W)$ are compact in the
finite-dimensional domain, and the maximum is attained at an
interior $W_*$.

Put $S=(C+\beta W_*)^{-1}$. Differentiating $\log A(W)^2$ in
arbitrary symmetric directions in each block gives
\begin{equation}\label{eq:rare:gauss-marginal}
 S_{BB}=W_{*,B}^{-1} \quad\text{for every }B.
\end{equation}
Let $g_B$ be the Gaussian density with precision $W_{*,B}$
and $Q_*=N(0,S)$, the normalized tilt of $p$ against
$g=\bigotimes_B g_B$. Its block marginals equal $g_B$
by~\eqref{eq:rare:gauss-marginal}. For an arbitrary product
density $v$, put $h_B=v_B/g_B$. The arithmetic--geometric mean
inequality gives
\[
 \frac{A(v)}{A(g)}
   =\E_{Q_*}\prod_B h_B(X_B)^{1/q}
   \ \le\ \frac1q\sum_B\E_{Q_*}h_B(X_B)=1 .
\]
Equality requires the $h_B(X_B)$ to agree $Q_*$-almost surely.
The tilted Gaussian has a strictly positive density on the full
product space, so by Fubini these functions of disjoint variables
must be a common constant almost everywhere. Normalization of
the $v_B$ makes the constant one. This proves global optimality
and uniqueness among product densities. The same argument
establishes optimality for any positive-definite $W$ satisfying
\eqref{eq:rare:gauss-marginal}. By~\eqref{eq:rare:kl-identity}
the joint tilt $Q_*$ also attains the KL minimum.
\end{proof}

\subsection{When the new bound is active}
\label{app:beyond:active}

\noindent\textbf{Proposition~\ref{prop:structure}(c).}
For three Gaussian views with nonnegative correlations, and for any
fixed number of views generated by one common Gaussian factor with
positive loadings and independent Gaussian noise, $L=I_*/2$.
There are four-view Gaussian laws with all pairwise correlations
positive for which $L$ is smaller than both $I_*/2$ and $J$.
For every Gaussian law with $0<J\le I_*/2$, some partition satisfies
$R_\cP<J$.

\begin{proof}
For a fixed Gaussian covariance $C\succ0$ and block-diagonal
reference covariance $V\succ0$, equation~\eqref{eq:rare:gauss-affinity}
rewritten using covariance determinants gives
\begin{equation}\label{eq:rare:gauss-det-objective}
 \begin{split}
 2(q-1)R_\cP
 =\min_V\Big\{
     q\log\det[(C+(q-1)V)/q]
       -\log\det C-(q-1)\log\det V\Big\}.
 \end{split}
\end{equation}
Indeed
$\alpha C^{-1}+(1-\alpha)V^{-1}
 = C^{-1}(\alpha V+(1-\alpha)C)V^{-1}$, and
Appendix~\ref{app:beyond:gaussian} justifies minimizing over
Gaussian references. For scalar blocks, write
$V=\operatorname{diag}(v_1,\ldots,v_q)$ and choose $S$ by
including each coordinate independently with probability $1/q$.
The principal-minor expansion gives
\[
 \det[(C+(q-1)V)/q]
    =\E\Big[\det C_{SS}\prod_{i\notin S}v_i\Big],
\]
using $\det C_{\varnothing\varnothing}=1$. Jensen's inequality for
the logarithm cancels the $v_i$ terms in
\eqref{eq:rare:gauss-det-objective}, and yields
\begin{equation}\label{eq:rare:gauss-jensen}
 2(q-1) R_\cP\ge q\,\E\log\det C_{SS}-\log\det C .
\end{equation}

\paragraph{Three nonnegatively correlated views.}
After rescaling the marginal variances to one, let
$m=\min_{i<j}(1-r_{ij}^2)$ and $\Delta=\det R>0$. There is
only one three-block partition. The minimum cut is
$I_*=\tfrac12\log(m/\Delta)$, so it suffices to show
\[
  g=\frac{\det[(R+2V)/3]}{(\det V)^{2/3}}\ \ge\ m^{1/3}
\]
for all positive diagonal $V$. Put $s_i=(2v_i)^{-1}$ and
$s=(s_1s_2s_3)^{1/3}$. The principal-minor expansion and
the arithmetic--geometric mean inequality give
\begin{equation}\label{eq:rare:three-views-lower}
 \begin{split}
 g
 &=\frac4{27s}\left[1+\sum_i s_i+
     \sum_{i<j}(1-r_{ij}^2)s_i s_j+\Delta s^3\right]\\
 &\ge\frac4{27}\big[s^{-1}+3+3ms+\Delta s^2\big].
 \end{split}
\end{equation}
The formula for the determinant is concave in each correlation
separately, so its minimum on the cube $[0,\sqrt{1-m}]^3$
is attained at a vertex. Substituting those vertices gives
\[
 \Delta\ge\min\{2m-1,\ 3m-2+2(1-m)^{3/2}\}.
\]
For $0<m\le1/2$, drop $\Delta s^2$ and minimize
$s^{-1}+3ms$ to get $g\ge(4/27)(3+2\sqrt{3m})$.
The ratio of this lower bound to $m^{1/3}$ is decreasing on
$(0,1/2]$ by differentiation. At $1/2$ the lower bound exceeds
$4/5$, whose cube is greater than $1/2$.

For $1/2\le m\le5/8$, the smaller of the two vertex bounds
is $2m-1$, since the second expression exceeds the first by
$(1-m)(2\sqrt{1-m}-1)\ge0$.
The function $H(m)=8/15+25m/48$ is the upper tangent to
$m^{1/3}$ at $64/125$. For fixed $s$, subtract $H(m)$ from
the lower bound in~\eqref{eq:rare:three-views-lower}
with $\Delta$ replaced by $2m-1$. The result is affine in
$m$. At the left endpoint the lower bound exceeds
$4/5>H(1/2)$. At the right endpoint use
$s^2\ge4s/3-4/9$ and then minimize over $s$. This gives the
lower bound
\[
 (4/27)(26/9+2\sqrt{53/24})>209/243>H(5/8),
\]
where $(35/24)^2<53/24$ gives the first inequality.
Both endpoints hold uniformly in $s$, so $g\ge H(m)$ on
this interval.

For $5/8\le m\le1$, both vertex bounds are at least $m^3$.
The first difference is $(1-m)(m^2+m-1)$ and the second is
$z^3(1-z)^2(z+2)$ with $z=\sqrt{1-m}$. Jensen applied to the
principal-minor expansion for $g$ gives
\[
 \log g \ge \frac{2}{27}\sum_{i<j}\log(1-r_{ij}^2)
                 +\frac1{27}\log\Delta
            \ge \tfrac13\log m .
\]
This proves the claimed bound for all $m$, and hence
$R_{1|2|3}\ge I_*/2$. The argument is strict for $m<1$.

\paragraph{One Gaussian factor.}
Write $X_j=\lambda_j Z+\epsilon_j$, where $Z$ is standard Gaussian
and the $\epsilon_j$ are independent Gaussian residuals with
variances $\Delta_{jj}>0$, independent of $Z$. For any partition,
whiten the residuals within each block and rotate the whitened
loading vector to the first axis. Each block $B_i$ then contains
one scalar with loading $\sqrt{x_i}$, where
$x_i=\sum_{j\in B_i}\lambda_j^2/\Delta_{jj}$, and independent
standard Gaussian noise coordinates. The blockwise projection
onto these scalars and the reverse channel that appends independent
noise both preserve block-product references. Data processing
for R\'enyi divergence in each direction shows that the minimum
in Definition~\ref{def:rare} is unchanged.

With $w_i=\sqrt{x_i}$ the retained covariance is
$C=I+ww^\top$. Let $T=\sum_i x_i$. The matrix determinant lemma
gives $\det C_{SS}=1+\sum_{i\in S}x_i$. Substitute
$\log(1+x)=\int_0^\infty e^{-t}(1-e^{-tx})\,dt/t$
in~\eqref{eq:rare:gauss-jensen}. At each $t>0$ activate each
block independently with probability $u_i=1-e^{-tx_i}$ and let
$N$ be the number activated. Conditional on the activated set,
the probability it meets $S$ is $1-(1-1/q)^N$. Thus the
integrand after factoring out $e^{-t}/t$ is the expectation of
\[
 q[1-(1-1/q)^N]-{\bf1}_{N>0}.
\]
The expression is zero for $N=0,1$. For $2\le N\le q$,
Bonferroni's inequality gives
\[
 q[1-(1-1/q)^N]-1
 \ \ge\ N-1-\binom N2/q\ \ge\ (q-1)N/(2q).
\]
The final difference, multiplied by $2q$, is $(N-2)(q-N)$.
Choose an index $b$ with $x_b$ minimal. The probability that
$b$ is activated and another block is activated is at most
$\E[N{\bf1}_{N\ge2}]/q$. For any $i\ne b$ the difference
between the analogous probability for $i$ and that for $b$ is
$(u_i-u_b)(1-\prod_{j\ne i,b}(1-u_j))\ge0$, so $b$ attains the
smallest of the $q$ probabilities.

The integral of the probability for $b$ is
$\log[(1+x_b)(1+T-x_b)/(1+T)]$ by the same logarithm
identity. It follows that
\[
 2(q-1)R_\cP\ge
 \tfrac{q-1}{2}\log\frac{(1+x_b)(1+T-x_b)}{1+T}
 = (q-1)I(X_{B_b};X_{B_b^c}) \ge (q-1)I_* .
\]
The equality uses the one-factor covariance determinant and the
independence of the discarded noise coordinates. This proves the
claim for every partition. If some residual variance is zero
while the full covariance is positive definite, add independent
noise of variance $\eta>0$ to each view. Data processing bounds
the original $R_\cP$ below by the perturbed $R_\cP$, and the
finitely many cut determinants converge as $\eta\downarrow0$.

\paragraph{Four strictly positively correlated views.}
Consider
\[
 R=\begin{pmatrix}
 1&a&b&b\\ a&1&b&b\\ b&b&1&a\\ b&b&a&1
 \end{pmatrix},\qquad a=1/5,\quad b=59/100.
\]
Its eigenvalues are $1-a,1-a,1+a-2b,1+a+2b$, with minimum
$1/50$. Write $D=\det R$ and $M=\det[(R+3I)/4]$. The
determinants are
\[
 D=(1-a)^2[(1+a)^2-4b^2]=476/15625,\qquad
 M=14663459/16000000.
\]
Every three-view minor is $(1-a)(1+a-2b^2)$. The products of
the block principal minors for a two-block partition are among
this number, $(1-a^2)^2$, and $(1-b^2)^2$. Their minimum is
$p_2=2519/6250$. For three blocks the only nontrivial block
has size two, giving a minimum product $p_3=1-b^2=6519/10000$.
Four singleton blocks have product one. Therefore
\[
 I_*/2=\tfrac14\log(p_2/D),\qquad
 J=\min\{I_*,\tfrac14\log(p_3/D),-\tfrac16\log D\}.
\]
The feasible singleton reference $V=I$ in
\eqref{eq:rare:gauss-det-objective} gives
$R_{1|2|3|4}\le U=\tfrac16\log(M^4/D)$. Comparing the
displayed rational numbers yields
$p_2>9D$, $p_3>9D$, and $M^4<27D<1$. The first two
inequalities and $27D<1$ imply $I_*/2,J>(\log3)/2$, while
$M^4<27D$ implies $U<(\log3)/2$. Thus
$L\le U<\min\{I_*/2,J\}$.

\paragraph{Numerical certificate.}
\nolinkurl{scripts/check_d3_meso.py} verifies
positive definiteness and the comparisons $p_2>9D$, $p_3>9D$,
and $M^4<27D<1$ using fractions. The
\texttt{positive\_certificate} row of
\nolinkurl{d3_meso/rows.csv} records the
corresponding values of $I_*/2$, $J$, and $U$.

\paragraph{A condition forcing a strict improvement over $J$.}
If $0<J\le I_*/2$, a partition attaining $J$ has $q\ge3$
because the two-block information rates are at least
$I_*\ge2J>J$. For a minimizing partition take
$v=\bigotimes_B p_B$. Its log likelihood ratio $\log(p/v)$
under $p$ has mean $D_\cP=(q-1)J>0$. The ratio is nonconstant
because both Gaussian densities are positive everywhere and a
constant ratio would give $p=v$. Applying strict Jensen to the
exponential of $(\alpha-1)\log(p/v)$ gives
$D_\alpha(p\|v)<D(p\|v)$. Hence
$R_\cP\le D_\alpha(p\|v)/(q-1)<J$.
\end{proof}

\subsection{The threshold for three negatively equicorrelated views}
\label{app:beyond:negative}

\noindent\textbf{Theorem~\ref{thm:three}.}
For three Gaussian views with equal correlations $r\in(-1/2,0)$,
$L(R)=R_{1|2|3}(r)$. For any fixed $\epsilon>0$, maximum likelihood recovers
all permutations with probability tending to one if
$d\ge (R_{1|2|3}^{-1}+\epsilon)\log n$. Every estimator has success
probability tending to zero if $d\le (R_{1|2|3}^{-1}-\epsilon)\log n$.
At $r=-1/3$, $R_{1|2|3}\approx0.09835$.

\begin{proof}
Write $R_3=R_{1|2|3}$ and let $\phi_v$ denote the centered scalar
Gaussian density of variance $v$. We first prove sufficiency at
$r=-1/3$, then extend the proof to the whole family.

\paragraph{The tilt and the two-sided test at $r=-1/3$.}
The optimal tilt and its normalizing constant are
\[
 q_3=Z_3^{-1}p^{2/3}(\phi_{v_3}^{\otimes3})^{1/3},
 \qquad v_3=(1+\sqrt3)/3,
 \qquad -\tfrac32\log Z_3=R_3.
\]
Indeed, the precision $2R^{-1}/3+I/(3v_3)$ has inverse with diagonal
$v_3$ and correlations $-(2-\sqrt3)$. Matching its marginals to the
product reference gives optimality by H\"older as in the proof of
Proposition~\ref{prop:structure}(b). The Gaussian normalizer gives
$R_3=\frac12\log(16/27)+\frac34\log[(3+2\sqrt3)/4]$.
We may take $d=\lceil\kappa\log n\rceil$ for
$R_3^{-1}<\kappa<\min\{10.2,R_3^{-1}+\epsilon\}$, using only
this many coordinates in our decoder.

Use the highest likelihood among the alignments for which every row
subset $S$ obeys
\begin{equation}
 \sum_{i\in S}\sum_{t=1}^d \log\frac{q_3(X_{it})}{p(X_{it})}
 \le t_{|S|},\qquad
 t_j=\log\binom nj+\eta j\log n,\qquad t_0=0.
 \label{eq:jneg-test}
\end{equation}
Choose an arbitrary output if no alignment qualifies. On the true table,
the exponential of the statistic for a fixed subset has expectation one.
Markov's inequality and a union bound give a rejection probability at
most $\sum_{j\ge1}n^{-\eta j}=o(1)$ for any fixed $\eta>0$.

Relabel so the true permutations are identities. Represent an incorrect
alignment by a bipartite graph between true and proposed rows. Combine
all views shared by the same two rows into one edge. The anchor edges
give a perfect matching, so a nontrivial connected component has $m$
vertices on each side. Its vertices have degrees two or three. The sums
of degrees agree, so both sides have the same number $k=fm$ of degree-three
vertices. An incorrect qualified alignment that ties or beats the qualified
true table has at least one component with nonnegative score difference.
Its subsets of degree-three vertices satisfy \eqref{eq:jneg-test} on both
sides, since each is a subset of its respective whole table.

For a fixed such component, the event of nonnegative score difference
lets us bound the true density by the geometric mean of the true and
proposed densities. Within the two tests on degree-three subsets we may
multiply that bound by $\exp(t_k)$ times the product of $(p/q_3)^{1/2}$
over those vertices and coordinates. At each degree-three vertex the
resulting factor is $Z_3^{3/2}q_3/\prod_{e}\sqrt{\phi_{v_3}}$.
At a degree-two vertex, the two views on its double-view edge can be
rotated and scaled to a standardized sum and difference. The difference
is independent of the sum and the third view. The two square root
densities for the difference on the same edge integrate to one.
Thus only the standardized sum and the third view remain, with
correlation $\rho=r\sqrt{2/(1+r)}$. Put
\[
 v_2=\sqrt{1-\rho^2},\quad y=\frac{|\rho|}{1+v_2},\quad
 R_2=\log\frac{1+v_2}{2v_2},\quad Z_2=e^{-R_2/2}.
\]
For their density $p_2$, the Gaussian
$q_2=Z_2^{-1}\sqrt{p_2\phi_{v_2}\phi_{v_2}}$ has variances $v_2$ and
correlation $-y$, as follows by inverting the two-dimensional precision.
The factor at such a vertex is $Z_2q_2/\prod_e\sqrt{\phi_{v_2}}$.
Dropping the event from the integral now bounds its probability by
\begin{equation}
 \exp\{t_k-d[(m-k)R_2+2kR_3]\}\,\mathcal I_G^d,\qquad
 \mathcal I_G=\int\prod_{v\in G}
 \frac{q_v(x_{\partial v})}
      {\prod_{e\ni v}\sqrt{\phi_{\nu_v}(x_e)}}\,\prod_e dx_e,
 \label{eq:jneg-branch}
\end{equation}
where $q_v=q_j$ and $\nu_v=v_j$ at a degree-$j$ vertex.

\paragraph{Gaussian integrals with matched edge marginals.}
For the integral in \eqref{eq:jneg-branch}, consider positive messages
proportional to $\sqrt{\phi_s}$ on directed edges. To send a message,
multiply the vertex factor by the incoming messages on the other edges
and integrate over those other variables. In the $\nu_v$ variance units,
an incoming message of variance $s$ adds precision
$H=(\nu_v/s-1)/2$. If the output variable then has variance $T$ in
these units, dividing by the remaining square root marginal density yields
output precision ratio $2/T-1$.

At $r=-1/3$ the box $39/40\le \nu_v/s_{v\to w}\le41/40$ is invariant.
To see this, write $C^0_v$ for the original correlation matrix at $v$.
If the remaining inputs have precision increments $D=\operatorname{diag}(H_i)$
on the set $O$ of edges other than output $e$, then
\[
 T=1-C^0_{eO}(I+D C^0_{OO})^{-1}D C^0_{Oe}.
\]
The closed forms for $v_2,v_3$ give $|H_i|<0.073$ and $H_i>-0.063$.
An input on any degree-three port or on a double-view port has
$H_i\ge-1/80$, by $v_3\ge v_2$ and common degree on double-view ports.
Using $\|C^0_{OO}\|\le1.32$ and $\|C^0_{Oe}\|^2\le2(0.268)^2$
bounds $|T-1|$ by
$0.073\cdot2(0.268)^2/(1-1.32\cdot0.073)<0.0117$.
This places $2/T-1$ in the box. All integrated precisions are positive,
so the update is continuous and Brouwer's theorem gives a fixed point.

At any such fixed point normalize the product of the two messages on
edge $e$ to obtain $b_e$ and the vertex factor times all incoming
messages to obtain $b_v$. The update implies $(b_v)_e=b_e$.
Writing $C_v$ for the resulting vertex correlation matrix, at
$r=-1/3$ we have $|(C_v)_{ij}|<0.278$ at degree three and $<0.330$
at degree two. At degree three put $x=2-\sqrt3$. The original
correlation precision is $aI+b\mathbf1\mathbf1^\top$ with
$a=1/(1+x)$ and $b=x/[(1+x)(1-2x)]$. Eliminating the third coordinate
$l$ after adding the message increments gives the two-variable
precision with diagonals $a+H_i+b'$ and off-diagonal
$b'=b(a+H_l)/(a+H_l+b)$. The absolute correlation is at most
$\bar b/(a-1/80+\bar b)$, where $\bar b$ evaluates $b'$ at $H_l=0.073$.
At degree two the absolute correlation is
$y/\sqrt{[1+(1-y^2)H_i][1+(1-y^2)H_j]}$.
The double-view port has $H_i\ge-1/80$, which yields the stated bound.

Here are two general identities for these fixed points. On multiplying
the identities defining $b_v$ and cancelling the messages against
$b_e$, the integrand equals $Z_{\mathrm B}\prod_v b_v/\prod_e b_e$.
Taking expectations of the logarithms of the local normalizations,
using $(b_v)_e=b_e$ to cancel the message terms, gives
\[
 \log Z_{\mathrm B}=\sum_v\left[-D(b_v\|q_v)+
       \tfrac12\sum_{e\ni v}D(b_e\|\phi_{\nu_v})\right]\le0 .
\]
For the inequality, let $Q,H$ be the local covariances of $q_v,b_v$,
$M=Q^{-1/2}HQ^{-1/2}$, and $u_i=Q^{1/2}e_i/\sqrt{Q_{ii}}$.
With $g_0(t)=(t-1-\log t)/2$ the joint entropy is
$\tr g_0(M)$ and the marginal entropy is
$g_0(u_i^\top M u_i)\le u_i^\top g_0(M)u_i$ by scalar Jensen in an
eigenbasis of $M$. Since $\lambda_{\max}(C_v^0)<2$, we have
$\sum_i u_i u_i^\top\preceq2I$. The sum bound follows from
$g_0(M)\succeq0$.

Standardize by $b_e$ on every edge, let $C=\operatorname{diag}_v C_v$ on
half-edges, let $\mathsf W$ swap the two half-edges of each edge, and let $D_0$
copy an edge variable to its two half-edges. The latter satisfies
$D_0 D_0^\top=I+\mathsf W$ and $D_0^\top D_0=2I$. Whenever $C\prec2I$,
the integral of $\prod_v b_v/\prod_e b_e$ is
\begin{equation}
 [\det C\,\det(D_0^\top C^{-1}D_0-I)]^{-1/2}
   =\det(I-(C-I)\mathsf W)^{-1/2}.
 \label{eq:jneg-bethe}
\end{equation}
The first expression results from integrating a Gaussian of precision
$D_0^\top C^{-1}D_0-I\succ0$. For the second, multiply $I-(C-I)\mathsf W$
by $\mathsf W$, then apply Sylvester's determinant identity to
$D_0D_0^\top-C$ and use $\det\mathsf W=(-1)^{|E(G)|}$.

\paragraph{Counting and charging cycles at $r=-1/3$.}
Choose a fixed even cutoff $\Gamma\ge4$. Given the support and all true
row partitions, insert the anchor blocks first, then the other blocks
in fixed left-label and view order. Mark an inserted edge whose
endpoints were already at distance at most $\Gamma-1$. A marked edge has
$O_{\Gamma}(1)$ possible targets, since degree is at most three. Every
short cycle contains a marked edge, by considering its last insertion.
Each remaining non-anchor block has at most $m$ targets, and there are
$m+k$ such blocks. With $h$ marks and $\theta=\log m/\log n$, the
support choice $\binom nm$ and the $\exp[O_{\Gamma}(m)]$ possibilities
for partitions and mark positions give
\begin{equation}
 \log N(m,k,h)\le (1+f\theta)m\log n-h\log m+O_{\Gamma}(m).
 \label{eq:jneg-count}
\end{equation}
 
Let $B=(C-I)\mathsf W$. It gives a non-backtracking turn followed by crossing
an edge. At the fixed law its absolute row sums are at most $a=0.556$
and its entries at most $b=0.330$. In particular $C\prec2I$, and
$-\log\det(I-B)=\sum_{j\ge1}\tr(B^j)/j$ converges absolutely.
For walks meeting a given mark, shift their origin to one oriented
crossing of it, paying at most $j$ shifts against the factor $1/j$.
From that oriented state, the final two steps of a return have at most
one realization once the first $j-2$ are fixed, since the graph is
simple. Thus each of the two orientations has return weight bounded
by $a^{j-2}b^2$. The graph is bipartite, so the positive lengths are
even and at least four. Including the factor $1/2$ of
\eqref{eq:jneg-bethe}, we obtain
\begin{equation}
 C_{\mathrm{edge}}=\frac{b^2 a^2}{1-a^2}<0.04873,
 \qquad 10.2 C_{\mathrm{edge}}<\tfrac12.
 \label{eq:jneg-fixed-cost}
\end{equation}
Walks avoiding marks have length greater than $\Gamma$. There are at
most $6m$ starting states, so their total cost is at most
$m\delta_{\Gamma}$ with $\delta_{\Gamma}=3a^{\Gamma+1}/(1-a)$. Consequently
$\log\mathcal I_G\le h C_{\mathrm{edge}}+m\delta_{\Gamma}$.

For $m\ge\sqrt n$, we have
$t_k\le[f(1-\theta)+\eta f]m\log n+O(m)$, using
$f\log(1/f)\le1/e$ and $t_0=0$. In the product of
\eqref{eq:jneg-count} and \eqref{eq:jneg-branch}, the cost
$d h C_{\mathrm{edge}}$ is covered by $h\log m$. Since
$R_2>R_3$ and $\kappa R_3>1$, taking $\eta$ small and then $\Gamma$
large leaves a uniform negative multiple of $m\log n$ in the log
probability for each $m,k,h$.

For $m\le\sqrt n$ use the original Bhattacharyya exponent $B_G$.
In units of $\sqrt{1-r}$, the variance in the row-mean direction is
$\delta=(1+2r)/(1-r)$ and the variance perpendicular to it is one.
Let $U,V$ collect the orthonormal mean directions of the true and
candidate rows. Then $M=U^\top V=W/3$, where $W$ records the number
of views on each edge. Each true--candidate principal angle pair
with cosine $\sigma$ contributes the midpoint covariance determinant
ratio $1+(1-\delta)^2(1-\sigma^2)/(4\delta)$, by the determinant of
the two-dimensional midpoint. It follows that, with $A=I-MM^\top$,
\begin{equation}
 B_G=\tfrac12\log\det(I+\xi A),\quad
 \xi=\frac{(1-\delta)^2}{4\delta},\quad
 \tr A=m(4+2f)/9.
 \label{eq:jneg-principal}
\end{equation}
The trace follows because the squared row norm of $M$ is $5/9$ at
degree two and $1/3$ at degree three. At $r=-1/3$, $\xi=9/16$.
Concavity on the spectrum $[0,1]$ yields
\begin{equation}
 B_G\ge m(2+f)\log(25/16)/9,\qquad
 (2/9)\log(25/16)>R_3 .
 \label{eq:jneg-smallfixed}
\end{equation}
The unmarked count in \eqref{eq:jneg-count} is bounded by
$\exp[(1+f/2)m\log n+O(m)]$ here. The fixed margin removes this
range as well. The polynomial number of counts at each $m$ is
absorbed by the uniform negative multiple of $m\log n$.
Thus the qualified decoder succeeds. Under the uniform prior, complete
data ML is Bayes optimal for exact recovery, so it also succeeds.

\paragraph{Certificate for the fixed law.}
The rational enclosures in \path{d1_k3_sharp/rows.csv},
generated by \path{scripts/check_d1_k3_sharp.py},
bound the variance box, both turn magnitudes, the edge charge in
\eqref{eq:jneg-fixed-cost}, and the gaps $R_2-R_3$ and
$(2/9)\log(25/16)-R_3$. The strict inequalities in this part use
the outward bounds.

\paragraph{Parameters throughout the family.}
Write $-x$ for the correlation of $q_3$, so $0<x<1/2$.
The tilted covariance has eigenvalues $v_3(1-2x)$ and $v_3(1+x)$,
the latter twice. Solving $R^{-1}=\frac32\operatorname{Cov}(q_3)^{-1}
-I/(2v_3)$ and imposing unit original variances gives
\begin{align}
 r&=-\frac{3x}{2+x+2x^2},
 &v_3&=\frac{(1+x)(2-x)}{2+x+2x^2},\\
 R_3&=-\tfrac14\log(1-2x)-\log[(1+x)(1-x/2)],
 &\delta&=\frac{(1-2x)(2-x)}{2(1+x)^2}.
 \label{eq:jneg-param}
\end{align}
The formula for $R_3$ follows from the Gaussian normalizer. Marginal
matching again gives optimality. The map $x\mapsto r$ decreases
strictly from zero to $-1/2$. For $s=-r$, put
$f_j(r)=\log[(1-r)^{j-1}(1+(j-1)r)]$. The product reference $N(0,I)$
gives $4R_3\le3f_3(r/3)-f_3(r)$, and
$4(I_*/2)=f_2(r)-f_3(r)$. The difference between the right sides is
$A(s)=f_2(-s)-3f_3(-s/3)$. Direct differentiation gives
$A'(s)/(2s)=-1/[(1+s)(1-s)]+1/[(1+s/3)(1-2s/3)]$.
It is positive up to $s=3/7$ and negative thereafter.
Since $A(0)=0$ and
$A(1/2)=\log[(3/4)/(49/54)^3]>0$, $R_3<I_*/2$ for the entire
open interval. Alternatively, substituting \eqref{eq:jneg-param} into
$I_*=\frac12\log[(1+r)/((1-r)(1+2r))]$ gives the identity
$e^{4(I_*/2-R_3)}=(2-x)^3(1+x)^2(1-x+x^2)(2+x+2x^2)/16$. Its right
side equals one at $x=0$ and has logarithmic derivative
$9x^2(1-x)(1+2x^2)/[(2-x)(1+x)(1-x+x^2)(2+x+2x^2)]>0$ on $(0,1/2)$.
Also $R_3>0$ by \eqref{eq:jneg-param}, or by
$R_3'(x)=9x(1-x)/[2(1-2x)(2-x)(1+x)]>0$.
The scalar cut formulas above give $I_*=-\log v_2$ and
$R_2\ge I_*/2$, since $(1+v_2)/2\ge\sqrt{v_2}$.
These facts prove $L(R)=R_3$ and reduce necessity to
Theorem~\ref{thm:rare}.

\paragraph{Small supports throughout the family.}
For this family put
\begin{gather*}
 L_0=\log(1+\xi),\quad g=L_0+e^{-L_0}-1,\quad
 S=L_0/3+2g/27-R_3,\\
 C_0=\min(R_3,S),\qquad \theta_s=C_0/R_3.
\end{gather*}
where $\xi$ is from \eqref{eq:jneg-principal}. Use the original
simple graph and mark edges as in \eqref{eq:jneg-count}, writing
$h_0$ for this mark count. For a four-cycle $F$ let $w(F)$ be the
product of its edge view multiplicities. Expanding
$81\tr[A(I-A)]=81\tr(MM^\top-(MM^\top)^2)$ gives
\[
 81\tr[A(I-A)]=12m-4\sum_{|F|=4} w(F).
\]
Indeed the diagonal contribution is $(20-2f)m$ from squared row
norms $5$ or $3$ of $W$. The off-diagonal terms with a common
column contribute $(8-2f)m$ from column types $(2,1)$ and $(1,1,1)$.
The terms with two distinct column indices count each four-cycle
four times.

For $m\ge4$, any edge is in four-cycles with total weight at most three.
An edge of multiplicity two gives at most one four-cycle. Its opposite edge
cannot have multiplicity two, since that would close a component with
$m=2$. For a unit edge incident to a degree-two vertex, taking its other
edge forces a double edge and then a unique outgoing unit edge at the next
vertex. This again gives at most one four-cycle, of weight at most two.
If both endpoints have degree three, the total weight is the sum of the
$2$ by $2$ submatrix between their other neighbors. Its row and column
sums are at most two, and equality at total four would saturate all six
vertices, giving a component with $m=3$. Thus
$\tr[A(I-A)]/m\ge (4/27)(1-h_0/m)$, since every four-cycle contains
a mark. For $0\le t\le1$,
$\log(1+\xi t)\ge t L_0+t(1-t)g$. Differentiating the difference
of the first two terms with respect to $\xi$ gives
$t(1-t)\xi/[(1+\xi)(1+\xi t)]\ge t(1-t)\xi/(1+\xi)^2$,
which integrates to the claim. Applying this bound to
\eqref{eq:jneg-principal} gives
\begin{equation}
 B_G/m\ge (2+f)L_0/9+(2/27)(1-h_0/m)g.
 \label{eq:jneg-small}
\end{equation}

For the endpoint comparisons define $F=2L_0/9+4g/81-R_3$ and
$H=R_3/2-2g/27$. Both are strictly positive. To verify the signs put
$u=x/(1-2x)$, $U=1+2u$, $V=1+3u$, $W_0=2+3u$, and
$D=4+15u+18u^2$. Substitution gives
$L_0=2\log D-2\log V-\log W_0-\log 8$,
$R_3=\frac94\log U-\log V-\log(W_0/2)$, and
$g'=L_0'81u^2U^2/D^2$ with derivatives in $u$.
The coefficients, in ascending power of $u$, of
$D^3UVW_0 F'$ and $D^3UVW_0 H'$ are respectively
\begin{gather*}
 (0,0,72,792,7785/2,20115/2,13851,9234,2268),\\
 (0,144,1764,9153,103275/4,170829/4,83349/2,22599,5346).
\end{gather*}
Both functions vanish at zero. At $\theta=1/2$ the difference between
\eqref{eq:jneg-small} and $R_3$ times the counting coefficient in
\eqref{eq:jneg-count} is bounded below by
\[
 (1+f/2)F+
 [2/27-(1+f/2)4/81]g+(h_0/m)H>0.
\]
At $\theta=0$ the Gaussian exponent bound for
Proposition~\ref{prop:structure}(a) gives $B_G/m\ge I_*/2>R_3$.
Its pointwise proof is uniform in $m$. The difference for any fixed graph is affine in $\theta$,
which covers $[0,1/2]$. For $\theta\ge1/2$, the mark term
adds $(h_0/m)(\theta R_3-2g/27)\ge0$. The remaining difference
is affine in $f$: at $f=0$ it is at least $F$ and at $f=1$
it is $S-\theta R_3$. The preceding comparison at
$\theta=1/2$ with $f=1,h_0=0$ also proves $\theta_s\ge1/2$.
This covers $\theta\le\theta_s$. The fixed dimension margin provides
a uniform saving since $B_G/m\ge I_*/2$. For $m=2,3$, the same
bound and the $O(n^m)$ count suffice.

\paragraph{Large supports and the common budget.}
The tilt, subset tests and branch bound
\eqref{eq:jneg-test}--\eqref{eq:jneg-branch} hold for every $x$.
Each double-view edge pairs two degree-two vertices. For $k>0$, each
maximal path of degree-two vertices between degree-three ports contains
$2\ell$ internal vertices, for some $\ell\ge0$. It joins opposite sides,
and all such paths compress to a connected cubic bipartite multigraph
with $k$ vertices per side. The sum of their $\ell$ is $m-k$.
Each parallel group has at most one direct edge, since the original
graph is simple. For $k>3$, no triple edge can occur since its endpoints
would be isolated from the component.

In the core count use arbitrary distinct labels for all $m$ right
vertices. The endpoint type and port colors cost $O(m)$ to record.
Order paths by their left port. The length choices cost $O(m)$ in
logarithm and the labels of internal vertices contribute
$((m-k)!)^2$. Encode each core endpoint by its right label unless the
edge is marked, in which case a bounded-length port sequence determines
it. The mark positions, port sequences and right port colors cost
$O_{\Gamma}(m)$ in logarithm. Divide by $m!$ to remove the right relabeling,
as the anchors identify the original right rows. The number is at most
\[
 \binom nm \exp[O_{\Gamma}(m)] k^{3k-h}((m-k)!)^2/m!,
\]
which gives \eqref{eq:jneg-count} with the core mark count $h$.
All schemes below remove nonbridges, so $h\le k+1$.
When $k\le7$, use the unmarked original count
$\exp[O(m)]n^m m^k$ and $B_G/m\ge I_*/2$ directly.

For $\theta\ge\theta_s$ it is enough to show
\begin{equation}
 \log\mathcal I_G\le h C_0+(m-k)(R_2-R_3)
    +\tau_{\Gamma}m+O_{x,\Gamma}(1),\qquad \tau_{\Gamma}\longrightarrow0.
 \label{eq:jneg-budget}
\end{equation}
Indeed $t_k\le[f(1-\theta)+\eta f]m\log n+O(m)$.
At $d/\log n=1/R_3$, the leading terms in
\eqref{eq:jneg-branch} times the count are nonpositive because
$\theta R_3\ge C_0$. Increasing $d/\log n$ by a fixed positive amount
saves at least that amount times
$[m(1+f)R_3-hC_0]\log n\ge (m-1)R_3\log n$,
before the remainder in \eqref{eq:jneg-budget}.
Choose $\eta$ small and then $\Gamma$ large. The remainder is absorbed
uniformly because $m\ge\sqrt n$. Summing the counts uses a uniform
negative multiple of $m\log n$. This calculation also applies when a
budget holds at equality.

\paragraph{Core bounds from non-backtracking walks.}
Let $H(v,t)$ denote the two-dimensional Gaussian of variance $v$
and correlation $t$ divided by the square roots of both margins.
As kernels in Lebesgue measure,
$H(v,t)H(v,t')=H(v,tt')$: after cancelling the common margin this is
composition of Gaussian conditional transitions. Thus a nonempty path
compresses to $H(v_2,y^{2\ell})$. Direct edges identify the endpoint
variables. For $v_2\le v'\le v_3$, the matched-marginal Gaussian
construction above applies to a mixture of the degree-three factors
and the two-dimensional path factors of variance $v'$.
In the common physical variance box $[v',v_3]$ the incoming precision
at a degree-three vertex satisfies $0\le H\le(v_3/v'-1)/2$.
If $H$ on the right denotes this upper endpoint, its output variance
ratio is between $1-2x^2H/[1+(1-x)H]\ge1/(1+H)$ and $1$.
The displayed lower inequality follows from
$1-2x^2+(1-x-2x^2)H>0$. At a binary vertex with absolute
correlation $t$, the output ratio is $1-t^2H/(1+H)$ for
$(v'/v_3-1)/2\le H\le0$, and lies in $[1,1/(1+H)]$.
Hence both send output variance back into the box. Brouwer's theorem
and the entropy estimate preceding \eqref{eq:jneg-bethe} apply.
If $b=v'/v_3$, the absolute degree-three turns are at most
\[
 P_b=\frac{x+x^2 z}{1-x^2 z},\qquad z=\frac{1-b}{1+b}.
\]
This is the correlation after eliminating the third coordinate at the
largest possible incoming increment on it, which adds covariance
$-x^2 z$ and subtracts $x^2 z$ on the two diagonals. Nonnegative
increments on the other two precisions can only shrink its absolute
correlation. The binary turn magnitude is at most
$t/[1-(1-t^2)(1-b)/2]$ by inverting its updated two-dimensional precision.

The degree-three turn block also has operator norm at most $2x$.
Indeed for nonnegative diagonal increments $D$ its covariance $S_1$
satisfies $S_1\succeq (1-2x)(I+(1-2x)D)^{-1}$ and
$(S_1)_{ii}\le 1/(1+D_{ii})$ in variance units. Its standardized
correlation matrix thus has minimum eigenvalue at least $1-2x$.
Its off-diagonals are negative, so the absolute turn block is
nonnegative symmetric and has norm at most $2x$ by Perron's theorem.
Together with the binary bound, which is less than one for $t<1$,
this gives $C\prec2I$ in \eqref{eq:jneg-bethe}.
Eliminating internal directed path states in the determinant by
Schur complements gives the product of their binary turn magnitudes as
an edge transmission. The eliminated blocks are unit triangular in
$I-B$, because each directed chain is acyclic. All closed core walks
have even length and therefore nonnegative weight after the
degree-three turn signs are removed. In the original uncompressed
integral the number of binary turns on each path is even. Splitting
each edge transmission between its two traversals into and out of a
state, as square roots on the two sides of the turn matrix, is a
diagonal similarity. The nonnegative core matrix $A_0$ and all its
edge-deleted principal submatrices have norm at most $2x$.
With $q_2$ as the binary vertex factor one may also perform this
argument before path compression, using $v'=v_2$ and binary magnitude
$y/[1-(1-y^2)(1-v_2/v_3)/2]$.

For a mark, remove its two orientations. By Jacobi's identity the
ratio of the reduced determinant to the previous determinant equals
the $2$ by $2$ orientation principal determinant of $(I-A_0)^{-1}$.
Its entries are nonnegative, so this ratio is at most the product of
its two diagonal entries. Each diagonal entry is one plus the total
weight of positive-length returns to that orientation.
Thus an upper bound $U_0$ on each of those weights gives a
charge $\log(1+U_0)$ to the half-log correction. After all marks are
deleted and the girth exceeds $\Gamma$, the remaining correction is at
most $3m(2x)^{\Gamma+1}/(1-2x)$ by the trace series.

\paragraph{The interval from zero to $x=0.45069846675$.}
First mark the extra edges of parallel groups, keeping a direct edge
if there is one. Then insert the remaining simple core in fixed order,
marking a short-cycle closing edge as before. Each parallel mark has
one already specified partner and saves an endpoint choice. For a
simple mark the bounded-radius argument gives the same saving.
The two sets of marks leave a connected graph of girth greater than
$\Gamma$. For a simple mark this follows by replacing every marked path
edge of its insertion-prefix witness by its own earlier witness.
Put $u=v_2/v_3$, $P=P_u$, $Q=y/[1-(1-y^2)(1-u)/2]$.
On the unmodified graph, every nonempty path has transmission at most
$Q^2$. The parallel mark has one possible two-step return. For even
length $j\ge4$, the first $j-2$ steps have at most $2^{j-2}$ choices,
the penultimate connection has total transmission at most $1+Q^2$,
and the root transmission is at most $Q^2$. For a simple mark there
are at most three four-step returns, since four would saturate an
isolated $K_{3,3}$. Each three-step terminal multiplicity vector has
squared norm at most $12$: there are at most eight paths, and at most
two coincident endpoints of multiplicity two. The coincidences can
occur only when the two distinct first steps lead to the same
second-step vertex, with at most one shared terminal arc from each
such vertex. Applying the norm bound $2x$ in the middle of longer
returns yields the respective charges
\begin{align}
 C_w&=\log\left[1+Q^2\left(P^2+
            (1+Q^2)\frac{4P^4}{1-4P^2}\right)\right],\\
 C_s&=\log\left[1+3P^4+\frac{12P^6}{1-4x^2}\right].
 \label{eq:jneg-first-cost}
\end{align}
For $0<x\le1/16$, $P,Q\le2x$. This follows from
$y\le6x/5$ and $u>9/10$ in this interval, by substituting
\eqref{eq:jneg-param} and the formula for $v_2$.
Consequently $C_s\le52x^4$ and $C_w\le18x^4$ by
$\log(1+t)\le t$, whereas $R_3\ge x^2$ by integrating
$R_3'(x)\ge2x$. Thus both charges are below $R_3/2\le C_0$.

\paragraph{Certificate for the first core bound.}
On $[1/16,7/20]$, the outward rational enclosures generated by
\path{scripts/check_d1_equicorr_family.py}, in
\path{d1_equicorr_family/rows.csv}, verify
$1-4P^2>0$ and $C_s,C_w<\min(S,R_3)$.
This and the preceding analytic bound verify
\eqref{eq:jneg-budget} for all $x\le7/20$.

For the next segment, normalize each nonempty path from variance $v=v_2$
to $v/a$, $u\le a\le1$. Put $T_\ell=(1-y^{2\ell})/(1+y^{2\ell})$.
Choose new correlation $t'$ so $(1-t')/(1+t')=aT_\ell$. In the
orthonormal sum and difference coordinates the kernel $H(v,t)$ has
precision entries $(1-t)/(2v(1+t))$ and its reciprocal ratio
$(1+t)/(2v(1-t))$, respectively, and prefactor
$[2\pi v(1-t^2)]^{-1/2}$. Hence the replacement leaves difference
precision unchanged, multiplies the prefactor by the inverse of
$(1+T_\ell)/(1+aT_\ell)$, and reduces sum precision by
$(1-a^2)T_\ell/(2v)$.
Let $d_3$ be the diagonal precision of the degree-three vertex factor
$q_3/\prod\sqrt{\phi_{v_3}}$, and $j=1/(2d_3)$, so
\[
 d_3=\frac{1}{v_3}
       \left[\frac{1}{1+x}-\frac12+
                  \frac{x}{(1+x)(1-2x)}\right].
\]
The sum directions of distinct paths are orthogonal, because different
nonempty paths use distinct port variables. For each such sum direction
$e$, the normalized Gaussian precision $P_0$ before restoring its
original term has $e^\top P_0e=d_3+a^2T_\ell/(2v)$. The endpoints
are at distinct degree-three vertices and prior restorations are
orthogonal to $e$. Restore the missing positive rank-one increment.
By the determinant lemma and
$e^\top P_0^{-1}e\ge1/(e^\top P_0 e)$, the resulting log integral
decreases by at least
$\frac12\log[(1+jT_\ell/v)/(1+j a^2 T_\ell/v)]$.
Thus the net per-path charge is at most
\[
 G_\ell(a)=\log\frac{1+T_\ell}{1+aT_\ell}
 -\frac12\log\frac{1+jT_\ell/v}{1+j a^2T_\ell/v}.
\]
For $j/v\le1$ it is at most $\ell G_1(a)$. Indeed, as a function
of $T=T_\ell$, the right side is the integral from $a$ to $1$
of $(1-bz)T/[(1+zT)(1+bz^2T)]\,dz$ with $b=j/v$.
This integrand is increasing and concave in $T\in[0,1]$,
since its derivative has a nonnegative decreasing numerator
$(1-bz)(1-bz^3T^2)$ and positive increasing denominator.
As $T_\ell=\tanh(\ell\log(1/y))$ is increasing and concave,
$G_\ell(a)$ is concave in $\ell$ and vanishes at zero.
Since $T_1=v$, this gives
\[
 G_1(a)=\log\frac{1+v}{1+av}
         -\tfrac12\log\frac{1+j}{1+j a^2}.
\]
Choose $a$ to make $G_1(a)\le\Delta=R_2-R_3$ while
minimizing $a$ in $[u,1]$. With
$H_c=4v^2e^{2R_3}/(1+j)$ and
$a_c=(H_c-1)/(v+\sqrt{v^2+(H_c-1)(v^2-H_c j)})$,
this is $a=\max(u,a_c)$: the equality $G_1(a_c)=\Delta$ solves
$(1+a_cv)^2=H_c(1+j a_c^2)$, and $G_1$ decreases in $a$.
Write $\sigma=\Delta-G_1(a)$. The fitted turns have
$P=P_{u/a}$ and nonempty path transmission at most
\[
 W=\frac{t}{1-(1-t^2)(1-u/a)/2},\qquad t=\frac{1-av}{1+av}.
\]
This follows from the matched two-dimensional bound and
$T_\ell\ge v$. Every marked weak edge (one containing a nonempty
path) uses a distinct pair of degree-two vertices. It can charge
$\sigma$ to that pair in addition to $C_0$.

We detail how to bound the marked simple edges. After removing one
weak edge per parallel group, first remove surviving partners whose
endpoints are still in an exterior four-cycle with the original
group. Such a cycle cannot contain another parallel group: adjacent
groups overfill a vertex, and opposite groups isolate the four
vertices. Each removed partner has both endpoint degrees at most two
in the remaining simple graph. Then repeatedly delete an edge on a
simple cycle of length at most $\Gamma\ge14$. Prefer an edge on no
four-cycle, then one on exactly one four-cycle. If neither can be
chosen, an edge on two four-cycles can be chosen in a $K_{2,3}$ or
a $K_{3,3}$ with one missing edge, as follows. An edge on three
four-cycles gives the incomplete $K_{3,3}$ on its two endpoints and
their four neighbors. Its only two ports to the outside are at the
missing-edge endpoints, and any outside edge there is on no
four-cycle. An edge at a missing-edge endpoint lies on two
four-cycles. If no edge is on three four-cycles, take a component of
the union of the four-cycles. If every vertex in it has degree three,
each edge belongs to exactly two faces when the four-cycles are
viewed as faces, and at each vertex each pair of edges belongs to
one face. The face adjacency graph is connected, so the incidence
boundary map from faces to edges has rank one less than the number
of faces over $\mathbb F_2$. Its image lies in the cycle space.
Writing $E=3V/2$ and $F=E/2$ gives $V-E+F\le2$, hence
$V\le8$. Such a saturated component is excluded by $k>7$.
Thus there is a vertex of degree two in the union. Its two incident
edges belong to two common four-cycles, so its two neighbors have
three common neighbors, forming $K_{2,3}$. Any edge leaving it is
on no four-cycle, as otherwise an incident internal edge is on at
least three. Choose an internal edge at either cubic center.

For each deletion, the other edges of its witnessing cycle give
a port sequence of length at most $\Gamma-1$. This makes the endpoint
encoding in \eqref{eq:jneg-count} valid even with adaptive order:
start with the unmarked edges, and restore any edge whose witness
can already be followed. Reverse deletion order guarantees at least
one such witness until the core is restored. Removed parallel edges
can be decoded from their now restored partners.

For a marked simple edge let $(a_4,b_6)$ be its counts of four-
and six-cycles, let $d_8$ bound its number of eight-step returns,
and let $K_4$ bound the product of its forward and backward
four-step terminal multiplicity norms. A simple bipartite
non-backtracking return of length six must be a cycle, and at
length eight it is an eight-cycle or the concatenation of two
four-cycles. A repeat of a vertex separates two closed portions,
each of length at least four. The norm bound in the middle of all
returns of length at least ten gives
\[
 U_s=a_4P^4+b_6P^6+d_8P^8
             +K_4P^8\frac{4x^2}{1-4x^2}.
\]
For the two chosen graphs with $a_4=2$ use
$(b_6,d_8,d_{10})=(0,6,16)$ and $(2,10,26)$ respectively, with
tail $d_{10}P^{10}+27P^8(4x^2)^2/(1-4x^2)$ in place of the last term.
For these counts the external ports are on no short cycle. An
excursion returning via a different port would form a short cycle
by closing inside the graph along a path of length at most three.
An excursion using the same port takes at least six steps, including
an exterior four-cycle. At length ten it can only be inserted into
one of the two internal root four-cycles. In $K_{2,3}$ there are
at most two exterior ports on either internal four-cycle; in the
incomplete $K_{3,3}$ there is one. Each external four-cycle has at
most four orientations and choices from the two available exterior
arcs, giving $16$ and $8$ exterior returns, respectively. Internal
counts come from iterating directed non-backtracking adjacency.
For a general pair $(a_4,b_6)$ the calculation of $K_4$ and $d_8$
enumerates the binary $2$ by $2$ arm adjacency, shared or private
neighbors of each arm, and remaining central edges subject to degree
three. Four-step terminal multiplicities from third-step
incoming counts $c_{uv}$ at $v$ are
$\sum_w c_{wv}-c_{uv}$; a missing outside neighbor can be counted
by completing its port, with the sum of squares bounded by aligning
the fresh ports of different arms. Eight-step returns concatenate
two three-step prefixes and one central edge, omitting joins that
reverse either adjacent arc. If both prefix ends are fresh,
the contribution of a central edge with prefix counts $c,d$ is
$cd\le(c^2+d^2)/2$. These rules give a finite exhaustive upper
enumeration, allowing every legal local graph and possibly extra
completions. The two $a_4=2$ cases have $K_4\le27$ by taking
the square root of the corresponding joint norm bounds.

For a weak root with its exterior four-cycle present, write
$U_w=WP^2+W(1+W)P^4+W(8+2W^2)P^6/(1-4x^2)$.
The first two terms use its parallel partner and the exterior
four-cycle. In a three-step prefix there are at most eight paths.
Two of them repeat the weak root internally and have the extra factor
$W$. The other six can share at most one terminal state, with
multiplicity two. Indeed after taking the partner followed by
the side port, the next alternatives are at the nearer exterior vertex;
a prefix that first takes the other side port can join these only by
reaching the same exterior vertex in two steps. If its two second
steps instead use an external parallel pair, at most one outgoing
arc from their common endpoint can coincide and that endpoint
cannot be the nearer exterior vertex, since the four core vertices
would then be saturated. This gives squared norm at most
$W(8+2W^2)P^6$ for each orientation. The surviving partner has
one initial step, at most one four- and one six-step return
(the latter via the two spare exterior ports), and four-step squared
terminal norm at most $12P^8$ by the earlier three-step norm bound.
So $U_d=P^4+P^6+12P^8/(1-4x^2)$ charges for that partner.

For a weak root without an exterior four-cycle let the two edges
beyond each of its two side neighbors be of type I if they split
and type II if parallel. In the two-step or four-step returns only
the parallel root pair can be used. In six steps there can also
be a three-step crossing between the two side neighbors, or a
two-edge cycle at one of those neighbors with return to the root.
For I--I the middle crossing contributes at most three edges,
since a fourth would saturate the eight-vertex component. For
I--II there is at most one middle edge with total weight $1+W$
once the choice of the parallel port is included, plus $2W$ for a
side two-cycle. For II--II there is no middle edge for $k>7$,
and at most $4W$ for the two side two-cycles. Set
$N_{\mathrm I}=6+2W^2$, $N_{\mathrm{II}}=3+3W+4W^2$.
The squared three-step norm of an orientation going to a type-I
side is at most $WP^6 N_{\mathrm I}$: the eight terminal states are
distinct, and two paths incur the repeated-root factor.
On the type-II side the parallel edges of weights
$\alpha\le1$, $\beta\le W$ give terminal weights at most
$\alpha\sqrt\beta,\beta\sqrt\alpha,\alpha+\beta$ after extracting
the common factor $\sqrt W P^3$. The other terms have squared sum
at most $2+2W^2$, giving $WP^6 N_{\mathrm{II}}$.
Put $b_{\mathrm{I,I}}=3$, $b_{\mathrm{I,II}}=1+3W$,
$b_{\mathrm{II,II}}=4W$. The resulting budget uses
\[
 U_{ij}=WP^2+W^2P^4+WP^6
 \left[W^2+b_{ij}+\sqrt{N_iN_j}\frac{4x^2}{1-4x^2}\right].
\]
These inequalities count walks before preceding deletions, which
only decrease the nonnegative weights. The needed budgets are
$\log(1+U_s)\le S$,
$\log(1+U_w)+\log(1+U_d)\le2S+\sigma$ and
$\log(1+U_{ij})\le S+\sigma$.

\paragraph{Finite checks for the short-cycle bounds.}
The integer enumeration above is recorded by
\path{scripts/check_d1_cycle6_motifs.py} in
\path{d1_cycle6_motifs/rows.csv} and its
\path{square_free/rows.csv} and \path{capsules/rows.csv} subpaths.
The general table is used for the single-four-cycle rows.
The tables cap $K_4^2$, the product of the two four-step squared norms,
at $36^2$. The cap binds only for three local graphs, all with
$(a_4,b_6)=(0,0)$. In each of them one or both orientations reach four
second-layer vertices, each by one path and each with two unexplored
ports. The script completes unexplored ports by aligning them, which
sends four of these eight unit arrivals into one fresh vertex and gives
squared norm $4^2+4^2+8=40$. A fresh vertex has at most three ports,
and the two ports of one vertex lead to distinct fresh vertices, so the
arrivals split at most as $3+3+2$ and the squared norm is at most
$3^2+3^2+2^2+8=30$. The other orientation has squared norm at most
$36$, so these products are at most $36\cdot30<36^2$.
The interval certificates in
\path{d1_short_cycle_budget/rows.csv} from
\path{scripts/check_d1_short_cycle_budget.py}
check all three budget inequalities, $S<R_3$, $j/v<1$, the
calibration domain, and $W<1$ on $[7/20,0.45069846675]$.
For the final part of the interval, the active I--I gap is
bounded using a negative derivative and its positive lower-endpoint
value at $0.45069846675$; all other gaps are verified by closed
subinterval bounds. Each extra $\sigma$ charge is used only once.
Thus the sum of path and mark charges verifies
\eqref{eq:jneg-budget} on this whole interval.

\paragraph{Electrical comparison for marked edges.}
For the remaining intervals use $a=u$, so the path variance is $v_3$,
the degree-three turn magnitude is exactly $x$, and the nonempty path
transmission is at most $w=(1-v_2^2/v_3)/(1+v_2^2/v_3)$.
We may increase core transmissions in an upper bound, since the
closed-walk series has nonnegative terms. All ordinary core edges
can have transmission one. On $[9/20,239/500]$ keep the attenuation
$w$ on any parallel-group mark. For $x\ge239/500$ we may increase
those transmissions to one as well.

Choose a representative from each parallel group, preferring a direct
edge whenever possible. Insert the representatives, then the edges
without parallel partners, then the extra parallel edges, with fixed
left-port order within each stage. The representatives form a matching
and all are retained. Retain the first edge that forms a cycle. After
that, mark and skip any edge whose endpoints are within distance
$\Gamma-1$ in the retained prefix. This procedure keeps the graph
connected, retains at least one cycle and at most one of length
$\le \Gamma$, and each mark has a bounded-radius target at its insertion.
Encoding each left port's insertion stage costs $O(k)$ bits, so
\eqref{eq:jneg-count} still holds. From either marked endpoint,
a shortest retained path to a cycle followed by repeated laps
lifts to a retained ray in its cavity. Each endpoint retains
one or two edges.
If the mark has a parallel partner, the partner is retained, and
at least one endpoint retains an additional edge, by connectedness
and $k>7$. Hence its endpoint types are $(1,2)$ or $(2,2)$.

Assign arc weight $\alpha_e=x$ to ordinary edges and $\alpha_e=xw$
to attenuated ones. For the unweighted arc non-backtracking matrix
$B=D^\top T-\mathsf W$, with $T,D$ the origin and destination incidence
matrices and $\mathsf W$ arc reversal, put $\mathcal A=\operatorname{diag}(\alpha_e)$.
Factor $I+\mathsf W\mathcal A$ from $I-B\mathcal A$. Using
$(I+\mathsf W\mathcal A)^{-1}=(I-\mathsf W\mathcal A)(I-\mathcal A^2)^{-1}$
and exchanging the rectangular matrix products in the remaining
determinant gives
\begin{equation}
 \det(I-B\mathcal A)=\prod_e(1-\alpha_e^2)\det M_0 .
 \label{eq:jneg-bass}
\end{equation}
Here $(M_0)_{vv}=1+\sum_{e\ni v}\alpha_e^2/(1-\alpha_e^2)$
and edge $uv$ gives $-\alpha_e/(1-\alpha_e^2)$ to the two
off-diagonal entries; parallel contributions add.
Dividing $M_0$ by $x/(1-x^2)$ gives a graph Laplacian plus
diagonal ground conductances. Unit edges have conductance one.
An attenuated edge has conductance $c=w(1-x^2)/(1-x^2 w^2)$
and adds ground conductance $\zeta$ to both endpoints, where
\[
 \mu=x^{-1}+2x-3>0,\qquad
 \zeta=\frac{(1-x)(1-w)}{1+xw}.
\]
Every vertex has the base ground conductance $\mu$. These formulas
follow by subtracting the total incident conductance from the
diagonal entry.

For clarity, the half-log loop correction can be compared with
finite lifts as follows. In an $N$-fold lift give every edge an
independent uniform matching among the $N$ copies of its endpoints.
Revealing a ball of any fixed radius needs a bounded number of
matchings and has collision probability $O(1/N)$ at any root.
There exists a sequence of lifts for which each such ball becomes
tree-like at a $1-o(1)$ fraction of the copies of each edge and
vertex. For a fixed term of the non-backtracking trace expansion
the lifted trace divided by $N$ therefore tends to zero. The
remaining trace tail, divided by $N$, is uniformly bounded using
$2x<1$. Thus the normalized lifted loop correction tends to zero.
Local edge and vertex factors in \eqref{eq:jneg-bass} cancel against
the finite graph factors. The original loop correction is therefore
one-half the limit of the lifted log determinant divided by $N$
minus the original log determinant, with each determinant of
the grounded Laplacian.

Interpolate all marked conductances simultaneously from zero to
their final values as $t$ varies in $[0,1]$, while retaining all
ground conductances and other edges. At $t=0$ this half-log
difference is at most $\tau_{\Gamma}m+O_{x,\Gamma}(1)$ with
$\tau_{\Gamma}\to0$. To see this, approximate $\log$ uniformly
on the common spectral interval $[\mu,\mu+6+\zeta]$ by a polynomial
of degree $j_0$. The diagonal of that polynomial depends only on
the $j_0$-neighborhood, with the vertex ground conductances as local
labels. Taking $\Gamma\ge4j_0+4$ ensures that such balls are tree-like
in the retained graph except within constant distance of its one
possible short cycle. The polynomial diagonals agree on all
tree-like balls, and each exceptional diagonal is bounded.
The uniform approximation can be made arbitrarily accurate, proving
the claim. For each edge of slope $a$ in the interpolation,
the derivative of the half-log difference is $a/2$ times the
difference of the lifted and finite graph effective resistances.
Indeed the derivative of $\log\det P(t)$ in direction
$a(e_u-e_v)(e_u-e_v)^\top$ is $a(e_u-e_v)^\top P(t)^{-1}(e_u-e_v)$.
An effective resistance is bounded by $2/\mu$ in any lift because
$P(t)\succeq\mu I$. The estimates below apply to arbitrarily deep
finite tree neighborhoods by keeping only their edges and their
ground conductances. Kirchhoff reduction or extending a unit flow
to the common ground bounds the resistance of the whole network
by these test subnetworks. Take the lift limit at fixed depth, then
let the depth increase. The bound $2/\mu$ permits passage to the
upper limit inside the $t$ integral.

\paragraph{Weighted resistance bounds on $[9/20,239/500]$.}
At most one attenuated edge touches any vertex. Write
$\operatorname{ser}(c,h)=ch/(c+h)$, with $\operatorname{ser}(0,0)=0$
by continuity. Put $H_*=(1-x)/x$, $\xi_0=3/500$ and $\mu_0=\mu-\xi_0$.
The interval gives $\mu_0>0$, $H_*\ge109/100$, $w\le17/20$
and $c\le4/5$ by direct substitution. For $0\le h\le H_*$,
\begin{equation}
 \zeta+\operatorname{ser}(tc,h+\zeta)
   \ge\operatorname{ser}(t,h)-\xi_0 .
 \label{eq:jneg-cavity}
\end{equation}
In fact $\operatorname{ser}(t,h)-\operatorname{ser}(tc,h+\zeta)$
is nondecreasing in $h$ since $c\le1$. At $H_*$, setting
$V=H_*+\zeta$ and using
$\operatorname{ser}(1,H_*)=1-x$ and
$\operatorname{ser}(c,V)=1-x-\zeta$, the difference less $\zeta$ is
\[
 \frac{\zeta(1-t)(2ct-H_*V)}{(t+H_*)(tc+V)}.
\]
Its positive part is at most
$\zeta(2c-H_*V)_+^2/(8cH_*V)$ by maximizing the quadratic numerator.
If nonzero, $w\ge c\ge H_*^2/2\ge a_*=(109/100)^2/2$.
Now $\zeta\le(11/20)(1-a_*)/(1+(9/20)a_*)$,
$2c-H_*V\le8/5-2a_*$ and $8cH_*V\ge16a_*^2$.
Their ratio bound is strictly less than $\xi_0$, proving
\eqref{eq:jneg-cavity}.

Let $H=\mu_0+2\operatorname{ser}(t,H)$ and
$Y=\mu_0+\operatorname{ser}(t,H)+\operatorname{ser}(1,Y)$
denote the positive solutions. They are at most $H_*$, which is
the solution at $t=1$ with base mass $\mu$. The first recurrence
gives a lower bound on a general cavity by comparing finite-depth
trees starting at the leaf conductance $\mu_0$. If an attenuated
edge enters the cavity, its root has extra ground conductance
$\zeta$ and both outgoing edges are ordinary. If one edge out is
attenuated, its child cavity has extra $\zeta$ and the root's extra
$\zeta$ covers the difference in \eqref{eq:jneg-cavity}, using at most
the mass decrement $\xi_0$. Thus the lower bounds are $H$ for
ordinary incoming edges and $H+\zeta$ for attenuated incoming edges.
Along a retained ray, the distinguished outgoing conductance is
one, so the second recurrence yields $Y$ and $Y+\zeta$, respectively.
The monotone finite-depth iterations converge to the positive
solutions. This follows by increasing depth from the leaf mass,
using the bound $H_*$, and solving the scalar quadratics.

An ordinary root edge endpoint with one retained edge has ground
conductance in its cavity at least $g_1=Y$; with two, it has at least
$g_2=\mu_0+\operatorname{ser}(1,Y)+\operatorname{ser}(1,H)$.
If it touches an attenuated mark, its two other edges are that
mark and its retained partner. The endpoint is a retained leaf, so
deleting it leaves the other end of the partner connected to a cycle.
The ray can therefore continue past the partner without backtracking.
Across the skipped weak edge, any retained ray is available. Thus use
$g_p=\mu_0+\zeta+\operatorname{ser}(1,Y)
+\operatorname{ser}(tc,Y+\zeta)$.
An attenuated root with endpoint retention counts $r_0,s_0$ uses
the bounds $g_{r_0}+\zeta,g_{s_0}+\zeta$.
For cavity lower bounds $g,h$, the lifted half-resistance times
conductance slope $a$ is at most
\[
 \frac{a}{2at+2/(g^{-1}+h^{-1})}.
\]
The path through the common ground is the series of the two
cavities, in parallel with the root edge.

For a finite ordinary root put potentials $a_0,-a_0$ at its
endpoints and $b_0,-b_0$ on the respective nonroot neighbor
sets, with zero elsewhere. Let $p_0$ be the number of root endpoints
incident to a different attenuated mark, and $d_0$ half the sum
of the conductances from the two root endpoints to their other
neighbors. An ordinary mark requires a retained cycle. Extra parallel
edges arrive later, each with its retained partner at distance one,
so each is marked. This gives nonroot conductance $1+t$, $2$ or $1+ct$
at either root endpoint. There are at most $4-p_0$ neighboring
vertices and $8-3p_0$ residual neighbor ports. There are at most
$j_{p_0}$ edges between the two neighbor sets, with
$j_0=3$ and $j_1=j_2=1$. Four cross-edges require two distinct
neighbors on each side, since a repeated neighbor has at most one
unused port. Those four edges would saturate six vertices and isolate
them in the core. Each cross-edge contributes at most
$4b_0^2$, of which $2b_0^2$ is already covered by its two ports.
Ground masses at the neighbors are at most $\mu+\zeta$.
The test energy is then at most
\begin{gather*}
 2\big[(\mu+p_0\zeta/2+2t+d_0)a_0^2-2d_0a_0 b_0+
             (d_0+A_{p_0})b_0^2\big],\\
 A_{p_0}=\frac{(4-p_0)(\mu+\zeta)+8-3p_0+2j_{p_0}}{2}.
\end{gather*}
Optimizing $b_0$ gives a finite graph half-resistance lower bound
$[\mu+p_0\zeta/2+2t+d_0-d_0^2/(d_0+A_{p_0})]^{-1}$,
by dividing the squared root potential difference by the energy.
For an attenuated parallel root each endpoint has one exterior
neighbor, with at most four aggregate spare ports and two cross-edges.
With $d_0=(r_0+s_0-2+(4-r_0-s_0)t)/2$, repeating the test with
the root pair of conductance $1+ct$ gives the lower bound on
half-resistance times slope
\[
 \frac{c}{\mu+\zeta+2(1+ct)+d_0-
                  d_0^2/(d_0+\mu+\zeta+4)} .
\]

\paragraph{Certificate for the weighted comparison.}
The outward rational interval computation
\path{scripts/check_e1_weighted_parallel.py}, with rows
\path{e1_weighted_parallel/rows.csv}, bounds the
integral of the difference of the preceding resistance bounds
for all six unordered ordinary endpoint cases and the two parallel
types. The lifted upper bound decreases with $t$ by conductance
monotonicity. The finite lower bound also decreases, since
$d_0-d_0^2/(d_0+A)=d_0A/(d_0+A)$ increases with $d_0$ and $A$.
Using left ends for the former and right ends for the latter gives
upper Darboux bounds on each closed parameter cell. It certifies
the ordinary charge $C_{\mathrm{ord}}<C_0$ and
\[
 0\le (C_{\parallel}-C_0)_+<R_2-R_3-G_1(u).
\]
The minimum certified gaps for the ordinary and parallel inequalities
are greater than $0.0127629695$ and $0.0002325468$, respectively.
There is at most one parallel mark per group, and each uses a
different pair of degree-two vertices. Hence the budget
\eqref{eq:jneg-budget} holds on the weighted interval.

\paragraph{Unit-conductance comparison on $[239/500,999/2000]$.}
Increase the weak transmissions to one and keep the same retention
rule. Remove $\zeta$ and the mass decrement in the cavity calculation.
Now put $H=\mu+2\operatorname{ser}(t,H)$ and
$Y=\mu+\operatorname{ser}(t,H)+\operatorname{ser}(1,Y)$, and set
$g_1=Y$, $g_2=\mu+\operatorname{ser}(1,Y)+\operatorname{ser}(1,H)$.
They bound cavity conductances by the same finite-depth argument.
For endpoint pair $(r_0,s_0)$ define
\[
 T_{r_0s_0}=[2t+2/(g_{r_0}^{-1}+g_{s_0}^{-1})]^{-1}.
\]
For an ordinary root use
$d=(r_0+s_0+(4-r_0-s_0)t)/2$ in the finite graph test.
Even if a neighboring pair of edges is parallel, at most four
neighbors and eight remaining ports are available. Four cross-edges
would saturate a six-vertex component when the local neighbors are
distinct, and a repeated neighbor has only one residual port.
The upper bound on the energy of those ports is $14b_0^2$.
For a parallel root there are only two outside neighbors.
Use $d'=(r_0+s_0-2+(4-r_0-s_0)t)/2$ and the upper bound
$8b_0^2$ for residual energy. The resulting two finite graph
half-resistance lower bounds are
\begin{align*}
 F_{r_0s_0}
 &=\big[\mu+2t+d-d^2/(2\mu+d+7)\big]^{-1},\\
 F_{r_0s_0}^{\parallel}
 &=\big[\mu+2(1+t)+d'-(d')^2/(\mu+d'+4)\big]^{-1}.
\end{align*}
They follow by optimizing the same test potential as in the weighted
case. The ordinary and parallel charges are the respective integrals
of $T_{r_0s_0}-F_{r_0s_0}$ and
$T_{r_0s_0}-F_{r_0s_0}^{\parallel}$ over $t\in[0,1]$.
Use $G_1(u)$ for the path charge on $[239/500,12/25]$.
For $x\ge12/25$ it suffices to use the prefactor charge
$A_1=\log[(1+v_2)/(1+v_2^2/v_3)]$ without the rank-one decrease.
In fact its general $\ell$-pair version
$A_\ell=-\log[(1+u+(1-u)y^{2\ell})/2]$ is concave in $\ell$
by log-sum-exp and is zero at $\ell=0$.

\paragraph{Certificates for the unit-conductance comparison.}
The outputs \path{e1_pure_trivalent/rows.csv}
from \path{scripts/check_e1_pure_trivalent.py}
verify the ordinary charge is below $C_0$ throughout
$[239/500,999/2000]$. For the parallel charge $C_{\parallel}$,
the same output verifies
$A_1+(C_{\parallel}-C_0)_+<R_2-R_3$ on $[97/200,999/2000]$.
On $[12/25,97/200]$ this inequality follows from
\path{e1_parallel_endpoint/rows.csv},
generated by \path{scripts/check_e1_parallel_endpoint.py}.
On $[239/500,12/25]$ the inequality with $G_1(u)$ in place of $A_1$
is in \path{e1_path_credit/rows.csv},
generated by \path{scripts/check_e1_path_credit.py}.
Its path-only rows certify $G_1(u)<R_2-R_3$ where it is used.
The same Darboux comparison applies: the cavity conductances
increase with $t$ and $\mu$, and each finite denominator increases.
The scripts bound both endpoints in closed $x$ cells with outward
rounding. The condition $j/v_2\le1$ ensuring concavity of the
rank-one-adjusted path charge holds here by
$2v_3 d_3=(1-x+2x^2)/
[(1+x)(1-2x)]\ge3/[5(1-2x)]$ and
$v_2/v_3\ge\sqrt{(1-2x)/3}$, so $2v_2d_3\ge1$ on $[9/20,1/2)$.
It applies in the weighted interval as well.
As before at most $m-k$ parallel marks consume the remaining
per-path slack, giving \eqref{eq:jneg-budget}.

\paragraph{The interval ending at the singular boundary.}
For $x\ge499/1000$, retained rays give a uniform comparison
independent of endpoint type, and the path-cost bound below holds.
The certificate locates the stricter cutoff imposed by the support budget $S$.
For every $t>0$, a side branch with all conductances at least $t$
provides ground conductance at least $t/2$ in the deep-tree limit.
Indeed a unit flow splitting equally down a branch of depth
$D$ has energy at most $2/t+1/(\mu 2^D)$, counting the ground
conductances of the leaves. Along a retained ray, use these
branches at each vertex, giving a cavity conductance at least
$Z$ satisfying $Z=t/2+Z/(1+Z)$, so
$Z=(t/2+\sqrt{t^2/4+2t})/2$. The two rays give lifted
half-resistance at most $1/(Z+2t)$. For the finite graph let
$b=e_u-e_v$ and $M_t=\mu I+\mathcal L_t$, with $\mathcal L_t$
the interpolated Laplacian.
Then $b^\top M_t b\le2\mu+4t+6$: of the four remaining root ports,
two at most use the same parallel partner, contributing at most
four together, and all others contribute at most one each.
Cauchy--Schwarz gives $b^\top M_t^{-1}b\ge4/(b^\top M_t b)$.
The edge charge is accordingly bounded by
\begin{align*}
 C_{\mathrm{end}}
 &=\int_0^1\left[\frac1{Z+2t}-\frac1{\mu+3+2t}\right]dt\\
 &=\tfrac9{10}\log 6-\tfrac12\log2
                      -\tfrac12\log\frac{5+\mu}{3+\mu}.
\end{align*}
The first integral evaluates by $t=2Z^2/(1+Z)$, which maps
$Z\in[0,1]$ to $t\in[0,1]$ and gives integrand
$-1/[2(1+Z)]+9/[2(1+5Z)]$ in $Z$.

The path prefactor cost is at most $\log2$ per degree-two pair.
Writing $e=1-2x$, we have $v_2=(1-r)\sqrt{\delta/(1+r)}
\le(5/4)\sqrt e$ on this interval, and
$(1+x)(1-x/2)>11/10$. Thus
$R_2-R_3\ge-\frac14\log e+\log(11/25)>\log2$,
using $e\le1/500$ and $500(11/25)^4>16$.
Also, $\delta<1/16$ implies
$L_0'=[(1-\delta)/(1+\delta)](-\delta'/\delta)>
(15/17)2/(1-2x)$. With $g'>0$ and
$R_3'<1/[2(1-2x)]$ this yields $S'>0$.
Since $\mu$ decreases, $C_{\mathrm{end}}$ decreases and $R_3$ increases.

\paragraph{Certificate for the terminal comparison.}
\path{d1_equicorr_endpoint/rows.csv} contains the
rational enclosures from \path{check_d1_equicorr_endpoint.py} in
\path{scripts/}. They verify $R_3>C_{\mathrm{end}}$
at $499/1000$ and bracket the unique root of $S-C_{\mathrm{end}}$
in $(0.49948994483,0.49948994485)$.
In particular, for $0.49948994485\le x<1/2$, we have
$C_{\mathrm{end}}\le C_0$. Combining this with the path budget
covers that range.

\paragraph{An additional comparison retaining the ground mass.}
For $x\ge4987/10000$, keep the ground mass $\mu$ in the covering-tree
cavities, using the unit-conductance recurrences for $H$ and $Y$ above.
The lifted half-resistance is at most $1/(Y+2t)$. With the same finite
graph lower bound, the edge charge is
\[
 C_{\mu}=\int_0^1\frac{dt}{Y(\mu,t)+2t}
          -\tfrac12\log\frac{5+\mu}{3+\mu}.
\]
For $e\le13/5000<(22/75)^4$, the preceding path bounds give
$R_2-R_3>\log(3/2)>\log(1+v_2)\ge A_1$.
On $[4987/10000,49949/100000]$, the rows in
\path{e1_negative_endpoint_mass/rows.csv}
from \path{certify_e1_negative_endpoint_mass.py} in
\path{scripts/} certify $C_\mu<\min(R_3,S)$.
In each $x$ cell a left sum on $t_i=(i/768)^2$ bounds the integral
above at the smallest $\mu$, by monotonicity of $Y$ in $t$ and $\mu$.
Use the largest $\mu$ for the subtracted logarithm and the left $x$
endpoint for the increasing budgets $R_3,S$. The right endpoint
$49949/100000>0.49948994485$ gives an additional overlap with the
terminal comparison.

\paragraph{Additional overlapping checks.}
On $[9/20,0.46544977415]$ there is also a short-cycle check
using $a=u$, with turn bound $P=x$ and $W<153/200$.
For the two-splitting root, let $b=0,1,2,3$ be the number
of edges between the two exterior neighbor pairs. In the
eight-step count, grouping the returns by whether they stay in the
root pair, cross the exterior graph once, or return to their own
exterior endpoint gives
\begin{align*}
 U_b={}&Wx^2+W^2x^4+Wx^6(W^2+b)\\
 &+Wx^8[W^3+(1+2W)b+E_b]
      +\frac{Wx^{10}M_b(W)}{1-4x^2}.
\end{align*}
Here $E_b$ bounds the joint count of five-step crossings and
four-step exterior returns. The values for $b=0,1,2,3$ are
$20,22,14,8$, from enumerating the common $2$ by $2$ adjacency
multiplicities subject to the two free stubs at each exterior neighbor.
The $\sqrt W x^5$-scaled five-step terminal norm squared is
bounded by $M_b(W)$. It is computed by grouping the free ports
at common or distinct endpoints and, for any unexplored
neighborhood, grouping residual arrivals in decreasing order in
triples. For a triple of incoming weights $z_i$, the outgoing
squared norm is $(\sum_i z_i)^2+\sum_i z_i^2$. Sorted groups
maximize the sum of the squares of the group totals by majorization.
This yields the last term from the two five-step terminal vectors
and the contraction $2x$ in the intervening steps.

The corresponding integer tables are \path{rows.csv} in
the \path{d1_weak_eight_fork/}, \path{d1_fork_five/} and
\path{d1_weak_eight_parallel/} subdirectories of the ancillary files.
They are produced by \path{check_d1_weak_eight_fork.py},
\path{check_d1_fork_five.py} and \path{check_d1_weak_eight_parallel.py}
in \path{scripts/}.
The parallel-side table additionally counts the eight-step
excursions around a parallel exterior pair and bounds the
four-step norms for the one- or two-parallel-side cases.
The script \path{scripts/check_d1_weak_eight_budget.py}
and its \path{d1_weak_eight_budget/rows.csv} verify
the weak root charges against $S+\sigma$,
and the paired square charge against $2S+\sigma$.
For the simple roots the script
\path{scripts/check_d1_simple_ten.py} records the
counts through ten steps and the five-step norm products in
\path{d1_simple_ten/rows.csv}. This check refines the deletion
rule: among the edges on exactly one four-cycle, one on the fewest
six-cycles is deleted first. Accordingly the script omits a root on
one four-cycle when another edge of its local graph provably lies on a
short cycle and is on no four-cycle, or is on one four-cycle and on
fewer six-cycles. For the remaining such roots it reads the table
\path{d1_cycle6_motifs/single_square/rows.csv}, which
\path{check_d1_cycle6_motifs.py} writes with the option
\path{--single-square-priority}. The covering argument above uses
neither this refinement nor that table. Its interval output
\path{d1_simple_ten/intervals/rows.csv} verifies the
simple charges against $S$, with the active root bracketed between
$0.46544977415$ and $0.46544977424$. These checks overlap the
weighted resistance argument, which already provides the bridge
across this region.

\paragraph{Splicing the intervals.}
The short-cycle bounds cover $0<x\le0.45069846675$, the weighted
resistance comparison covers $9/20\le x\le239/500$, and the
unit-conductance comparison covers $239/500\le x\le999/2000$.
The terminal comparison covers $0.49948994485\le x<1/2$, so
the last two intervals overlap on $[0.49948994485,999/2000]$.
The mass-retaining estimate supplies a separate overlap.
Each comparison by itself supplies \eqref{eq:jneg-budget}, with
its own mark encoding. The small- and large-support estimates
therefore cover all candidates for each fixed $x$. The true
table passes the subset tests with probability $1-o(1)$, so
the qualified estimator succeeds. Full-data ML has at least
its success probability by Bayes optimality. This completes the
fixed-margin sufficiency on $(-1/2,0)$.
\end{proof}

\subsection{Three views with positive equal correlations}\label{app:beyond:positive}

\noindent\textbf{Theorem~\ref{thm:cases}(i).}
For three views with equal correlations $0<r<1$, let
\[
L=R_*:=\frac{I_*}{2}
=-\frac14\log\frac{(1-r)(1+2r)}{1+r}.
\]
If $\liminf d/\log n>1/R_*$, maximum likelihood recovers all
permutations with probability tending to one. If
$\limsup d/\log n<1/R_*$, the success probability of every estimator
tends to zero. The three-block rate is at least $R_*$ by the nonnegative
correlation calculation in Proposition~\ref{prop:structure}(c). The lower side follows from
Theorem~\ref{thm:exact}(b).

\paragraph{Small components for every $r>0$.}
Relabel so the true permutations are identities. In a connected
component with $m$ true rows and $m$ candidate rows, let $W_{ij}$ count
the views shared by the two rows, $k=fm$ be the number of degree-three
vertices on either side, and $h$ be the number of short-cycle marks in
the original simple graph. The two sides have the same degree counts
because their degree sums are equal. The marks are assigned by inserting
the anchor edges first, then the remaining edges in a fixed order,
marking an edge when a path of length at most $\Gamma-1$ already connects
its endpoints. Fix an even $\Gamma\ge4$ and write $\theta=\log m/\log n$.
The support costs at most $(1-\theta)m\log n+O(m)$, row partitions
and mark locations cost $O(m)$, and the $m+k$ nonanchor endpoints
cost at most $\log m$ each, or $O_\Gamma(1)$ at a mark. Thus for fixed
$(m,k,h)$ the logarithmic count is at most
\begin{equation}\label{eq:jpos-count}
m\log n(1+\theta f-\theta h/m)+O_\Gamma(m).
\end{equation}
The unmarked graph stays connected: recursively replace each removed
edge by its earlier path, so $h\le k+1$.

Put $M=W/3$, $S=MM^\top$, $A=I-S$ and
$\delta=(1+2r)/(1-r)$. Since $M$ averages the three restricted
permutation matrices, $0\preceq A\preceq I$. After scaling the transverse
variance to one, the true and false covariances are
$I+(\delta-1)UU^\top$ and $I+(\delta-1)VV^\top$, where the columns
of $U,V$ are the normalized row sums and $U^\top V=M$. At a
principal angle with cosine $\sigma$, the arithmetic mean of
the two covariances has determinant
$\delta+(\delta-1)^2(1-\sigma^2)/4$ on the corresponding plane.
Consequently, writing $\xi=(\delta-1)^2/(4\delta)$, the
one-coordinate Bhattacharyya exponent is
$B_G=\frac12\log\det(I+\xi A)$.

We will use two bounds on the second spectral moment. Direct expansion
gives
\begin{equation}\label{eq:jpos-moments}
 \frac{\tr A}{m}=\frac{4+2f}{9},\qquad
 \frac{\tr(A-A^2)}{m}\ge
 \max\left\{\frac4{27}(1-h/m),\frac{10-6f}{81}\right\}
 \quad(m\ge4).
\end{equation}
Indeed $9S_{ii}$ is five or three according to row degree. In
$81\tr(S-S^2)$, diagonal terms contribute $(20-2f)m$ and the
off-diagonal same-column terms subtract $(8-2f)m$. The remaining
terms subtract four times the sum of products of $W_e$ around
undirected four-cycles. Each such cycle contains a mark and each
edge carries at most three units of this cycle weight. If an edge
has weight two, or has weight one and a degree-two endpoint, the
degree-two endpoint forces the other path around the cycle, and
a cycle of weight four would fill a component of size two. At an
edge connecting two degree-three vertices, the sum is the weight
of the $2\times2$ matrix between the other neighbors on each side.
A weight of four fills those four neighbors and makes a component
of size three. This gives the first second-moment bound. For the
other, the off-diagonal entries $9S_{ij}$ in a degree-two row sum
to four, and no entry is four because this again isolates a size-two
component. Their squared sum is at most ten. In a degree-three
row they sum to six and are at most three. At most one is three. Two entries of three would force both other
rows to be supported on the same three columns. The three rows would
saturate those columns and form an isolated component. The squared sum is at most fourteen.
Subtracting these bounds from $9(9S_{ii})-(9S_{ii})^2$ gives ten
and four, respectively.

Set $\Lambda=\log(1+\xi)$ and $g=\Lambda+e^{-\Lambda}-1$.
For $0\le z\le1$,
$\log(1+\xi z)\ge z\Lambda+z(1-z)g$, as follows by
differentiating the difference in $\xi$ and integrating from zero.
By \eqref{eq:jpos-moments},
\[
 \frac{B_G}{m}\ge\frac{2+f}{9}\Lambda+\frac{2}{27}(1-h/m)g.
\]
The quantities $F=2\Lambda/9+4g/81-R_*$ and $H=R_*/2-2g/27$
are strictly positive. To see this, set $y=r/(1-r)$ and
$U_1=1+y$, $U_2=1+2y$, $U_3=1+3y$, $D_0'=2+3y$. Then
$R_*=\frac14\log(U_1U_2/U_3)$ and
$\Lambda=\log((D_0')^2/(4U_3))$.
Differentiation in $y$ gives
\begin{align*}
4(D_0')^3U_1U_2U_3 F' &= 2y^3(4+12y+7y^2),\\
8(D_0')^3U_1U_2U_3 H' &= 2y(16+96y+192y^2+144y^3+33y^4).
\end{align*}
Both functions are zero at $y=0$. At $\theta=1/2$, the margin over
$R_*(1+f/2-h/(2m))$ is at least
\[
 (1+f/2)F+
 \left[\frac2{27}-(1+f/2)\frac4{81}\right]g+(h/m)H>0.
\]
At $\theta=0$ Proposition~\ref{prop:structure}(a) gives $B_G\ge mR_*$. Interpolating
affinely gives the comparison with \eqref{eq:jpos-count} for
$0\le\theta\le1/2$. For $m=2,3$, use Proposition~\ref{prop:structure}(a) and the count
$O(n^m)$. Since $B_G\ge mR_*$ in all cases, a fixed margin in
$d/\log n$ absorbs $O_\Gamma(m)$ and makes the sum over $m,k,h$ vanish.
This rules out all components with $m\le\sqrt n$.

\paragraph{Large components for $0<r\le9/20$.}
Let $x$ be the off-diagonal correlation in the three-singleton
optimal tilt, and $v_3$ its marginal variance. Write $\rho$ for the
correlation of a singleton and the standardized sum of the other two
views. Their within-block contrast is independent and integrates out.
The stationarity equation for the three-singleton Gaussian tilt,
diagonalized in the sum and contrast directions, gives
\[
 r=\frac{3x}{2-x+2x^2},\qquad
 v_3=\frac{(1-x)(2+x)}{2-x+2x^2},\qquad
 \rho=r\sqrt{\frac2{1+r}}.
\]
In the two-block tilt the marginal variance is
$w=\sqrt{1-\rho^2}$ and the correlation is $e=\rho/(1+w)$, as
is seen by inverting the tilted two-by-two precision
$\frac12 P^{-1}+\frac1{2w}I$. The two-block rate
$R_2=\log[(1+w)/(2w)]$ is at least $R_*=-\frac12\log w$ by
the arithmetic--geometric mean inequality. The same inequality
$R_3\ge R_*$ for the three-singleton rate follows from Proposition~\ref{prop:structure}(c).
From the displayed formulas,
\[
 \beta^2:=(w/v_3)^2
 =1-\frac{6x^2(1-x)}{(2+x)(2+2x+2x^2)} .
\]
The map $x\mapsto r$ is increasing for $0<x<1$, and
$r(2/7)=21/46$. Hence $x\le2/7$, $9/10\le\beta\le1$,
$w\ge4/5$, and $3x/2\le r\le8x/5$.

Apply the subset test \eqref{eq:jneg-test} on both sides of a
candidate, using the positive-correlation three-singleton tilt. By
\eqref{eq:jneg-branch} the rate before the Gaussian integral is
$(m-k)R_2+2kR_3$, and the logarithmic test expense is at most
$(1-\theta+\eta)k\log n+O(m)$. We bound the integral as follows.
Use incoming messages proportional to square roots of Gaussian
densities with variances in $[w,v_3]$. In the intrinsic standardized
coordinates they add precisions $\Delta_i=(v/\nu_i-1)/2$, where $v=w$
or $v_3$ is the intrinsic variance and $\nu_i$ the message variance.
After integrating all ports except the output, write $t$ for its
standardized marginal variance. Its outgoing variance is
$v/(2/t-1)$. At a degree-three vertex the other two $\Delta_i$ are in
$[0,\Delta_0]$, where $\Delta_0=(1/\beta-1)/2$. The smallest $t$ occurs
when both incoming precisions equal $\Delta_0$, and
\[
 1\ge t\ge1-\frac{2x^2\Delta_0}{1+(1+x)\Delta_0}\ge\frac1{1+\Delta_0}.
\]
The last comparison follows from
$1-2x^2+(1+x-2x^2)\Delta_0\ge0$. At a degree-two vertex the incoming
precision increment $\Delta$ lies in $[(\beta-1)/2,0]$, and
$1\le t=1-e^2\Delta/(1+\Delta)\le1/(1+\Delta)$.
Both updates stay in the box. The full precision is positive
because the intrinsic correlations have largest eigenvalues below
two and the added standardized precision is greater than
$-I/2$. Brouwer's theorem gives a fixed point.

After adding only the third degree-three precision increment $\Delta$,
the correlation between the other two ports is $(x-x^2\tau)/(1-x^2\tau)$
with $\tau=\Delta/(1+\Delta)$, and lies in $[0,x]$. Additional nonnegative diagonal
precisions on those ports can only reduce their correlation, by
two-by-two inversion. At degree two the same inversion bounds the
correlation by
\[
 q=\frac{e}{1-(1-e^2)(1-\beta)/2}<\frac{4x}{3},
 \qquad e\le\frac{8\sqrt2}{9}x .
\]
The degree-three absolute turn matrix has norm at most $2x$.
Since the intrinsic correlation eigenvalues are below two, the
normalization in the Gaussian cycle integral satisfies $Z_{\rm B}\le1$
by the entropy argument in Appendix~\ref{app:beyond:negative}.

In a component containing degree-three vertices, compress each
maximal degree-two path. There are an even number of degree-two
vertices on the path, paired by the double-view edges. The resulting
core is a bipartite cubic multigraph on $k$ vertices per side. A
compressed edge with an internal pair has transmission at most $q^2$;
a direct edge has transmission one, and parallel edges contain at
most one direct edge. Eliminating the directed internal path states
from the loop determinant multiplies the local turns by these
transmissions: the eliminated block is unit triangular along the
path. The compressed operator has norm at most $2x$: its factors
are the turn blocks, an edge-routing permutation, and a diagonal
of transmissions bounded by one. A triple parallel class would
saturate both endpoints and hence require $k=1$. Mark all but one
edge in each parallel class, retaining its direct edge if present.
Insert the remaining simple core in order, marking short-cycle
closers and retaining them in the prefix for counting. Then insert
the parallel marks, whose partners are already present. Write $h_w,h_s$ for the
two mark counts and $h=h_w+h_s$. Removing marks preserves
connectivity and leaves girth greater than $\Gamma$, so $h\le k+1$.

For clarity, the compressed-core encoding counts choices of the
right-side row types using $m$ auxiliary right labels. Pairing the
degree-two double-view ports takes at most $(m-k)!$ choices.
Placing the resulting pairs in the $3k$ oriented core paths costs
at most another $(m-k)!$ times a weak-composition count
$\exp(O(m))$. Sending the core paths to right ports costs at most
$\log(3k)$ per unmarked edge and $O_\Gamma(1)$ per mark. Divide by
$m!$ for the auxiliary labels, which can be assigned in $m!$
ways to every anchored candidate. Including the support and the
row partitions yields
$\log N\le m\log n+(k-h)\log m+O_\Gamma(m)$.

Delete weak parallel marks first. When $k>3$ each weak marked edge
has at most one parallel partner. For a fixed orientation, length
two returns cost at most $q^2x^2$. In length $j\ge4$ the first
$j-2$ choices have at most two successors each, the penultimate
return to the start has total transmission at most $1+q^2$,
and the final weak transmission is at most $q^2$. Next delete
marks in the simple core. There are at most three four-step
returns from a given edge: four would use all degrees of a
$K_{3,3}$ component. Three-step paths beginning at a directed
edge are at most eight. Their terminal directed edges are
distinct within either choice of the first step, and at most two
terminal states occur in both choices. For the last assertion,
the two neighbors reached on the first step would have to share
the same next vertex, which then has just one remaining edge
different from those two; there are at most two such common
vertices. Thus the squared three-step row and column norms are
at most $12x^6$. A length $j\ge6$ return has weight at most
$12x^6(2x)^{j-6}$ by Cauchy--Schwarz and the core norm bound.

For either type, the deletion charges half the log determinant
of the two orientation states in the resolvent before deletion.
Its determinant is at most the product of the diagonal entries,
and each diagonal equals one plus the return sum. Summing the
even return lengths and using $\log(1+t)\le t$, we obtain
\begin{align*}
 C_w &\le q^2\left[x^2+(1+q^2)\frac{4x^4}{1-4x^2}\right],\\
 C_s &\le 3x^4+\frac{12x^6}{1-4x^2}
      =\frac{3x^4}{1-4x^2}.
\end{align*}
The residual high-girth integral has log at most $m\varepsilon_\Gamma$,
where $\varepsilon_\Gamma\to0$ by the norm bound. The simple-edge
coefficient $C_s/x^2$ is at most $4/11$, and the weak-edge
coefficient is at most
$(16/9)(4/49)[1+(1+64/441)(16/33)]$. Both are less than $45/116$,
whereas $R_*\ge\rho^2/4\ge (45/58)x^2$. Thus each mark costs
less than $R_*/2$.

For $m\ge\sqrt n$ set $d=\lceil\kappa\log n\rceil$ with fixed
$\kappa>1/R_*$ inside the assumed margin, discarding extra
coordinates if needed. The count, test, and rate give log
probability at most
\[
 (m+k-h\theta+\eta k)\log n
  -d[(m+k-h/2)R_*-m\varepsilon_\Gamma]+O_\Gamma(m).
\]
Here $m+k-h/2\ge m-1$. For $k\le3$ use the unmarked count:
the bounded core norm gives a log integral $O_r(1)$, and if
$k=0$ the degree-two component is a single cycle, also with
bounded log integral. A small fixed $\eta$ followed by a large
fixed $\Gamma$ makes the exponent at most $-c m\log n$ for some
$c>0$, uniformly for large $n$. Summing the count classes and
including the true table's test failure probability gives success
of the feasible likelihood rule. Full-data maximum likelihood
has at least this success probability under the uniform prior.

\paragraph{Large components for $9/20\le r\le7/8$.}
Again use $d=\lceil\kappa\log n\rceil$ with $\kappa>1/R_*$.
For a given true support of size $m\ge\sqrt n$, let $T_0,T_\perp$
be the sums of squares across its $d$ coordinates in the longitudinal
and transverse row subspaces after scaling by $1/\sqrt{1-r}$.
Set $X=T_0/(md\delta)$, $Y=T_\perp/(2md)$,
$\psi(t)=(t-1-\log t)/2$, and $\Psi=\psi(X)+2\psi(Y)$.
Under the true law, $X,Y$ are independent chi-squares divided by
their degrees of freedom $md,2md$. For $U\sim\chi_p^2/p$,
Chernoff bounds on the upper and lower tails give
$\Prob(p\psi(U)>a)\le2e^{-a}$. Thus each term in $md\Psi$ is
stochastically dominated by $\log 2$ plus an independent
unit exponential, and
$\Prob(\Psi>z)\le4(1+mdz)e^{-mdz}$. Put
$z=(1-\theta+\eta)/\kappa$. Taking the union over supports and
over $m\ge\sqrt n$ shows that all such supports pass $\Psi\le z$
with probability tending to one.

Set $B=-\min(9r/10,33/50)$,
$C=\min(13r/50,4/25)$, $b=f B$ and $c=f C$. Multiplying the
true density by $\exp[-(bT_0/\delta+cT_\perp)/2]$ has normalization
$[(1+b)(1+c)^2]^{-md/2}$ and tilts the two variances to
$v_\parallel=\delta/(1+b)$ and $v_\perp=1/(1+c)$. Since the likelihood
ranks candidates by longitudinal square sum, a candidate beats
the truth only if its square sum $T_F$ satisfies $T_F\ge T_0$.
Put $t=v_\parallel/v_\perp$ and $\xi_t=(t-1)^2/(4t)$.
Under the tilted law multiply by $\exp[s(T_F-T_0)/2]$, where
$s=(v_\parallel-v_\perp)/(2v_\parallel v_\perp)>0$.
In a principal-angle plane of cosine $\sigma$, the resulting
normalized precision determinant equals
\[
 1+[s(v_\parallel-v_\perp)-s^2v_\parallel v_\perp](1-\sigma^2)
 =1+\xi_t(1-\sigma^2).
\]
The transverse principal minor is at least $1-sv_\perp>0$, so
the Gaussian integral is finite. The negative log moment of the
product of the two exponential factors, divided by $md$, is
\[
 \tfrac12[\log(1+b)+2\log(1+c)]
 +\tfrac1{2m}\log\det(I+\xi_t A).
\]
On $T_F\ge T_0$ the product is at least
$\exp[-md(bX+2cY)/2]$. Since
$\psi(u)-a'u/2\ge\frac12\log(1-a')$ for $a'<1$, on $\Psi\le z$
we have
$(bX+2cY)/2\le fz-f[\log(1-B)+2\log(1-C)]/2$.
The candidate probability intersected with the support test is
therefore at most $\exp[-md(G_G-fz)]$, with
\begin{align*}
 G_G &= \frac{\log\det(I+\xi_t A)}{2m}+\Phi,\\
 \Phi &= \tfrac12\big\{\log(1+b)+2\log(1+c)
                +f[\log(1-B)+2\log(1-C)]\big\}.
\end{align*}

Write $\ell=(4+2f)/9$, $u=h/m$, $u_0=(1+3f)/6$,
$a=1-4/(27\ell)$ and $a_u=a+[4/(27\ell)]\min(u,u_0)$.
By \eqref{eq:jpos-moments}, $\tr A^2/\tr A\le a_u$.
The function $\log(1+\xi_t v)/v=\int_0^{\xi_t}(1+\omega v)^{-1}d\omega$
is decreasing and convex. Applying Jensen with weights
$\lambda_i/(m\ell)$ to the eigenvalues of $A$ gives
\[
 \frac{1}{2m}\log\det(I+\xi_t A)
 \ge \frac{\ell}{2a_u}\log(1+\xi_t a_u).
\]
The tangent lower bound at $\min(u,u_0)=0$ has slope $-D_t$,
where
\[
 D_t=\frac{2}{27a^2}
 \left[\log(1+\xi_t a)-\frac{\xi_t a}{1+\xi_t a}\right].
\]
Adding $R_*\theta u$, with $\theta\ge1/2$ and $u\le f+1/m$,
therefore gives
\begin{equation}\label{eq:jpos-mid-certificate}
 G_G+R_*\theta u
 \ \,\ge\,\,
 \frac{\ell}{2a}\log(1+\xi_t a)+\Phi
 -\min(f,u_0)(D_t-R_*/2)_+-O_r(1/m).
\end{equation}

\paragraph{Computer-assisted inequality.}
The script \path{scripts/certify_e1_positive_mid.py}
and the table \path{e1_positive_mid/rows.csv} certify
that the explicit expression in \eqref{eq:jpos-mid-certificate},
before the $O_r(1/m)$ term, exceeds $(1+f)R_*$ uniformly on
$[9/20,7/8]\times[0,1]$. The script splits at the breakpoints
$8/13,11/15$ in $r$ and $1/3$ in $f$, then covers each subrectangle
by closed rational rectangles. It checks both possible branches of
the positive part using interval substitution or the mean value
theorem with interval derivative bounds. The table records the
outward lower bound of the gap and the number of rectangles in
each subdomain.

Combining the test-adjusted probability with
\eqref{eq:jpos-count} gives log probability per row, divided by
$\log n$, at most
$1+f-\theta u-\kappa G_G+f\eta+o(1)$.
As $1+f-\theta u\ge1-1/m$, the finite positive dimension
margin gives a uniform negative bound for small enough $\eta$.
The sum over $m,k,h$ tends to zero. A failed support test also has
probability $o(1)$, and the fixed-coordinate estimator transfers
to the full data by optimality of maximum likelihood.

\paragraph{Large correlations $7/8\le r<1$.}
First fit each nonanchor view to the anchor by two-view maximum
likelihood. Set $v=\sqrt{1-r^2}$, $a_0=(1-v)/(1+v)$,
$I_{\rm pair}=-\log v$ and $Q_0=\log[(1+v)/(2v)]$.
The Bhattacharyya exponent of a two-view wrong cycle of length
$\ell\ge2$ is
\[
 E_\ell=\ell Q_0+\log(1-a_0^\ell),\qquad E_2=I_{\rm pair}.
\]
Indeed the two-block tilt gives a $2\ell$-vertex loop with
constant turn $r/(1+v)$. The local normalization is one after
factoring out $e^{-\ell Q_0}$, since the incoming messages
are the intrinsic square-root marginal densities. The two
orientations of the loop each contribute
$1-[r/(1+v)]^{2\ell}=1-a_0^\ell$ to the determinant of
the loop matrix, which yields the formula.

A wrong cycle in a maximizer must individually beat or tie the
true score, since it could otherwise be reverted. At most
$n^\ell/\ell$ such cycles have length $\ell$. Consequently with
$d\ge\kappa\log n$, if $Q_0>R_*$ then the expected number of
errors is at most
\[
 \sum_{\ell\ge2}n^{\ell-\kappa E_\ell}
 \le \frac{n^{2-\kappa I_{\rm pair}}}{1-n^{1-\kappa Q_0}},
\]
using monotonicity of $\log(1-a_0^\ell)$. To check the rates
set $j=\sqrt{1+2r}/(1+r)$, so $I_*=-\log(jv)$.
The inequality $Q_0>R_*$ is equivalent to $4v/(1+v)^2<j$.
For $7/8\le r\le9/10$, the left side is below $22/25$ using
$v(7/8)<97/200$, and $j$ is above $22/25$ using its value
at $9/10$. For $r\ge9/10$, $v(9/10)<11/25$ and
$4(11/25)/(1+11/25)^2<\sqrt3/2\le j$.
Also $I_{\rm pair}>3I_*/4$ is equivalent to $v<j^3$,
which follows from $\sqrt{15}/8<(\sqrt3/2)^3$.

Choose $\max(0,2-\kappa I_{\rm pair})<\theta_0<1/2$.
Markov's inequality puts each true permutation within Hamming
distance $n^{\theta_0}$ of the respective two-view fit with
probability tending to one. Maximize the joint likelihood over
the product of these two Hamming balls, rounded up. Every
candidate in the product ball differs from the truth in at most
$4\lceil n^{\theta_0}\rceil<\sqrt n$ true rows for all large $n$.
The small-component argument excludes any such erroneous candidate
with nonnegative score difference. Thus the constrained rule
succeeds and so does the Bayes-optimal full likelihood rule.

\subsection{Four views with correlation \texorpdfstring{$-1/4$}{-1/4}}\label{app:beyond:four-view}

\noindent\textbf{Theorem~\ref{thm:cases}(ii).}
Let $K=4$ with all off-diagonal correlations equal to $-1/4$, and
let $\rho=R_{1|2|3|4}$. Then $L=\rho$, and with fixed margins
$d/\log n>1/\rho$ gives exact recovery by maximum likelihood,
whereas $d/\log n<1/\rho$ gives vanishing success probability for
every estimator. Here $1/\rho\approx9.8512$, and the exchange
and shared-information conditions together only require
$d/\log n>8.511$ to the stated precision.

\paragraph{Local Gaussian factors.}
Divide each view by $\sqrt{5/4}$. Given a partition with block
sizes $b_i$ and $q$ blocks, keep the normalized sum of each block.
All orthogonal within-block contrasts are independent, standard
normal, and independent of the vector of block sums. The reduced
covariance and precision are
\[
 P_b=I-uu^\top/5,\qquad P_b^{-1}=I+uu^\top,
 \qquad u_i=\sqrt{b_i},\qquad \det P_b=1/5 .
\]
The same coordinate transform applies at both ends of each edge.
The independent contrast densities integrate to one in the tilted
graph integral, and incorporating these contrasts with their
original marginals preserves the reference optimizer. In fact a
Gaussian reference built this way and satisfying tilted marginal
consistency is optimal by Proposition~\ref{prop:structure}(b). For $\beta=1/q$,
$\alpha=1-\beta$ and reference variances $\nu_i$ in the sum
directions, that consistency and the Gaussian integral give
\[
 Q_b^{-1}=\alpha P_b^{-1}+\beta\operatorname{diag}(1/\nu_i),
 \qquad (Q_b)_{ii}=\nu_i.
\]
\begin{equation}\label{eq:jfour-rate}
 R_\pi=\frac{\log\det Q_b^{-1}+\alpha\log(1/5)
                  +\beta\sum_i\log\nu_i}{2\alpha}.
\end{equation}
There are four shapes, denoted $1111,112,13,22$. For a partition
of shape $b$, $D_\pi=\frac12\log[5\prod_i(1-b_i/5)]$, by the
ratio of marginal determinants to the joint determinant.
The two cuts give $I_*/2=\frac14\log(8/5)$, and minimizing
$D_\pi/(q-1)$ over the four shapes gives
$J=\frac16\log(256/125)>I_*/2$. Thus the two older conditions
require the maximum of $2/I_*$ and $1/J$, equal to
$4/\log(8/5)$. The certificate below verifies $\rho<I_*/2$
and $R_{112}>\rho$, so the rare-subpopulation converse proves the
lower side.

For the upper side we apply the large-component criterion
\eqref{eq:jcrit-budget} in Appendix~\ref{app:beyond:criterion}
with $\theta_0=29/50$. The factor at a vertex from the
theory there is $F=q_b/\prod_i\sqrt{\phi_{\nu_i}}$, where
$q_b$ is the density with covariance $Q_b$ and $\phi_\nu$ is
the centered scalar density of variance $\nu$. Its precision is
$\operatorname{diag}(a_i)+\alpha uu^\top$, with
$a_i=\alpha+(\beta-1/2)/\nu_i$. In this paragraph a Gaussian
message has precision $t_i$ in density units (half the
square-root density convention used in
Appendix~\ref{app:beyond:criterion}). A Schur complement gives
the update toward port $i$
\begin{equation}\label{eq:jfour-message}
 t_i'=a_i+\frac{\alpha b_i}{1+\alpha\sum_{j\ne i}b_j/(a_j+t_j)} .
\end{equation}
The message boxes indexed by block size are
$[0.662,0.791]$, $[0.949,1.119]$, and $[1.548,1.582]$.
The certificate checks $a_i>0$ and verifies that the monotone
updates \eqref{eq:jfour-message} map the boxes into themselves.
The first inequality also implies the intrinsic standardized
covariance of $Q_b$ is below $2I$, by writing the precision of
$F$ as $Q_b^{-1}-\frac12\operatorname{diag}(1/\nu_i)$.
After adding all incoming messages, remove all ports except
$i,j$ by a Schur complement. The retained two-port precision has
diagonal part $\operatorname{diag}(a_i+t_i,a_j+t_j)$ plus
$\lambda u_{ij}u_{ij}^\top$, with
$\lambda=\alpha/(1+\alpha\sum_{k\notin\{i,j\}}b_k/(a_k+t_k))$.
Consequently its squared correlation is
\[
 \frac{\lambda^2 b_i b_j}
 {(a_i+t_i+\lambda b_i)(a_j+t_j+\lambda b_j)} .
\]
Increasing $\lambda$ or decreasing the retained diagonal entries
bounds this expression from above, so the certificate uses only
box endpoints.

\paragraph{Small-component endpoint.}
On a component of size $m$, the true and candidate row spaces
have overlap $M=W/4$, where $W_{ij}$ counts shared views. Their
principal-angle determinant gives
$B_G=\frac12\log\det[I+\frac45(I-MM^\top)]$ by the same
two-dimensional calculation as in
Appendix~\ref{app:beyond:positive}, with $\delta=1/5$.
By concavity of the logarithm this is at least
\[
 \sum_{v\,{\rm true}}
 \left[1-\sum_{i\in\pi_v}b_i^2/16\right]\frac{\log(9/5)}2 .
\]
The certificate verifies that each summand is greater than
$[1+\theta_0(q_v-2)]\rho$. This gives the small-component endpoint
of the criterion even without savings from marked edges.

\paragraph{Computer-assisted inequalities.}
The script \path{scripts/check_e2_four_view.py}
and the table \path{e2_four_view/rows.csv} verify
four groups of comparisons. The tilted marginal update
$\nu\mapsto\operatorname{diag}[(\alpha P_b^{-1}
+\beta\operatorname{diag}(1/\nu_i))^{-1}]$ maps the prescribed
rational variance boxes into themselves by monotonicity. A fixed
point follows by continuity. The rate enclosures from
\eqref{eq:jfour-rate} establish binding of $\rho$ and place
$1/\rho$ in $(9.8511971,9.8512024)$ with the endpoints rounded
outward. The message boxes and squared-correlation caps above give
absolute turn row sum at most $0.711$. Inserting the caps into
the two-step endpoint formula \eqref{eq:jcrit-cost} verifies for
each shape
\[
 \frac12\sum_{B\in\pi}(C^1(\pi,B)-\theta_0\rho)_+
 \le \frac{q-1}{2}(R_\pi-\rho).
\]
Finally the small-component margin inequality just stated holds for
each of the four shapes. The log enclosures use the 48-term series
for $2\operatorname{arctanh}t$, $t=(z-1)/(z+1)$, with absolute
error at most $2|t|^{97}/[97(1-t^2)]$. The comparisons are made with
rational endpoints before decimal display. The criterion then
proves the upper side.

\subsection{Three-view Gaussian chains}\label{app:beyond:chains}

\noindent\textbf{Theorem~\ref{thm:cases}(iii).}
For a positive definite three-view Gaussian chain, relabel so that
$r_{13}=r_{12}r_{23}$. Then
$L=I_*/2=\min(I_{12},I_{23})/2$, and the two-view cut threshold
gives exact recovery with fixed margins.

For a Gaussian vector the correlation identity gives conditional
independence of views one and three given view two. Hence its
density is $p_{12}p_{23}/p_2$. We can anchor on view two since
the uniform relative-permutation model is invariant to changing
the anchor. The log likelihood of a proposed alignment is the sum
of the individual edge-$12$ and edge-$23$ log likelihoods, plus
a constant across permutations. Maximizing each edge likelihood
independently therefore gives a joint maximum, whose relative
permutations with respect to the original anchor are obtained by
composition. The terminal cut isolating view one has information
$I_{12}$, and the other terminal cut has information $I_{23}$,
by the chain rule and the conditional independence. The middle
cut has information $I(X_2;X_1,X_3)$ at least as large as both.
Thus $I_*=\min(I_{12},I_{23})$. Flip the signs of either
terminal view as needed to make both path correlations nonnegative.
Their product is the terminal-to-terminal correlation, so all three
are then nonnegative. These invertible changes preserve the rates,
and Proposition~\ref{prop:structure}(c) gives $R_{1|2|3}\ge I_*/2$.

When both path correlations are nonzero, the established two-view
cycle bound gives probability of any incorrect maximum tending to
zero on each edge as soon as $d I_{ij}>(2+\epsilon)\log n$ for
some fixed $\epsilon>0$. In detail, the Bhattacharyya exponent of
a cycle of length $\ell$ is at least $\ell I_{ij}/2$ by the
cycle-information norm bound in Proposition~\ref{prop:structure}(a), and the count of cycles
is at most $n^\ell/\ell$. Every incorrect maximum has a wrong
cycle with nonnegative score difference. Summing these bounds on
both edges proves joint success. Theorem~\ref{thm:exact}(b)
gives failure on the lower side. If either path correlation is
zero, one terminal view is independent of the other two, so its
relative permutation remains uniform and exact success probability
is at most $1/n!$, in agreement with $L=0$.

\subsection{A finite-dimensional sufficient criterion}\label{app:beyond:criterion}

Fix $K$ and a positive definite Gaussian covariance $R$ with $L>0$,
where $L$ is as in Conjecture~\ref{conj:c1prime}. The local
conditions below on block matrices of size at most $K$ imply
success with probability tending to one when
$\liminf d/\log n>1/L$. Together with the converses this gives
the conjectured threshold for such $R$. The criterion holds in a
neighborhood of the identity whenever $I_*>0$, which is
Theorem~\ref{thm:cases}(iv), and at the two negative
equal-correlation examples, $K=3$ with $r=-1/3$ and $K=4$ with
$r=-1/4$. It fails along the nearly singular negative
equal-correlation family.

\paragraph{Local factors and message boxes.}
For each partition $\mathcal P$ into $q\ge2$ blocks use two
sets of parameters, $\alpha^0=1/2$ and $\alpha^1=(q-1)/q$.
In either set write
\[
 V=\prod_{B\in\mathcal P}v_B,\qquad
 Z=\int p^\alpha V^{1-\alpha},\qquad
 Q=Z^{-1}p^\alpha V^{1-\alpha}.
\]
For set 1 take the optimal product Gaussian reference from
Proposition~\ref{prop:structure}(b). For set 0 it suffices to maximize affinity within
centered block-product Gaussians. A maximum within this class is
attained in the positive definite interior. Indeed if $V_0$ is
the block-diagonal reference covariance, the affinity is
\[
 \frac{(det R)^{(1-\alpha)/2}(det V_0)^{\alpha/2}}
 {\det((1-\alpha)R+\alpha V_0)^{1/2}} .
\]
Bounding the denominator by the determinant with $R$ replaced
by $\lambda_{\min}(R)I$, the affinity tends to zero if any
reference eigenvalue tends to zero or infinity. The other
eigenvalue factors each have a finite upper bound. Differentiating
the Gaussian normalization with respect to the block precisions
gives $M_B=S_{BB}$ at the maximum, where $S$ is the covariance
of $Q$ and $M_B$ that of $v_B$. The corresponding rates are
\[
 T^0_{\mathcal P}=-2\log Z^0,\qquad
 T^1_{\mathcal P}=-q\log Z^1=(q-1)R_{\mathcal P}.
\]
For two-block partitions the two rates agree and are at least
$I_*/2$ by the two-block comparison in Appendix~\ref{app:beyond}.

Each possible block $B$ has a single coordinate system used
at both endpoints, for example coordinates whitening $R_{BB}$.
In either parameter set the local factor is
$F=Q/\prod_B\sqrt{v_B}$. A message along a block-$B$ edge
is proportional to a Gaussian density square root with precision
$W_B$. Choose a common Loewner interval
$[W_B^-,W_B^+]$ containing the intrinsic precisions $M_B^{-1}$
for every partition with that block. Define
$\Delta_D=(W_D-M_D^{-1})/2$ and let $H$ be the covariance
with all incoming messages,
$H^{-1}=S^{-1}+\operatorname{diag}_D\Delta_D$.
Let $H_{\setminus B}$ denote the covariance with $\Delta_B$ set
to zero. The checks for each set of parameters are
\begin{enumerate}[leftmargin=*]
\item $H^{-1}$ and each $H_{\setminus B}^{-1}$ are positive
definite at the lower endpoints.
\item For an output block $B$, the two values of
$2[(H_{\setminus B})_{BB}]^{-1}-M_B^{-1}$ at the remaining
lower and upper endpoints both lie in $[W_B^-,W_B^+]$.
\item The intrinsic block-standardized covariance, formed by
whitening each marginal of $S$ with its $M_B$, is at most $2I$.
\end{enumerate}
The expression in check 2 is obtained by integrating all other
blocks in the local factor times their incoming messages. It is
monotone in each incoming precision by the order-reversing
property of inversion. Therefore every finite candidate graph
has a continuous message update from a compact convex box into
itself, and Brouwer's theorem yields a fixed point. Normalize
$F$ times the incoming messages at vertex $v$ to a density
$b_v$ and the product of the two opposite messages on edge $e$
to $b_e$. The update equations make the corresponding marginals
of $b_v$ equal to $b_e$.

For any $B\ne D$ in a partition, let
$b(\mathcal P;B,D)=b(\mathcal P;D,B)$ be an upper bound on the
operator norm of the $BD$ block after whitening the covariance
$H$ by its block marginals. This bound can be checked by requiring
\[
 \begin{pmatrix}
 b(\mathcal P;B,D) H_{BB} & H_{BD} \\
 H_{DB} & b(\mathcal P;B,D) H_{DD}
 \end{pmatrix}\succeq0
\]
throughout the message box. A useful coarser bound, if
$\ell I\preceq H^{-1}\preceq u I$, is
\begin{equation}\label{eq:jcrit-bound}
 b(\mathcal P;B,D)=u\|\operatorname{offblock}(S^{-1})\|/\ell^2
\end{equation}
when the right side is used as the chosen cap. To see this,
subtract from $H$ the inverse of the block-diagonal part of
$H^{-1}$. The difference is the product of $-H$, the
off-block precision, and that block-diagonal inverse. Its norm
is at most $\ell^{-2}\|\operatorname{offblock}(S^{-1})\|$.
The left and right marginal whitening factors increase the bound
by at most $u$.

For set $j=0,1$, compute the following quantities from its caps:
\begin{align}
 a&=\max_{\mathcal P,B}\sum_{D\ne B}b(\mathcal P;B,D),&
 W(B)&=\max_{\mathcal P\ni B}\sum_{D\ne B}b(\mathcal P;B,D)^2,
 \notag\\
 D(\mathcal P,B)&=\sum_{D\ne B}b(\mathcal P;B,D)^2 W(D),&
 C^j(\mathcal P,B)&=s_B\log\!\left(1+
                  \frac{D(\mathcal P,B)}{1-a^2}\right).
 \label{eq:jcrit-cost}
\end{align}
Here $s_B$ is the dimension carried on the edge. If a block
splits into a correlated direction and independent contrasts
with the same marginals at both ends, the contrasts integrate to
one and may be omitted before applying these checks. Require
$a<1$ for each set used and choose $\theta_0\in(0,1)$ such that
for all partitions,
\begin{equation}\label{eq:jcrit-budget}
\begin{aligned}
 T^0_{\mathcal P}&\ge L[1+\theta_0(q-2)]
        +\sum_{B\in\mathcal P}(C^0(\mathcal P,B)-\theta_0 L)_+,\\
 T^1_{\mathcal P}&\ge (q-1)L
        +\sum_{B\in\mathcal P}(C^1(\mathcal P,B)-\theta_0 L)_+ .
\end{aligned}
\end{equation}
Set 0 and its inequality may be replaced by a direct
Bhattacharyya lower bound as described below. There are finitely
many partitions and blocks. The covariance checks and caps can
be certified with finite interval bounds on fixed-size matrices,
for example using the monotone endpoint checks, the explicit cap
\eqref{eq:jcrit-bound}, or interval Schur complements.

\paragraph{The Gaussian loop integral.}
Relabel the true permutations to identities. In a connected
candidate component, merge the views shared by each pair of
true and false rows into one edge. This is a simple bipartite
graph, with $m$ rows on each side and every degree between two
and $K$. Edges carry their view blocks. Let $\mathcal I_G$ be
the integral of the product of its local factors, with one
variable per edge. Denote by $C_v$ the block-standardized
covariance of $b_v$, by $C$ its direct sum over vertices, and
by $\mathsf W$ the swap of the two half-edges of each edge. We claim
\begin{equation}\label{eq:jcrit-loop}
 \mathcal I_G=Z_{\rm B}\det[I-(C-I)\mathsf W]^{-1/2},
 \qquad Z_{\rm B}\le1.
\end{equation}

The fixed point cancels messages between the product of beliefs
and the product of factors:
$\prod_v F_v=Z_{\rm B}\prod_v b_v/\prod_e b_e$.
Taking logarithms of the normalization constants and
expectations under the consistent beliefs gives
\[
 \log Z_{\rm B}=
 \sum_v\left[-D(b_v\|Q_v)
              +\tfrac12\sum_{B\in\mathcal P_v}
                     D((b_v)_B\|v_B)\right].
\]
In fact each incoming message appears in one vertex normalizer
and in one edge normalizer and has the same expectation in both.
To bound the displayed sum, let
$A=S^{-1/2}\operatorname{Cov}(b_v)S^{-1/2}$.
The block covariance normalized relative to $M_B$ is
$P_B^\top A P_B$, where $P_B$ is the isometric embedding
$S^{1/2}E_B M_B^{-1/2}$ and $E_B$ inserts the coordinates of
block $B$. With $g(t)=(t-1-\log t)/2$, Jensen's inequality
applied to each eigenvector of this compression gives
$\tr g(P_B^\top A P_B)\le\tr(P_B^\top g(A)P_B)$.
Check 3 says $\sum_B P_BP_B^\top\preceq2I$ because this sum
has the same spectrum as the block-standardized $S$.
Thus the block entropies sum to at most twice the joint
entropy and $Z_{\rm B}\le1$.

Whiten the common edge marginals $b_e$. Write $D_E$ for
duplication of every edge variable onto its half-edges, so
$D_ED_E^\top=I+\mathsf W$ and $D_E^\top D_E=2I$.
The block norm caps give $\|C-I\|\le a<1$, which makes
$D_E^\top C^{-1}D_E-I$ positive definite. Direct Gaussian
integration gives the inverse square root of
$\det C\det(D_E^\top C^{-1}D_E-I)$. The determinant identity
$\det(I+XY)=\det(I+YX)$ and $D_ED_E^\top=I+\mathsf W$ identify this
product with $\det[I-(C-I)\mathsf W]$, proving \eqref{eq:jcrit-loop}.

\paragraph{Cost of marked edges.}
After inserting anchor edges, add nonanchor edges in a fixed
order and mark any edge whose endpoints are then connected by
a path of length at most $\Gamma-1$, with $\Gamma\ge4$ even. Write $h$
for the number of marks, $U=|E|-m$, and
$\theta=\log m/\log n$. With the support fixed and the true
row partitions given there are $U$ nonanchor endpoints to send.
Each has at most $m$ choices, or $O_{K,\Gamma}(1)$ at a marked edge.
Including the support, partitions and mark locations gives
\begin{equation}\label{eq:jcrit-count}
 \log N(m,U,h)\le[m+\theta(U-m-h)]\log n+O_{K,\Gamma}(m).
\end{equation}
Deleting all the marks leaves girth greater than $\Gamma$, since
the last edge of a surviving short cycle would have been marked.
The remaining graph is connected, because each mark has an
earlier alternate path and earlier marks on that path can
recursively be replaced. Thus $h\le U-m+1$.

To bound the log determinant in \eqref{eq:jcrit-loop}, replace
the block norm of each turn by its scalar bound. The resulting
nonnegative directed nonbacktracking matrix $A_+$ is a symmetric
local turn matrix times a permutation, so $\|A_+\|\le a$.
For any oriented edge $u\to v$ carrying $B$, two-step
paths lead to distinct terminal oriented edges. The starting
edge's destination and the terminal edge's origin uniquely
determine the intermediate edge, since the graph is simple.
The squared two-step row norm is therefore at most
$D(\mathcal P_v,B)$ and the squared column norm at most
$D(\mathcal P_u,B)$. Cauchy--Schwarz bounds the return at
length $j\ge4$ by
$a^{j-4}\sqrt{D(\mathcal P_u,B)D(\mathcal P_v,B)}$.
Returns can occur only at even $j\ge4$, and all bounds
continue to hold after deleting edge states.

Charge a closed walk in the log determinant expansion to the
first marked edge it visits in a fixed deletion order. Rotate
its starting point to one of the two states of that edge.
The absolute block trace is at most $s_B$ times the product of
scalar turn bounds. The scalar sum for walks charged to that
edge is exactly the reduction in $-\tfrac12\log\det(I-A_+)$
when the two states are removed from the current principal
submatrix. Jacobi's identity expresses the reduction as half
the log determinant of the corresponding two-state submatrix
of the resolvent. Its determinant is bounded above by the
product of its diagonal entries, since the Neumann series is
entrywise nonnegative. Either diagonal is at most
$1+\sqrt{D(\mathcal P_u,B)D(\mathcal P_v,B)}/(1-a^2)$.
The inequality $1+\sqrt{xy}\le\sqrt{(1+x)(1+y)}$ then bounds
the charge by $[C^j(\mathcal P_u,B)+C^j(\mathcal P_v,B)]/2$.
The uncharged walks have length greater than $\Gamma$ and their
trace series is bounded by $O_K(m)\sum_{j>\Gamma}a^j$.
It follows that
\begin{equation}\label{eq:jcrit-integral}
 \log\mathcal I_G\le
 \frac12\sum_{e=uv\,{\rm marked}}
       [C^j(\mathcal P_u,B_e)+C^j(\mathcal P_v,B_e)]
 +m\varepsilon_\Gamma,\qquad \varepsilon_\Gamma\longrightarrow0 .
\end{equation}

\paragraph{Combining the two size ranges.}
For set 0 the square root of the original density at a vertex
is $Z^0F$ and its negative log constant is $T^0_{\mathcal P}/2$.
Thus \eqref{eq:jcrit-budget} and \eqref{eq:jcrit-integral}
give the exponent
$B_G\ge L[m+\theta_0(U-m-h)]-m\varepsilon_\Gamma$
at the size endpoint $\theta_0$.
This same bound from the original covariance determinants
may be used in place of set 0. At the size endpoint zero,
Proposition~\ref{prop:structure}(a) gives $B_G\ge m I_*/2\ge mL$.
Interpolate the two lower bounds for the same component to
obtain $B_G\ge L[m+\theta(U-m-h)]-m\varepsilon_\Gamma$ for
$0\le\theta\le\theta_0$.
Combining this with \eqref{eq:jcrit-count} excludes all these
components with fixed dimension margin. Explicitly, use
$d'=\lceil\kappa\log n\rceil$ coordinates for a fixed
$\kappa>1/L$ in the margin, and reserve $d'-L^{-1}\log n$
coordinates at rate at least $mL$. A sufficiently large fixed
$\Gamma$ absorbs the $m\varepsilon_\Gamma$ term and gives a negative
exponent of order $m\log n$.

For the other sizes, accept a table only if for every partition
$\mathcal P$ with $q\ge3$ and every subset of $s$ rows
the sum of $\log(Q^1_{\mathcal P}/p)$ over those rows and
all coordinates is at most
$t_s=\log\binom ns+\eta s\log n$. The true table passes
all tests with probability tending to one: the exponential
moment for a fixed test is one, and after summing over the
subsets the failure bound is a finite multiple of
$\sum_{s\ge1}n^{-\eta s}$.
A feasible competitor of at least true likelihood has a
component of nonnegative score difference. Group its rows
on each side by their respective partitions. The geometric
mean of the true and candidate densities dominates the true
density on that score event. For a group of size $s$ with
$q$ blocks, the test permits inserting into the geometric
mean $(p/Q^1)^{(q-2)/2}$ at each vertex at cost
$(q-2)t_s/2$ in the exponent. The resulting vertex integrand is
\[
 p^{1/2}(p/Q^1)^{(q-2)/2}=(Z^1)^{q/2}F.
\]
On each side $\sum_v(q_v-2)=U-m$. The sum of the group
expenses is at most $(1-\theta+\eta)(U-m)\log n+O_K(m)$:
use $\log\binom ns\le s\log(n/m)+s\log(m/s)+s$ and
$s\log(m/s)\le m/e$. With \eqref{eq:jcrit-count}, the
count and test expenses are thus at most
$(U-\theta h+\eta(U-m))\log n+O_{K,\Gamma}(m)$.
Meanwhile $-\log (Z^1)^{q/2}=T^1_{\mathcal P}/2$.
Combining the second inequality in \eqref{eq:jcrit-budget}
with \eqref{eq:jcrit-integral} gives a one-coordinate exponent
at least $L(U-\theta_0h)-m\varepsilon_\Gamma$.

For $\theta\ge\theta_0$ the count is paid at the threshold
apart from the test slack and long-cycle tail, and
$U-\theta_0h\ge m-1$. The fixed dimension margin absorbs
these terms for sufficiently small fixed $\eta$ and sufficiently
large fixed $\Gamma$. Summing over $m,U,h$ gives vanishing probability
of a feasible winning competitor. The feasible maximum-likelihood
rule therefore succeeds, and maximum likelihood on the full data
has at least its success probability under the uniform prior.

\paragraph{Weak correlations.}
Standardize the view variances to one and set
$\gamma=\|R-I\|_{\rm op}$. For each partition, after whitening
its blocks, write the covariance as $I+E_{\mathcal P}$, where
$E_{\mathcal P}$ has zero block diagonal and norm $O_K(\gamma)$.
For either parameter set,
\begin{align*}
 M_B&=I+O_K(\|E_{\mathcal P}\|^2),\\
 S&=I+\alpha E_{\mathcal P}+O_K(\|E_{\mathcal P}\|^2),\\
 -\log Z&=\frac{\alpha(1-\alpha)}4
       \|E_{\mathcal P}\|_F^2(1+O_K(\gamma)).
\end{align*}
For the first two formulas, the block consistency equation is
$S_{BB}-M_B=0$, where
$S^{-1}=\alpha(I+E_{\mathcal P})^{-1}
 +(1-\alpha)\operatorname{diag}(M_B^{-1})$.
Its derivative in the block-diagonal unknown at the identity
is $-\alpha I$, and its derivative in the off-block input
vanishes on the diagonal blocks. The implicit function
theorem gives the local branch. The Gaussian affinity
superlevel sets are uniformly compact near the identity
by the eigenvalue bound above. Any sequence of maximizers
as $E_{\mathcal P}\to0$ therefore limits to the unique
independence optimizer and is eventually on this branch.
In the log affinity formula the linear block-diagonal terms
cancel. The quadratic log-determinant expansion in the
off-block perturbation gives the third formula; its error
is relative $O_K(\gamma)$.

Use common precision boxes
$[1-A_K\gamma^2,1+A_K\gamma^2]I$ with a sufficiently large
fixed $A_K$. Their full and partial precisions are positive
definite. With intrinsic inputs the output equals its intrinsic
precision. Perturbing the other blocks of the incoming
precision by $O_K(\gamma^2)$ changes the output Schur complement
by $O_K(\gamma^4)$, since the perturbation is between two
off-diagonal blocks of norm $O_K(\gamma)$.
This proves box invariance for small enough $\gamma$.
The intrinsic standardized covariances are $I+O_K(\gamma)$,
and \eqref{eq:jcrit-bound} gives
$b(\mathcal P;B,D)=O_K(\|E_{\mathcal P}\|)$, with $a<1$.
Consequently
$C^j(\mathcal P,B)=O_K(\gamma^2)T^j_{\mathcal P}$,
a relative bound that remains valid for exceptionally weak
cuts.

Let $c_*$ be the least sum of squared correlations across
a cut, and $c_{\mathcal P}$ the sum across blocks of the
partition, counting each cut pair once. Isolating each block
gives $2c_{\mathcal P}\ge q c_*$. From the expansion above
and the Gaussian cut determinant,
$T^0_{\mathcal P}=(1+O_K(\gamma))c_{\mathcal P}/4$ and
$I_*/2=(1+O_K(\gamma))c_*/4$.
Hence for $q\ge3$,
$T^0_{\mathcal P}\ge(1-O_K(\gamma))q L/2$.
For $q=2$ it is at least $L$ exactly. Take $\theta_0=1/4$.
Both no-loop budget terms in \eqref{eq:jcrit-budget}
are at most their corresponding $T=T^j_{\mathcal P}$,
using the preceding bounds for set 0 and the definition of
$L$ for set 1. If $T\le2KL$, every $C^j$ is below
$\theta_0 L$ for small $\gamma$. If $T>2KL$, the no-loop
term is at most $KL<T/2$, and the sum of the $C^j$ is
$O_K(\gamma^2)T<T/2$. This proves the criterion uniformly
over the strength of the least cut.
If $I_*>0$ then $L>0$ as required: zero rare rate would,
by the attained optimizer and the equality case of the
affinity bound, make $p$ a product across a nontrivial cut.

\paragraph{Negative examples.}
For equicorrelation, write $\delta=(1+(K-1)r)/(1-r)$ and
$\xi=(1-\delta)^2/(4\delta)$. The principal-angle formula
has $M=W/K$. Concavity gives a lower bound for $B_G$ that
sums over true rows the quantity
\[
 b_{\mathcal P}
 =\tfrac12\log(1+\xi)
     \left[1-\sum_{B\in\mathcal P}(|B|/K)^2\right].
\]
It is enough at the small-component endpoint to verify
$b_{\mathcal P}\ge L[1+\theta_0(q-2)]$ for every partition.
The checks in Appendix~\ref{app:beyond:four-view} give
this and the large-component bound for $K=4,r=-1/4$.

For $K=3,r=-1/3$ take $\theta_0=1/2$. The two possible
values of $b_{\mathcal P}$ are
$(2/9)\log(25/16)$ and $(1/3)\log(25/16)$.
In the large-component calculation let
$x=2-\sqrt3$, let $v_3$ and $v_2$ be the intrinsic
degree-three and degree-two marginal variances in
Appendix~\ref{app:beyond:negative}, and put
$e=|\rho|/(1+v_2)$ for the intrinsic degree-two correlation.
Messages in the physical variance interval $[v_2,v_3]$
obey the same update as in
Appendix~\ref{app:beyond:positive}. Here
$b=v_2/v_3\le1$, $H_0=(1/b-1)/2$, and the degree-three
variance lower bound is $1-2x^2H_0/[1+(1-x)H_0]$.
It is at least $1/(1+H_0)$ because
$1-2x^2+(1-x-2x^2)H_0>0$. The degree-two update is
unchanged, so the box is invariant. The intrinsic largest
correlation eigenvalues are $1+x$ and $1+e$, both below
two. With $z=(1-b)/(1+b)$, the degree-three turn cap from
conditioning on the remaining port is
$(x+x^2z)/(1-x^2z)$, and the degree-two cap is
$e/[1-(1-e^2)(1-b)/2]$.

\paragraph{Computer-assisted inequalities.}
The script \path{scripts/check_e3_three_view.py}
and the table \path{e3_three_view/rows.csv}
certify turn caps $0.274$ and $0.334$ for degree three and
two, respectively. They certify positivity of both
small-component margins above $L[1+\theta_0(q-2)]$ and
both large-component margins $L/2-C^1(\mathcal P,B)$
using \eqref{eq:jcrit-cost} and $a=0.548$.
Thus there is no loop excess at any degree. Since
$T^1_{\mathcal P}\ge(q-1)L$, \eqref{eq:jcrit-budget}
holds after using the small-component endpoint in place of set 0.
The four-view numerical inequalities are given in
Appendix~\ref{app:beyond:four-view}.

\paragraph{Failure near a linear constraint.}
Fix $K\ge3$ and rescale the negative equal-correlation law
to $I-(1-\delta)uu^\top$, where
$u=\mathbf 1/\sqrt K$ and $\delta\downarrow0$.
In the space of normalized block sums the covariance has one
eigenvalue $\delta$ and all other eigenvalues one, and each
proper principal minor is bounded below. For a partition
with $q\ge3$, the optimal product reference may be reduced
to these sums by the independent-contrast argument of
Appendix~\ref{app:beyond:four-view}. Write its covariance as
$V_0=\operatorname{diag}(v_i)$ and $\beta=1/q=1-\alpha$.
The ratio
$\det(\beta R+\alpha V_0)/(\prod_i v_i)^\alpha$ in the
reduced space is bounded below uniformly over positive $v_i$.
In the determinant's expansion in the $v_i$, all coefficients
are nonnegative. For $\prod_i v_i\le1$ retain the term
linear in $\max_i v_i$, whose coefficient is uniformly
positive by the principal minor bound. Then use
$\max_i v_i\ge(\prod_i v_i)^{1/q}\ge(\prod_i v_i)^\alpha$.
For $\prod_i v_i>1$ retain the product term.
The choice $V_0=I$ supplies a uniform upper bound on the
infimum of this ratio. The affinity formula gives
\[
 R_{\mathcal P}=\frac{\log(1/\delta)}{2(q-1)}+O_K(1).
\]
The least cut has $I_*/2=\tfrac14\log(1/\delta)+O_K(1)$,
since its numerator consists of proper principal minors.
Consequently the singleton partition eventually attains
$L$ for $K>3$. For $K=3$ the strict comparison $R_{1|2|3}<I_*/2$ in
Appendix~\ref{app:beyond:negative} gives the same conclusion.

By permutation invariance and uniqueness the singleton
reference is $vI$. The marginal consistency equation is
\[
 1=\frac{K-1}{K(\alpha v+\beta)}
       +\frac{\delta}{K(\alpha v+\beta\delta)},
 \qquad \alpha=1-1/K .
\]
The first term implies $v\ge(K-2)/(K-1)>0$.
Passing to the limit gives $v\to(K-2)/(K-1)$,
while the tilted longitudinal variance is of order $\delta$.
At the intrinsic precision, which the common box contains,
the singleton tilted correlation off the diagonal is
$-1/(K-1)+O_K(\delta)$ by the two eigenvalues of that tilt.
Thus the absolute turn row sum is $1-O_K(\delta)$.
Any valid cap with $a<1$ must have $1-a^2=O_K(\delta)$,
and the two-step quantity $D(\mathcal P,B)$ has a positive
lower bound. Hence $C^1(\mathcal P,B)\ge\log(1/\delta)-O_K(1)$.
For this partition the base budget $(q-1)L$ already
equals $T^1_{\mathcal P}$, so the second inequality of
\eqref{eq:jcrit-budget} fails. This restriction on the
criterion is compatible with the sharp threshold for three views
established by a different estimate in
Appendix~\ref{app:beyond:negative}.

\subsection{Critical-window bounds at correlation \texorpdfstring{$-1/3$}{-1/3}}\label{app:beyond:window}

\noindent\textbf{Proposition~\ref{prop:window}.}
For three views with equal correlations $-1/3$, set
$R_3=R_{1|2|3}$ and $s_n=dR_3-\log n$.
If $s_n\to+\infty$, maximum likelihood succeeds with
probability tending to one. If
$s_n+\tfrac34\log\log n\to-\infty$, every estimator has
vanishing success probability.

\paragraph{Upper side.}
If needed, use only
$\lceil(\log n+\log\log n)/R_3\rceil$ coordinates. This keeps
$s_n\to+\infty$ and ensures $d/\log n\to1/R_3$.
Replace the true and candidate subset test thresholds
in \eqref{eq:jneg-test} by
$t_k=\log\binom nk+a_nk$, with $a_n=s_n/4$ and $t_0=0$.
The true table fails with probability at most
$\sum_{k\ge1}e^{-a_nk}=o(1)$ by the same first moment
as before. For all $m\le\sqrt n$ the fixed positive
small-component margin in \eqref{eq:jneg-smallfixed}
suffices.

For $m\ge\sqrt n$ take the even short-cycle cutoff
$\Gamma=A\log\log n+O(1)$, with $A$ a sufficiently large
constant. The long-cycle bound of
Appendix~\ref{app:beyond:negative} has
$\delta_\Gamma=O(0.556^\Gamma)$, so $d\delta_\Gamma=o(1)$.
In \eqref{eq:jneg-count} the dependence of the
remainder on $\Gamma$ is $O(m)+O(\Gamma h)$: support, row partitions
and mark positions cost $O(m)$ after pulling out the
displayed terms, and the number of choices for each marked
endpoint is at most the size of a radius-$\Gamma$ ball in a graph
of maximum degree three. For the fixed-law edge cost
$C_{\rm edge}$ in \eqref{eq:jneg-fixed-cost},
$C_{\rm edge}/R_3<1/2$. Thus for a sufficiently large $n$
every mark contributes at most
$dC_{\rm edge}-\log m+O(\Gamma)\le0$ to the log probability.

\paragraph{Reused computer-assisted inequalities.}
For the upper side,
\path{scripts/check_d1_k3_sharp.py} and
\path{d1_k3_sharp/rows.csv} certify
$R_2>R_3$, $(2/9)\log(25/16)>R_3$, and
$C_{\rm edge}/R_3<1/2$. The last follows from the two
outward comparisons $1/R_3<10.2$ and $10.2C_{\rm edge}<1/2$.
These are the fixed-law inequalities used in
Appendix~\ref{app:beyond:negative}.

By \eqref{eq:jneg-branch},
$\log\binom nk\le k(1-\theta)\log n+O(m)$, and
$R_2\ge R_3$, the remaining log probability is bounded by
\begin{align*}
 &(m+k)(\log n-dR_3)+a_n k+O(m)+md\delta_\Gamma\\
 &\hspace{2em}\le -ms_n/2
\end{align*}
for sufficiently large $n$. Summing over the polynomially
many classes of $m,k,h$ with $m\ge\sqrt n$ gives $o(1)$.
The feasible likelihood rule succeeds, and by Bayes optimality
the full-data maximum-likelihood rule succeeds as well.

\paragraph{Lower side.}
Let $P,V,Q$ be the true law, the optimal singleton product
reference, and $Q=Z^{-1}p^{2/3}v^{1/3}$, respectively. Write
$D=D(Q\|P)$, $T=D(Q\|V)$,
$\ell=\log(q/p)$ and $\tau=\log(q/v)$.
The pointwise equality
\[
 \ell+\tau/2=-\tfrac32\log Z=R_3
\]
implies $D+T/2=R_3$. Here $T>0$ since the tilted Gaussian
covariance has nonzero off-diagonal terms. For a sufficiently
large fixed $w$, let
$A_d=\{dD\le\sum_{j=1}^d\ell_j\le dD+w\}$.
Under $Q$, the log ratio $\ell$ has positive variance
and finite third absolute central moment, since it is
a nonconstant Gaussian quadratic form. Berry--Esseen
bounds the difference from the corresponding normal interval
probability by a constant times $d^{-1/2}$. That normal
probability is asymptotic to
$w/[\sqrt{2\pi d}\operatorname{sd}_Q(\ell)]$, so choosing
$w$ large enough gives $Q^{\otimes d}(A_d)\asymp d^{-1/2}$.

For the lower side we may increase $d$ to at least
$\lfloor(\log n-\log\log n)/R_3\rfloor$. The lower-side
premise still holds, this can only help the estimator, and
now $d\sim\log n/R_3$. Independently mark each true row
with conditional probability $e^{-dD-w}q^{\otimes d}/p^{\otimes d}$
on $A_d$ and zero outside. It is at most one by the upper
edge of the band. Thus a marked row has law
$Q_A=Q^{\otimes d}(\cdot\mid A_d)$ and the mean number of
marked rows is $n e^{-dD-w}Q^{\otimes d}(A_d)$.
Since $D<R_3$, this grows as a positive power of $n$
up to subpolynomial factors. Binomial concentration gives
\[
 \log m=\log n-dD-\tfrac12\log d+O(1)
\]
with probability tending to one.

Reveal the marked sets in all views and the full
correspondences of all unmarked rows. The remaining relative
permutations are two independent uniform permutations of
the marked rows. Conditioned on the side information, the
three-view tuples at their true matching are independent
with law $Q_A$: the independent marks condition each row
separately and the original uniform permutations are
independent of the unpermuted data and marks. The pointwise
identity on $A_d$ gives $\sum_j\tau_j\le dT$.
It follows that the likelihood ratio of $Q_A$ to
$V^{\otimes d}$ is at most $\exp(dT+\tfrac12\log d+O(1))$.
Under the reference, the two hidden permutations
remain uniform and independent of the observations,
since $V$ factors over the three views. An estimator
therefore succeeds with probability at most $(m!)^{-2}$.
Changing measure and using Stirling's formula gives the
conditional success bound
\[
 \min\{1,\exp[m(dT+\tfrac12\log d+O(1))-2\log(m!)]\}.
\]
On the concentration event the exponent is
$2m[s_n+\tfrac34\log d+O(1)]$.
Since $\log d=\log\log n+O(1)$, this tends to minus infinity
under the stated lower-side premise. The success probability
therefore tends to zero even with the extra information.

\subsection{The axially symmetric three-view band: statement and proof outline}\label{app:beyond:axial}

This subsection and the next state two further results on Conjecture~\ref{conj:c1prime} and describe their proofs. Unlike Appendices~\ref{app:beyond:rare}--\ref{app:beyond:window}, they do not contain complete proofs: we give the structure of each argument, its new ingredients and the formulas on which it rests, and we omit the derivations.

Let $K=3$ with unit variances, $r_{12}=r_{13}=r$ and $r_{23}=s=r+\delta$; positive definiteness is $|s|<1$ and $e=1+s-2r^2>0$. Write $u=r^2$, $I_i=I(X_i;X_{[3]\setminus\{i\}})$, so $I_1=-\frac12\log\frac{e}{1+s}$ and $I_2=I_3=-\frac12\log\frac{(1-s)e}{1-u}$, and $T_\alpha=\min_VD_\alpha(P\Vert V)$ over product references, so $R_{1|2|3}=T_{2/3}/2$ and $L=\min\{T_{2/3}/2,I_1/2,I_2/2\}$. The optimal reference is $\mathrm{diag}(y,z,z)$, with $z$ the unique root in $[v_{(2)}^2,1]$ of a cubic obtained from the stationarity equations; this is the case of Proposition~\ref{prop:structure}(b) with the symmetry of views $2$ and $3$.

\begin{theorem}[Axially symmetric band; proof in outline]\label{thm:axial}
For every fixed $r\in[-11/30,-3/10]$ and every $\delta$ with $2r^2-1-r<\delta<1-r$, the maximum-likelihood estimate recovers both permutations with probability tending to one whenever $\liminf d/\log n>1/L$. Together with Theorems~\ref{thm:exact}(b) and~\ref{thm:rare}, $1/L$ is the first-order threshold. At $r=-1/3$ the minimum in $L$ is $I_1/2$ for $\delta\le\delta_a$, $R_{1|2|3}$ on $[\delta_a,\delta_b]$, $I_2/2$ on $[\delta_b,2/3]$ and $I_1/2$ on $[2/3,4/3)$, where $\delta_a\approx-0.0251207582$ and $\delta_b\approx0.0439919612$.
\end{theorem}

The argument follows the scheme of Appendix~\ref{app:beyond:criterion}: a decoder that also runs the subset tests of Section~\ref{sec:beyond}, edge-by-edge marking of short cycles with the $\log m$ saving per mark, a Gaussian integral at a belief-propagation fixed point, and a two-level count in $\theta=\log m/\log n$. The new ingredients, and the parameter regions in which each is used, are as follows.

\paragraph{Two tilts and a common message box.} Two configurations are used: the square-root likelihood ($\alpha=1/2$ at three-singleton rows) and the order-$2/3$ subset test, each with the order-$1/2$ tilt at rows split into two blocks. After rescaling each view by the tilted marginal variance, messages stay in a common interval of variances, the turning weights are bounded by explicit $p_{ij},q_i$, and a non-backtracking comparison matrix of norm $a<1$ sums all returns longer than two steps. Each marked port is charged $\log(1+d/(1-a^2))$, with $d$ the squared norm of its two-step vectors. With anchor colour $1$ or $2$ the marks are paid by the larger of these port costs, less the excess rate $P_i-L$, $P_i=\log\frac{1+w_i}{2w_i}$, of two-block rows. If $A=T_{1/2}-L>0$, the square-root cost $C_0\le A$ and $\max\{A-L,\ A+2(R_{1|2|3}-L)-C_1\}\ge0$, every component size is covered, since the square-root bound handles $\theta\le\min\{L,A\}/L$ and the subset test the rest.

\paragraph{Coloured short cycles.} Near $\delta=-0.32$ the two-step charge is too crude. Enumerating, for each of the four root types, the coloured neighbourhoods of the root up to three steps gives the joint weight of all four- and six-cycles through the root and a bound for longer returns from three-step vectors; this extends the square-root argument down to $\delta\approx-0.3586$ at $r=-1/3$.

\paragraph{The singular end: coloured resistances.} For $s<r<0$ write $x=-s$ and represent $R=\mathrm{diag}(D)^{1/2}(I-zz^\top)\mathrm{diag}(D)^{1/2}$ with $z_1^2=u/(u+x)$, $z_2^2=z_3^2=x/(1+x)$ and $\tau=\|z\|^2<1$. Keep the edges of a candidate that carry the same pair of rows as parallel edges of three colours. Then
\[
2B_G=m\log\xi+\log\det(\mu I+\mathcal L_G),\qquad \xi=\frac{\tau^2}{4(1-\tau)},\quad\mu=\frac{2(1-\tau)}{\tau},
\]
where $\mathcal L_G$ is the Laplacian with colour conductances $z_i^2/\tau$; after dividing by the light conductance these are $(1,Q,Q)$ with grounding $M$, $Q=\frac{x(u+x)}{u(1+x)}\ge1$ and $M=\frac{2xe}{u(1+x)}$. Comparing $G$ with locally tree-like lifts, $2(mT_{1/2}-B_G)$ is the limit of the difference of the normalized log-determinants of lift and base. Marks are restored colour by colour; the derivative of the log-determinant in a mark's conductance is its effective resistance, bounded above on the covering tree by the worst alternating ray, with conductance given by the fixed point of $g_p=\mu+S(q,g_q)$, $g_q=\mu+S(p,g_p)$, $S(a,b)=ab/(a+b)$, and below in the base graph by Dirichlet test potentials. For a light mark with a parallel heavy partner, potentials $\pm v$, $v=Q/(M+3Q+2)$, at the two outer neighbours give the base half-resistance bound $(2t+M+3Q-Q^2/(M+3Q+2))^{-1}$. Each mark is then charged less than $T_{1/2}-F$, $F=I_1/2$, and when $T_{1/2}>2F$ the count closes with margin $(m-h)(T_{1/2}-2F)$ at $\theta=1$. Next to $e=0$ a single ray bound $\log(1+2\sqrt2Q)<F$ charges all marks at once. Here $T_{1/2}-2F=\frac12\log\min_{0<z\le1}\frac{(z+2-D)^2(z+D)(z+e)}{16(2-D)Dz^2}$ with $D=1+s$, which is decreasing in $D$ and increasing in $e$.

\paragraph{Common marks and intermediate orders.} Where neither configuration alone suffices, both use the same marks; for a candidate with $k$ three-singleton rows and $h$ marks, $t=\min(h,k)/k$ and $y=L\log m/\log n$, the margins per three-singleton row are
\[
M_0=A-y+t(y-a_0),\qquad M_\alpha=T_\alpha-(1+\lambda)L-(1-\lambda)y+t(y-a_\alpha),\qquad\lambda=\frac{2\alpha-1}{1-\alpha},
\]
the second from the subset test at order $\alpha\in[1/2,2/3]$ via the identity $\sqrt p\,(p/q_\alpha)^\gamma=e^{-T_\alpha/2}q_\alpha/\sqrt v$, $(1+\gamma)(1-\alpha)=1/2$. It suffices that $a_0\le A$ and $(L-A)(a_\alpha-A)_+\le(T_\alpha-2L)(L-a_0)$. Orders $11/20$, $3/5$ and $13/20$ are used near $r=-11/30$; on the three-colour graph the order-$\alpha$ integral is again a grounded Laplacian determinant, with conductances $(1,Qj_1/j_2,Qj_1/j_2)$ and grounding $Mj_1$, $j_i=1+(1/(2\alpha)-1)/\tilde v_i$.

\paragraph{The positive end.} When $s^2(1+s)\ge2r^2$, equivalently $I(X_2;X_3)\ge I_1$, views $2$ and $3$ are aligned first by two-view maximum likelihood and the aligned pair is then matched to view $1$; both steps need at most $2/I_1\le1/L$. This covers $\delta\ge4/5$ on the band.

\paragraph{Certificates.} The finite-dimensional inequalities are verified on closed rational boxes in $(r,\delta)$ by outward-rounded interval arithmetic. The scripts and their recorded tables are in the ancillary files, and the \path{README} there lists them: they are the scripts whose names contain \path{d1_axial}, together with \path{check_d1_outer_left_short.py} and \path{check_d1_thirtieth_determinant.py}. Rerunning the scripts reproduces the recorded tables byte for byte. The interval endpoints in $r$ are chosen by the certificates, and the theorem is a pointwise statement for each fixed covariance.

\subsection{Analytic arguments and open neighbourhoods at the ends: statement and proof outline}\label{app:beyond:analytic}

For the negatively equicorrelated family parametrize $r=-3x/(2+x+2x^2)$, $0<x<1/2$, as in Theorem~\ref{thm:three}. Appendix~\ref{app:beyond:negative} proves that theorem, and with it Conjecture~\ref{conj:c1prime}, on the whole family, in part with certified numerics. Here we describe arguments that use no certified numerics at the two ends, and the stability of the threshold under small perturbations of $R$. As in Appendix~\ref{app:beyond:axial}, only an outline is given.

\begin{proposition}[Analytic arguments at the ends; proof in outline]\label{prop:analytic-ends}
If $0<x\le1/3$ or $0<1-2x\le2^{-10}$, then $L=R_{1|2|3}<I_*/2$ and maximum likelihood recovers both permutations with probability tending to one when $\liminf dR_{1|2|3}/\log n>1$, by an argument with explicit rational constants. Each such equicorrelated $R$ has a neighbourhood in the space of three-view correlation matrices on which, for every fixed matrix, $L=R_{1|2|3}$ and the same conclusion holds.
\end{proposition}

\emph{Near the singular end} ($0<1-2x\le2^{-10}$), the tilt is applied at every row. On a three-coloured expansion of a candidate with $m$ rows per side, the Gaussian integral is $\det(I-xB)^{-1/2}$ for the signed non-backtracking matrix $B$, and the weighted Ihara--Bass formula gives
\[
\det(I-xB)=(1-x^2)^mx^{2m}\det(\mu I+L),\qquad \mu=\frac1x+2x-3,
\]
equivalently $[(1-x^2)^2(1-4x^2)]^m\det(I+\zeta\mathcal A)$ with $\zeta=9x^2/((1-x^2)(1-4x^2))$. In components with a three-singleton row the seed edges form a forest, and a comparison with a large-girth lift through cavity conductances bounds the charge of each mark by an explicit function of $x$; the resulting closing inequality is proved for $1-2x\le2^{-10}$ by elementary estimates, and numerically it persists up to $1-2x\approx2.5\times10^{-3}$.

\emph{Near zero} ($0<x\le1/3$), rows split into two blocks use the square-root factor and three-singleton rows the tilt, so the integrand mixes two precisions. A signed-trace expansion of the mixed normalization shows that only even returns of length at least four enter, with cost at most $MC_m+h_0C_h$ for $M$ marks on cycles through three-singleton rows and $h_0$ other marks, where $C_m<\Delta$ and $C_h<C_0$ for explicit functions $\Delta,C_0$ of $x$, by polynomial inequalities proved on the whole range; for example $I_*/2-R_{1|2|3}>\frac{7}{50}x^3$ on this range.

\emph{Open neighbourhoods.} The optimal product reference depends continuously on $R$, so all rates and the strict budgets persist nearby; the turning weights of a perturbed matrix are dominated termwise by those of an equicorrelated comparison law with slightly larger $x$, and the integral identities above are unchanged because every colour uses the same variance at both ends of an edge. The argument gives no explicit radius.

\emph{Limits of the method.} At $x=7/20$, that is $r=-70/173$, both estimates fail: closed walks around parallel pairs force $\log\mathcal I\ge(m-k)P$ in the full-support tilt, and the mixed cost exceeds its budget ($C_m\approx0.0354>\Delta\approx0.0261$). This is a limitation of these two estimates, not a counterexample; the conjecture holds there by Theorem~\ref{thm:three}.

\begingroup
\let\ladderdisplaystart\[
\let\ladderdisplayend\]
\def\[#1\]{%
  \ladderdisplaystart
    \adjustbox{max width=\displaywidth}{$\displaystyle #1$}%
  \ladderdisplayend
}
\section{Proof of Theorem~\ref{thm:ladder}}\label{app:ladder}

Throughout this appendix we use the standardized model: every object has, in every coordinate $\ell\in[d]$, the $K$-vector law $N(0,R_K(\rho))$ with $\rho=\rho_n\to0$, $d/\log n\to\infty$ and $b_n=d\rho^2/\log n\to b$. The model of~\citet{wee2026geometric} reduces to this one by scaling all views by the same constant, which does not change the maximizers of the pairwise sum. Put $H=\sqrt{d\log n}$; then $d\rho=\sqrt{b_n}\,H$.

\subsection{Tree decoders}

For $K=2$, $J=I(X^1;X^2)=(\frac12+o(1))d\rho^2$, so $J/\log n\to b/2$. If $b>2$, Theorem~\ref{thm:main}(a) gives a two-view estimator with $o(n)$ errors for each tree edge. Composition along the path from view $1$ to view $a$ makes at most the sum of the errors on that path. If $b<2$, Theorem~\ref{thm:pairwise} with $K=2$ shows that every two-view estimator of $\sigma_{ac}$ has vanishing expected overlap; this applies to the estimate on each tree edge.

\subsection{Types and the maximum of the pairwise sum}

Fix an anchor row $i$ holding object $o$, and write $T(t)=\sum_{a<c}\langle Y^a_{j_a},Y^c_{j_c}\rangle$ for a candidate $t$. The identity partition $\cP$ of $t$, called its \emph{type}, has an anchor block $B_0\ni1$ of size $s$ holding $o$ and $r=|\cP|-1$ free blocks holding distinct objects $o_1,\ldots,o_r\ne o$. For a pair of views $e=\{a,c\}$, let $S_e$ be the set of free blocks containing $a$ or $c$, so that $S_e=\varnothing$ exactly when $e\subseteq B_0$. Let $m_0(\cP)=\sum_{B\in\cP}\binom{|B|}2$ be the number of pairs inside blocks, and $M=\binom K2$. Define
\begin{equation}\label{eq:VP}
V(\cP)=\max\Big\{\sum_{e:S_e\ne\varnothing}x_e:\ \tfrac12\sum_{e:\,\varnothing\ne S_e\subseteq U}x_e^2\le|U|\ \text{ for all }\varnothing\ne U\subseteq\{1,\ldots,r\}\Big\},
\end{equation}
with $V(\cP)=0$ if $r=0$.

\begin{lemma}\label{lem:types}
For every type $\cP$, $\max_{t\text{ of type }\cP}T(t)/H=m_0(\cP)\sqrt{b}+V(\cP)+o_P(1)$.
\end{lemma}

\begin{proof}
\emph{Conditioning on the anchor.} Condition on the features $z=(z^a)_{a\in[K]}$ of object $o$. Each $\langle z^a,z^c\rangle$ has variance $O(d)$, so with probability $1-O((\log n)^{-1/2})$ we have $|\langle z^a,z^c\rangle-d(\mathbf 1_{a=c}+\rho\mathbf 1_{a\ne c})|\le\sqrt d\,(\log n)^{1/4}$ for all $a,c$; we call such $z$ typical. Since $\sqrt d\,(\log n)^{1/4}=o(H)$, pairs $e\subseteq B_0$ then contribute $\binom s2\sqrt{b_n}H+o(H)$ to $T$, which is the same for all candidates of the type. For the other pairs, let $X_e$ be the inner product minus its conditional mean, divided by $H$. The conditional mean is $d\rho$ for a pair inside a free block and $0$ otherwise.

\emph{Moderate deviations.} Let $\mathcal E$ be the set of distinct inner products appearing in one candidate, or in two candidates whose free objects come from disjoint pools for distinct blocks. We claim that for every fixed $t\in\R^{\mathcal E}$, uniformly over typical $z$,
\begin{equation}\label{eq:mgf}
\frac1{\log n}\log\E\Big[\exp\Big(\log n\sum_{e\in\mathcal E}t_eX_e\Big)\,\Big|\,z\Big]\to\frac12\sum_{e\in\mathcal E}t_e^2 .
\end{equation}
Coordinates are independent given $z$. In coordinate $\ell$, the free variables form a Gaussian vector $g_\ell$ with covariance $C=I+O(\rho)$, and $\sum_et_eX_e$ restricted to coordinate $\ell$ equals $H^{-1}\big(g_\ell^\top Bg_\ell-\tr(BC)+c_\ell^\top g_\ell\big)$. Here $B$ is a fixed symmetric matrix carrying $t_e/2$ in the two positions of each quadratic pair, and $c_\ell$ is linear in $z_\ell$. Put $v=\log n/H=\sqrt{\log n/d}\to0$, $M_0=C^{1/2}BC^{1/2}$ and $\tilde c_\ell=C^{1/2}c_\ell$. Completing the square, the conditional log-MGF of coordinate $\ell$ is
\[
-v\tr M_0-\tfrac12\log\det(I-2vM_0)+\tfrac12v^2\tilde c_\ell^\top(I-2vM_0)^{-1}\tilde c_\ell=v^2\tr M_0^2+\tfrac12v^2\|\tilde c_\ell\|^2+O\big(v^3(1+\|z_\ell\|^2)\big).
\]
Summing over $\ell$ and dividing by $\log n=v^2d$ gives $\tr M_0^2+(2d)^{-1}\sum_\ell\|\tilde c_\ell\|^2+O(v)$. As $\rho\to0$, $\tr M_0^2\to\tr B^2=\frac12\sum t_e^2$ over quadratic pairs. On typical $z$, $(2d)^{-1}\sum_\ell\|c_\ell\|^2\to\frac12\sum t_e^2$ over pairs joining the anchor block to a free block, because $\langle z^a,z^{a'}\rangle/d\to\mathbf 1_{a=a'}$ and distinct such pairs into the same free variable come from distinct anchor views. This proves~\eqref{eq:mgf}. By the usual tilting argument, for a box of side $\eta$ around $x$ we get $\Prob(X\in\mathrm{box}\mid z)=n^{-\frac12\|x\|^2+O(\eta\|x\|)+o(1)}$, uniformly over typical $z$.

\emph{Upper bound.} With $\eta$ small, there are at most $n^{K-1}$ candidates and each $|X_e|$ exceeds a constant $C$ with probability $n^{-C^2/2+o(1)}$. Hence all candidates lie in a fixed cube with high probability. For every nonempty $U\subseteq[r]$, the subvector $(X_e)_{\varnothing\ne S_e\subseteq U}$ depends only on the objects of the blocks in $U$, which can be chosen in at most $n^{|U|}$ ways. Cover the cube by small boxes. A box on which $\frac12\sum_{\varnothing\ne S_e\subseteq U}x_e^2>|U|+\eta$ for some $U$ contains no candidate with high probability, by the union bound over these choices. All candidates therefore lie near the feasible set of~\eqref{eq:VP} enlarged by $\eta$. Letting $\eta\to0$ and using continuity of the value gives the upper bound.

\emph{Lower bound.} Let $x^*$ attain $V(\cP)$ and $x=(1-\alpha)x^*$, so that $I_U=\frac12\sum_{\varnothing\ne S_e\subseteq U}x_e^2<|U|$ for every $U$. Assign each free block a disjoint pool of $\lfloor(n-1)/r\rfloor$ objects other than $o$, and let $Z$ count candidates from the pools whose vector lies in a small box around $x$. Then $\E Z=n^{r-I+O(\eta)+o(1)}$ with $I=I_{[r]}<r$. Two candidates that share objects exactly on the blocks of $U$ share exactly the inner products with $S_e\subseteq U$; the disjoint pools exclude any other coincidence. By~\eqref{eq:mgf} applied to their union, such a pair lies in the box with probability $n^{-2I+I_U+O(\eta)+o(1)}$, and there are $O(n^{2r-|U|})$ such pairs. Pairs sharing no object are conditionally independent given $z$. Hence $\Var Z/(\E Z)^2\le o(1)+\sum_{U\ne\varnothing}n^{I_U-|U|+O(\eta)+o(1)}\to0$ for $\eta$ small, so $Z>0$ with high probability. Letting $\alpha,\eta\to0$ gives the lower bound, and removing the conditioning on typical $z$ completes the proof.
\end{proof}

\begin{lemma}\label{lem:move}
Let $\cP$ be a type whose anchor block has size $s\ge2$, and let $\cP'$ be obtained by moving the $s-1$ non-anchor views of the anchor block into a new free block. Then $m_0(\cP')=m_0(\cP)-(s-1)$ and $V(\cP')\ge V(\cP)+\sqrt{s(s-1)}$.
\end{lemma}

\begin{proof}
The $s-1$ pairs joining the anchor to the moved views leave the blocks, while the $\binom{s-1}2$ pairs among the moved views stay inside a block. Take an optimal $x$ for $\cP$ and keep its values on the old pairs. The $\binom s2$ pairs of the old anchor block now have $S_e=\{\text{new}\}$; give each the value $\sqrt{2/\binom s2}$, which uses exactly one unit of budget and adds $\sqrt{2\binom s2}=\sqrt{s(s-1)}$ to the objective. Old pairs can only gain the new block in their $S_e$. Constraints not involving the new block therefore only lose terms. A constraint for $U\ni\text{new}$ involves old pairs with $S_e\subseteq U\setminus\{\text{new}\}$, costing at most $|U|-1$, plus the new pairs, costing $1$.
\end{proof}

\begin{proof}[Proof of Theorem~\ref{thm:ladder}(ii)]
\emph{Below $K/(K-1)$.} For a type with $s\ge2$, Lemmas~\ref{lem:types} and~\ref{lem:move} show that the best candidate of type $\cP'$ exceeds the best of type $\cP$ by at least $\sqrt{s(s-1)}-(s-1)\sqrt b-o_P(1)$, in units of $H$. This margin is positive because $b<K/(K-1)\le s/(s-1)$. Since $\cP'$ has anchor block $\{1\}$ and there are finitely many types, the row maximizer has anchor block $\{1\}$ with probability tending to one; that is, every non-anchor entry is wrong.

\emph{Completion.} Let $F$ be the table of raw proposals and $U_a\subseteq[n]$ the set of rows of view $a$ that no anchor row proposes. Each proposed row of view $a$ keeps exactly one anchor row, so the rows to be filled in view $a$ are exactly $U_a$. Suppose conflict resolution and completion are functions of $F$ and of independent randomness. Relabeling the rows of view $a$ by any permutation $\tau$ supported on a fixed set $U$ preserves the joint law of the data and the truth. On the event $U_a=U$ it also leaves $F$ unchanged, because the raw maximization is equivariant, ties have probability zero, and no proposal points into $U$. Hence, conditionally on $F$ and the randomness, an object whose true row in view $a$ lies in $U_a$ has that row uniformly distributed over $U_a$. The $|U_a|$ unmatched anchor rows therefore receive their true rows at most once in expectation. Kept proposals are correct at most as often as raw proposals, so by exchangeability of anchor rows $\E\,\ov(\widehat\pi_a)\le\Prob(\text{the maximizer of a given row is correct in view }a)+1/n\to0$.

\emph{Two non-anchor views.} For $a,c\ge2$, count the hits of $\widehat\pi_c\circ\widehat\pi_a^{-1}$ through anchor rows. Rows filled by completion in view $a$ or in view $c$ contribute at most one hit per view in expectation, by the symmetry above. Rows kept in both views whose proposals come from different objects contribute nothing. Rows kept in both views whose proposals come from the same wrong object $o'$ need $a$ and $c$ in the same free block of the winning type. Every optimal solution of~\eqref{eq:VP} has all coordinates bounded below by some $\kappa_K>0$: if a coordinate vanished, increasing it by $\delta$ and rescaling by $(1+\delta^2/2)^{-1/2}$ would keep every constraint, since $|U|\ge1$, and increase the objective. By the proof of Lemma~\ref{lem:types} the winner lies within $o_P(1)$ of the optimal set of its type. Hence, outside a set of rows of expected size $o(n)$, such a hit forces $\langle Y^a_{\pi_a(o')},Y^c_{\pi_c(o')}\rangle\ge d\rho+\kappa_KH/2$. By Chebyshev's inequality each object satisfies this with probability $O(1/\log n)$, and each object yields at most one hit. The composed estimates therefore have vanishing expected overlap as well.

\emph{Above $K/(K-1)$.} This follows from Theorem~2.5 of~\citet{wee2026geometric}. It can also be read off from Lemma~\ref{lem:types}. By Cauchy--Schwarz, $V(\cP)\le\sqrt{2r(M-\binom s2)}\le\sqrt{2rM}$, while a wrong type loses $e(\cP)\sqrt b$ in the mean, where $e(\cP)\ge r(2K-r-1)/2$ is its number of separated pairs. The ratio $8M/(r(2K-r-1)^2)$ is largest at $r=1$ or $r=K-1$, where it equals $K/(K-1)$ and $4/K$. Hence for $b>K/(K-1)$ every wrong type loses with high probability, and the bad rows number $o_P(n)$, which any completion tolerates.
\end{proof}

\begin{remark}[Completion that rereads the data]\label{rem:completion}
Take $K=3$ and $4/3<b<3/2$, and normalize by $H$. By Lemma~\ref{lem:types}, the true tuple scores $3\sqrt b$, tuples keeping one correct non-anchor view score at most $\sqrt b+2$, tuples using two different wrong objects score at most $\sqrt{12}$, and the best tuple pairing the anchor with one wrong object in both other views scores $\sqrt b+\sqrt6$, the largest of these. Since the cross terms $\langle Y^1_i,Y^2_{\pi_2(j)}+Y^3_{\pi_3(j)}\rangle/H$ are at most $2+o_P(1)$ over all wrong objects $j\ne o$, the winning tuple of a typical row has an internal inner product between views $2$ and $3$ of at least $d\rho+cH$ for some fixed $c>0$. Only a vanishing fraction of objects have such large internal inner products, so in each non-anchor view the raw proposals cover only $o(n)$ distinct rows. A completion may first compute the partition-test solution of Theorem~\ref{thm:main} on the whole table and then enforce the $o(n)$ kept proposals by swapping output rows. Each swap changes at most two anchor rows and never disturbs a proposal enforced earlier, since kept proposals use distinct rows. The result is almost exact. Statements about the final output of Algorithm~1 must therefore specify the completion rule.
\end{remark}

\begin{remark}[Anisotropic spectra]
With unequal $\rho_\ell$, the unweighted pairwise sum sees $(\sum_\ell\rho_\ell)^2/(d\log n)$, which can be much smaller than $A/\log n$. Weighting coordinate $\ell$ by $\rho_\ell$ restores the signal $A/\log n$. The analysis above then applies with $H^2=A\log n$, provided $h\to0$ and $h\sqrt{\log n/A}\to0$; the latter holds automatically when $A/\log n$ tends to a positive constant. Without such a spread condition the conclusion can fail, for instance when a single coordinate carries all the correlation.
\end{remark}

\endgroup

\section{Proof of Theorem~\ref{thm:screen}}\label{app:screen}

\subsection{Screening the seed views}

Fix $3\le m\le K$ and write $A=\sum_\ell\rho_\ell^2$ and $h=\max_\ell\rho_\ell$. For $2\le s\le m$, let $P_s$ be the law of a true tuple of the first $s$ views. For each nontrivial partition $\cP$ of $[s]$, set
\[
Q_{s,\cP}=\bigotimes_{B\in\cP}P_B,\qquad
L_{s,\cP}=\log\frac{dP_s}{dQ_{s,\cP}},\qquad
c(\cP)=\sum_{a<b}\mathbf 1\{a,b\text{ in different blocks of }\cP\}.
\]
At level $s$ a tuple passes if every $L_{s,\cP}$ exceeds $\alpha c(\cP)\log n$, where $2/m<\alpha<b/2$. For each anchor row and view $a=2,\ldots,m$, form the list of rows passing the analogous two-view test on views $1,a$. Discard any anchor row if one of its lists has length greater than $D_{\max}=\lceil n^{1-2/m}\rceil$. Starting from the remaining anchors, extend each surviving prefix by the next view's anchor list and apply the tests for the new length. The test at length two was already applied when making the list for view $2$. At an intermediate level $s<m$, abort with identity permutations if more than $2n^{u_s}$ prefixes survive, where $u_s=\max\{1,s-s(s-1)/m\}$. At level $m$ scan the passing tuples and keep each tuple whose rows are all unused. This gives a maximal set of disjoint tuples. Complete the output by matching the unused rows of each view to the unused anchor rows in index order.

\paragraph{True tuples.} At coordinate $\ell$, let $C_\ell$ and $B_\ell$ be the covariance matrices under $P_s$ and $Q_{s,\cP}$. The two matrices agree on the diagonal blocks of $\cP$, so $\tr(B_\ell^{-1}C_\ell)=s$. A second-order expansion of the Gaussian divergence gives
\begin{align*}
\E_{P_s}L_{s,\cP}
  &=\tfrac12\sum_\ell\log(\det B_\ell/\det C_\ell)\\
  &=\tfrac12 c(\cP)A+O_m(hA).
\end{align*}
Indeed, the differences from the identity have zero trace, and their squared Frobenius norms differ by $2c(\cP)\rho_\ell^2$ at coordinate $\ell$. The remainder is $O_m(\rho_\ell^3)$, which sums to $O_m(hA)$. By Lemma~\ref{lem:llr} the variance is $O_m(A)$. Since $A/\log n\to b$ and $\alpha<b/2$, each true test has mean exceeding its threshold by a fixed positive multiple of $\log n$. Chebyshev's inequality and a union bound over the tests, including all the anchor-list tests, give failure probability $o(1)$ for each true tuple. By Markov's inequality the number of true tuples failing any of these tests is $o_{\Prob}(n)$.

\paragraph{Anchor lists.} Conditional on the hidden permutations, a wrong anchor pair has the product law $Q$ of the two row marginals. Its log-likelihood ratio $L$ has $\E_Q e^L=1$, so the probability of passing is at most $n^{-\alpha}$. Each list has at most one true entry and at most $n^{1-\alpha}$ expected false entries. The probability that an anchor is discarded is $O_m(n^{-(\alpha-2/m)})$ by Markov's inequality on each list and a union bound. Thus only $o_{\Prob}(n)$ anchors are discarded.

\paragraph{Prefix counts.} Conditional on the hidden permutations, a wrong tuple with identity partition $\cP$ has law $Q_{s,\cP}$. If $\cP$ has $p\ge2$ blocks, there are at most $n^p$ tuples with that identity partition. Passing the test for $\cP$ alone has probability at most $n^{-\alpha c(\cP)}$, regardless of the tests on its prefixes. Among partitions of $s$ views into $p$ blocks, the cross-block pair count is smallest for sizes $s-p+1,1,\ldots,1$, so
\[
c(\cP)\ge(p-1)(s-p/2).
\]
The function $p\mapsto p-(2/m)(p-1)(s-p/2)$ is convex, with maximum $u_s$ on $[1,s]$ attained at an endpoint. Because $\alpha>2/m$, each nontrivial partition contributes an expected $n^{p-\alpha c(\cP)}=o(n^{u_s})$ false tuples. Summing over the finitely many partitions gives the same bound for the expected number of false survivors at level $s$. This count includes all index tuples passing the tests at that level and hence bounds the count reached by the algorithm. Since there are at most $n$ true tuples and $u_s\ge1$, the probability of hitting any intermediate cap is $o(1)$ by Markov's inequality and a union bound.

\paragraph{Selection.} Since $u_m=1$, only $o_{\Prob}(n)$ false tuples pass at the final level. With probability tending to one there is no abort, and all true tuples other than those with failed tests or discarded anchors reach the final selection. A passing true tuple omitted from the maximal disjoint set intersects a selected false tuple because distinct true tuples are disjoint. Each selected false tuple can block at most $m$ true tuples. Thus the first $m$ views are aligned on a common set of $n-o_{\Prob}(n)$ anchor rows.

\subsection{Attaching the remaining views}

For each anchor row, sum the corresponding rows of the first $m$ views after alignment and divide coordinate $\ell$ by $\sqrt{m(1+(m-1)\rho_\ell)}$. In the oracle data with the first $m$ views truly aligned, this produces a standard Gaussian row. For any fixed remaining view its true paired row has coordinate correlation
\[
r_\ell=\frac{\sqrt m\,\rho_\ell}{\sqrt{1+(m-1)\rho_\ell}}.
\]
The two-view log-likelihood ratio has mean $I_m$ and variance $\sum_\ell r_\ell^2$ under the true paired law, where
\[
I_m=-\tfrac12\sum_\ell\log(1-r_\ell^2)=(m/2+O_m(h))A.
\]
In the oracle covariance, the target view couples to the first $m$ views only in the all-ones direction. The seed contrasts are independent of the sum and of the target, so the sum retains their mutual information with that view.
Use a fixed threshold $\beta\log n$ with $1<\beta<mb/2$. In the oracle data, by Chebyshev's inequality and Markov's inequality the threshold graph omits $o_{\Prob}(n)$ true pairs. Since a false pair has the product law, the chance that it passes is at most $n^{-\beta}$. The expected number of false edges in the oracle graph is therefore at most $n^{2-\beta}=o(n)$.

The actual graph agrees with the oracle graph on every anchor row correctly aligned across the first $m$ views. A maximal matching can select at most one false edge incident to each contaminated anchor row. Every other selected false edge is in the oracle graph. Consequently it selects $o_{\Prob}(n)$ false edges. All but $o_{\Prob}(n)$ true edges are present on clean rows, and each selected false edge can block at most two such edges. The matching keeps $n-o_{\Prob}(n)$ true edges, which are preserved by completion. This comparison is instancewise and uses unconditional bounds on the oracle graph. Applying it to each of the finitely many remaining views gives a common set of $n-o_{\Prob}(n)$ correctly aligned anchors.

\subsection{Arithmetic cost}

With the whitened rows given, a fixed Gaussian test score is a constant plus weighted row squares and weighted pairwise inner products. Its coefficients depend only on the partition and the known coordinate correlations. Precomputing the tables for all tests on the first $m$ views costs $O_m(n^2d)$, including the anchor-list tests. Each candidate extension is then checked in $O_m(1)$ operations, with the constant accounting for all partitions at its level. A level $s<m$ has $O(n^{u_s})$ prefixes, where $u_1=1$, and each can be extended in at most $D_{\max}$ ways. The prefix cap gives this bound on every input. The maximum of $u_s$ for integer $1\le s<m$ equals $\lfloor(m+1)^2/4\rfloor/m$, by maximizing $s(m+1-s)/m$. The aggregate cost of extending prefixes and selecting disjoint final tuples is therefore bounded by $O_m(n^{\gamma_m})$, with
\[
\gamma_m=\max\left\{2,\ 1-\frac2m+\frac{\lfloor(m+1)^2/4\rfloor}{m}\right\}.
\]
The remaining views cost $O((K-m)n^2d)$ to score and match by scanning their edges. The total is $O_m(Kn^2d+n^{\gamma_m})$. For $m=3,4$, $\gamma_m=2$, yielding the quadratic bounds and the stated thresholds. Computing or applying whitening transformations is outside this arithmetic count.

\section{Proof of Theorem~\ref{thm:recursive} and the four-view designs}\label{app:recursive}

\subsection{Recursive weakest-partition decoding}

\begin{proof}[Proof of Theorem~\ref{thm:recursive}]
\emph{Inheritance of the margin.}
For a node $U$ and partition $\cP$ of $U$, write $D_U(\cP)=D(P_U\Vert\bigotimes_{B\in\cP}P_B)$. Suppose $\cP$ attains $J_U$ and has $p$ blocks. Refining it by a partition $\mathcal Q$ of one block $B$ into $q\ge2$ blocks gives a partition $\cP'$ into $p+q-1$ blocks. The chain rule for divergence gives
\[
D_U(\cP')=D_U(\cP)+D_B(\mathcal Q).
\]
The left side is at least $(p+q-2)J_U$, while $D_U(\cP)=(p-1)J_U$. Therefore $D_B(\mathcal Q)\ge(q-1)J_U$, and minimizing over $\mathcal Q$ gives $J_B\ge J_U$. Singleton blocks have $J_B=\infty$ by convention. Each internal node inherits the root's margin. A partition of the $p$ super-views with law $P_U$ is a coarsening of $\cP$ on $U$ with the same number of blocks and the same divergence. Its normalized divergence is at least $J_U$, with equality for the partition into $p$ separate super-views. Thus the shared information at this decoding step is exactly $J_U$.

\emph{One decoding step.}
For each node fix a representative view, and use it also as the representative of the child containing it. The output at a node is indexed by the representative view's row, with each original row of every member view used once. A leaf outputs its observed rows. For the analysis at a node with $p$ children, relabel each child's rows by the object in its representative view. A row of child $B_j$ is contaminated if any of its member rows belongs to a different object. Let $E_{B_j}$ count these rows, and put $c=\sum_{j=1}^p E_{B_j}$.

On the oracle data, in which each child's rows are correctly concatenated, let $F$ be the number of true tuples failing a partition test and let $T$ be the number of false tuples passing them all. For a nontrivial partition $\mathcal R$ of the children, let $L_{\mathcal R}$ be the corresponding log-likelihood ratio and let $D_{\mathcal R}$ be its true mean. The inherited margin gives $D_{\mathcal R}\ge(1+\epsilon)(|\mathcal R|-1)\log n$, so the gap to the threshold is at least $(\epsilon-\gamma)D_{\mathcal R}/(1+\epsilon)$. Since $\Var(L_{\mathcal R})\le2K^2D_{\mathcal R}$ by Lemma~\ref{lem:llr}, Chebyshev's inequality and a union bound give $\E F=O_{K,\epsilon,\gamma}(n/\log n)$.

Condition on the true permutations. A false oracle tuple with identity partition $\mathcal R$ has the product law $Q_{\mathcal R}$ of the corresponding block marginals. Under it, $\E_{Q_{\mathcal R}} e^{L_{\mathcal R}}=1$, so the probability of passing the test for its own partition is at most $n^{-(1+\gamma)(|\mathcal R|-1)}$. There are at most $n^{|\mathcal R|}$ such tuples, and summing over partitions gives $\E T=O_K(n^{1-\gamma})$. Both estimates concern the oracle data under the original law and require no conditioning on any child decoder's output.

In the actual data at least $n-F-c$ true tuples have all rows clean and pass the tests. The selected tuples are disjoint. Among the selected false tuples, at most $c$ use contaminated rows and all the others are counted by $T$. Any passing clean true tuple absent from the maximal selection intersects a selected false tuple. Each such false tuple intersects at most $p$ true tuples. Thus the node's output retains at least $n-F-c-p(T+c)$ clean true tuples. Completion preserves them, giving the instancewise bound
\[
E_U\le F+pT+(p+1)\sum_{j=1}^p E_{B_j}.
\]
From $E_{\{a\}}=0$ at the leaves and the two oracle counts, induction over the fixed-size tree yields $\E E_{[K]}=O_{K,\epsilon,\gamma}(n/\log n+n^{1-\gamma})$. Markov's inequality gives $E_{[K]}=o_{\Prob}(n)$, which implies simultaneous almost-exact recovery.

\emph{Arithmetic cost.}
Enumerating partitions of subsets of $[K]$ and computing their Gaussian divergences costs $O_K(d^3)$, where $d=\sum_a d_a$. At a node, every Gaussian test score is a constant plus one quadratic term per child row and one bilinear term per pair of child rows. For a coefficient block $H$ and row matrices $X,Y$, tabulate the bilinear terms by computing $(XH)Y^\top$. Summed over all tests and nodes, the table construction takes $O_K(nd^2+n^2d)$ operations. The coefficients are determined by the known covariance, and each node's score tables use its available child rows. A scan of $n^p$ tuples at a node with $p$ children tests each candidate and updates the maximal selection with $O_K(1)$ operations using the tables. Hence the total scan cost is $O_K(n^{p^*})$. When every split is binary, there are exactly $K-1$ internal nodes, each using a two-view threshold test and maximal matching.

\emph{Recovering the tree by greedy merges.}
For an internal node $U$ with unique weakest bipartition $U_L\mid U_R$, write $\lambda_U=J_U=I(X^{U_L};X^{U_R})$. If its child $W$ is internal, refining $U_L\mid U_R$ by the weakest split of $W$ produces a competing partition into three blocks. Its normalized divergence is $(\lambda_U+\lambda_W)/2$ by the chain rule and exceeds $\lambda_U$ by uniqueness. Thus $\lambda_W>\lambda_U$.

Maintain the invariant that the greedy procedure's current clusters are disjoint nodes of the desired tree covering all leaves. It holds for the singleton start. For current nonsibling nodes $A,B$, let $U$ be their lowest common ancestor. By data processing under projection,
\[
I(X^A;X^B)\le I(X^{U_L};X^{U_R})=\lambda_U.
\]
At least one child of $U$ is still covered by two or more current clusters. In that child's subtree a deepest unmerged internal node $W$ has both children present as current clusters. Their merge score is $\lambda_W>\lambda_U$. Hence every nonsibling pair has strictly smaller score than some available sibling pair, so a maximizing merge preserves the invariant. After $K-1$ steps all internal nodes of the desired tree have been merged.
\end{proof}

\begin{proposition}[Decoding along a two-block partition]\label{prop:app-twoblock}
Let $\Sigma\succ0$ and let $\{B_1,B_2\}$ partition $[K]$. Suppose $J_{B_1}$, $J_{B_2}$, and $D_{\{B_1,B_2\}}$ are each at least $(1+\epsilon)\log n$ for a fixed $\epsilon>0$ and all large $n$. Align both blocks almost exactly, concatenate each block's aligned rows, and keep a maximal matching of pairs whose block log-likelihood ratio exceeds $(1+\gamma)\log n$, where $0<\gamma<\epsilon$. After one-to-one completion, the result is simultaneously almost exact. If the bipartition is weakest, the sufficient condition reduces to $J\ge(1+\epsilon)\log n$.
\end{proposition}

\begin{proof}
The within-block margin guarantees decoders by Theorem~\ref{thm:main}(a). In the one-step proof above, the only test for two children has mean $D_{\{B_1,B_2\}}$ and threshold $(1+\gamma)\log n$. Its oracle counts satisfy $\E F=O_{K,\epsilon,\gamma}(n/\log n)$ and $\E T=O(n^{1-\gamma})$. The same instancewise bound, now $E_{[K]}\le F+2T+3(E_{B_1}+E_{B_2})$, gives the claim for any split satisfying these conditions. When the split is weakest, $D_{\{B_1,B_2\}}=J$, and inheritance of the margin gives $J_{B_1},J_{B_2}\ge J$.
\end{proof}

\subsection{Thresholds and decoders for the four-view designs}

\begin{proposition}[Thresholds for the clustered example]\label{prop:app-cluster-thresholds}
In the uniform design of Example~\ref{ex:cluster}, the best tree decoder, row-wise pairwise sum with the per-view completion described in the proof of Theorem~\ref{thm:ladder}(ii), and partition tests have almost-exact thresholds $12$, $8$, and $6$ in $S$. In the clustered design their thresholds are $72$, $22.5$, and $18$. The corresponding exact-recovery thresholds are $8$ and $36$. The within-cluster pairs in the clustered design are matchable for $S>4.5$, and for $S<18$ every cross pair has vanishing overlap under any estimator.
\end{proposition}

\begin{proof}
In the uniform design every pair has $A_{ab}/\log n=S/6$. The best tree needs $S/6>2$, the pairwise sum needs $S/6>4/3$, and $J/\log n\to(K/4)\,S/6=S/6$ in the weak limit.

In the clustered design the strong pairs have $A/\log n=S/2.25$ and the four cross pairs $S/36$. Every spanning tree uses a cross pair, which needs $S/36>2$, that is, $S>72$. By Proposition~\ref{prop:strength}, $J/\log n$ tends to the strength of the graph with weights $\frac12A_{ab}/\log n$. The cut $\{AB,CD\}$ has weight $\frac12\cdot4\cdot S/36=S/18$. Every other partition cuts a strong pair of weight $2S/9$ and has at most four blocks, so its normalized weight is at least $2S/27>S/18$. Hence the minimizing partition is $\{AB,CD\}$ for large $n$, and $J/\log n\to S/18$, with threshold $18$. The pair $AB$ alone has $I_{AB}/\log n\to\frac12S/2.25$, which exceeds $1$ when $S>4.5$.

For the pairwise sum, consider the type $\{A,B\}\mid\{C,D\}$ with anchor $A$. Relative to the true tuple it loses the means of the four cross pairs, $4\cdot\frac14\sqrt{S/2.25}$ in units of $H$. Its five non-fixed pairs $AC,AD,BC,BD,CD$ all depend on one free object, so~\eqref{eq:VP} gives $V=\sqrt{10}$. With probability tending to one, this type beats the true tuple when $\sqrt{S/2.25}<\sqrt{10}$, that is, $S<22.5$, and loses to it when $S>22.5$; the limits say nothing at $S=22.5$ itself. Lemma~\ref{lem:types} extends to this design, whose correlations are nonnegative and identical across coordinates, with the mean term replaced by the sum of $\sqrt{A_{ab}/\log n}$ over pairs inside blocks. In units of $\mu=\sqrt{S/2.25}$ the strong pairs have mean $1$ and the cross pairs $1/4$, and, with probability tending to one, a type beats the true tuple if $V>\mu L$ and loses to it if $V<\mu L$, where $L$ is the mean it loses. By Cauchy--Schwarz with $U=[r]$, $V^2\le2r(6-\binom s2)$. The other two-block types have $L\ge3/2$ and $V^2\le12$, three-block types have $L\ge2$ and $V^2\le24$, and the four-block type has $L=3$ and $V^2\le36$. Their ratios $V^2/L^2$ are at most $16/3$, $6$ and $4$, below the value $10$ for $\{A,B\}\mid\{C,D\}$. Hence above $22.5$ the true tuple beats every type. Below $22.5$, a typical anchor proposes at least one wrong target, so only $o(n)$ raw proposals are fully correct in expectation. During completion, each view keeps one proposer per proposed target and fills the remaining entries using only the raw proposals and independent randomness. The raw maximization is equivariant under row relabeling. By the symmetry argument in the proof of Theorem~\ref{thm:ladder}(ii), completion contributes at most one expected hit per non-anchor view. Each newly correct tuple needs at least one of these hits, so expected complete-tuple overlap tends to zero.

For exact recovery, a bipartition's mutual information is asymptotic to its total cut weight by Proposition~\ref{prop:strength}. In the uniform design every edge has weight $S/12$ in units of $\log n$. A smallest cut has three edges, so $I_*/\log n\to S/4$. In the clustered design the cut $\{AB,CD\}$ has weight $S/18$, and every other cut crosses a strong edge of weight $2S/9>S/18$. Thus $I_*/\log n\to S/18$ and this cut attains the minimum for large $n$. The threshold $I_*=2\log n$ in Theorem~\ref{thm:exact} gives $S=8$ in the uniform design and $S=36$ in the clustered design.

The cross-pair claim for $S<18$ follows from Theorem~\ref{thm:pairwise}: since the unique weakest partition separates the clusters, each cross pair has $J^*_{ab}\le J$. For $S>4.5$, the within-cluster pairs are each above the two-view almost-exact threshold.
\end{proof}

\begin{proposition}[Decoding the clustered design]\label{prop:app-cluster}
For $S>18$ in the clustered design of Example~\ref{ex:cluster}, align $A$ to $B$ and $C$ to $D$ by maximum-weight assignments with inner-product scores, then match the summed rows of each resulting cluster by an almost-exact two-view decoder. This recovers all relative permutations simultaneously almost exactly. In particular, a third maximum-weight assignment suffices and the three assignments run in $O(n^3+n^2d)$ time. Simultaneous almost-exact recovery is impossible for $S<18$.
\end{proposition}

\begin{proof}
Let $s$ be the correlation inside the clusters, so that $S=\frac94ds^2/\log n$ and $s\to0$, and index rows by objects. For $j\ne i$, conditionally on $A_i$ the vector $B_i-B_j$ is Gaussian with mean $sA_i$ and covariance $(2-s^2)I_d$. Hence, for $\lambda>0$,
\begin{align*}
\E e^{-\lambda\langle A_i,B_i-B_j\rangle}&=\big(1-2c(\lambda)\big)^{-d/2},\\
c(\lambda)&=-\lambda s+\tfrac12\lambda^2(2-s^2),
\end{align*}
and $\lambda=s/(2-s^2)$ gives $\Prob(\langle A_i,B_i-B_j\rangle\le0)\le(1-s^2/2)^{d/2}\le e^{-ds^2/4}=n^{-S/9}$. A union bound over the $n(n-1)$ ordered pairs in each cluster shows that for $S>18$, with probability at least $1-2n^{2-S/9}\to1$, every row of $A$ prefers its true partner in $B$ to every other row, and likewise for $C$ and $D$. Then the true alignment is the unique maximum-weight assignment, so step (i) is exact in both clusters.

Consider the oracle that receives the true within-cluster alignments and forms $U_i=A_i+B_i$ and $V_i=C_i+D_i$. These databases have i.i.d.\ rows whose coordinates have variance $2(1+s)$ and covariance $4\cdot s/4=s$, that is, correlation $\gamma=s/(2(1+s))$. Hence $I(U;V)/\log n=(\frac12+o(1))\,d\gamma^2/\log n\to S/18>1$, so the two-view partition test is almost exact by Theorem~\ref{thm:main}(a). A maximum-weight assignment with weights $\langle U_i,V_j\rangle$ is also almost exact. Indeed, the Gaussian pair log-likelihood is a positive multiple of this inner product plus separate row terms whose sum is invariant under assignment. Thus the maximum-weight assignment is maximum likelihood, and the balanced two-view Gaussian result applies since $\gamma\to0$ and the information exceeds $\log n$ by a fixed margin~\citep[Theorem~3.1]{dai2023gaussian}. For the full four-view block likelihood ratio, the within-cluster sum and difference coordinates diagonalize the within-cluster covariance. The cross-cluster covariance couples only the sums, because it is proportional to the $2\times2$ all-ones matrix. The differences are jointly independent of the sums and of one another, so the likelihood ratio across the split depends only on the summed rows.

On the event that the first two assignments are exact, the decoder and its oracle counterpart coincide. The three assignments cost $O(n^3)$ and their score computation costs $O(n^2d)$. The lower bound for $S<18$ follows from Proposition~\ref{prop:app-cluster-thresholds} and Theorem~\ref{thm:main}(b).
\end{proof}

\paragraph{Three assignments in the uniform design.} In the uniform design with a fixed $S>12$, the three maximum-weight assignments used in Appendix~\ref{app:numerics} are simultaneously almost exact. Index rows by objects and write $U_i=A_i+B_i$, $V_j=C_j+D_j$, $W_{ij}=\langle U_i,V_j\rangle$ and $\rho=\sqrt{S\log n/(6d)}$. Each within-cluster pair has $I/\log n\to S/12>1$, so the maximum-weight assignments $\alpha$ and $\delta$, which are the two-view maximum-likelihood assignments here, err on $o(n)$ rows~\citep[Theorems~3.1 and~3.2]{dai2023gaussian}. Let $\mathcal E$ be the set of objects whose row in $A$ or in $C$ is affected by these errors, so $|\mathcal E|=o(n)$ in probability. The third assignment $\sigma$ maximizes $\sum_i\widetilde W_{i\sigma(i)}$, where $\widetilde W_{ij}=W_{ij}$ unless $i\in\mathcal E$ or $j\in\mathcal E$. Every entry of $W$ and $\widetilde W$ has the form $\langle A_i+B_k,C_j+D_l\rangle$; Bernstein's inequality and a union bound over the $n^4$ such products show that all of them are at most $M=C_0\sqrt{d\log n}$ in absolute value, for a constant $C_0$, with probability tending to one. Conditionally on $U_i$, the difference $W_{ii}-W_{ij}$ for $j\ne i$ is Gaussian with mean $(4+o(1))\rho d$ and variance $(8+o(1))d$, uniformly on the event that $\|U_i\|^2=(2+o(1))d$ for all $i$. Hence, if $c>0$ is small enough that $(\sqrt{S/6}-c/4)^2>2$, a union bound over the $n^2$ ordered pairs shows that $W_{ii}-W_{ij}\ge c\sqrt{d\log n}$ for all $i\ne j$ with probability tending to one. On these events, comparing $\sigma$ with the identity gives
\begin{align*}
c\sqrt{d\log n}\,\big|\{i:\sigma(i)\ne i\}\big|
&\le\sum_i\big(W_{ii}-W_{i\sigma(i)}\big)\\
&\le\sum_i\big(\widetilde W_{ii}-\widetilde W_{i\sigma(i)}\big)
    +8M|\mathcal E|\\
&\le8M|\mathcal E|,
\end{align*}
because $W$ and $\widetilde W$ differ only in rows and columns indexed by $\mathcal E$, which enter at most $2|\mathcal E|$ of the terms. So the third assignment errs on $o(n)$ rows, and so does each composed map.

\section{Barriers for natural decoders}\label{app:barriers}

Throughout, $\Lambda=\sqrt{\log n}$. Parts~(a) and~(b) of Proposition~\ref{prop:barriers} are proved in full. For part~(c) we state the three lemmas on which the proof rests and outline their proofs and the deduction; the outlines give the main steps, not every estimate.

\subsection{Explicit enumeration}

\begin{proof}[Proof of Proposition~\ref{prop:barriers}(a)]
Fix a target view $a\in\{2,3\}$ and let $\ell_{i,a}$ be the log-likelihood ratio of the true pair $(Y^1_i,Y^a_{\pi_a(i)})$ against independent rows. Per coordinate it equals $-\frac12\log(1-\rho^2)+\frac\rho2(G_+^2-G_-^2)$ with independent standard $G_\pm$, so $\E\ell_{i,a}=(b/2+o(1))\log n$ and $\Var\ell_{i,a}=d\rho^2=O(\log n)$. Fix $0<\varepsilon<\min\{1-b/2,\delta/2\}$, $w=n^{1-b/2-\varepsilon}$ and $u=(b/2+\varepsilon/2)\log n$. By Chebyshev's and Markov's inequalities, $\#\{i:\ell_{i,a}>u\}=o_\Prob(n)$.

Fix $\eta\in(0,1)$ and a set $D$ of $m=\lceil\eta n\rceil$ anchor rows, and condition on $\pi_a$ outside $D$. Under the reference law in which the pairs in $D$ are replaced by independent rows, the remaining bijection is uniform given $(Y^1,Y^a)$ and the algorithm's randomness, because the lists read nothing else. So the event $E_D$ that every $i\in D$ has $|L_{i,a}|\le w$ and $\pi_a(i)\in L_{i,a}$ has reference probability at most $w^m/m!$. The likelihood ratio of the true law against this reference is $\exp\sum_{i\in D}\ell_{i,a}$, so on $F_D=\{\ell_{i,a}\le u,\ i\in D\}$,
\[
\Prob(E_D\cap F_D)\le e^{mu}w^m/m!=\exp\big(-\tfrac\varepsilon2m\log n+O_\eta(m)\big).
\]
A union over at most $2^n$ sets $D$ shows that $B_{a,n}=\#\{i:|L_{i,a}|\le w,\ \pi_a(i)\in L_{i,a}\}=o_\Prob(n)$. Deterministically, a scanned true triple has a list of length at most $w$ in some view or costs at least $w^2$ visits, so $H_n\le B_{2,n}+B_{3,n}+T_n/w^2$, and $T_n/w^2\le n^{1-\delta+2\varepsilon}=o(n)$.

For the upper side with $b>0$, take $0<\kappa<b/2$ and lists of the $\lceil n^{1-b/2+\kappa}\rceil$ largest raw inner products. Given the anchor row $x$, the normalized true score is $N(\rho\|x\|,1-\rho^2)$ and the $n-1$ wrong scores are independent standard normals; with $t<\sqrt b$ and $t^2/2>b/2-\kappa/2$, the expected number of wrong scores above a true score $\ge t\Lambda$ is at most $n^{1-t^2/2}$, so Markov's inequality drops the true row from the list with probability at most $n^{-\kappa/2}$.
\end{proof}

\subsection{Half-edge dual prices}

The linear program has a variable $z_v\ge0$ for every tuple $v=(i_1,\dots,i_K)$, unit row constraints in each view, and objective $\sum_vz_v\sum_{a<c}s_{ac}(i_a,i_c)$ with $s_{ac}(i,j)=(X^a_i)^\top R_{ac}X^c_j$. The prices $u_{a,i}=\frac12\sum_{c\ne a}s_{ac}(i,i)$ make every true dual constraint tight.

\begin{proof}[Proof of Proposition~\ref{prop:barriers}(b)]
Let $S|S^c$ be a minimizing cut, $W=R_{SS^c}$, $A=\|W\|_F^2$ and $h=\|R-I\|_{\rm op}$. Whitening within blocks and expanding $-\frac12\log\det(I-G^\top G)$ gives $A=(2+O(h))I_*=(2c+o(1))\log n$. For objects $i\ne j$, the false tuple taking object $i$ in the views of $S$ and $j$ in those of $S^c$ violates its dual constraint by
\[
Z_{ij}=F_{ij}-\tfrac12D_i-\tfrac12D_j,\qquad F_{ij}=(X_i^S)^\top WX_j^{S^c},\quad D_i=F_{ii},
\]
since the within-block scores cancel; $\E Z_{ij}=-A$. As a quadratic form in the stacked rows, $Z_{ij}$ has coefficient matrix of squared Frobenius norm $\frac A2(1+\frac14+\frac14)=\frac{3A}4$, and for two pairs sharing one object in the same role the sum has $\frac A2(1+1+1+\frac14+\frac14)=\frac{7A}4$ and mean $-2A$; all operator norms are $O(h)$. Hence the cumulant generating function of such a form is $t\E L+t^2\|M\|_F^2+O(hA)$ uniformly for bounded $t$. Chernoff with $t=2/3$ and the exponential tilt at $t=2/3+\alpha$ give $\Prob(Z_{ij}>0)=e^{(-1/3+o(1))A}$, and Chernoff with $t=4/7$ gives $\Prob(Z_{ij}>0,Z_{kl}>0)\le e^{(-4/7+o(1))A}$ for pairs sharing one object.

Split the objects into halves $I_1,I_2$ and let $N=\sum_{(i,j)\in I_1\times I_2}\mathbf 1\{Z_{ij}>0\}$. Disjoint pairs are independent, and $n_1n_2(n_1+n_2-2)$ ordered pairs share one object, in the same role. Then $\E N=n^{2-2c/3+o(1)}\to\infty$ and
\[
\frac{\Var N}{(\E N)^2}\le\frac1{\E N}+O(n^{-1})e^{(2/3-4/7+o(1))A}=\frac1{\E N}+n^{-1+4c/21+o(1)}\to0
\]
for $c<3$, so $N>0$ with probability tending to one.
\end{proof}

\subsection{Sinkhorn soft averaging: proof outline}

Write $\mathcal N(H)$ for $T$ alternating row and column normalizations of $e^H$, ending with columns, so that every table has unit column sums; the initial tables are $Q_a^0=\mathcal N(\beta_0S_{a1})$ and an update of view $a$ is $Q_a=\mathcal N(\beta F_a)$ with $F_a=S_{a1}+\sum_{c\ne a}S_{ac}Q_c$, read in the Gauss--Seidel order $2,3,4,5$, and $\beta/\Lambda\to\sqrt b$. The field reads $F_a(j,i)=\langle X^a_j,P_i\rangle/\sqrt d$ with prototype $P_i=X_i^1+\sum_{c\ne1,a}\sum_rQ_c(r,i)X_r^c$. Relabel rows by true objects for the analysis.

\begin{lemma}[Squared mass of the initial tables]\label{lem:bar-mass}
For every fixed number of normalizations, $\E\|Q_c^0\|_F^2/n\le n^{-(1-b)/2+o(1)}$, and the same holds for training-coordinate tables with $b$ replaced by the training signal.
\end{lemma}

\begin{proof}[Outline of proof]
For the likelihood kernel $A_{ji}=e^{\ell(x_i,y_j)}/n$, where $\E_{\rm ind}e^{2\ell}=M_2=(1-\rho^2)^{-d}=n^{b+o(1)}$, the row and column sums deviate from one by $O(\sqrt{M_2/n}+M_2/n)$ on average, one normalization step satisfies $\|B'-A\|_1\le2\|B-A\|_1+D(A)$, and $\|Q\|_F^2\le\|\min(A,1)\|_F^2+2\|Q-A\|_1$ for tables with entries in $[0,1]$. The raw kernel differs from the likelihood kernel by row and column factors whose logarithms are $o(\log n)$ uniformly, and each normalization at most doubles the logarithm of the entrywise ratio.
\end{proof}

\begin{lemma}[Wrong scores in rectangles]\label{lem:bar-rect}
Fix $0<\tau<\sqrt2$ and $\varepsilon>0$. There is $\xi>0$ such that for $0<\delta<\xi$ and fixed $\eta\in(0,1/2)$, with probability tending to one, for all nonnegative column-stochastic tables $Q_3,Q_4,Q_5$ (which may depend on all data) and all disjoint sets $A,B$ of at least $\eta n$ labels, some column $i\in A$ has at least $c_\eta n^{1-\tau^2/2-\delta}$ rows $j\in B$ with $F(j,i)>(\tau-4\varepsilon)\Lambda$. On the same event, every column $i\in[n]$ has at least $c\,n^{1-\tau^2/2-\delta}$ rows $j\ne i$ with $F(j,i)>(\tau-4\varepsilon)\Lambda$, for a constant $c>0$; here the rows range over all of $[n]\setminus\{i\}$, not over a prescribed set.
\end{lemma}

\begin{proof}[Outline of proof]
Condition on the four source views; the target rows are $\mu_j+\sigma G_j$ with independent standard $G_j$. For fixed $(i,c,r)$, the events $\{S_{21}(j,i)>\tau\Lambda,\ S_{2c}(j,r)<-\varepsilon\Lambda\}$ are independent in $j$ with probability at most $n^{-\tau^2/2-\varepsilon^2/2+o(1)}$, so by a Chernoff bound and a union over $O(n^2)$ triples, at most $n^{1-\tau^2/2-\xi}$ rows have both. Splitting $A$ into blocks of $n^{\tau^2/2+\delta}$ columns, Sudakov--Fernique and the Gaussian Poincar\'e inequality show that a block has no anchor score above $\tau\Lambda$ in a given row with probability $O(1/\log n)$; independence of rows and a union over at most $4^n$ rectangles give many high anchor scores in some column of $A$ for every $(A,B)$. For a single column $i$, the anchor scores of the rows $j\ne i$ are independent given the sources and exceed $\tau\Lambda$ with probability at least $n^{-\tau^2/2-\delta/2}$, so a Chernoff bound and a union over the $n$ columns give the count over all rows. Since every table has unit column sums, the total weight that the three tables put on scores below $-\varepsilon\Lambda$, summed over these rows, is $o$ of their number, and on the remaining rows the three back-flow terms sum to at least $-4\varepsilon\Lambda$. This argument is deterministic given the two counting events, so it holds for all tables at once.
\end{proof}

\begin{lemma}[True scores]\label{lem:bar-true}
In each of the first two passes, the updated tables give uniform true labels weight tending to zero in probability, and the field of the first updated view satisfies $F(I,I)/\Lambda\to\sqrt b$ in probability for a uniform label $I$.
\end{lemma}

\begin{proof}[Outline of proof]
Three facts are combined. (i) \emph{Row diffusion.} For some $\kappa>0$, after every intermediate column normalization of the initial tables and of the first-pass table of view $2$, a uniform row has largest entry at most $n^{-\kappa}$ with probability tending to one; for the initial tables this follows from Lemma~\ref{lem:bar-mass}, and for view $2$ from a rectangle lower bound on the column denominators that holds simultaneously for all positive row factors. (ii) \emph{Resampling stability.} If one row of a logit matrix changes and both versions have row maxima at most $\eta<1/2$ at every intermediate stage, the column potentials change by at most $4\eta$ per step; a full-matrix change of size $\Delta=o(1)$ changes columns by $O_T(\Delta)$ in $\ell_1$. Resampling the Gaussian residual of a uniform target row therefore changes every table read by the field by $O_\Prob(n^{-\kappa}(\log n)^3)$ in column $\ell_1$, and scoring the original residual against the resampled prototype gives
\[
F(J,I)=\rho\sqrt d\Big(\mathbf 1_{\{J=I\}}+\sum_{c}q_c(J,I)\Big)+O_\Prob(1)+O_\Prob(\rho\Lambda).
\]
(iii) \emph{Weight comparison.} If $F(I,I)/\Lambda\to\sqrt b$ and only $o_\Prob(n^2)$ entries are below $-\delta\Lambda$ for each fixed $\delta$, then after any positive row factors and column normalization the uniform true weight is at most $n^{-\lambda}$ for every $\lambda<1-b$: sorting labels by row factor, a low-factor column has linearly many competitors from high-factor rows with scores at least $-\delta\Lambda$. Induction along the update order, starting from Lemma~\ref{lem:bar-mass}, gives the claims; in the second pass $F_2^2(I,I)-S_{21}(I,I)=o_\Prob(\Lambda)$.
\end{proof}

\begin{proof}[Outline of the proof of Proposition~\ref{prop:barriers}(c)]
Choose $\sqrt b<\tau<\sqrt{2(1-\omega)}$, $\varepsilon$ with $\tau-4\varepsilon>t_0>\sqrt b$ and $\delta<\min\{\xi,1-\tau^2/2-\omega\}$. By Lemma~\ref{lem:bar-true} only $o_\Prob(n)$ true scores exceed $t_0\Lambda$. If a fixed fraction $\gamma$ of the remaining columns kept the true row among the top $W$ after scaling by row factors $r_j$ and column factors, split those labels into low-factor columns $A$ and high-factor rows $B$; Lemma~\ref{lem:bar-rect} gives a column of $A$ with more than $Cn^\omega\ge W$ rows of $B$ whose scores and row factors both beat the true entry, a contradiction. Lemma~\ref{lem:bar-rect} uses only column-stochastic tables, so it applies to the fields of both passes. Completion edges are correct only at fixed points of a uniform relative permutation, $O_\Prob(1)$ of them, so the rounded overlap vanishes. The first target's table is not updated again before rounding in a two-pass run, and in the default run it is updated in the third pass, which is not covered. The second-pass wrong-score bound is Lemma~\ref{lem:bar-rect} for all columns.

For the held-out statement, let $G$ be the training data of the other views, their permutations and the algorithm's randomness. Given $G$, the held-out prototypes have covariance $V=(1-\rho)(I+\sum_cq_c^\top q_c)+\rho\mathcal A^\top\mathcal A$ and cross-covariance $\rho\mathcal A$ with the target, where $\mathcal A=I+\sum_cq_c$. Replacing the cross-covariance by $\rho I$ yields a reference experiment equivalent to a clean two-view problem with correlation $\rho/\sqrt{1-\rho}$ plus independent noise. Whitening by $V^{-1/2}$, both cross blocks have operator norm $O(\sqrt\rho)$, so the relative entropy per coordinate is $O(\rho^2\|\mathcal A-I\|_F^2)$, and by Lemma~\ref{lem:bar-mass} the total is $o(n\log n)$. In the reference experiment, the change-of-measure count in the proof of part~(a), with lists of length one, shows that matching $\eta n$ rows has probability $e^{-\Omega_\eta(n\log n)}$. The data-processing bound $D(\Prob\Vert\mathbb Q)\ge p\log(1/q)-\log2$ then forces the probability of matching $\eta n$ rows under the true law to vanish, for every fixed $\eta$.
\end{proof}

\section{Almost-exact matching with known observation laws}\label{app:f1-known}

Keep the sampling and permutation model of Section~\ref{sec:model} with fixed $K$.
The feature of a view may take values in any measurable space, and the common tuple law $P_n$ on the product of these spaces is known.
Write $P_{n,B}$ for its marginal on $B\subseteq[K]$.
For a partition $\cC$ of $B$ into at least two blocks, set
\[
Q_{B,\cC}=\bigotimes_{A\in\cC}P_{n,A},\qquad
D_{B,\cC}=D(P_{n,B}\Vert Q_{B,\cC}),\qquad
J_B=\min_{\cC}\frac{D_{B,\cC}}{|\cC|-1}.
\]
For singleton $B$, set $J_B=\infty$. As in the main text, $J=J_{[K]}$ and
$J^*_{ab}=\max_{B\supseteq\{a,b\}}J_B$.
Let $f$ and $g$ be the densities of $P_{n,B}$ and $Q_{B,\cC}$ with respect to their sum, and write $L_{B,\cC}=\log(f/g)$.
Take $L_{B,\cC}=-\infty$ on $\{f=0\}$ and $L_{B,\cC}=+\infty$ on $\{f>0,\ g=0\}$.
We use the following condition on true tuples, required for each partition $\cC$ of the indicated set $B$ and every fixed $\eta>0$:
\begin{equation}\label{eq:f1-tail}
P_{n,B}\!\left(L_{B,\cC}<\min\{D_{B,\cC},K\log n\}-\eta\log n\right)=o(1).
\end{equation}

\begin{theorem}[Known observation laws]\label{thm:f1-known}
\begin{enumerate}[label=(\alph*),leftmargin=*]
\item Let $\cC$ partition $[K]$ into $m\ge2$ blocks. Choose a representative $r_A$ of each block $A$, with $r_{A_1}=1$ for the block $A_1$ containing view $1$. For every $n\ge2$ and every estimator,
\begin{equation}\label{eq:f1-converse}
\frac{\sum_{A\ne A_1}\E M_{r_A}}{(m-1)n}
\ge 1-\frac{D_{[K],\cC}}{(m-1)\log n}-\frac{2}{\log n}.
\end{equation}
The bound is vacuous for $D_{[K],\cC}=\infty$. If $\limsup_n J/\log n<1$, simultaneous almost-exact recovery is impossible.

\item On a fixed set $B$ with at least two views, suppose $\gamma_n>0$ satisfies $\gamma_n\log n\to\infty$ and, for every partition $\cC$ of $B$ into at least two blocks,
\begin{equation}\label{eq:f1-test}
P_{n,B}\!\left(L_{B,\cC}\le (1+\gamma_n)(|\cC|-1)\log n\right)=o(1).
\end{equation}
The partition-test decoder with permutation completion is then simultaneously almost exact on $B$.
In particular, $\liminf_n J_B/\log n>1$ and~\eqref{eq:f1-tail} on $B$ suffice.

\item Assume~\eqref{eq:f1-tail} for every subset with at least two views.
If $\liminf_n J^*_{ab}/\log n>1$, a permutation estimate of $\sigma_{ab}$ has overlap tending to one in probability.
If $\limsup_n J^*_{ab}/\log n<1$, then $\sup_f\E\,\ov(f)\to0$, where the supremum is over randomized maps using all views.

\item Suppose $\limsup_n J/\log n<1$ and, for every sufficiently large $n$, some minimizing partition of $[K]$ separates $a$ and $b$.
If~\eqref{eq:f1-tail} holds for the partitions of $[K]$, then $\sup_f\E\,\ov(f)\to0$ for this pair.
\end{enumerate}
\end{theorem}

Parts~(a)--(d) together give Theorem~\ref{thm:known}: condition~\eqref{eq:f1-tail} on every set of at least two views is condition~\eqref{eq:lowertail}.

\subsection{The finite-sample converse and partition tests}

\begin{proof}[Proof of Theorem~\ref{thm:f1-known}(a)]
Assume $D_{[K],\cC}<\infty$. Reveal each within-block relative permutation
$\pi_a\pi_{r_A}^{-1}$ for $a\in A$, collectively denoted by $G$. Use $G$ to align each block to its representative and call the resulting data $Y'$.
Let $U=(\pi_{r_A})_{A\ne A_1}$. The original permutations are in bijection with $(U,G)$, so $U$ is uniform on $\Sym_n^{m-1}$ and independent of $G$. The original data can be reconstructed from $(Y',G)$.

Let $Q$ be the product law of the block databases, with independent rows of law $P_{n,A}$ in each block $A$. Conditional on $(U,G)$, undoing the row permutations of the blocks turns $Y'$ into $n$ independent tuples of law $P_n$, and leaves $Q$ invariant.
Thus $D(\mathcal L(Y'\mid U,G)\Vert Q)=nD_{[K],\cC}$ for each conditioning value.
Averaging over $(U,G)$ and factoring the density through $\mathcal L(Y'\mid G)$ gives
\[
I(U;Y'\mid G)+\E D(\mathcal L(Y'\mid G)\Vert Q)=nD_{[K],\cC}.
\]

Given a map $[n]\to[n]$, fix the $t$ inputs at which a true permutation differs from it. If the retained images repeat, there is no compatible permutation. Otherwise the $t$ free columns admit at most $t!$ assignments. There are at most $\binom nt t!\le n^t$ possible permutations. For a vector $\widehat U$ of estimated maps, encode the error counts followed by the true permutations:
\begin{equation}\label{eq:f1-hamming}
H(U\mid\widehat U,G)
\le (m-1)\log(n+1)+(\log n)\sum_{A\ne A_1}\E M_{r_A}.
\end{equation}
The estimator's random seed is independent, so
$I(U;\widehat U\mid G)\le I(U;Y'\mid G)$.
Combining these inequalities with $H(U\mid G)=(m-1)\log(n!)$,
$\log(n!)\ge n\log n-n$ and $\log(n+1)\le n$ yields~\eqref{eq:f1-converse}.
For $\limsup_n J/\log n<1$, a minimizing partition has a positive fixed gap in~\eqref{eq:f1-converse}. Simultaneous almost-exact recovery would make each bounded error fraction tend to zero in expectation and hence the left side tend to zero.
\end{proof}

\begin{proof}[Proof of Theorem~\ref{thm:f1-known}(b)]
For a fixed partition, abbreviate $Q=Q_{B,\cC}$ and $L=L_{B,\cC}$. With $f,g$ as above,
\[
\E_Q e^L=\int_{\{g>0\}}f\,d(P_{n,B}+Q)\le1,
\qquad Q(L>t)\le e^{-t}.
\]
The singular part of $P_{n,B}$ is supported on $\{L=+\infty\}$, which has $Q$-measure zero.

For each anchor row, the decoder accepts a candidate tuple when it exceeds the threshold in~\eqref{eq:f1-test} for every partition. The true tuple fails with probability $o(1)$ by a union bound over the finitely many partitions. Conditional on the permutations, a false tuple whose rows come from distinct objects across the blocks of $\cC$ and the same object within each block has law $Q_{B,\cC}$. There are at most $n^{|\cC|-1}$ such candidates per anchor row because the anchor object is fixed and one chooses an object for each other block. The probability that any of these candidates is accepted is at most $n^{-\gamma_n(|\cC|-1)}$. Summing over partitions gives $o(1)$.

Each anchor row with an accepted tuple proposes one. A row is good if its true tuple is accepted and every false tuple is rejected. The expected number of bad rows is $o(n)$. In each view, retain one proposal per column and complete the assignment with an arbitrary bijection on the remaining rows and columns. The proposed true columns of the good rows are distinct. Each bad row can displace at most one good row per view. Thus every view has at most twice the number of bad rows in error, which proves simultaneous almost-exact recovery by Markov's inequality.

For the fixed-gap implication, choose a fixed $\gamma>0$ with $J_B\ge(1+2\gamma)\log n$ eventually and $(1+2\gamma)(K-1)\le K$. For $s=|\cC|-1$, the capped divergence in~\eqref{eq:f1-tail} is at least $(1+2\gamma)s\log n$. Taking $\eta=\gamma/2$ places the threshold $(1+\gamma)s\log n$ strictly below the bound in~\eqref{eq:f1-tail}, proving~\eqref{eq:f1-test}.

For finite positive $D=D_{B,\cC}$, the variance condition $\Var_{P_{n,B}} L_{B,\cC}=o(D^2)$ suffices for~\eqref{eq:f1-tail}. The lower-tail event requires a deviation from the mean $D$ of at least $\eta\log n$ when $D\le2K\log n$, and at least $D/2$ otherwise. Chebyshev's inequality gives $o(1)$ in each case. At $D=0$ the information density is zero almost surely.
\end{proof}

For Gaussian laws with finite divergence, Lemma~\ref{lem:llr} gives $\Var_{P_{n,B}}L_{B,\cC}\le 2K^2D_{B,\cC}$, so the two Chebyshev bounds above are $O(1/\log n)$. At infinite divergence the Gaussian support argument in Appendix~\ref{app:main} gives $L_{B,\cC}=+\infty$ almost surely under $P_{n,B}$. Thus~\eqref{eq:f1-tail} holds for Gaussian laws, and Theorems~\ref{thm:main}(a) and~\ref{thm:pairwise} follow from Theorem~\ref{thm:f1-known}.

\subsection{A fixed coordinate law}

\begin{proof}[Proof of Corollary~\ref{cor:product}]
Fix $B,\cC$. Write $p_B$ for the single-coordinate marginal, $q=\bigotimes_{A\in\cC}p_A$, and $\ell$ for their extended log likelihood ratio. The negative part is integrable. Indeed, writing $f$ for the density of the absolutely continuous component of $p_B$ relative to $q$ gives
\[
\E_{p_B}\ell^-=\int_{\{0<f<1\}}f\log(1/f)\,dq\le e^{-1}.
\]
After regrouping by coordinate, the block-product reference for $d$ coordinates is $q^{\otimes d}$. Using $(p_B+q)^{\otimes d}$ as the common dominating measure shows that the row information density is $\sum_{j=1}^d\ell_j$ for independent copies under the true law. Each true summand lies in $(-\infty,+\infty]$.

If $D(p_B\Vert q)<\infty$, both parts of $\ell$ are integrable. The weak law gives $L_{B,\cC}=dD(p_B\Vert q)+o_{\Prob}(d)$.
If $D(p_B\Vert q)=\infty$, then $\ell\wedge A$ is integrable for fixed $A>0$, and $\E(\ell\wedge A)\uparrow+\infty$ as $A\to\infty$. For each fixed finite target choose $A$ so that this mean exceeds the target, then apply the weak law to the truncated sum. The original sum dominates it, giving $L_{B,\cC}/d\to+\infty$ in probability.
Together with $d/\log n\to c\in(0,\infty)$, these limits give~\eqref{eq:f1-tail} for every $B,\cC$.

The same product representation gives $D_{B,\cC}=dD(p_B\Vert q)$ in the extended reals. Hence $J_B(P_n)=dJ_B(p)$ and $J^*_{ab}(P_n)=dJ^*_{ab}(p)$. Applying Theorem~\ref{thm:f1-known}, with its pairwise claims proved in Section~\ref{app:f1-pairwise}, yields both thresholds.
\end{proof}

\subsection{Pairwise matchability}\label{app:f1-pairwise}

\paragraph{Achievability.}
Choose a set $B\supseteq\{a,b\}$ attaining $J^*_{ab}$ and apply the
partition test within $B$, with view $a$ as the anchor. The estimate
for view $b$ is a permutation with overlap tending to one. There are
finitely many choices of $B$, so the conclusion also holds if the
maximizing set changes with $n$.

\paragraph{Coarsening a partition.}
Let $\mathcal C$ refine $\mathcal D$, both partitions of $B$. The KL
chain rule gives
\begin{equation}\label{eq:f1-coarsen}
D_{B,\mathcal C}=D_{B,\mathcal D}
 +\sum_{T\in\mathcal D}
 D\left(P_{n,T}\,\middle\Vert\,
      \bigotimes_{A\in\mathcal C:A\subseteq T}P_{n,A}\right).
\end{equation}
We set $D_{B,\{B\}}=0$. To justify the identity when the left side
is finite, put $P=P_{n,B}$, $V=\bigotimes_{T\in\mathcal D}P_{n,T}$ and
$Q_{\mathcal C}=\bigotimes_{A\in\mathcal C}P_{n,A}$. Projection to $T$
bounds each summand by $D(P\Vert Q_{\mathcal C})$. Thus $dV/dQ_{\mathcal C}$
is a product of the coarse-block marginal density ratios. The zero set
of each factor has zero mass under the corresponding $P_{n,T}$, so
$dV/dQ_{\mathcal C}>0$ $P$-almost surely. Divide $dP/dQ_{\mathcal C}$ by
this product to obtain $dP/dV$. Both logarithms in the quotient are
integrable under $P$, by finiteness of the stated divergences and
integrability of the negative part of each log density ratio.
Integrating proves~\eqref{eq:f1-coarsen}. When the terms on the right
are known to be finite, $P\ll V\ll Q_{\mathcal C}$ gives the same
identity by multiplying the density ratios and integrating their
logarithms.

Suppose $S\subseteq B$ meets $v\ge2$ blocks of a weakest partition
$\mathcal C_B$ with $m_B$ blocks and finite $J_B$. Merge those $v$ blocks.
The coarsened divergence is at least $(m_B-v)J_B$, with value zero
when only one block remains. By~\eqref{eq:f1-coarsen}, the joint
divergence among the $v$ merged blocks against their product is at
most $(v-1)J_B$. Projecting each block onto its intersection with
$S$ can only decrease that divergence. The resulting $v$-block
partition of $S$ therefore gives $J_S\le J_B$. In particular, if a
finite weakest partition of $[K]$ separates $a,b$, every $S$ containing
the pair satisfies $J_S\le J$. Taking $S=[K]$ gives $J^*_{ab}=J$.
This separation property also follows from the structure of optimal
partitions in~\citet[Theorems~5.2--5.3 and Proposition~5.3]{chan2015multivariate}.

\paragraph{Reduction and tree.}
For the converse, take a fixed $\delta>0$ such that
$J^*_{ab}\le(1-2\delta)\log n$ for all sufficiently large $n$.
Suppose that, along a subsequence, estimators that use all the
original observations have expected overlap at least $\eta>0$.
They may be converted into permutation estimates. Given a map
$g:[n]\to[n]$, keep one uniformly chosen preimage of each used
column and complete the partial assignment by a bijection on the
remaining rows and columns. If $g$ has $C$ correct entries, their
columns form a set $S_g$ of size $C$. With $k_j$ denoting the
number of preimages of column $j$, the expected number of correct
entries retained, conditional on $g$ and the truth, is
\[
\sum_{j\in S_g}\frac{1}{k_j}
\ge \frac{C^2}{\sum_{j\in S_g} k_j}
\ge \frac{C^2}{n}
\]
for $C>0$ by Cauchy--Schwarz. The bound is zero for $C=0$.
Jensen's inequality gives a permutation estimate $f$ with
$\E\,\ov(f)\ge \eta^2=:u$.

Starting at $[K]$, split each nonsingleton $B$ with
$J_B\le(1-\delta)\log n$ into the $m_B$ blocks of a weakest
partition $\mathcal C_B$. Every node containing both $a,b$ is split,
so $a$ and $b$ lie in different leaves. Use a priority order with
$a$ first and $b$ second, and choose the highest-priority view as the
representative of each node. One child inherits its parent's
representative. There are finitely many trees for fixed $K$, so pass
to a further subsequence on which the tree is fixed.

Anchor the experiment at $a$ and reveal the true correspondences
$G$ within the leaves. The finite permutation bijection used in the
converse for simultaneous recovery shows that the $r-1$ non-anchor
leaf representative permutations $\Pi_{\mathcal L}$, where $r$
is the number of leaves, are independent and uniform, and are
independent of $G$ and the unshuffled tuples. Reanchoring preserves
these laws because it composes every permutation with the inverse
of the new anchor permutation and reorders the i.i.d.\ tuples.
After aligning within each leaf, the leaf databases and $G$
determine the original observations.

\paragraph{Augmenting a node.}
Fix $\varepsilon>0$, to be chosen below. At each internal node $B$,
for each unordered pair of children and each object, attach the
same uniform label from $[M_B]$ to the two child representatives,
where
\[
M_B=\left\lceil
\exp\left(\frac{2}{m_B}\big((1+\varepsilon)\log n-J_B\big)\right)
\right\rceil .
\]
The labels are independent across nodes, child pairs, and objects,
and independent of the original tuples. Each label moves with
the row of its representative.

Suppose the true alignments within the children of $B$ are given.
The decoder at $B$ uses the aligned original features and the
labels attached at $B$. A partition of the children into $v\ge2$
groups corresponds to a coarsening of $\mathcal C_B$. Its original
divergence satisfies
\[
(v-1)J_B\le D_{\mathrm{old}}\le(m_B-1)J_B<K\log n .
\]
The lower bound follows from the definition of $J_B$ and the
upper bound from~\eqref{eq:f1-coarsen}. The lower-tail
condition~\eqref{eq:f1-tail} therefore gives
$L_{\mathrm{old}}\ge D_{\mathrm{old}}-o_{\Prob}(\log n)$ under the
true law. For a child pair separated by the partition, its two
labels agree under the true law and are independent uniform
labels under the block product. That pair contributes $\log M_B$
to the true information density. A pair inside one group contributes
zero. Hence $e$ separated child pairs contribute $e\log M_B$ to
both the information density on true tuples and the divergence.

For fixed $v$, the number of separated pairs is smallest when one
group has $m_B-v+1$ children and the other groups are singletons.
Indeed, moving a child from any other group of size at least two
to a designated largest group increases the number of within-group pairs.
Consequently
\[
e\ge \binom{m_B}{2}-\binom{m_B-v+1}{2}
  =(v-1)(m_B-v/2)\ge (v-1)m_B/2 .
\]
The combined true information density is at least
$(v-1)(1+\varepsilon)\log n-o_{\Prob}(\log n)$.
The acceptance threshold in~\eqref{eq:f1-test} with
$\gamma_n=\varepsilon/2$ thus gives expected error $o(n)$
in each child representative permutation. When all children are
in different groups, the combined divergence equals
$(m_B-1)(1+\varepsilon)\log n+O_K(1)$ because
$e=\binom{m_B}{2}$ and
$0\le\log\lceil x\rceil-\log x\le\log 2$ for $x\ge1$.

\paragraph{An upper bound on the augmented divergence.}
Let $Y$ include the augmented leaf databases and $G$. Write
$P_{\mathrm{mix}}$ for its planted law and $Q$ for the product of
the leaf database marginals with an independent copy of $G$.
Iterating~\eqref{eq:f1-coarsen} gives
$\sum_B(m_B-1)J_B$ for the original one-object leaf divergence,
where the sum runs over internal nodes. The child representatives
of any internal node lie in distinct leaves. Its label layer
therefore contributes its singleton divergence to the leaf partition.
Since $\sum_B(m_B-1)=r-1$, the augmented one-object leaf divergence is
\[
D^{\mathrm{aug}}_{\mathcal L}
  =(r-1)(1+\varepsilon)\log n+O_K(1).
\]
Conditional on $\Pi_{\mathcal L}$, undoing the shuffles gives $n$
independent augmented tuples, and preserves $Q$. Averaging the
conditional divergence gives
\begin{equation}\label{eq:f1-pairwise-kl}
D(P_{\mathrm{mix}}\Vert Q)+I(\Pi_{\mathcal L};Y)
  =nD^{\mathrm{aug}}_{\mathcal L}.
\end{equation}

Let $U_B$ be the vector of $m_B-1$ relative permutations from the
representative of $B$ to the representatives of its other children.
These vectors determine $\Pi_{\mathcal L}$ by composition along the
tree and are determined by it. Expand
$H(\Pi_{\mathcal L}\mid Y)$ in an order with descendants before
ancestors. At node $B$, conditioning on the preceding vectors
and on $Y$ supplies the alignments within each child from the
descendants' permutations and $G$. The node estimator uses the
original data and the labels added at $B$. The joint marginal law
of its input and $U_B$ is the node experiment just analyzed. This
follows by reanchoring at the representative of $B$ and using
independence of the permutations and the i.i.d.\ augmented tuples.
Choose deterministic tie-breaking and completion. The resulting
$\widehat U_B$ depends only on $Y$ and preceding vectors and has
expected error $o(n)$ in each component. The Hamming count
in~\eqref{eq:f1-hamming} bounds the node's conditional entropy
by $H(U_B\mid\widehat U_B)=o(n\log n)$. There are finitely many
nodes, hence $H(\Pi_{\mathcal L}\mid Y)=o(n\log n)$. Using
$H(\Pi_{\mathcal L})=(r-1)\log(n!)$ in~\eqref{eq:f1-pairwise-kl}
gives
\begin{equation}\label{eq:f1-pairwise-kl-upper}
\frac{D(P_{\mathrm{mix}}\Vert Q)}{n\log n}
 \le (r-1)\varepsilon+o(1).
\end{equation}

\paragraph{Positive overlap forces a lower bound.}
Let $B_0$ be the lowest node containing both $a$ and $b$.
By the priority order, they represent two different children of
$B_0$ and their respective leaves. Let $T$ count equalities of
their shared labels at $B_0$ when rows are paired by the
permutation estimate $f$ computed from the original data.
Under $P_{\mathrm{mix}}$, every correct pairing gives an equality,
so $\E_{P_{\mathrm{mix}}}T\ge nu$. Under each leaf marginal, the
original features and auxiliary labels are independent. Conditional
on the common shuffle, the two independent i.i.d.\ arrays retain
their laws, which do not depend on the shuffle. The leaves are
independent under $Q$. Thus, conditional on the original data, $G$
and the independent estimator seed, the two label arrays are
independent and uniform across rows.
Since $f$ is a permutation, the equality indicators use disjoint
row pairs and $T\sim\operatorname{Bin}(n,1/M_{B_0})$.

Adding the same independent seed to both laws preserves their
divergence. The nonnegativity of KL relative to the law obtained
by tilting $Q$ by $\exp(tT)$ gives
\[
D(P_{\mathrm{mix}}\Vert Q)\ge
t\E_{P_{\mathrm{mix}}}T-\log\E_Q e^{tT}.
\]
For $t=\log M_{B_0}$, the last term is
$n\log(2-1/M_{B_0})\le n\log 2$. Since
$J_{B_0}\le J^*_{ab}\le(1-2\delta)\log n$ and $m_{B_0}\le K$, we obtain
\begin{equation}\label{eq:f1-pairwise-kl-lower}
\frac{D(P_{\mathrm{mix}}\Vert Q)}{n\log n}
 \ge \frac{4u\delta}{K}-o(1).
\end{equation}
Choose $0<\varepsilon<4u\delta/[K(K-1)]$. This contradicts
\eqref{eq:f1-pairwise-kl-upper} since $r\le K$, proving that the
supremum of the expected overlap tends to zero.

\paragraph{Pairs separated by a weakest partition.}
Suppose $J\le(1-2\delta)\log n$ eventually for some fixed $\delta>0$
and a weakest partition separates $a,b$ at each sufficiently large
$n$. Use that partition as the sole split and reveal the
correspondences within its blocks. In the proof above the only
old information densities used by a node are now the densities
for coarsenings of this global partition. Their divergences are
finite and at most $(K-1)J$. The root rate $J$ gives the same
lower bound on $\log M_B$. The label augmentation and
bounds~\eqref{eq:f1-pairwise-kl-upper}
and~\eqref{eq:f1-pairwise-kl-lower} yield vanishing optimal
expected overlap under~\eqref{eq:f1-tail} for the global partitions
alone. This proves the last part of Theorem~\ref{thm:f1-known}.

\subsection{A sparse family with a changing law}\label{app:f1-sparse}

Let every view observe the same vector $X\in\{0,1\}^d$ of independent Bernoulli$(1/n)$ bits. Since the views coincide, every partition into $q$ blocks has $D_\cP=(q-1)H(X)$, so $J=H(X)=d\,h(1/n)$ with $h$ the binary entropy, and $J=(1+o(1))(d/n)\log n$ when $d\ge n$. In particular $J\ge(1+\epsilon)\log n$ once $d\ge\kappa n$ for a fixed $\kappa>1$.

\begin{proposition}\label{prop:f1-sparse}
In this family simultaneous almost exact recovery holds if $d/n\to\infty$. If $d\le cn$ along a subsequence with $c<\infty$ fixed, then for every pair of views and every estimator the expected overlap is at most $1-e^{-c}+o(1)$.
\end{proposition}

\begin{proof}
Objects with the same vector are exchangeable given the data: conditional on the observed multisets, the restriction of the relative permutation to a group of $N$ identical rows is uniform. For any map $f$, the expected number of correct images within the group is at most $1$, since each target row in the group is correct with conditional probability $1/N$. The all-zero group has size $N_0\sim\mathrm{Bin}(n,(1-1/n)^d)$, so along the subsequence $\E N_0/n=(1-1/n)^d\ge(1-1/n)^{cn}=e^{-c}+o(1)$, and at most $1$ of these rows is matched correctly in expectation. This proves the second statement.

If $d/n\to\infty$, match every row whose vector is unique among the $n$ objects to the row with the same vector in each view. A fixed object has the all-zero vector with probability $(1-1/n)^d\to0$. Given that its vector has exactly one $1$, another object shares it with probability at most $n\cdot n^{-1}(1-1/n)^{d-1}\to0$; given that it has $w\ge2$ ones, another object shares it with probability at most $n\cdot n^{-w}\le n^{-1}$. Hence all but $o_\Prob(n)$ objects have unique vectors, and they are matched correctly.
\end{proof}

Here $L_{B,\cC}$ equals $(q-1)\log(1/p(X))$ for the realized vector $X$, which for the all-zero vector is only $(q-1)d\log\frac{n}{n-1}=O(d/n)$, although $J\sim(d/n)\log n$; this is the failure of~\eqref{eq:f1-tail} behind the proposition.

\section{Partially observed objects}\label{app:overlap}

\begin{proof}[Proof of Proposition~\ref{prop:overlap}]
\emph{Converse.} Reveal to the statistician the type $M_i$ of every row and all correspondences outside a fixed type $M$ with $\mu(M)>0$. Given the masks and the row positions of each type, the rows of type $M$ in the views of $M$ are independently and uniformly shuffled copies of $N_M$ i.i.d.\ tuples with law $P_{n,M}$, independent of everything revealed. Any estimator restricted to these rows can be completed to permutations without losing correct pairs, so Theorem~\ref{thm:main}(b), applied with $N_M$ objects, bounds its expected number of correct pairs. With probability tending to one $\mu(M)n/2\le N_M\le n$, so $\log N_M=\log n+O(1)$ uniformly on this event. If $J_M<(1-\epsilon)\log n$, the bound~\eqref{eq:converse} for a weakest partition of $M$ gives $\Omega(n)$ expected missed same-object pairs, and since the normalized loss is bounded it cannot be $o_\Prob(n)$. The pairwise converse is Theorem~\ref{thm:pairwise}(b) applied in the same way with $J^*_{ab}(M)$.

\emph{Decoder.} For every set $B$ of at least two views, test every candidate tuple of rows, one from each view of $B$, against every partition $\cC$ of $B$ into $q\ge2$ blocks at level $(q-1)(\log n+\sqrt{\log n})$, with the support test of Appendix~\ref{app:main} for infinite divergences. Given the masks and the shuffles, a candidate whose rows come from $q$ different objects has the product of the corresponding block marginals as its law, and for each identity partition there are at most $n^q$ such candidates. By Markov's inequality applied to the likelihood ratio of the identity partition, the expected number $F$ of accepted false candidates is $O_K(ne^{-\sqrt{\log n}})=o(n)$. If $J_B\ge(1+\epsilon)\log n$, the variance bound of Lemma~\ref{lem:llr} and Chebyshev's inequality show that the true sub-tuple on $B$ of an object whose type contains $B$ fails some test with probability $O(1/\log n)$.

Join any two rows that lie in a common accepted tuple, keep each connected group that contains at most one row per view, and split the other groups into single rows. A group that contains rows of two objects contains an accepted false candidate, since any path leaving an object passes through one; hence at most $KF$ objects lie in groups that mix objects. Each row has at most $K-1$ same-object partners in the truth and in the estimate, so these objects contribute $O_K(F)$ to the loss. Every other object whose true tuple on $M_i$ was accepted is output exactly. Taking $B=M$ for every positive-mass type gives the first statement, and taking for $B$ a set attaining $J^*_{ab}(M)$ gives the pairwise statement, in expectation $o(n)$ and hence in probability by Markov's inequality.

\emph{The averaged formula.} For a partition $\cP$ of $[K]$ and a type $M$ write $\cP|M$ for the induced partition of $M$, with $q_\cP(M)$ blocks. The natural average is
\[
J_\mu=\min_\cP\frac{\sum_M\mu(M)D_{\cP|M}}{\sum_M\mu(M)(q_\cP(M)-1)_+},
\]
a weighted average of ratios $D_{\cP|M}/(q-1)\ge J_M$, so $J_\mu\ge\min_MJ_M$. Take $K=3$, each view present independently with probability $1/2$, and $d$ i.i.d.\ coordinates whose three views are equicorrelated with correlation $1/2$. Per coordinate the pairwise information is $I=\frac12\log\frac43$ and the total correlation of three views is $T=\frac12\log2>2I$. The singleton partition gives the ratio $(3I+T)/5$ and each bipartition gives $(I+T)/3$, which is larger, so $J_\mu/d=(3I+T)/5\in(I,T/2)$, while $J_{123}/d=T/2$. For $d=\lceil c\log n\rceil$ with $5/(3I+T)<c<1/I$, the average exceeds $\log n$, the type $\{1,2,3\}$ has $J_{123}>\log n$ and is recovered almost exactly, and each two-view type has $J^*_{ab}(M)=dI<\log n$, so every estimator has vanishing overlap on it.
\end{proof}

\section{Unknown covariance}\label{app:unknown}

\subsection{Unknown marginal covariances}

\begin{proof}[Proof of Proposition~\ref{prop:unknown}(a)]
\emph{Calibration.} Reserve $m=\lceil\sqrt{nu_n/c_n}\rceil=o(n)$ predetermined rows in each view and form their sample covariances about the known zero mean. Write $\widehat S_a$ for these estimates and $F_a=S_a^{-1/2}\widehat S_aS_a^{-1/2}-I$. For a fixed unit direction the normalized sample quadratic form is $\chi_m^2/m-1$, and a $\chi^2$ tail bound on a $1/4$-net of size at most $9^{d_a}$ gives $\max_a\|F_a\|_{\rm op}\lesssim\sqrt{(d+\log n)/m}$ with probability $1-o(1)$.

For one view let $A_t=[S^{1/2}(I+tF)S^{1/2}]^{1/2}$ and $T_t=A_t^{-1}S^{1/2}$, $0\le t\le1$. Then $T_t^\top T_t=(I+tF)^{-1}$, so $\|T_t\|_{\rm op}$ is bounded on the calibration event. Differentiating $A_t^2$ in its eigenbasis gives $(A_t^{-1}A'_t)_{ij}=\frac{a_j}{a_i+a_j}(T_tFT_t^\top)_{ij}$, where $a_i$ are the eigenvalues of $A_t$. The multiplier is a logistic average of masks $\mathbf 1\{\log a_i+z\le\log a_j\}$, each a compression of upper triangular truncation in dimension $2d_a$, whose norm is at most $1+\lceil\log_2(2d_a)\rceil$. Integrating $T'_t=-A_t^{-1}A'_tT_t$ gives $\|T_1-I\|_{\rm op}\lesssim\|F\|_{\rm op}\log(2d_a)$. For bounded condition number, the integral representation of the Sylvester equation $A_1Z+ZA_0=S^{1/2}FS^{1/2}$ gives $\|T_1-I\|_{\rm op}\lesssim\kappa(S)\|F\|_{\rm op}$ instead. Hence the block-diagonal map $T=\mathrm{diag}(\widehat S_a^{-1/2}S_a^{1/2})$ satisfies $e=\|T-I\|_{\rm op}=O(\sqrt{u_n/m})$, and $q_n=u_n/(mc_n)=o(1)$.

\emph{Partition scores.} Fix a partition $\cP$ into $p\ge2$ blocks, let $B$ be the block-diagonal part of $R$, $U=R-B$, $H=R^{-1}-B^{-1}$, $C=B^{-1/2}RB^{-1/2}$, $C'=R^{1/2}B^{-1}R^{1/2}$ and $D=D_\cP=-\frac12\log\det C$. Both $R$ and $B$ lie between $c_nI$ and $KI$, and the isospectral matrices $C,C'$ have trace $d$ and eigenvalues at most $p$. Hence $\|C-I\|_F^2\le4K^2D$ and $\|U\|_F\lesssim_K\sqrt D$. The identities
\[
H=R^{-1/2}(I-C')R^{-1/2}=-B^{-1}UB^{-1}+V,\qquad V=R^{-1/2}(C'-I)^2R^{-1/2},
\]
give $\|H\|_F\lesssim_K\sqrt D/c_n$ and $\tr V\lesssim_KD/c_n$.

Condition on the hidden permutations and the reserved rows. At most $Km$ objects occur among the reserved rows; call a tuple clean if it avoids them. A clean false tuple with identity partition $\cP$ has $x\sim N(0,B)$ in true whitened coordinates, and its plug-in score is $L_\cP(Tx)=D-(Tx)^\top H(Tx)/2$. With $F_B=B^{-1/2}(T-I)B^{1/2}$ and $\Delta=F_B+F_B^\top+F_BF_B^\top$, the vector $B^{-1/2}Tx$ has covariance $I+\Delta$, where $\Delta$ is block diagonal with $\|\Delta\|_{\rm op}\lesssim_Ke/\sqrt{c_n}=o(1)$, and
\begin{equation}\label{eq:unknown-mgf}
\log\E_Be^{L_\cP(Tx)}=-\tfrac12\log\det[I+(I-C)\Delta]\le C_KDe^2/c_n,
\end{equation}
because $A=(I-C)\Delta$ has trace zero, $\|A\|_{\rm op}=o(1)$ and $|\tr A^j|\le\|A\|_F^2\|A\|_{\rm op}^{j-2}$ with $\|A\|_F\le\|I-C\|_F\|\Delta\|_{\rm op}$. For a true tuple, $x\sim N(0,R)$; the linear change of the mean vanishes because $RH=-UB^{-1}$ is off the block diagonal while $T-I$ is block diagonal, and the quadratic change is at most $\tr V\,\|EBE^\top\|_{\rm op}+\|H\|_Fe^2\|U\|_F$ with $E=T-I$. Thus the true score has mean $D+O_K(De^2/c_n)$ and variance $O_K(D)$.

\emph{Decoding.} Excluding the reserved rows, accept a tuple if $L_\cP(Tx)\ge(1-\eta_n)D_\cP$ for every partition, with $\eta_n=\sqrt{q_n}+(\log n)^{-1/4}$. If $J\ge(1+\epsilon)\log n$, a clean true tuple is rejected with probability $o(1)$ by Chebyshev's inequality, and by~\eqref{eq:unknown-mgf} and Markov's inequality a clean false tuple with $p$-block identity partition is accepted with probability at most $n^{-(1+\epsilon/2)(p-1)}$, so $o_\Prob(n)$ clean false tuples pass. A maximal set of disjoint accepted tuples then contains $n-o_\Prob(n)$ true tuples: at most $K^2m$ selected tuples touch reserved objects, and each selected false tuple blocks at most $K$ true ones. Completion to permutations keeps these matches.
\end{proof}

\subsection{Unknown noise level}

\begin{proof}[Proof of Proposition~\ref{prop:unknown}(b)]
In the model of~\citet{wee2026geometric}, $\Omega_{ac}$ is a negative multiple of the identity for $a\ne c$ whatever $\tau$ is (Remark~\ref{rem:posterior}), so the maximum-likelihood estimate maximizes $\sum_{a<c}\sum_i\langle Y^a_{\pi_a(i)},Y^c_{\pi_c(i)}\rangle$ and does not depend on $\tau$; Theorem~\ref{thm:exact}(a) therefore applies verbatim, and the converses do not use the decoder.

For the partition tests write $v=1+\tau^2$ and $w=v-1$. Each coordinate of an object's $K$ observations has covariance $wI_K+\mathbf 1\mathbf 1^\top$, and $v\to\infty$ because $\tau^4=d/(b\log n)$ and $d/\log n\to\infty$. Inverting $wI_m+\mathbf 1\mathbf 1^\top$ for $m=K$ and for the blocks of a partition $\cP$ gives, for a tuple $y=(y^a)_{a\in[K]}$ with block sums $s_B=\sum_{a\in B}y^a$,
\[
L_\cP(y;v)=\frac1{2w}\Big[\frac{\|s_{[K]}\|^2}{w+K}-\sum_{B\in\cP}\frac{\|s_B\|^2}{w+|B|}\Big]+\frac d2\,g_\cP(w),
\]
where $g_\cP(w)=\sum_{B\in\cP}\log(w+|B|)-\log(w+K)-(|\cP|-1)\log w$. For $1\le m\le K$ the function $w\mapsto1/(w(w+m))$ has a derivative of absolute value at most $2/w^3$, and $g_\cP'(w)=-w^{-1}\sum_{B}|B|(K-|B|)/((w+|B|)(w+K))$, so $|g'_\cP(w)|\le K^2/w^3$. Let $E_n$ be the event that $\|Y_i^a\|^2\le2vd$ for every row of every view. Since $\|Y_i^a\|^2/v\sim\chi^2_d$ and $d/\log n\to\infty$, a $\chi^2$ tail bound and a union over the $Kn$ rows give $\Prob(E_n)\to1$. On $E_n$ every candidate tuple has $\|s_B\|^2\le|B|\sum_{a\in B}\|y^a\|^2\le2K^2vd$ for every set $B$ of views. Hence, for every partition and every $v'$ with $v'-1\ge w/2$,
\[
\Big|\frac{\partial}{\partial v}L_\cP(y;v')\Big|\le C_K\,\frac{vd}{w^3}=O_K\Big(\frac d{v^2}\Big).
\]
Estimate $v$ by $\widehat v=(nd)^{-1}\sum_i\|Y_i^1\|^2$. Since $nd\,\widehat v/v\sim\chi^2_{nd}$ has variance $2nd$, Chebyshev's inequality gives $|\widehat v/v-1|\le\sqrt{\log n/(nd)}$ with probability $1-o(1)$. This bounds the relative error; the absolute error $|\widehat v-v|$ is larger by the factor $v$, which tends to infinity. On the intersection of the two events and for large $n$, every $v'$ between $v$ and $\widehat v$ has $v'-1\ge w/2$, and the mean value theorem gives, uniformly over all candidate tuples and all partitions,
\[
\big|L_\cP(y;\widehat v)-L_\cP(y;v)\big|\le C_K\,\frac d{v^2}\cdot v\sqrt{\frac{\log n}{nd}}=C_K\sqrt{\frac d{v^2}\cdot\frac{\log n}{n}}\le C_K\sqrt b\,\frac{\log n}{\sqrt n}=o(1),
\]
because $d/v^2\le d/\tau^4=b\log n$. Neither event refers to a candidate, so reusing the data for $\widehat v$ is harmless, and the proof of Theorem~\ref{thm:main}(a) goes through for the plug-in scores with every threshold shifted by $o(1)$.
\end{proof}

\subsection{Unknown orthogonal cross-covariance}

Let $\Pi$ be uniform on $\Sym_n$ and, independently, let $O$ be Haar on $O(d_0)$, fixed across rows. Given $O$, the matched pairs are independent with covariance $\Sigma_O=\big(\begin{smallmatrix}I&\rho O\\\rho O^\top&I\end{smallmatrix}\big)$. Write $\ov$ for the fraction of rows with $\widehat\Pi(i)=\Pi(i)$.

\begin{lemma}\label{lem:haar}
For every estimator, possibly randomized and knowing $\rho$, and $r=\lceil\alpha n\rceil$ with $0<\alpha\le1$,
\[
\Prob(\ov\ge\alpha)\le\min\Big\{1,\ \frac{(1-\rho^4)^{-n(n+1)/4}}{\sqrt{r!}}\Big\}.
\]
\end{lemma}

\begin{proof}
Let $Q_0$ be the law of two independent standard Gaussian views. Fix $\Pi=\pi$ and align the rows under $\pi$. Put $A_O=\Sigma_O-I$ and let $\ell_O$ be the density ratio of one pair. The precision $H=\Sigma_O^{-1}+\Sigma_{O'}^{-1}-I$ of the Gaussian integral of $\ell_O\ell_{O'}$ satisfies $H\succeq[2/(1+\rho)-1]I\succ0$ and $\Sigma_OH\Sigma_{O'}=I-A_OA_{O'}$, whose diagonal blocks $I-xOO'^\top$ and $I-xO^\top O'$, $x=\rho^2$, have equal positive determinants. So the one-pair integral is $\det(I-xO^\top O')^{-1}$, and averaging over independent Haar $O,O'$,
\[
1+\chi^2(P_{Y\mid\Pi=\pi}\Vert Q_0)=\int_{O(d_0)}\det(I-xU)^{-n}\,dU,
\]
the integral of~\citet[equation~(3.34)]{li2025shuffled}. Let $\mathcal H_k$ be the homogeneous polynomials of degree $k$ in $n$ vectors of $\R^{d_0}$. Then $\det(I-xU)^{-n}=\sum_kx^k\tr(U\mid\mathcal H_k)$, termwise integration is justified by $\sum_kx^k\dim\mathcal H_k=(1-x)^{-nd_0}$, and Haar averaging projects each $\mathcal H_k$ onto its invariants. By the first fundamental theorem of invariant theory for the orthogonal group~\citep{weyl1939classical,lehrer2012second}, the invariants are generated by the $M=n(n+1)/2$ inner products, of degree two, so the integral is at most $(1-x^2)^{-M}$ coefficientwise.

Under the reference $\mathrm{Unif}(\Pi)\otimes Q_0$, conditionally on the data and the estimator's randomness, a given set of $r$ rows is matched correctly with probability at most $(n-r)!/n!$, so $\ov\ge\alpha$ has reference probability at most $1/r!$. The squared likelihood ratio of the joint law of $(\Pi,Y)$ has reference expectation at most $(1-x^2)^{-M}$, and the Cauchy--Schwarz inequality gives the lemma.
\end{proof}

\begin{proof}[Proof of Proposition~\ref{prop:unknown}(c)]
Take $\alpha=1$: since $\log n!=n\log n+O(n)$ and $-\log(1-\rho^4)=\rho^4+O(\rho^8)$, the condition $n\rho^4\le(2-\epsilon)\log n$ gives the bound $\exp[-(\epsilon/4+o(1))n\log n]$. If $n\rho^4=o(\log n)$, the numerator is $e^{o(n\log n)}$ while $r!=e^{(\alpha+o(1))n\log n}$, so $\Prob(\ov\ge\alpha)\to0$ for each fixed $\alpha$. Both bounds are averages over $O$, so for every estimator they also hold along a sequence of fixed matrices $O$. In the final example $J=-\frac{d_0}2\log(1-\rho^2)\sim b\log n$ and $n\rho^4=4b^2n(\log n)^2/d_0^2=o(\log n)$.
\end{proof}

\providecommand{\Part}{\mathcal P}\providecommand{\Cov}{\operatorname{Cov}}
\section{Proof of the general converse}\label{app:f2-converse}

\begin{lemma}[A moment bound under the geometric tilt]\label{lem:f2-tilt}
Let \(f,g\) be probability densities, \(t\in[1/K,1-1/K]\), and
\(Z=\int f^t g^{1-t}>0\). Under the law \(Q\) with density
\(f^t g^{1-t}/Z\), put \(Y=\log(f/g)\) on the common support. With
\(b=-\log Z\) one has \(\E_QY^2\le C_K(b+b^2)\).
\end{lemma}

\begin{proof}
Writing \(\beta=1-t\), direct integration gives
\(\E_Q e^{\beta Y}\le e^b\) and \(\E_Q e^{-tY}\le e^b\).
The integrals can be restricted to the common support. The function
\(t e^{\beta y}+\beta e^{-ty}\) equals \(1\) at zero, has derivative
zero there, and has second derivative at least
\(\min(t\beta^2,\beta t^2)>0\) for all \(y\).
Taking expectations gives \(\E_QY^2=O_K(b)\) if \(b\le1\).
If \(b>1\), the same exponential moments imply
\(\Prob_Q(|Y|>s)\le 2e^{b-s/K}\). Integrating the tail of \(Y^2\)
from \(s=K(b+\log 2)\), and bounding the lower part by the
square of that cutoff, gives \(O_K(b^2)\).
\end{proof}

\subsection{Transpositions across a cut}\label{app:f2-swap}

Reveal the within-side alignments and combine each side into a single
view. The original permutations can be written bijectively as the
within-side relative permutations and the cross-side permutation of
two representative views. The latter remains uniform and independent of
the features after the revelation. Recovering any cross-side pair with
the revealed alignments would determine this permutation.

By relabeling, assume the true permutation is the identity.
Write \(A=Z_S^*Z_S\). The affinity of the square roots of the
true and swapped densities on two objects is
\(\tr(A^2)=e^{-\chi}\). Indeed, a cycle of length \(k\) pairs each
left feature with two successive right features in the square-root
integrand. Integrating out the left features gives the cyclic integral
of \(k\) copies of the kernel
\(A(y,y')=\int\sqrt{p(x,y)p(x,y')}\,dx\), namely \(\tr(A^k)\).
The eigenvalues of \(A\) are nonnegative and sum to one, so \(\chi\)
is finite for any fixed density.

Let \(f\) and \(g\) be the single-coordinate true and swapped
densities. Under their normalized midpoint tilt, the log ratio
\(\log(g/f)\) is symmetric since the swap is an involution.
Lemma~\ref{lem:f2-tilt} bounds the variance of its \(d\)-coordinate
sum \(\Lambda\) by \(C_Kd(\chi+\chi^2)=o((\log n)^2)\).
Choose \(s_n=o(\log n)\) so that the tilted probability of
\([-s_n,s_n]\) is at least \(3/4\), using Chebyshev's inequality.
The tilted probability of \([0,s_n]\) is then at least \(3/8\).
Changing back to the true law, a specified swap ties or beats the
truth with probability at least
\[
  a_n\ge \tfrac38 e^{-d\chi-s_n/2}.
\]

Swaps on disjoint objects are independent under the truth. For two
swaps sharing one object, their joint event has probability at most
the affinity between the two swapped densities: on the event, the
true density is bounded by their geometric mean. The relative
permutation of the two swaps is a three-cycle, with single-coordinate
affinity \(\tr(A^3)\le(\tr(A^2))^{3/2}\).
If \(N\) counts swaps tying or beating the truth, then
\[
  \E N=\binom n2 a_n\to\infty,\qquad
  \frac{\Var N}{(\E N)^2}
       \le\frac1{\E N}+O\bigl(n^{-1}e^{d\chi/2+s_n}\bigr)=o(1)
\]
under the fixed margin. Thus \(N\to\infty\) in probability. Uniform
tie-breaking among MAP maximizers is Bayes optimal and equivariant
under row relabeling. Given the truth and the data, its success
probability is at most \(1/(1+N)\): if the truth maximizes the
likelihood, all \(N\) counted swaps are also maximizers. This upper
bound tends to zero in expectation.

\subsection{Rare subpopulations}\label{app:f2-rare}

For a \(q\)-block partition and any product reference \(V\), let
\(\alpha=(q-1)/q\), \(\beta=1/q\), and
\(r=D_\alpha(P\|V)/(q-1)\). We show that recovery fails if
\begin{equation}\label{eq:f2-rare-condition}
  dr\le(1-\epsilon)\log n,\qquad d(r+r^2)=o((\log n)^2).
\end{equation}
With \(Z=\int p^\alpha v^\beta\), let \(Q\) have density
\(p^\alpha v^\beta/Z\). On the common support, the log ratios
\(\log(dQ/dP)\) and \(\log(dQ/dV)\) are affine in \(\log(p/v)\)
with slopes \(-\beta\) and \(\alpha\), respectively, and common
constant \(-\log Z=\alpha r\). Thus for
\(D_0=D(Q\|P)\) and \(D_1=D(Q\|V)\),
\[
  D_0+\frac{D_1}{q-1}=r.
\]
The variances of both log ratios are \(O_K(r+r^2)\) by
Lemma~\ref{lem:f2-tilt}.

Fix \(\eta=\epsilon/16\). Let \(W\) denote the event for a row that
its two \(d\)-coordinate log ratios are at most their respective
means \(dD_0,dD_1\) plus \(\eta\log n\). By
\eqref{eq:f2-rare-condition} and Chebyshev,
\(Q^{\otimes d}(W)=1-o(1)\). Independently for each row, mark it
with conditional probability
\[
  \mathbf 1_W \exp\!\left(
     \log\frac{dQ^{\otimes d}}{dP^{\otimes d}}
       -dD_0-\eta\log n\right).
\]
This is at most one by the first constraint in \(W\). The conditional
law of a marked row is \(Q^{\otimes d}(\,\cdot\mid W)\). The marked
count \(m\) is binomial with mean
\(n e^{-dD_0-\eta\log n}Q^{\otimes d}(W)\ge
n^{\epsilon-\eta}(1-o(1))\), because \(D_0\le r\). Consequently,
with probability \(1-o(1)\),
\(\log m=\log n-dD_0-\eta\log n+O(1)\).

Reveal the mark pattern in the anchor, the within-block alignments,
the marked positions in each representative view, and all unmarked
correspondences. Also condition on the unmarked data. The uniform
permutations split bijectively into the within-block relative
permutations and the representative permutations. Conditional on
the disclosed parts, the remaining \(q-1\) bijections between marked
positions are independent and uniform. Because marks are generated
independently by row before permutation, the marked latent rows are
still independent with law \(Q^{\otimes d}(\,\cdot\mid W)\) and are
independent of these bijections.

Compare this conditional experiment with one that draws the marked
latent rows from \(V^{\otimes d}\), with the same uniform bijections.
Block independence under \(V\) makes the shuffled data independent of
the remaining bijections, so the reference success probability is
at most \((m!)^{-(q-1)}\). The joint likelihood ratio on the marked
latent rows and permutations is bounded by
\(\bigl[e^{dD_1+\eta\log n}/Q^{\otimes d}(W)\bigr]^m\).
The second constraint in \(W\) gives this bound row by row.
Using \(\log(m!)\ge m\log m-m\), the log of the true conditional
success probability is at most
\[
  m\bigl[(q-1)(dr-\log n)+q\eta\log n+O_K(1)\bigr]
\]
on the high-probability count event. The bracket is bounded above by
a negative constant times \(\log n\). The exceptional count event
contributes \(o(1)\) to the unconditional success probability.

\subsection{Combining the two obstructions}

If the cut term attains \(L(P_n)\), apply the transposition bound.
If a partition term attains it, choose a product reference whose
rate \(r\) exceeds the infimum by at most
\(\min\{1,\epsilon\log n/(2d)\}\). This preserves a fixed margin in
\eqref{eq:f2-rare-condition}. For fixed \(P\), the active cut
information or near-optimal reference rate is bounded, since there
are finitely many cuts and partitions. Its product with \(d\) is
\(O(\log n)\), which also gives the required second-moment condition.
For varying \(P_n\) and \(d\to\infty\), either active quantity \(a\)
satisfies \(d(a+a^2)=O(\log n+(\log n)^2/d)=o((\log n)^2)\).
The bounds apply on the corresponding indices and prove
Theorem~\ref{thm:f2-converse}.

For completeness, the two-block R\'enyi rate is already at least
the corresponding transposition rate. The supremum of the affinity
over product references at order \(1/2\) is \(\|Z_S\|_{\mathrm{op}}\):
the square roots of the reference densities are nonnegative unit
vectors, and taking absolute values cannot reduce a pairing with
\(Z_S\). Thus \(R_{S|S^c}=-\log\lambda_{\max}(Z_S^*Z_S)
\ge I_2^\circ(S)/2\).

\section{Proof of the weak-dependence threshold}\label{app:f2-weak}

In this section \(h=h_n=o(1)\). All \(O_K(h)\) bounds are uniform
over cuts, partitions, candidates, and underlying coordinate laws.

\subsection{Partition divergences and their quadratic approximation}
\label{app:f2-quadratic}

Let \(g=\log(dP/dU)\) and decompose \(g=\sum_{A\subseteq[K]}g_A\)
by the orthogonal ANOVA decomposition under \(U\). Explicitly,
\[
  g_A=\sum_{C\subseteq A}(-1)^{|A|-|C|}
                         \E_U[g\mid X_C].
\]
Each nonconstant component integrates to zero in any of its
variables and has sup norm \(O_K(h)\). Put
\[
  c(\Part)=\sum_{A:\,A\text{ meets at least two blocks of }\Part}
                 \|g_A\|_{2,U}^2 .
\]
Write \(V=V_\Part^0\), \(t=\log(dP/dV)\), and \(z=e^t-1\).
Marginalization under \(U\) gives \(dV/dU=1+O_K(h)\), so
\(\|t\|_\infty=O_K(h)\). Since \(P\) and \(V\) have the same
block marginals, \(\E_V[z\mid X_B]=0\) for each block \(B\).
The bound \(|t-z|\le C_Kh|t|\), followed by conditional Jensen,
gives \(\|\E_V[t\mid X_B]\|_{2,U}\le C_Kh\|t\|_{2,U}\).
The conditional law of the other blocks under \(V\) has density
\(1+O_K(h)\) relative to their law under \(U\). The same bound
therefore holds with \(\E_U[t\mid X_B]\).

The cross-block projection of \(t\) is exactly the cross-block
projection of \(g\), since \(\log(dV/dU)\) is a sum of block
functions. The squared norm of the within-block and constant
projection is at most \(\sum_B\|\E_U[t\mid X_B]\|_{2,U}^2\).
It follows that \(\|t\|_{2,U}^2=(1+O_K(h))c(\Part)\).
Under \(V\), \(\E_V z=0\). Expanding
\((1+z)\log(1+z)\) and \((1+z)^\alpha\), and using
\(z^2=(1+O_K(h))t^2\), gives
\begin{equation}\label{eq:f2-quadratic}
  D_\Part=(1+O_K(h))\frac{c(\Part)}2,\qquad
  D_\alpha(P\|V)=(1+O_K(h))\alpha D_\Part
\end{equation}
for the finitely many orders \(\alpha=(q-1)/q\) used here.
The Taylor remainders are bounded by \(O_K(h)\E_V z^2\).
These are relative estimates even if the divergence of a particular
partition is much smaller than \(h^2\).

\subsection{Converse at the partition threshold}

If \(dB(P)\le(1-\epsilon)\log n\), take a minimizing partition
with \(q\) blocks and use its marginal product \(V_\Part^0\) as
the reference in Appendix~\ref{app:f2-rare}. By
\eqref{eq:f2-quadratic}, the reference rate there satisfies
\[
  r=\frac{D_{(q-1)/q}(P\|V_\Part^0)}{q-1}
    =(1+O_K(h))\frac{D_\Part}{q}.
\]
In particular \(dr\le(1-\epsilon/2)\log n\) eventually, and
\(r=O_K(h^2)\), so \(d(r+r^2)=O_K(\log n)\).
The reference-specific converse gives success probability \(o(1)\),
including when \(d\) is bounded.

\subsection{Likelihood scores and joint concentration}
\label{app:f2-joint}

Relabel the rows so that the truth is the identity. In a candidate
alignment, the views of each source row \(i\) are partitioned by
the candidate tuples to which they are assigned. Call the list of
these partitions the profile \(H\). A source row is split when
its partition has at least two blocks. A wrong profile has \(m\ge2\)
split rows. If the number of blocks at a split row is \(q_i\), put
\(Q_H=\sum_i(q_i-1)\), summing over split rows. The number of profiles
with \(m\) split rows is at most \(\binom nm\mathrm{Bell}(K)^m\).
Every nonsplit row must stay in its own tuple because its anchor
view is fixed. Within the resulting set of \(m\) tuples, the block
containing the anchor has a known destination, and each other
block has at most \(m\) destinations. Thus a profile has at most
\(m^{Q_H}\) candidates.

At one coordinate, the likelihood difference of a candidate
\(\tau\) and the truth is \(F_\tau-T_H\). Here \(T_H\) sums the
true interaction terms \(g_A\) lost under the candidate, and
\(F_\tau\) sums the new candidate terms. The view marginal
likelihoods cancel because each view's rows are permuted.
The lost terms are exactly those that cross a source row's
partition. For each \(A\), the number of lost terms equals
the number of new terms, because both alignments have \(n\)
terms of that type and the common terms cancel.

Under the independent reference \(U^{\otimes n}\), the remaining
terms are centered and pairwise orthogonal. Their sets of
\((\text{source row},\text{view})\) entries are distinct, so a
variable present in only one of two terms integrates to zero.
Consequently both marginal score variances are
\(A_H=\sum_i c(\Part_i)\), where the sum is over split rows,
and their covariance is zero.

Under the true law, connect two local terms if they share a source
row. This is a dependency graph with degree bounded in terms of
\(K\). Each source row supplies entries to at most \(K\) candidate
tuples, and each tuple has at most \(2^K\) interactions. The true
density on the at most \(2K\) source rows involved in a local term
or an adjacent pair is \(1+O_K(h)\) relative to independence.
Thus a term's variance changes by at most \(O_K(h)\) times its
reference variance, and an adjacent covariance is bounded by
\(O_K(h)\) times the product of the reference standard deviations.
Summing by \(2ab\le a^2+b^2\) and the degree bound gives
\[
  \Var(xF_\tau+yT_H)=(1+O_K(h))(x^2+y^2)A_H.
\]
For \(|x|,|y|\le2\), the centered local terms in this linear
combination have sup norm \(O_K(h)\), and the sum of their
variances is \(O_K((x^2+y^2)A_H)\).
Lemma~\ref{lem:f2-dep} therefore gives the two-sided estimate
\begin{equation}\label{eq:f2-joint}
 \log\E\exp\!\left\{x(F_\tau-\E F_\tau)+y(T_H-\E T_H)\right\}
 =\frac{(x^2+y^2)A_H}{2}\bigl(1+O_K(h)\bigr).
\end{equation}

The true and candidate laws have positive densities relative to
the independent reference. Hence \(\E e^{F_\tau-T_H}=1\).
Taking \((x,y)=(1,-1)\) in \eqref{eq:f2-joint} yields
\(\E(T_H-F_\tau)=(1+O_K(h))A_H\), uniformly over the profile.
For the \(d\)-coordinate scores, let \(\mu\) be the minimum of the
mean differences within the profile. Let \(T\) be the centered
sum of the true terms plus \(\mu\), and \(F\) be the centered
candidate sum. If the candidate ties or beats the truth, then
\(F\ge T\). This choice makes \(T\) common to all candidates
in the profile. By independence across coordinates,
\[
  \log\E e^{xF+y(T-\mu)}\le \tfrac12 V(x^2+y^2)
  \quad (|x|,|y|\le2),\qquad V=d(1+C_Kh)A_H.
\]
In the supercritical regime \(dB(P)\ge(1+\epsilon)\log n\),
\eqref{eq:f2-quadratic} and \(\sum_i q_i=m+Q_H\) imply
\begin{equation}\label{eq:f2-effective}
  D_H:=\mu^2/V\ge 2(1+\gamma)(m+Q_H)\log n,\qquad
  0<\mu/V\le2
\end{equation}
for some fixed \(\gamma>0\) and all sufficiently large \(n\).

\subsection{Two-level counting}\label{app:f2-count}

Put \(w=m\log n\), \(r=Q_H/m\ge1\), \(\theta=\log m/\log n\),
\(\ell=(1-\theta)w\), and \(M=\theta r w\).
The number of profiles is bounded by \(e^{\ell+O_K(m)}\), and
each profile has at most \(e^M\) candidates.
Choose \(0<\delta\le\gamma/2\).

If \(M\le\ell\), then \(\theta\le1/(1+r)\), and
\[
  w\le\ell+M=[1+\theta(r-1)]w
       \le\tfrac{2r}{1+r}w\le\tfrac{1+r}{2}w.
\]
For a single candidate, use \(x=-y=\mu/(2V)\) in the joint
bound to get \(\Prob(F\ge T)\le e^{-D_H/4}\).
By \eqref{eq:f2-effective}, a union over its profile is at most
\(e^{-\ell-\delta w}\).

If \(M>\ell\), the identity \(\ell+M/r=w\) and
Cauchy--Schwarz give
\[
 (\sqrt{\ell+\delta w}+\sqrt{M+\delta w})^2
 \le(1+r)(1+\delta+\delta/r)w\le D_H/2.
\]
Let \(a=\sqrt{2(\ell+\delta w)/D_H}\) and \(s=1-a\).
Since \(\ell<rw/(1+r)\), \eqref{eq:f2-effective} gives
\(a^2\le(1/4+\delta/2)/(1+\gamma)\le1/4\), hence \(s\ge a\).
The common event \(T<s\mu\) has probability at most
\(e^{-D_Ha^2/2}=e^{-\ell-\delta w}\), by taking
\(x=0,\ y=-a\mu/V\). On \(F\ge T\ge s\mu\) one has
\[
  sF-a(T-\mu)\ge(s^2+a^2)\mu .
\]
Take \(x=s\mu/V,\ y=-a\mu/V\) to bound this event by
\(e^{-D_H(s^2+a^2)/2}\le e^{-\ell-M-2\delta w}\).
The last inequality follows from
\(s\sqrt{D_H/2}\ge\sqrt{M+\delta w}\).
All parameters lie in the range of the joint moment bound
by \eqref{eq:f2-effective}. Summing the candidate bounds
and counting the common lower tail once gives at most
\(2e^{-\ell-\delta w}\) per profile.

In both regimes, summing over all profiles and all \(m\ge2\) yields
\[
2\sum_{m=2}^n e^{O_K(m)-\delta m\log n}=o(1).
\]
Thus no wrong candidate ties or beats the truth with probability
\(1-o(1)\), which proves the maximum likelihood upper bound.

\subsection{Comparison of the two rates}\label{app:f2-compare}

For a cut \(S\), let \(Y\) be the log likelihood ratio of a
swap to the truth on two objects at one coordinate. Then
\(\|Y\|_\infty=O_K(h)\). Under \(U^{\otimes2}\), the lost and
new terms are orthogonal, each contributing variance
\(2c(S|S^c)\). Changing to the true law thus gives
\(\E Y^2=(1+O_K(h))\,4c(S|S^c)\). Since \(\E e^Y=1\),
Taylor expansion gives \(\E Y=-\tfrac12\E Y^2+O_K(h)\E Y^2\)
and \(\E e^{Y/2}=1-\tfrac18(1+O_K(h))\E Y^2\).
Taking the negative log and using \eqref{eq:f2-quadratic} yields
\[
  I_2^\circ(S)=(1+O_K(h))D_{S|S^c}.
\]

For a partition of \(q\) blocks and \(\alpha=(q-1)/q\), we also
need the lower bound on the optimized R\'enyi rate. Work relative
to \(U\), and let \(v_B\) be a block reference density relative to
\(U_B\). Holding the other block densities fixed, the affinity is
maximized in \(v_B\) by taking \(v_B\) proportional to
\[
 \left(\int (dP/dU)^\alpha
            \prod_{C\ne B} v_C^{1-\alpha}\,dU_{-B}\right)^{1/\alpha}.
\]
This is equality in H\"older's inequality. Before the power,
the expression in parentheses is \(1+O_K(h)\) times a positive
constant independent of \(X_B\). Updating each block once therefore
gives \(v_B=1+O_K(h)\) for every block, without decreasing the
affinity. Thus the optimum can be approached using only uniformly
bounded references of this form. For such \(V\), the cross-block
projection of \(\log(dP/dV)\) is the cross-block projection of \(g\).
Expanding \(D_\alpha(P\|V)\) as above therefore gives the lower
bound \((1-O_K(h))\alpha c(\Part)/2\). The marginal-product
reference gives the matching upper bound by
\eqref{eq:f2-quadratic}. Hence
\(R_\Part(P)=(1+O_K(h))D_\Part/q\).
Combining the cut and partition comparisons proves
\(L(P)=(1+O_K(h))B(P)\).

\section{A dependency-graph moment bound}\label{app:f2-mgf}

\begin{lemma}\label{lem:f2-dep}
Let centered real variables \(Y_i\) have a dependency graph of
maximum degree \(\Delta\), and suppose \(|Y_i|\le b\).
A dependency graph means that two disjoint vertex sets with no
edge between them index independent families. Put
\(W=\sum_iY_i\) and \(S=\sum_i\E Y_i^2\). There are constants
\(c_\Delta,C_\Delta>0\) such that, for real \(t\) with
\(|t|b\le c_\Delta\),
\[
 \left|\log\E e^{tW}-\tfrac12t^2\Var W\right|
       \le C_\Delta b|t|^3 S .
\]
\end{lemma}

\begin{proof}
The case \(b=0\) is immediate. For a subset \(U\) of the graph
vertices and complex \(z\), put
\[
 Z_U(z)=\E e^{z\sum_{i\in U}Y_i},\qquad
 w(A)=\E\prod_{i\in A}(e^{zY_i}-1).
\]
Expand \(Z_U\) and factor the terms by the connected components
of the chosen vertex set. For \(i\in U\), grouping by the
component containing \(i\) gives the recursion
\[
 Z_U=Z_{U\setminus\{i\}}+
       \sum_{\substack{i\in A\subseteq U\\A\text{ connected}}}
            w(A)Z_{U\setminus N[A]},
\]
where \(N[A]\) is the closed neighborhood of \(A\).
With \(M_0=\Delta+1\), the number of connected \(k\)-sets
through one root is at most \(M_0^{2(k-1)}\). A traversal
of a spanning tree that returns to the root records such a set
in \(2(k-1)\) steps.

Set \(R=(4\cdot2^{M_0+1}M_0^2 b)^{-1}\). For \(|z|\le R\),
\(|w(A)|\le (2|z|b)^{|A|}\). Inductively on \(|U|\),
every deletion ratio \(Z_U/Z_{U\setminus\{i\}}\) lies within
\(1/2\) of one. Indeed, the ratios on smaller sets give
\(|Z_{U\setminus N[A]}/Z_{U\setminus\{i\}}|\le2^{M_0|A|}\)
by deleting one vertex at a time. The recursion and the
connected-set count then bound the distance to one by
\[
 M_0^{-2}\sum_{k\ge1}(M_0^2 2^{M_0+1}|z|b)^k\le 1/3.
\]
This proves the induction and the absence of zeros in the disk.

Write \(\sigma_i^2=\E Y_i^2\). Centering improves the one-vertex
bound to \(|w(\{i\})|\le C|z|^2\sigma_i^2\). For \(|A|\ge2\),
applying Cauchy--Schwarz to two factors and bounding the rest
uniformly gives
\[
  |w(A)|\le C|z|^2(2|z|b)^{|A|-2}\sum_{i\in A}\sigma_i^2.
\]
Sum the logs of deletion ratios along a fixed deletion order,
using \(|\log(1+u)|\le2|u|\) for \(|u|\le1/2\).
Each connected set appears only when its first vertex is deleted.
For a given size \(k\), the sum of \(\sum_{i\in A}\sigma_i^2\)
over connected \(k\)-sets is at most \(M_0^{2(k-1)}S\).
Summing the resulting geometric series gives
\(|\log Z_U(z)|\le C_\Delta |z|^2 S\) in the disk, with the
logarithm chosen to vanish at zero. For the full vertex set,
the analytic function \(z^{-2}\log Z_U(z)\) takes the value
\(\Var W/2\) at zero. Cauchy's coefficient estimate on
\(|z|=R/2\) bounds its difference from that constant at
\(|t|\le R/4\) by \(C_\Delta |t|S/R\).
Multiplication by \(t^2\) proves the lemma.
\end{proof}

\section{Additional remarks}\label{app:remarks}

\subsection{A near-linear Gaussian constraint}\label{app:nearconstraint}

We prove the threshold in Remark~\ref{rem:general} for fixed
\(c>0\). Write \(a=K/(K-1+\delta)\).
A principal submatrix of size \(r<K\) has
eigenvalues \(a\), with multiplicity \(r-1\), and
\(a[1-(1-\delta)r/K]\). Its log determinant is \(O_K(1)\). The full
covariance has determinant \(a^K\delta\). The determinant formula in
Section~\ref{sec:model} thus gives, for every nontrivial partition,
\[
D_\cP=-\tfrac12\log\delta+O_K(1)=c\log n+O_K(1).
\]
The minimum over bipartitions gives \(I_*=c\log n+O_K(1)\). The
normalization by the number of blocks minus one is smallest to leading
order for the singleton partition, so
\(J=c\log n/(K-1)+O_K(1)\). If \(c<\max\{2,K-1\}\), either the
converse in Theorem~\ref{thm:exact} applies or Theorem~\ref{thm:main}
rules out simultaneous almost exact recovery. For the latter case,
the stronger conclusion that exact success tends to zero also follows
from the likelihood bound below.

Reveal the within-block alignments of a minimizing partition with
\(p\) blocks. The remaining hypothesis is uniform over
\(N=(n!)^{p-1}\) cross-block permutations. Under a reference law with
independent blocks and rows, the hypothesis is independent of the data.
The true law relative to this reference has log likelihood ratio \(L\)
equal to a sum of \(n\) independent copies of \(L_\cP\) under the true
law. For any \(t>0\), splitting the success event according to
\(L\ge\log N-t\) bounds its probability by
\(\Prob(L\ge\log N-t)+e^{-t}\). Lemma~\ref{lem:llr} bounds
\(\Var(L)\le2K^2 nD_\cP\). If \(J<(1-\epsilon)\log n\), use
\(t=\epsilon(p-1)n\log n/3\) and
\(\log(n!)=n\log n-O(n)\). Chebyshev's inequality makes the first
probability tend to zero as well.

For the upper side, relabel the true alignment to the identity.
For a competing alignment \(\widetilde\Pi\), form the graph on row
indices with an edge between \(i\) and \(\widetilde\pi_a(i)\) for
every view \(a\). The
likelihood difference is additive across its connected components,
since each permutation preserves each component. A competitor with
likelihood at least the truth must have a moved component of size
\(m\ge2\) whose likelihood is at least that of the truth on the
component.

For such a component let \(U,V\) be the matrices whose columns carry
the true and candidate constraint directions, respectively. Each
column has \(K\) entries equal to \(K^{-1/2}\) at the observations
placed in one tuple and is zero elsewhere. Hence \(U^\top U=V^\top V=I_m\)
and \(M=U^\top V=K^{-1}\sum_{a=1}^K T_a\), where \(T_a\) are the
candidate's permutation matrices on the component with \(T_1=I_m\).
The two covariances are
\(C_0=a[I-(1-\delta)UU^\top]\) and
\(C_1=a[I-(1-\delta)VV^\top]\), with equal determinants. Let
\(b_0=(1-\delta)/2\) and let \(s_1\ge\cdots\ge s_m\) be the singular
values of \(M\). Applying the Sylvester determinant identity to
\([U\ V]\) gives
\[
\begin{aligned}
\frac{\det((C_0+C_1)/2)}{\det C_0}
&=\delta^{-m}
  \det\begin{pmatrix}
  (1-b_0)I_m & -b_0M\\
  -b_0M^\top & (1-b_0)I_m
  \end{pmatrix}\\
&=\prod_{j=1}^m
  \left[1+\frac{(1-\delta)^2}{4\delta}(1-s_j^2)\right].
\end{aligned}
\]
The matrix \(I_m-MM^\top\) is the Laplacian of a graph with
edge weights \((MM^\top)_{ij}\) for \(i\ne j\). Its graph contains
all the component's edges because \(M\) has positive diagonal.
Every positive edge weight is at least \(1/K^2\). The weighted
matrix-tree theorem therefore yields
\(\prod_{j=2}^m(1-s_j^2)\ge mK^{-2(m-1)}\).
The Gaussian square-root likelihood-ratio moment is the inverse square
root of the determinant ratio above. By Markov's inequality, the
probability that this competitor equals or beats the truth is at most
\[
m^{-1/2}\left(\frac{2K\sqrt{\delta}}{1-\delta}\right)^{m-1}.
\]
There are at most \(\binom nm(m!)^{K-1}\le n^m m^{(K-2)m}\)
choices. For each fixed \(m\ge2\), the product of this count and the
probability tends to zero if \(c>2\), since \(m-c(m-1)<0\).
If \(c>K-1\), choose a fixed \(\xi>0\) and \(m_0\) so that
\(c(m-1)\ge(K-1+\xi)m\) for all \(m\ge m_0\). Using \(m\le n\)
in the count, the sum over these sizes is at most
\(\sum_{m\ge m_0}(C_K n^{-\xi})^m=o(1)\).
This proves the upper side. For example, \(K=4\), \(c=5/2\)
gives \(I_*/\log n\to5/2\) and \(J/\log n\to5/6\).

\subsection{Growing numbers of views}\label{app:growingviews}

The finite-sample bound~\eqref{eq:converse} already holds for every
\(K\). If the maximum error fraction vanishes in probability, its
expectation vanishes by boundedness, and the left side of
\eqref{eq:converse} also vanishes. For exact recovery, revealing the
within-side alignments of a minimizing bipartition reduces the problem
to two-view matching. The exchange bounds in the proof of
Theorem~\ref{thm:exact}(b) depend on \(n\) and \(I_*\) only,
so they also apply when \(K\) grows.

\paragraph{Uniform tails for partition tests.}
For a partition with \(p\) blocks and finite divergence, whiten in
the support of the product of the block marginals. The joint
whitened covariance has eigenvalues \(0<\lambda_j\le p\). The upper
bound follows by applying Cauchy--Schwarz to the standard deviations
of a linear form on the \(p\) blocks, whose marginal covariances are
identities. Put \(x_j=\lambda_j-1\) and let \(g\ge1\) bound all
\(\lambda_j\). Lemma~\ref{lem:llr} gives, under a true tuple,
\[
L_\cP-D_\cP=\tfrac12\sum_j x_j(Z_j^2-1),
\qquad
\sum_j x_j^2\le4gD_\cP,
\]
with independent standard normal \(Z_j\). The second bound uses
\(x-\log(1+x)\ge x^2/[2\max\{1,1+x\}]\) for \(x>-1\).
For \(0<t\le1/2\), \(tx_j\ge-1/2\) and
\(v-\log(1+v)\le v^2\) for \(v\ge-1/2\), so the
log moment of \(-t(L_\cP-D_\cP)\) is at most \(2gt^2D_\cP\).
Using \(t=\kappa/(4g)\), for any fixed \(0<\kappa<1\) we get
\[
\Prob_P(L_\cP<(1-\kappa)D_\cP)
  \le\exp[-\kappa^2D_\cP/(8g)].
\]
Suppose that \(J\ge(1+\epsilon)\log n\). Choose \(\kappa>0\) small
enough that
\((1-\kappa)D_\cP-(p-1)\log n\ge c_\epsilon D_\cP\)
for all \(\cP\), with a fixed \(c_\epsilon>0\). At threshold
\((1-\kappa)D_\cP\), the probability of accepting any false tuple
of identity partition \(\cP\) for a given anchor is at most
\(e^{-c_\epsilon D_\cP}\), by Markov's inequality under the product
law and the count \(n^{p-1}\). Infinite-divergence tests use support
membership as in Appendix~\ref{app:main}. Their true rejection
and false acceptance probabilities are zero.

With \(g=p\), both error bounds are \(n^{-\Omega_\epsilon(1)}\),
so a union bound over at most \(K^K\) partitions suffices when
\(K\log K=o(\log n)\). In particular, the number \(N_{\rm bad}\)
of anchors whose true tuple fails or some false tuple passes is
\(o_{\Prob}(n)\). Resolve proposal conflicts and complete separately
in each view. Each bad anchor can displace at most one correct
proposal per view, so \(\max_a M_a\le2N_{\rm bad}\).

\paragraph{A cut-weight count for weak laws.}
Let \(R_\cP\) be the block diagonal part of \(R\) on \(\cP\),
and put \(c_R(\cP)=\sum_{a<b\text{ across }\cP}\|R_{ab}\|_F^2\).
The spectrum of \(R_\cP\) is in \([1-h,1+h]\). The matrix
\[
H_\cP=R_\cP^{-1/2}(R-R_\cP)R_\cP^{-1/2}
\]
has zero trace, \(\|H_\cP\|_{\mathrm{op}}\le 2h/(1-h)\),
and \(\|H_\cP\|_F^2=(1+O(h))\,2c_R(\cP)\).
The last estimate follows by bounding the singular values of the
two whitening factors between \((1+h)^{-1/2}\) and
\((1-h)^{-1/2}\). Expanding \(-\log\det(I+H_\cP)/2\) yields
\[
D_\cP=(1+O(h))\,c_R(\cP)/2
\]
with an absolute error constant uniform over the partitions and \(K\).
Write \(c_*\) for the minimum bipartition weight. In particular
\(c_*=(2+O(h))I_*\).

For \(u\ge1\), there are at most \(K^{9u}\) partitions of weight at
most \(uc_*\) when \(c_*>0\). To see this, contract randomly selected
edges with probabilities proportional to their weights until
\(t=\lceil4u\rceil\) vertices remain. If \(t\ge K\), the bound
follows from the total partition count \(K^K\). At a step with
\(r>t\) vertices, the total edge weight is at least \(rc_*/2\):
each contracted vertex defines an original cut of weight at least
\(c_*\). Conditional on a specified partition of weight at most
\(uc_*\) surviving thus far, the next step destroys it with
probability at most \(2u/r\). The survival probability is at least
\(\exp(-4u\sum_{r=t+1}^K r^{-1})\ge K^{-4u}\).
A uniformly chosen partition of the final vertices agrees with it
with conditional probability at least \(K^{-t}\).
The outputs corresponding to distinct original partitions are
disjoint, and \(4u+t\le9u\).

If \(c_*\gtrsim\log n\), this count gives, for any fixed \(c_0>0\),
\begin{equation}\label{eq:growing-cut-sum}
\sum_{\cP:\,|\cP|\ge2}e^{-c_0c_R(\cP)}
  \le \sum_{r\ge1}K^{9(r+1)}e^{-c_0rc_*}
  =o(1)
\end{equation}
when \(\log K=o(\log n)\), by grouping the ratios \(c_R(\cP)/c_*\)
into intervals \([r,r+1)\). For the almost-exact upper side,
\(I_*\ge J\) gives \(c_*\gtrsim\log n\). The weak covariance
bound allows \(g=2\) for all sufficiently large \(n\), and the
true and false error bounds are both \(e^{-\Omega_\epsilon(c_R(\cP))}\).
Equation~\eqref{eq:growing-cut-sum} replaces the union bound based on
\(K^K\).

\paragraph{Exact recovery for weak laws.}
We give the uniform estimates needed to extend the two-level
count in Appendix~\ref{app:exact}. For a candidate with moved
source rows \(i\), let \(\cP_i\) partition the views by the candidate
tuple receiving the observations of object \(i\); set
\(m=|\{i:|\cP_i|\ge2\}|\ge2\) and
\(Q=\sum_i(|\cP_i|-1)\). A row outside the moved set has all its
views paired with its anchor and is fixed in every view.
For a fixed moved set and fixed pattern of partitions, a candidate
assigns at most one target from the moved set to each non-anchor
block. There are therefore at most \(m^Q\) candidates for this
pattern. Write \(A_H=\sum_i c_R(\cP_i)\).

Let \(W\) be \(-R^{-1}\) with its diagonal view blocks set to
zero. Its operator norm is at most \(2h/(1-h)\), since deleting
the diagonal blocks of \(I-R^{-1}\) subtracts a block-diagonal
compression of norm at most \(\|I-R^{-1}\|_{\mathrm{op}}\).
The matrices of the discarded true quadratic score \(T\) and of
the introduced false score \(F\) are obtained by deleting the
appropriate block-diagonal compressions of the true and candidate
score matrices. Both thus have operator norm \(O(h)\), uniformly
in \(K\). They have disjoint supports, because edges in \(T\)
are within true objects and those in \(F\) join different objects.
For a partition \(\cP_i\), the identity
\[
R^{-1}=R_{\cP_i}^{-1}
  -R_{\cP_i}^{-1}(R-R_{\cP_i})R^{-1}
\]
shows that the cross-block parts of \(W\) and \(R\) differ in
Frobenius norm by at most \(O(h)\|R-R_{\cP_i}\|_F\). Indeed,
\(R_{\cP_i}^{-1}\) is block diagonal, and the two inverse factors
in the second term are within \(O(h)\) of the identity in operator
norm. The mean of \(T\) is therefore
\(\mu_H=(1+O(h))A_H\); \(F\) has mean zero. The squared Frobenius
norms of the two quadratic-form matrices both equal
\((1+O(h))A_H/2\). For \(F\), this uses that for each view pair,
the number of false edges equals the number of discarded true edges.

The data covariance has spectrum in \([1-h,1+h]\), independently of
\(K\). The Gaussian quadratic-form moment formula and the
\(\log\det\) expansion give, for \(|x|,|y|\le2\), a constant
\(V_H=(1+O(h))A_H\) for which
\[
\log\E\exp\{xF+y(T-\mu_H)\}\le V_H(x^2+y^2)/2 .
\]
Specifically, multiply the coefficient matrix by the covariance
square roots on both sides. Its operator norm stays \(O(h)\),
and its squared Frobenius norm is at most
\((1+O(h))A_H(x^2+y^2)/2\) by the disjoint supports.
The centered log moment is at most \(1+O(h)\) times this squared
Frobenius norm. Set \(D_H=\mu_H^2/V_H\).
A \(p\)-block partition cuts at least \(pc_*/2\) by summing the
cuts of its blocks and dividing by two. Consequently,
\(I_*\ge(2+\epsilon)\log n\) gives, for some fixed
\(\epsilon_0>0\),
\[
D_H=(1+O(h))A_H
  \ge 2(1+\epsilon_0)(m+Q)\log n .
\]

To sum the pattern bounds, let \(w=m\log n\), \(q=Q/m\),
\(\alpha=\log m/\log n\), \(L=(1-\alpha)w\), and \(M=\alpha qw\).
There are at most \(e^{L+m}\) moved sets and \(e^M\)
candidates per pattern. If \(M\le L\), then
\(\alpha\le1/(1+q)\), \(q\ge1\), and
\(L+M\le(1+q)w/2\le D_H/[4(1+\epsilon_0)]\).
The joint moment bound gives \(\Prob(F\ge T)\le e^{-D_H/4}\).
Summing over candidates gives at most \(e^{-L-\delta_0D_H}\)
for a sufficiently small fixed \(\delta_0>0\).
If \(M\ge L\), Cauchy--Schwarz gives
\[
\begin{aligned}
\sqrt{L+\delta_0D_H}+\sqrt{M+\delta_0D_H}
&\le \sqrt{(1+q)w}+2\sqrt{\delta_0D_H}\\
&\le \sqrt{D_H/2},
\end{aligned}
\]
again for small enough \(\delta_0\) depending only on \(\epsilon_0\).
Let \(a_0=\sqrt{2(L+\delta_0D_H)/D_H}\) and \(s_0=1-a_0\).
Then \(a_0\le1/2\) for sufficiently small \(\delta_0\), since
\(L\le qw/(1+q)\) and \(q/(1+q)^2\le1/4\).
One bound on the shared true score gives
\(\Prob(T<s_0\mu_H)\le e^{-L-\delta_0D_H}\).
On \(F\ge T\ge s_0\mu_H\),
\(s_0F-a_0(T-\mu_H)\ge(s_0^2+a_0^2)\mu_H\).
The joint moment bound with parameters
\((s_0\mu_H/V_H,-a_0\mu_H/V_H)\) then gives a probability at most
\(\exp[-D_H(s_0^2+a_0^2)/2]\le e^{-L-M-2\delta_0D_H}\).
All the moment parameters used lie in \([-2,2]\) for large \(n\).
Summing over candidates shows that a pattern fails with probability
at most \(2e^{-L-\delta_0D_H}\) in either case.

Finally \(D_H\ge c_1\sum_i c_R(\cP_i)\) for an absolute
\(c_1>0\). Summing the pattern bound over moved sets and patterns
therefore gives at most
\(\sum_{m\ge2}2(eZ_n)^m=o(1)\), with
\(Z_n=\sum_{\cP:\,|\cP|\ge2}e^{-\delta_0 c_1c_R(\cP)}=o(1)\)
by~\eqref{eq:growing-cut-sum}. This proves the claimed sufficient
condition for exact recovery.

\paragraph{A star with polynomially many views.}
Take \(K=\lceil n^\beta\rceil\) and \(K-1\) independent weak
two-view Gaussian channels of mutual information
\(I=(2+\eta)\log n\), where \(0<\eta<\beta\). The anchor view
contains the concatenated anchor ends, and each leaf contains
only the other end of its channel. In feature order the
covariance is a direct sum of the channels, so \(h=o(1)\)
and the minimum cut is \(I_*=I\). The posterior on the leaf
permutations factors over channels under the independent prior.
The optimal full success probability is therefore the product
of the individual success probabilities.

For a leaf, write \(p\) for the probability that a specified
transposition strictly improves the true likelihood. The exchange
bounds in Appendix~\ref{app:exact} give
\(e^{-I-O(\sqrt I)}\lesssim p\le e^{-I}\), and the intersection
probability for two transpositions sharing an index is at most
\(e^{-3I/2}\). Disjoint transpositions are independent. The
first two terms of inclusion-exclusion give a failure probability
at least \(\binom n2p-O(n^4p^2+n^3e^{-3I/2})\).
Both subtracted terms are \(o(n^2p)\), so the single-leaf failure
probability is at least \(n^{-\eta-o(1)}\).
The product success probability tends to zero because \(\eta<\beta\).

\subsection{Further computational results}\label{app:computationalremarks}

\paragraph{Heterogeneous layered screening.}
For a set of views \(S\) and its partition \(\cP\), let
\(C_S(\cP)=\sum_{a<b\text{ across }\cP} w_{ab}\),
where \(w_{ab}=\lim_n I(X^a;X^b)/\log n\). For fixed \(K\),
Proposition~\ref{prop:strength} shows that the true mean of the
partition score is \([C_S(\cP)+o(1)]\log n\).
Fix a small \(\eta>0\) and test at level
\(\theta_S(\cP)\log n\), where
\(\theta_S(\cP)=(C_S(\cP)-\eta)_+\).
Skip tests whose level is zero. A true tuple fails an enabled
test with probability \(o(1)\) by Lemma~\ref{lem:llr}, since the
variance is \(O_K(\log n)\) and the mean is separated from
the threshold by at least \((\eta/2)\log n\) for large \(n\).
There are at most \(n^p\) tuples of identity partition \(\cP\)
with \(p\) blocks. Each passes its own partition's test
with probability at most \(n^{-\theta_S(\cP)}\) under the corresponding
product law; the bound is 1 when the test is skipped.
Therefore the expected number of false tuples of that type
passing all tests is at most \(n^{p-\theta_S(\cP)}\).
The same bounds apply when the known finite-\(n\) weights
\(I(X^a;X^b)/\log n\) are used in the procedure.

Choose and order a seed \(S_m\) such that
\(\min_\cP C_{S_m}(\cP)/(|\cP|-1)>1\). Denote its first
\(s\) views by \(S_s\), with \(a_s\) the last of them. At step
\(s\), pick a source view \(a\in S_{s-1}\) maximizing
\(w_{aa_s}\). For each source row precompute the list passing
its two-view test with \(a_s\). With threshold exponent
\(\theta=(w_{aa_s}-\eta)_+\) the expected list length is
at most \(1+n^{1-\theta}\). If the list exceeds
\(\min\{n,\lceil2n^{(1-\theta)_++\eta}\rceil\}\), discard it.
Markov's inequality makes the probability of discarding a specified
true object's source list \(o(1)\); if the cap equals \(n\) there is
no discard. Extend the retained prefixes by these lists and
test each extension on \(S_s\). Abort if an intermediate output
exceeds \(2n^{v_s+\eta}\), with
\(v_s=\max\{1,\max_\cP (|\cP|-\theta_{S_s}(\cP))\}\),
where the maximum is over nontrivial partitions of \(S_s\).
The bound on the expected number of retained wrong tuples and
Markov's inequality show that aborts have probability \(o(1)\).
At the last level, choose \(\eta\) small enough that
\(p-\theta_{S_m}(\cP)<1\) for every nontrivial partition.
There are then \(n-o_{\Prob}(n)\) retained true tuples and
\(o_{\Prob}(n)\) retained false tuples. A maximal disjoint selection
followed by completion is simultaneously almost exact on the seed:
each selected false tuple can block at most \(m\) true tuples.

Define
\[
\begin{aligned}
u_1&=1,\qquad
u_s=\max\{1,\max_\cP(|\cP|-C_{S_s}(\cP))\}\quad(s\ge2),\\
\lambda_s&=(1-\max_{a\in S_{s-1}}w_{aa_s})_+ .
\end{aligned}
\]
At step \(s\), the prefix cap and list cap contribute a scanning
exponent at most \(u_{s-1}+\lambda_s+4\eta\).
Thus the dense lists and scans cost \(O_K(n^{\Gamma+\xi})\) for
any \(\xi>0\), by choosing \(\eta\) sufficiently small, with
\(\Gamma=\max\{2,\max_{2\le s\le m}(u_{s-1}+\lambda_s)\}\).
If each scanning exponent without slack is strictly below 2,
the cost is \(O_K(n^2)\). The partition likelihood scores can be
assembled from per-row and per-pair scores. With \(d=\sum_a d_a\),
forming their weights and precomputing scores costs
\(O_K(d^3+nd^2+n^2d)\), as in Theorem~\ref{thm:recursive}.
In a shared coordinate model with independent coordinates the
weight construction and per-row transforms instead cost \(O_K(nd)\).

Attach a new view \(b\) to the current aligned views \(S\) when
\(\sum_{a\in S}w_{ab}>1\). Under the true alignment the
two-view log likelihood for \(S\) concatenated against \(b\)
has mean \([\sum_{a\in S}w_{ab}+o(1)]\log n\).
A test at \((1+\gamma)\log n\), for fixed \(\gamma>0\) below
the margin, misses \(o_{\Prob}(n)\) true edges and allows
\(o_{\Prob}(n)\) false edges. In the actual concatenation,
\(o_{\Prob}(n)\) rows are contaminated. In a maximal disjoint
selection at most one edge can use each contaminated row.
On the other rows the reference tests apply unchanged, so the
selection and completion still have \(o_{\Prob}(n)\) errors.
Iterating gives simultaneous recovery.

In the example in Remark~\ref{rem:compute}, the three seed
bipartition weights are \(1.53,1.50,1.47\), and the singleton
partition has normalized weight \(2.25/2=1.125\).
In order \(1,2,3\), \(u_2=2-0.78=1.22\),
\(\lambda_2=0.22\), and \(\lambda_3=0.25\).
The two scans have exponents \(1.22\) and \(1.47\) before slack.
View 4 has weight \(0.58+0.48+0.02=1.08\) to the seed, so
attach it in one dense two-view scan. All six single edges have
weight less than 1. Each two-view edge estimate has vanishing
overlap by Theorem~\ref{thm:pairwise} applied to the two-view
model. To realize the weights, assign independent coordinate
groups to the six pairs, with two correlated coordinates per
dimension in the group's two views. Choose the number of
coordinates per group and the correlations so the mutual
information divided by \(\log n\) tends to the specified weight
and the maximum correlation tends to zero. The covariance is a
direct sum of weak pair channels, up to reordering coordinates.

\paragraph{Partial alignment and exact linear programming.}
In the weak regime the cut double count just used gives
\(D_\cP\ge (1+O(h))pI_*/2\) for a \(p\)-block partition.
If \(I_*\ge(2+\epsilon)\log n\), choose a fixed \(\eta>0\) so
that each \(D_\cP\ge(p+2\eta)\log n\). Test all tuples at
levels \((p+\eta)\log n\). For each false type, the probability
of accepting any tuple is at most \(n^p n^{-p-\eta}\).
The true miss probability for a specified object is
\(n^{-\Omega_\epsilon(1)}\) by the partition tail bound above.
A union bound over false types and Markov's inequality for
the number of missed true objects give, with high probability,
a partial alignment with no errors and \(O(n^{1-c_\epsilon})\)
missing objects for a fixed \(c_\epsilon>0\).

For the higher-signal guarantee, use the observed scores
\(s_{ab}(i,j)=(Y_i^a)^\top R_{ab}Y_j^b\) in whitened coordinates,
and set \(s_{ba}(j,i)=s_{ab}(i,j)\). For each tuple
\(v=(i_1,\ldots,i_K)\) let \(s(v)=\sum_{a<b}s_{ab}(i_a,i_b)\).
Use the linear program
\[
\begin{aligned}
\text{maximize }&\sum_v s(v)z_v\\
\text{subject to }& z_v\ge0,\qquad
  \sum_{v:\,i_a=i}z_v=1\quad(a\in[K],\,i\in[n]).
\end{aligned}
\]
To analyze its optimum, relabel the rows so the truth is the
identity. The dual prices
\(u_{a,i}=\tfrac12\sum_{b\ne a}s_{ab}(i,i)\)
make each constraint for a true tuple tight. For a false tuple
of identity partition \(\cP\), with source row \(j_B\) in block
\(B\), the dual slack is \(T-F\), where
\[
\begin{aligned}
T&=\tfrac12\sum_{B\in\cP}\sum_{\substack{a\in B\\ b\notin B}}
    s_{ab}(j_B,j_B),\\
F&=\sum_{\substack{a<b\\ a,b\text{ across }\cP}}s_{ab}(i_a,i_b).
\end{aligned}
\]
The within-block scores cancel, and
\(\E T=A_\cP\), \(\E F=0\), with \(A_\cP=c_R(\cP)\).
As quadratic forms of the independent objects in the tuple,
\(T\) and \(F\) have disjoint matrix supports. For each cross-block
edge the coefficient in \(T\) is halved and used on two objects,
while \(F\) uses it once across two objects. The coefficient
matrices have squared Frobenius norms \(A_\cP/4\) and \(A_\cP/2\),
respectively, and operator norm \(O_K(h)\). The log moment of
\(t(F-T+A_\cP)\) for bounded \(t\) is thus at most
\((1+O_K(h))3t^2A_\cP/4\) by the Gaussian quadratic-form
bound used above. For fixed \(0<r<1\), using \(t=2(1-r)/3\) gives
\[
\Prob(T-F\le rA_\cP)
 \le \exp\big\{-[(1-r)^2/3-O_K(h)]A_\cP\big\}.
\]
Since \(A_\cP\ge(1-O(h))pI_*\), for
\(I_*\ge(3+\epsilon)\log n\) we can choose \(r>0\)
small enough that this probability is at most \(n^{-p-c}\)
for some fixed \(c>0\). Summing over at most \(n^p\) false
tuples of each type proves strict dual slack, of order at least
\(\log n\), for every false tuple with high probability.
Complementary slackness then forces the unique optimum to be
the integral true alignment.

The LP has \(n^K\) variables and \(Kn\) equality constraints.
For a bit-complexity implementation, the objectives may be
rounded to integers at bounded absolute error: distribute
each true tuple's rounding error equally among its \(K\)
dual variables, changing any slack only by a bounded amount.
The strict margin above is preserved. Gaussian row norm tails
give \(\max_{a,i}\|Y_i^a\|=O_K(\sqrt{d+\log n})\) with high
probability on whitened data, so coefficients of the rounded
objective use \(O_K(\log(n+d))\) bits. Computing scores to
bounded absolute precision and solving the resulting rational LP
is polynomial for fixed \(K\).

\paragraph{The low-degree lower bound.}
Consider the equal-correlation model with \(d\) independent
coordinates of common correlation \(\rho\ge0\). Let
\(\Gamma^*_{ai,cj}\) be the same-object indicator for rows
\(i,j\) of views \(a\ne c\). The planted degree-\(D\) risk
is the infimum of the expected squared error summed over these
entries, divided by \(K(K-1)n^2\), with each entry estimated by
a polynomial of total degree at most \(D\).
The constant polynomial \(1/n\) gives risk \(1/n-1/n^2\).

Divide the observations by \(\sqrt{1-\rho}\), which preserves
the polynomial degree. In the resulting model each shared center
has law \(N(0,\lambda^2 I_d)\), \(\lambda^2=\rho/(1-\rho)\),
and each observation adds independent \(N(0,I_d)\) noise.
This is Definition~1 of~\citet{even2025computational} with
their \((K,M,p)\) equal to our \((n,K,d)\). Fixing an anchor
only chooses representatives of the same relative alignments,
and does not change the observation law or the same-object labels.
Their Theorem~2.2 bounds our risk by
\[
\begin{aligned}
R_D&\ge \frac1n-\frac{1}{n^2}
 \left(1+64(D+1)D^{36}\frac{\zeta}{1-\sqrt{\zeta}}\right),\\
\zeta&=64D^{40}\frac{b\log n}{(1-\rho)^2}\max(1,K/n),
\end{aligned}
\]
provided \(n\ge2(D+2)^2\) and \(\zeta<1\).
For \(D=\lfloor(\log n)^{1+\eta}\rfloor\),
\(b=(4+\epsilon)/K\), and
\((\log n)^{41+40\eta}\ll K=n^{o(1)}\),
one has \(\zeta=o(1)\) and \(D^{37}=o(n)\).
Therefore \(R_D=(1-o(1))/n\), the risk of chance-level
estimation to first order. By symmetry under view and row
relabelings, the same bound holds for any specified pair of views,
using all the views as input. This is a squared-error guarantee
for polynomial outputs in the planted model.
On the other hand, \(d\gg K\log n\) implies \(K\rho=o(1)\) at
this signal. In the uniform cut graph a \(p\)-block partition
cuts at least \(K(p-1)/2\) pairs, with equality for the singleton
partition. The uniform divergence expansion therefore gives
\(J=(1+o(1))Kd\rho^2/4\). By
Appendix~\ref{app:growingviews}, almost exact recovery is
information-theoretically possible there.

\paragraph{A logarithmic-degree estimator at fixed signal.}
Suppose \(b=d\rho^2/\log n\to b_0>0\), \(\rho\to0\),
and \(K\to\infty\). Fix an integer \(q>16/b_0\) with \(q\ge2\).
For each target view pair use a fixed subset of \(q\) views
that contains this pair. For a candidate tuple let \(L_\cP\)
be its partition log likelihood ratio on these views.
The minimum normalized cut bound just used, with \(K\) replaced
by \(q\), shows
\(\E_P L_\cP\ge(4-o(1))(|\cP|-1)\log n\).
Write
\(\ell_\cP=\tfrac52(|\cP|-1)\) and \(u_\cP=3(|\cP|-1)\).
We approximate a threshold with transition from \(\ell_\cP\)
to \(u_\cP\) applied to \(L_\cP/\log n\).

Uniformly over the finite set of identity partitions \(\cT\)
of the tuple, including the true one-block partition, for every
fixed \(t>0\) there is a constant \(C_t\) such that
\(\E_{Q_\cT}e^{t|L_\cP|}\le n^{C_t}\).
Indeed, in one coordinate \(L_\cP\) has a log determinant term
of order \(O_q(\rho^2)\) by zero trace of the covariance
perturbation. Its quadratic form after whitening under \(Q_\cT\)
has operator norm \(O_q(\rho)\), squared Frobenius norm
\(O_q(\rho^2)\), and mean \(O_q(\rho^2)\). The last bound follows
by expanding the two inverses to first order and using trace zero
of their first-order difference. The Gaussian moment formula
bounds the log moment in either sign by \(O_{q,t}(\rho^2)\)
per coordinate, and \(d\rho^2=O(\log n)\). Also, on a true
tuple, \(L_\cP/\log n\) has variance \(O_q(1/\log n)\).

Fix \(M>\max_\cP u_\cP\) and consider on \([-M,M]\) the
function equal to 0 below \(\ell_\cP\), linear on
\([\ell_\cP,u_\cP]\), and equal to 1 above \(u_\cP\).
Writing \(x=M\cos\theta\) turns this into an even periodic
function with Lipschitz constant \(O_q(M)\). Convolve it with
the normalized nonnegative Jackson kernel
\((\sin(j\theta/2)/\sin(\theta/2))^4\).
Before normalization, its integral is \(\Theta(j^3)\) and
its absolute first moment is \(O(j^2)\), by bounding the
kernel by \(C\min\{j^4,|\theta|^{-4}\}\) on \([-\pi,\pi]\)
and from below by \(cj^4\) on \([-1/j,1/j]\).
Taking \(j=O_q(M)\) with a sufficiently large constant
produces an even trigonometric polynomial of degree \(2(j-1)\)
within \(1/8\) uniformly. As a polynomial of \(x/M\) it
takes values in \([0,1]\) on \([-M,M]\).
Compose this polynomial with the probability of a majority
in \(k=O_q(\log n)\) independent Bernoulli trials with this
success parameter. A binomial tail bound gives a polynomial
\(p_\cP\) of degree \(O_q(M\log n)\), still bounded by 0 and 1
on the interval, at most \(n^{-3q}\) there below \(\ell_\cP\),
and at least \(1-n^{-3q}\) there above \(u_\cP\).

Let \(W=\prod_\cP p_\cP(L_\cP/\log n)\) and
\(U=\max_\cP|L_\cP|/\log n\).
A degree-\(k\) polynomial bounded by 1 on \([-M,M]\) is bounded
by \((8|x|/M)^k\) for \(|x|\ge M\), by expanding in Chebyshev
polynomials on the rescaled interval. Thus on \(U>M\),
\(W^2\le \exp(CU\log n)\) for a \(C\) independent of \(M\),
using \(\log(8U/M)\le C' U/M\) and the degree bound above.
Take \(t=C+1\) in the moment estimate. Increasing the fixed
\(M\) if necessary gives uniformly in the identity partition
\[
\E[W^2\mathbf{1}_{\{U>M\}}]+\Prob(U>M)\le n^{-6q}.
\]
On a true tuple, the concentration of its partition scores
therefore gives \(\|W-1\|_{L^2}=o(1)\).
For a false tuple with identity partition \(\cP\) of \(p\)
blocks, its own score passes \(\ell_\cP\log n\) with probability
at most \(n^{-5(p-1)/2}\), by Markov's inequality under \(Q_\cP\).
The polynomial suppression on the rest of the interval and
the tail bound give \(\|W\|_{L^2}=O(n^{-5(p-1)/4})\).

To estimate the indicator vector of the row matching a fixed
anchor row in the target view, sum \(W e_j\) over all candidate
tuples for this anchor, with \(j\) the target row of each tuple
and \(e_j\) the corresponding standard basis vector. The true
term differs from the true vector by \(o(1)\) in \(L^2\) norm.
There are at most \(n^{p-1}\) candidates of each false type,
so their aggregate norm is \(O(n^{-(p-1)/4})\) by the
\(L^2\) triangle inequality. Summing over the finitely many
types gives \(o(1)\) mean squared error per anchor row.
Each score is a quadratic polynomial in the observations.
The total degree, with \(q,M\) fixed, is \(O_{b_0}(\log n)\).
The bound is uniform over the target view pair, giving
\(R_D=o(1/n)\) at this degree.

\section{Numerical details}\label{app:numerics}

\newcommand{\NumericsPredictions}{33}
\newcommand{\NumericsSupported}{32}
\newcommand{\NumericsOriginalSHA}{c8a6e1ef02c0422c916877eb8aee5c568f8b789da39c75713bd3f0d66116173f}
\newcommand{\NumericsSupplementSHA}{a2224efdb57e1143f6dc03d76c341c32eaa0f5f2f8ab499dfe1f0192bfe25de9}

\subsection{Settings and analysis}

Table~\ref{tab:num-settings}
gives the sizes, feature dimensions and repeat counts. G1, G2, G4
and G5 use equicorrelated Gaussian coordinates. G1, G2 and G4 vary
$b=d\rho^2/\log n$, with additional G2 cells calibrated to exact
$J/\log n$. G5 calibrates $I_*/\log n$ directly. G3 uses clustered,
hierarchical and path covariances. G6 uses three views with fixed
pair information for $AB$ and varies how strongly $C$ is correlated
with both. All methods within a repeat use the same observations and
permutations. G6 also shares the $AB$ observations across designs.

\begin{table}[htbp]
\centering\small
\caption{Full-run grids. Decoders share the same independent draw in
each repeat. In G3, $n=8$ occurs only in the mixed-branch design.}
\label{tab:num-settings}
\begin{tabular}{lrlrr}
\toprule
Group & $K$ & $n$ & $d$ & Repeats/cell \\
\midrule
G1 & 3 & 256, 512, 1024, 2048 & 16384 & 16 \\
G2 & 3, 4, 6, 8 & 1024, 4096, 8192 & 4096 & 8 \\
G3 & 4, 6, 8 & 8, 256, 1024, 4096, 8192 & 4096 & 4 \\
G4 & 3 & 256, 1024, 4096, 8192 & 4096 & 4 \\
G5 & 3 & 256, 512, 1024, 2048, 4096 & 2048 & 8 \\
G6 & 3 & 256, 1024, 4096, 8192 & 4096 & 16 \\
\bottomrule
\end{tabular}
\end{table}

Complete-tuple overlap is the fraction of anchor rows whose partners
are all correct. Pair overlap counts only the partners for the
specified pair. We first compute each statistic within a repeat and
then average over repeats at fixed size, design and signal. Gains
subtract the baseline within each repeat before averaging. We use
repeats as the unit for standard errors. Under the recorded rules,
a directional difference is resolved if it exceeds twice its
combined standard error in magnitude. A supplemental point
prediction fails if more than \(5\%\) of its cells fall outside
the recorded tolerance or any cell deviates by more than \(0.10\).
Interval predictions are checked cell by cell.
For G2 we combine the first 616 records from the original run with
two 28-record shards. The last 21 records from the original run
duplicate shard records and are omitted.

\begin{figure}[tbp]
\centering
\includegraphics[width=\linewidth]{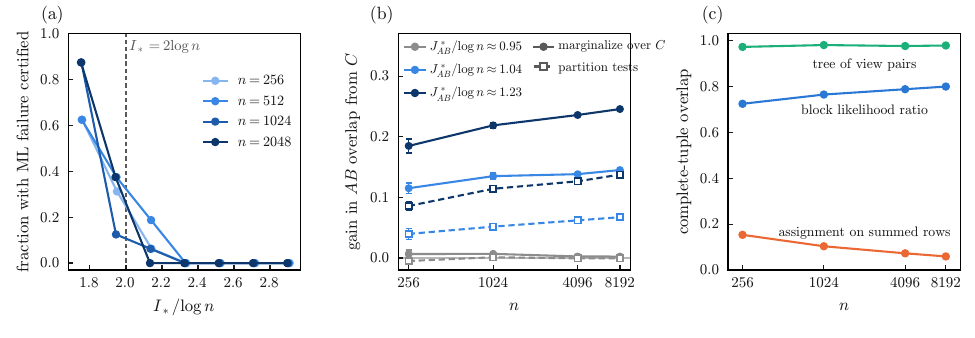}
\caption{Finite-$n$ checks of the main thresholds; points are means over replicates.
(a)~G1, $K=3$, $d=16384$: fraction of the 16 replicates per cell in which a feasible
assignment strictly beating the planted one was found, against the realized $I_*/\log n$.
Such an assignment certifies that maximum likelihood fails. The opposite certificate never
closed, so a zero is not evidence of success.
(b)~G6, $K=3$, $d=4096$, $I_{AB}/\log n=0.95$: paired change in the $A$--$B$ overlap when
view $C$ is read, with one standard error. The two designs with $J^*_{AB}>\log n$ show larger gains
(Theorem~\ref{thm:pairwise}).
(c)~G3 path design, $K=4$, $dr^2/\log n=3$, four replicates: the block likelihood ratio
keeps the informative channel across the root cut, while summed rows dilute it
(Remark~\ref{rem:rowsum}). A tree of original view pairs also does well on this design.}
\label{fig:num-theorems}
\end{figure}

Figure~\ref{fig:num-theorems} collects the three comparisons that bear directly on Theorems~\ref{thm:exact} and~\ref{thm:pairwise} and on Remark~\ref{rem:rowsum}; the tables below give the cell values.

\subsection{Exact recovery: failure certificates and winning swaps}

G1 scores the planted alignment and seeks a feasible assignment with
a strictly larger score. Such an assignment certifies failure of
maximum likelihood. The opposite certificate requires a valid upper
bound that closes at the planted score. Table~\ref{tab:num-g1} counts
failures and unresolved instances separately. The failure counts
fall as the realized $I_*/\log n$ increases. The dual certificate
never closed on this grid, so a cell with zero certified failures
provides no positive lower bound on the probability of maximum
likelihood recovery.

\begin{table}[htbp]
\caption{G1. Equicorrelated failure certificates at $K=3$ and $d=16384$. Each size and signal cell has 16 replicates and shows the number of instances with a feasible assignment beating the truth. The $I_*/\log n$ column gives the range across sizes. The truth was certified as a maximum-likelihood assignment in 0 of 448 instances.}
\label{tab:num-g1}
\centering
\small
\setlength{\tabcolsep}{4pt}
\begin{tabular}{@{}rrrrrrrr@{}}
\toprule
$b$ & $I_*/\log n$ & 256 & 512 & 1024 & 2048 & Failures & Unresolved \\
\midrule
1.8 & 1.751--1.758 & 10 & 10 & 14 & 14 & 48 & 16 \\
2.0 & 1.943--1.951 & 5 & 6 & 2 & 6 & 19 & 45 \\
2.2 & 2.134--2.143 & 1 & 3 & 1 & 0 & 5 & 59 \\
2.4 & 2.325--2.335 & 0 & 0 & 0 & 0 & 0 & 64 \\
2.6 & 2.516--2.527 & 0 & 0 & 0 & 0 & 0 & 64 \\
2.8 & 2.706--2.719 & 0 & 0 & 0 & 0 & 0 & 64 \\
3.0 & 2.896--2.910 & 0 & 0 & 0 & 0 & 0 & 64 \\
\bottomrule
\end{tabular}
\end{table}

G5 counts strict likelihood-improving two-row exchanges across each
singleton cut, as in the converse in Appendix~\ref{app:exact}.
Table~\ref{tab:num-g5} reports observed counts and the exact
expectation from the Gaussian swap probability, under the same
cut and repeat aggregation. In this isotropic cut model the
canonical correlation is $r=\sqrt{1-\exp(-2I_*/d)}$. Rotating the
row-difference pairs to their independent sums and differences
gives expected count
$\binom n2\Prob(B<(1-r)/2)$ for $B\sim\mathrm{Beta}(d/2,d/2)$.
G1 can also find better assignments that move more than two rows.

\begin{table}[htbp]
\caption{G5. Beneficial transpositions for $K=3$ and $d=2048$. The columns index the cut information $I(X^S;X^{S^c})/\log n$. Cells give the mean observed count and the mean exact expectation, separated by a slash and averaged over the 3 singleton cuts in each of 8 independent replicates. The last row sums counts and their expectations over sizes, cuts and replicates.}
\label{tab:num-g5}
\centering
\small
\setlength{\tabcolsep}{4pt}
\begin{tabular}{@{}rrrrrr@{}}
\toprule
$n$ & 1.0 & 1.5 & 1.8 & 2.0 & 2.2 \\
\midrule
256 & 14.6 / 14.2 & 0.917 / 0.741 & 0.0833 / 0.129 & 0 / 0.0406 & 0 / 0.0128 \\
512 & 28.2 / 27 & 1 / 0.995 & 0.125 / 0.141 & 0 / 0.0385 & 0 / 0.0106 \\
1024 & 58.1 / 51.6 & 1.5 / 1.34 & 0.125 / 0.154 & 0.0417 / 0.0367 & 0.0417 / 0.00878 \\
2048 & 89.3 / 98.9 & 1.33 / 1.82 & 0.0833 / 0.17 & 0 / 0.0352 & 0 / 0.00732 \\
4096 & 183 / 190 & 2.12 / 2.47 & 0.167 / 0.187 & 0.0417 / 0.0338 & 0.0417 / 0.00612 \\
\midrule
Total & 8960 / 9169.1 & 165 / 176.8 & 14 / 18.8 & 2 / 4.4 & 2 / 1.1 \\
\bottomrule
\end{tabular}
\end{table}

\subsection{Pairwise matchability: the gain from a third view}

In G6, a direct two-view assignment provides the $AB$ baseline.
We compare it with two methods reading $C$: local marginalization
over possible third rows before the $AB$ assignment, and a
three-view partition-test decoder. Each candidate's log-likelihood
ratio must exceed $(|\cP|-1+0.05)\log n$ for every partition $\cP$.
Each anchor proposes
the passing tuple with the largest singleton-partition log-likelihood
ratio, if one passes. The decoder greedily keeps disjoint triples in
decreasing order of that score, runs a maximum-weight $AB$ assignment
on the remaining rows, and matches the remaining $C$ rows to the
unmatched anchors in index order. Table~\ref{tab:num-g6} reports
paired changes in $AB$ overlap and the information level
$J^*_{AB}/\log n$. With the weakest $C$ both gains stay near zero.
At $n\ge1024$, the marginalization gain ranges from $0.002$ to
$0.007$ and exceeds two paired standard errors in each cell.
The two stronger designs put $J^*_{AB}$ above $\log n$ and show
larger gains on this grid, illustrating Theorem~\ref{thm:pairwise}
at finite $n$.

\begin{table}[htbp]
\caption{G6. Matching $A$ to $B$ with and without view $C$ at $K=3$, $d=4096$ and $I_{AB}/\log n=0.95$. Each replicate uses the same $AB$ data across strengths of $C$. Gains are paired $AB$ overlap differences relative to a direct two-view assignment, for local marginalization over $C$ and for three-view partition decoding. Entries give means over 16 replicates with one standard error for each gain.}
\label{tab:num-g6}
\centering
\small
\setlength{\tabcolsep}{4pt}
\begin{tabular}{@{}rrrrrr@{}}
\toprule
$n$ & $J/\log n$ & $J^*_{AB}/\log n$ & Direct $AB$ & Marginalization gain & Partition gain \\
\midrule
\multicolumn{6}{@{}l}{$I_{AC}/\log n=I_{BC}/\log n=0.2$, weakest partition \(AB\mid C\)} \\
256 & 0.381 & 0.950 & 0.722 & \(+0.0063\pm0.0062\) & \(-0.0051\pm0.0029\) \\
1024 & 0.379 & 0.950 & 0.698 & \(+0.0067\pm0.0023\) & \(+0.0010\pm0.0012\) \\
4096 & 0.377 & 0.950 & 0.678 & \(+0.0024\pm0.0007\) & \(-0.0008\pm0.0004\) \\
8192 & 0.376 & 0.950 & 0.668 & \(+0.0022\pm0.0007\) & \(-0.0008\pm0.0004\) \\
\addlinespace
\multicolumn{6}{@{}l}{$I_{AC}/\log n=I_{BC}/\log n=0.6$, weakest partition \(A\mid B\mid C\)} \\
256 & 1.046 & 1.046 & 0.722 & \(+0.1150\pm0.0085\) & \(+0.0396\pm0.0092\) \\
1024 & 1.043 & 1.043 & 0.698 & \(+0.1349\pm0.0050\) & \(+0.0516\pm0.0039\) \\
4096 & 1.041 & 1.041 & 0.678 & \(+0.1382\pm0.0019\) & \(+0.0620\pm0.0011\) \\
8192 & 1.039 & 1.039 & 0.668 & \(+0.1450\pm0.0015\) & \(+0.0672\pm0.0013\) \\
\addlinespace
\multicolumn{6}{@{}l}{$I_{AC}/\log n=I_{BC}/\log n=0.8$, weakest partition \(A\mid B\mid C\)} \\
256 & 1.237 & 1.237 & 0.722 & \(+0.1848\pm0.0112\) & \(+0.0857\pm0.0067\) \\
1024 & 1.233 & 1.233 & 0.698 & \(+0.2188\pm0.0048\) & \(+0.1141\pm0.0041\) \\
4096 & 1.229 & 1.229 & 0.678 & \(+0.2359\pm0.0024\) & \(+0.1264\pm0.0021\) \\
8192 & 1.228 & 1.228 & 0.668 & \(+0.2457\pm0.0016\) & \(+0.1375\pm0.0014\) \\
\bottomrule
\end{tabular}
\end{table}

\subsection{Recursive decoding and the choice of super-view score}

G3 compares likelihood-ratio recursion, a tree using only original
view pairs, and summed-row assignment. The original and tighter
binary hierarchies differ in how far child information exceeds
parent information. The mixed-branch design uses both binary and
ternary nodes, and its oracle root score measures root matching
given correct child alignments. In the path design the root block
likelihood uses the informative channel across the cut, while
summing all coordinates in a block dilutes that channel. The fitted
hierarchies and the path measurements are in
Table~\ref{tab:num-g3}. The path contrast is between two choices of
super-view statistic. A tree of original view pairs can do well on
that design.

\begin{table}[tbp]
\centering
\caption{Mean complete-tuple overlap for recursive partition tests, an original-view tree, and maximum-weight assignments on block sums. We show all signals at the largest tested $n$ for each hierarchy and for the mixed-branch design, and all tested sizes for the path. Signal is $I_{\mathrm{root}}/\log n$ for the first three designs and $dr^2/\log n$ for the path. A dash marks an unmeasured method.}
\label{tab:num-g3}
\begin{tabular}{lrrrrrr}
\toprule
Design & $K$ & $n$ & Signal & Recursive & Tree & Sum assignment \\
\midrule
Original hierarchy & 6 & 8192 & 0.75 & 0.204 & 0.010 & 0.463 \\
 & 6 & 8192 & 1 & 0.433 & 0.022 & 0.717 \\
 & 6 & 8192 & 1.25 & 0.626 & 0.040 & 0.890 \\
 & 6 & 8192 & 1.5 & 0.765 & 0.063 & 0.976 \\
 & 8 & 8192 & 0.75 & 0.205 & 0.002 & 0.462 \\
 & 8 & 8192 & 1 & 0.433 & 0.007 & 0.720 \\
 & 8 & 8192 & 1.25 & 0.625 & 0.017 & 0.893 \\
 & 8 & 8192 & 1.5 & 0.762 & 0.028 & 0.980 \\
\midrule
Tight hierarchy & 6 & 8192 & 0.75 & 0.019 & 0.000 & 0.167 \\
 & 6 & 8192 & 1 & 0.141 & 0.003 & 0.585 \\
 & 6 & 8192 & 1.25 & 0.365 & 0.010 & 0.878 \\
 & 6 & 8192 & 1.5 & 0.595 & 0.025 & 0.977 \\
 & 8 & 8192 & 0.75 & 0.008 & 0.000 & 0.126 \\
 & 8 & 8192 & 1 & 0.099 & 0.000 & 0.570 \\
 & 8 & 8192 & 1.25 & 0.314 & 0.000 & 0.876 \\
 & 8 & 8192 & 1.5 & 0.557 & 0.002 & 0.977 \\
\midrule
Mixed branch & 6 & 1024 & 1.125 & 0.216 & 0.152 & --- \\
 & 6 & 1024 & 1.2 & 0.278 & 0.210 & --- \\
\midrule
Path & 4 & 256 & 3 & 0.725 & 0.973 & 0.153 \\
 & 4 & 1024 & 3 & 0.765 & 0.981 & 0.104 \\
 & 4 & 4096 & 3 & 0.788 & 0.977 & 0.073 \\
 & 4 & 8192 & 3 & 0.800 & 0.979 & 0.059 \\
\bottomrule
\end{tabular}
\end{table}

\subsection{Layered screening at finite size}

\begin{figure}[tbp]
\centering
\includegraphics[width=0.99\linewidth]{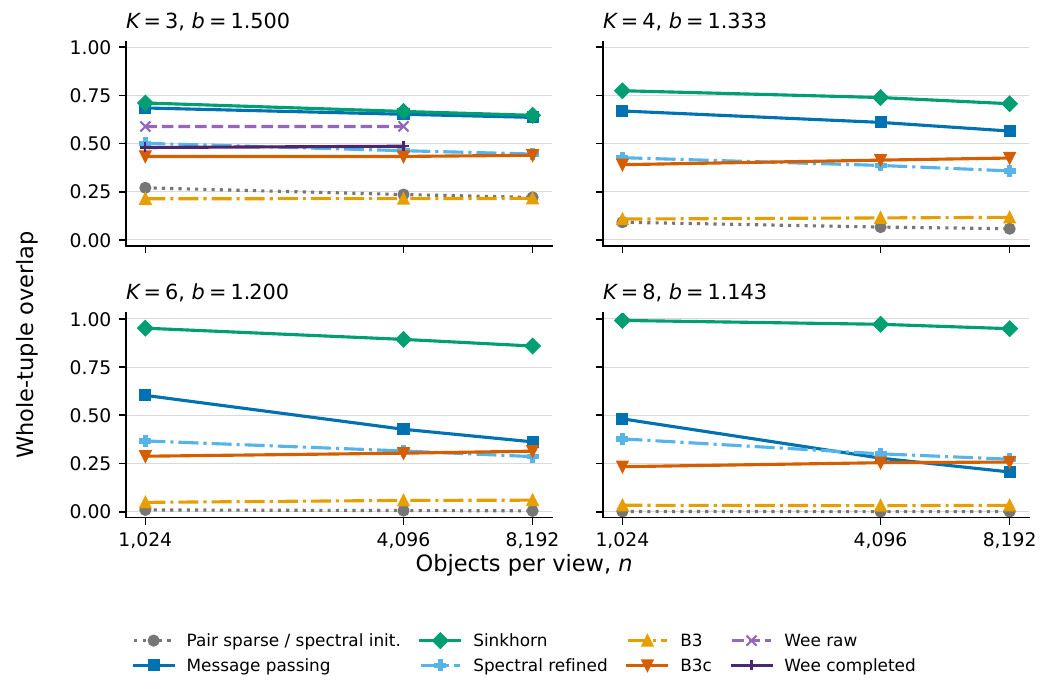}
\caption{G2 whole-tuple overlaps across sizes at the nominal row-wise boundary \(b=K/(K-1)\). Points are means of 8 replicates. The two Wee series are shown only where evaluated. Pair sparse and spectral initialization coincide in these rows.}
\label{fig:num-g2}
\end{figure}

G2 compares the screening decoder of Appendix~\ref{app:screen}
(B3), its variant B3c with adjusted thresholds and truncation,
the row-wise pairwise sum of~\citet{wee2026geometric}, and other
polynomial-time decoders. Table~\ref{tab:num-g2} and
Figure~\ref{fig:num-g2} report complete-tuple overlap. Raw proposals
from the row-wise sum can reuse rows and are reported separately
from its completed output. The lists and prefix caps limit
coverage at moderate $n$. In particular, a margin that is fixed
in the asymptotic scaling can be small compared with finite-sample
score fluctuations. Sinkhorn is an empirical comparator here and
has no recovery guarantee in this setting.

G2 uses \(d=4{,}096\) and 8 paired replicates per setting. Our outcome is the fraction of anchor rows whose entire tuple is correct. B3 uses the midpoint thresholds of layered screening, while B3c uses the finite-sample thresholds and keeps the highest-scoring prefixes when an intermediate layer exceeds its cap. For \(K>4\), their displayed overlaps include the remaining views attached to the four-view seed. Pair sparse, message passing, Sinkhorn, and the initial and refined spectral outputs were evaluated throughout the grid.

Table~\ref{tab:num-g2} shows the signal sweep at its middle size. B3c's mean finished overlap exceeds B3's in 84 of 84 settings. At \(K=4,b=2\) it rises from 0.712 to 0.760 as \(n\) grows from \(1{,}024\) to \(8{,}192\). At \(b=2\) the mean fraction of true seed tuples with all entries in the retained B3c lists ranges from 0.688 to 0.805 for \(K=3,4\), so list screening limits overlap even before the final tests. On the calibrated \(J/\log n\) grids, the largest B3c mean is 0.494. Message passing and Sinkhorn reach 0.744 and 0.774, respectively.

Figure~\ref{fig:num-g2} compares sizes at the nominal row-wise boundary. At \(K=8,b=8/7\), Sinkhorn's means are 0.992, 0.972, 0.950 in increasing size order, compared with 0.234, 0.254, 0.258 for B3c. At the weaker nominal \(b=0.821\), Sinkhorn falls from 0.617 to 0.376. These are finite-size descriptions of the implementations.

The Wee row-wise rule was evaluated for \(K=3\) at \(n=1{,}024,\, 4{,}096\), on both grids. Its raw whole-tuple overlap exceeds its completed overlap by 0.058--0.117 across cell means. The raw rule chooses partners independently for each anchor row. The completed rule resolves collisions and fills unmatched rows in index order. Section~\ref{sec:num-predictions} discusses the prediction for the raw proposals.

\begin{table}[tbp]
\centering
\caption{G2 whole-tuple overlaps at \(n=4{,}096\). Each cell is the mean of 8 replicate overlaps. Nominal \(b\) and calibrated \(J/\log n\) denote different grids. The dash denotes an unrun method. PS/SI are pair sparse and spectral initialization, which have identical overlaps in these rows. MP is message passing and SR is spectral refinement. Wee raw allows row reuse. All other columns report the output after enforcing a permutation in each view.}\label{tab:num-g2}
\small
\setlength{\tabcolsep}{3.5pt}
\begin{tabular}{rlrrrrrrrr}
\toprule
\(K\) & Signal & PS/SI & MP & Sinkhorn & SR & B3 & B3c & \multicolumn{2}{c}{Wee} \\
\cmidrule(lr){9-10}
 & & & & & & & & Raw & Completed \\
\midrule
3 & \(b=1.333\) & 0.160 & 0.512 & 0.509 & 0.349 & 0.160 & 0.334 & 0.491 & 0.374 \\
 & \(b=1.417\) & 0.197 & 0.582 & 0.589 & 0.407 & 0.195 & 0.384 & 0.545 & 0.435 \\
 & \(b=1.500\) & 0.236 & 0.652 & 0.667 & 0.463 & 0.215 & 0.433 & 0.589 & 0.486 \\
 & \(b=1.750\) & 0.373 & 0.812 & 0.857 & 0.628 & 0.273 & 0.579 & 0.709 & 0.630 \\
 & \(b=2.000\) & 0.528 & 0.905 & 0.954 & 0.764 & 0.331 & 0.700 & 0.800 & 0.741 \\
 & \(J/\log n=1.050\) & 0.209 & 0.607 & 0.617 & 0.426 & 0.202 & 0.402 & 0.561 & 0.453 \\
 & \(J/\log n=1.150\) & 0.279 & 0.713 & 0.744 & 0.525 & 0.234 & 0.486 & 0.633 & 0.539 \\
\addlinespace[3pt]
4 & \(b=1.000\) & 0.015 & 0.296 & 0.304 & 0.132 & 0.041 & 0.187 & -- & -- \\
 & \(b=1.167\) & 0.034 & 0.457 & 0.525 & 0.242 & 0.081 & 0.299 & -- & -- \\
 & \(b=1.333\) & 0.067 & 0.610 & 0.739 & 0.387 & 0.114 & 0.414 & -- & -- \\
 & \(b=1.667\) & 0.195 & 0.811 & 0.969 & 0.680 & 0.191 & 0.612 & -- & -- \\
 & \(b=2.000\) & 0.387 & 0.904 & 0.998 & 0.871 & 0.269 & 0.748 & -- & -- \\
 & \(J/\log n=1.050\) & 0.027 & 0.404 & 0.452 & 0.204 & 0.072 & 0.261 & -- & -- \\
 & \(J/\log n=1.150\) & 0.043 & 0.509 & 0.598 & 0.288 & 0.092 & 0.337 & -- & -- \\
\addlinespace[3pt]
6 & \(b=0.667\) & 0.000 & 0.042 & 0.068 & 0.010 & 0.001 & 0.024 & -- & -- \\
 & \(b=0.933\) & 0.001 & 0.210 & 0.451 & 0.090 & 0.014 & 0.131 & -- & -- \\
 & \(b=1.200\) & 0.006 & 0.428 & 0.894 & 0.315 & 0.059 & 0.304 & -- & -- \\
 & \(b=1.600\) & 0.054 & 0.700 & 0.998 & 0.752 & 0.135 & 0.548 & -- & -- \\
 & \(b=2.000\) & 0.215 & 0.848 & 1.000 & 0.944 & 0.229 & 0.709 & -- & -- \\
 & \(J/\log n=1.050\) & 0.000 & 0.093 & 0.170 & 0.027 & 0.003 & 0.055 & -- & -- \\
 & \(J/\log n=1.150\) & 0.000 & 0.144 & 0.288 & 0.051 & 0.007 & 0.085 & -- & -- \\
\addlinespace[3pt]
8 & \(b=0.500\) & 0.000 & 0.003 & 0.011 & 0.001 & 0.000 & 0.001 & -- & -- \\
 & \(b=0.821\) & 0.000 & 0.078 & 0.438 & 0.040 & 0.003 & 0.059 & -- & -- \\
 & \(b=1.143\) & 0.001 & 0.278 & 0.972 & 0.300 & 0.031 & 0.254 & -- & -- \\
 & \(b=1.571\) & 0.015 & 0.598 & 1.000 & 0.817 & 0.099 & 0.516 & -- & -- \\
 & \(b=2.000\) & 0.126 & 0.798 & 1.000 & 0.975 & 0.194 & 0.679 & -- & -- \\
 & \(J/\log n=1.050\) & 0.000 & 0.011 & 0.050 & 0.003 & 0.000 & 0.007 & -- & -- \\
 & \(J/\log n=1.150\) & 0.000 & 0.022 & 0.110 & 0.007 & 0.000 & 0.014 & -- & -- \\
\bottomrule
\end{tabular}
\end{table}

\subsection{True-row ranks}

In G4, each candidate's log-likelihood ratio must exceed
$(|\cP|-1+0.05)\log n$ for every partition $\cP$.
Each anchor proposes the passing tuple with
the largest pairwise sum, if one passes. In each non-anchor view,
the first proposer in index order keeps a contested row, and the
remaining anchors and unused rows are matched in index order.
In the weak equicorrelated limit with $K=3$, this margin gives the
unpruned partition-test decoder a threshold of $b=41/30$
(Appendix~\ref{app:main}), above the tested value $b=4/3$.

In the three-view rank-pruning experiment, we kept the leading candidates by pair score in each non-anchor view before applying partition tests. The list widths were $\lceil n^{1-b/4}\rceil$ and half that width rounded up. Across the sampled cells, the mean fraction of true tuples retained by both lists ranged from $92.3\%$--$95.7\%$ at nominal width and $82.7\%$--$92.9\%$ at half width. At $n=8192$ and the weakest tested signal, the lists generated $0.247\%$ and $0.062\%$ of all candidate tuples, respectively. The experiment also measured unpruned enumeration at $n=256$. There, the largest mean complete-tuple overlap losses relative to enumeration were $0.008$ at nominal width and $0.030$ at half width.

\subsection{Predictions recorded before the runs}\label{sec:num-predictions}

The original and supplemental prediction files were recorded before
their corresponding runs. The SHA-256 digests of the two prediction
files are listed in the \path{README} of the ancillary files.
Under the recorded rules, \NumericsSupported{} of
\NumericsPredictions{} predictions held. The sole exception was
the finite-size prediction for the row-wise sum. The recorded
simulation omitted the $n-1$ candidate tuples that
pair an anchor with the two other views of the same wrong object.
Those two rows have mean inner product $d\rho$, which makes these
impostor tuples stronger competitors than candidates whose two
rows come from different wrong objects. A diagnostic simulation at
$n=1024$ and $4096$ retained these tuples. The simulated minus
observed raw per-anchor accuracy ranged from $-0.028$ to $+0.025$.
Impostor tuples won at $16\%$--$37\%$ of simulated anchors. The original
prediction remains a failure under its recorded rule. This is
the impostor mechanism in Theorem~\ref{thm:ladder}(ii).

\end{document}